\documentclass{report}
\usepackage{graphicx} % Required for inserting images

\usepackage{blindtext}
\usepackage[a4paper, total={6in, 8in}]{geometry}
\usepackage{authblk}
\usepackage{hyperref}
\usepackage[numbers]{natbib}

\usepackage{graphicx}      % for \includegraphics
\usepackage{makeidx}
\usepackage{multibib}
\usepackage{ae}            % nicer fonts in PDF
\usepackage{times}          
\usepackage{dsfont}        % Symbol R for real numbers
\usepackage{exscale}       % automatically enlarges nested brackets. You must       
\usepackage{psfrag}        % replace text in eps with latex formulas
\usepackage{amsmath}  % nicer equation arrays, e.\,g. with align, etc
\usepackage{amsfonts}
\usepackage{amssymb}
\usepackage{exscale}
\usepackage{dsfont}
\usepackage{subfigure}
\usepackage{ae}
\usepackage{url}      % for web addresses in citations
\usepackage{color} % color for hyperlinks and tables
\usepackage{colortbl}
\usepackage{hyperref} % for hyperlinks in document
\usepackage[german,american]{babel}

\usepackage{algorithmic}
\usepackage{algorithm}
\usepackage{subfigure}

\usepackage{epsfig}                                                     % Grafik (.eps)
\usepackage{graphics}                                           % Grafik (.jpg, .tif. pdf)
\usepackage{pstricks}                                           % for drawings
\usepackage{pst-node}                                           % for nodes
\usepackage{pst-plot}                                           % for plot
\usepackage{pst-tree}                                           % for trees
\usepackage{pst-3d}                                                     % for 3D
\usepackage{pstricks-add}

\usepackage{moreverb}                                           % f<9f>r verbatimtab
\usepackage{multicol}                                           % f<9f>r mehrere Spalten
\usepackage{multirow}

\definecolor{darkred}{rgb}{0.5,0,0}
\definecolor{darkgreen}{rgb}{0,0.5,0}
\definecolor{darkblue}{rgb}{0,0,0.5}

\definecolor{highallergenic}{rgb}{0.7,0.0,0.0}     % for text
\definecolor{midallergenic}{rgb}{1.0,0.4,0.0}     % for text
\definecolor{lowallergenic}{rgb}{0.8,0.6,0.0} % for text

\definecolor{highallergenicbg}{rgb}{0.90,0.80,0.80}     % for background
\definecolor{midallergenicbg}{rgb}{1.0,0.9,0.8}     % for background
\definecolor{lowallergenicbg}{rgb}{0.95,0.95,0.70}     % for background
\definecolor{nonallergenicbg}{rgb}{0.85,0.85,0.85} % for background

\definecolor{lightyellow}{rgb}{1.0,1.0,0.7}  % for highlighing

\hypersetup{linktocpage
,bookmarksnumbered=true
,colorlinks
,linkcolor=darkblue
,filecolor=darkgreen
,urlcolor=darkred
,citecolor=darkblue
,breaklinks=true
,pdfauthor={Janis Fehr (fehr@informatik.uni-freiburg.de)}
,pdftitle={Local Invariant Features for 3D Image Analysis}
,pdfsubject={Dissertation, 2009}
,raiselinks
}
\usepackage{longtable}
\usepackage{multirow}
\usepackage{booktabs}  % nicer tables (\toprule, \midrule,
\usepackage{xspace}   % to insert correct spaces after commands, like \Alnus

\renewcommand{\cite}[1]{\citep{#1}}

\renewcommand{\vec}[1]{\boldsymbol{\mathbf{#1}}}
\providecommand{\doi}[1]{\href{http://dx.doi.org/#1}{\nolinkurl{doi:#1}}}

\def\mycleardoublepage{\clearpage\ifodd\thepage\hbox{}\clearpage\fi}

\makeindex
\makeatletter

\newcommand*{\chapterthumbwidth}{2em}
\newcommand*{\chapterthumbheight}{1em}

\newcommand*{\chapterthumbboxcolor}{black}
\newcommand*{\chapterthumbtextcolor}{white}

\newcommand*{\putchapterthumb}{%
  \begingroup
    \Large
    \setlength{\@tempdima}{\@oddheadshift}% (internal from scrpage2)
    \setlength{\@tempdima}{-\@tempdima}%
    \addtolength{\@tempdima}{\paperwidth}%
    \addtolength{\@tempdima}{-\oddsidemargin}%
    \addtolength{\@tempdima}{-1in}%
    \rlap{%
      \hspace*{\@tempdima}%
      \vbox to 0pt{%
        \setlength{\@tempdima}{\chapterthumbwidth}%
        \multiply\@tempdima by\value{chapter}%
        \addtolength{\@tempdima}{-\chapterthumbwidth}%
        \addtolength{\@tempdima}{-\baselineskip}%
        \vspace*{\@tempdima}%
        \llap{%
          \rotatebox{90}{\colorbox{\chapterthumbboxcolor}{%
              \parbox[c][\chapterthumbheight][c]{\chapterthumbwidth}{%
                \centering
                \textcolor{\chapterthumbtextcolor}{%
                  \strut\thechapter}\\
              }%
            }%
          }%
        }%
        \vss
      }%
    }%
  \endgroup
}

\title{On Invariance, Equivariance, Correlation and Convolution of Spherical Harmonic Representations\\ for Scalar and Vectorial Data.\\[2cm]
\includegraphics[scale=0.2]{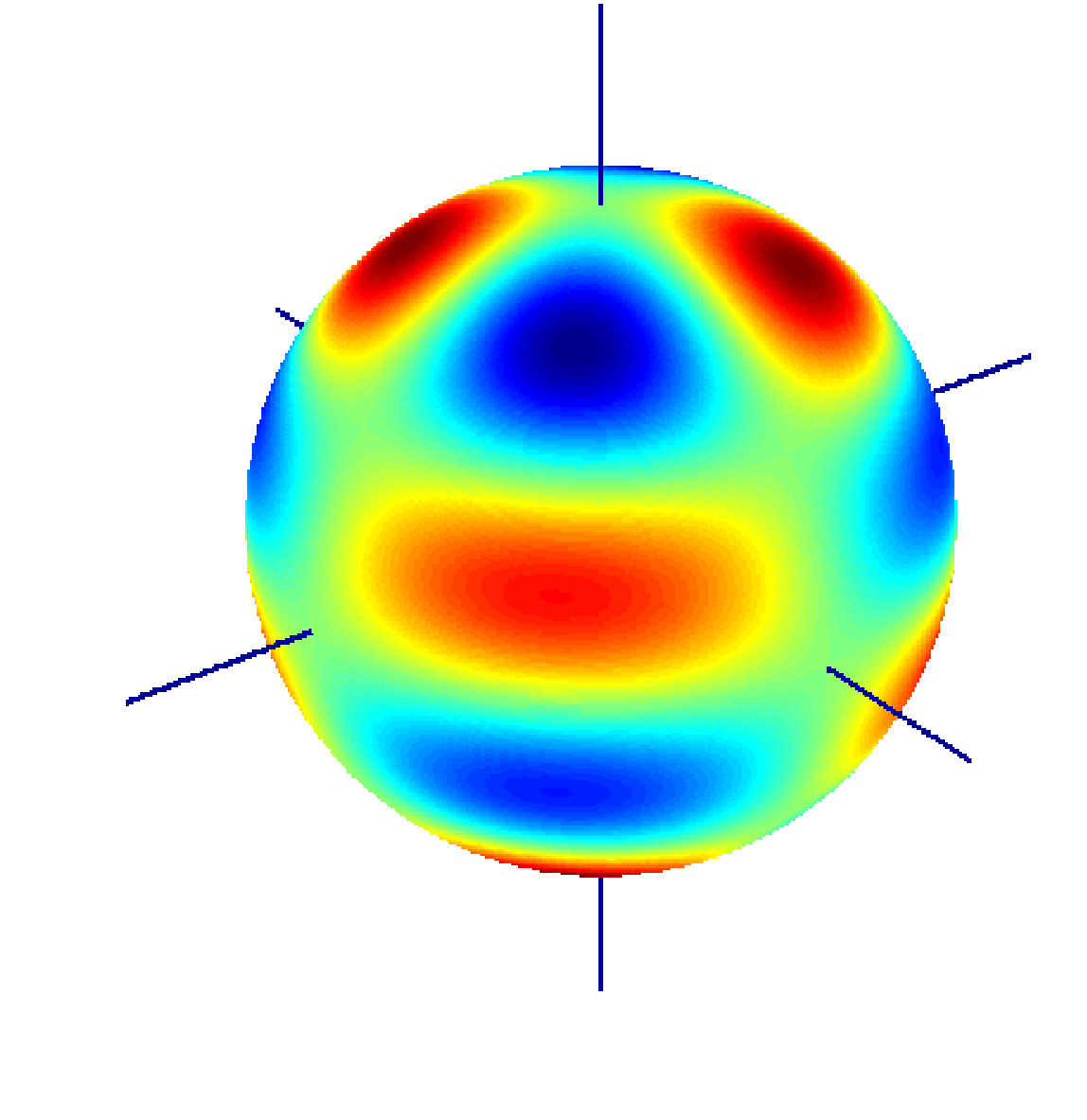}
\includegraphics[scale=0.2]{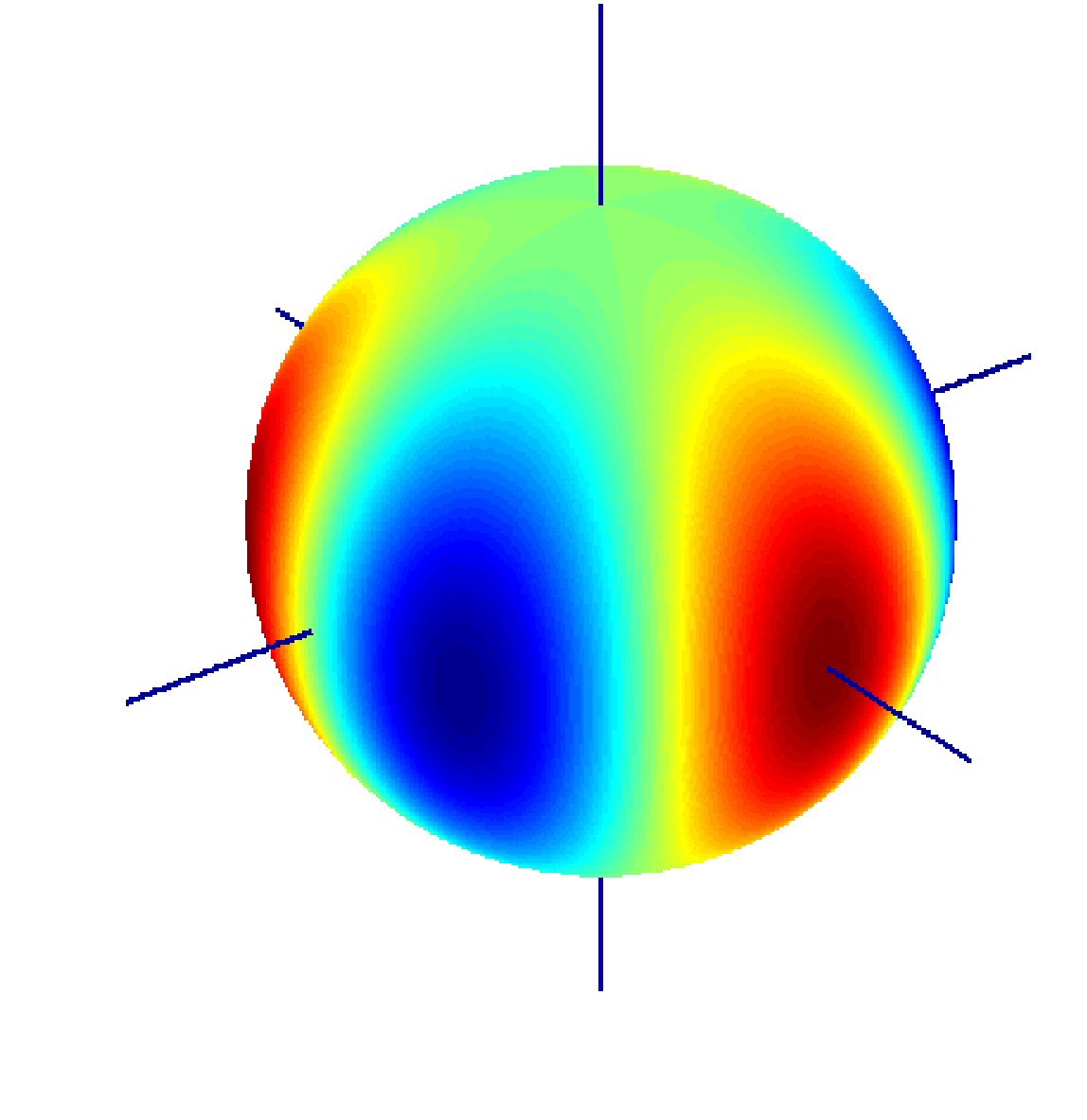}
\includegraphics[scale=0.2]{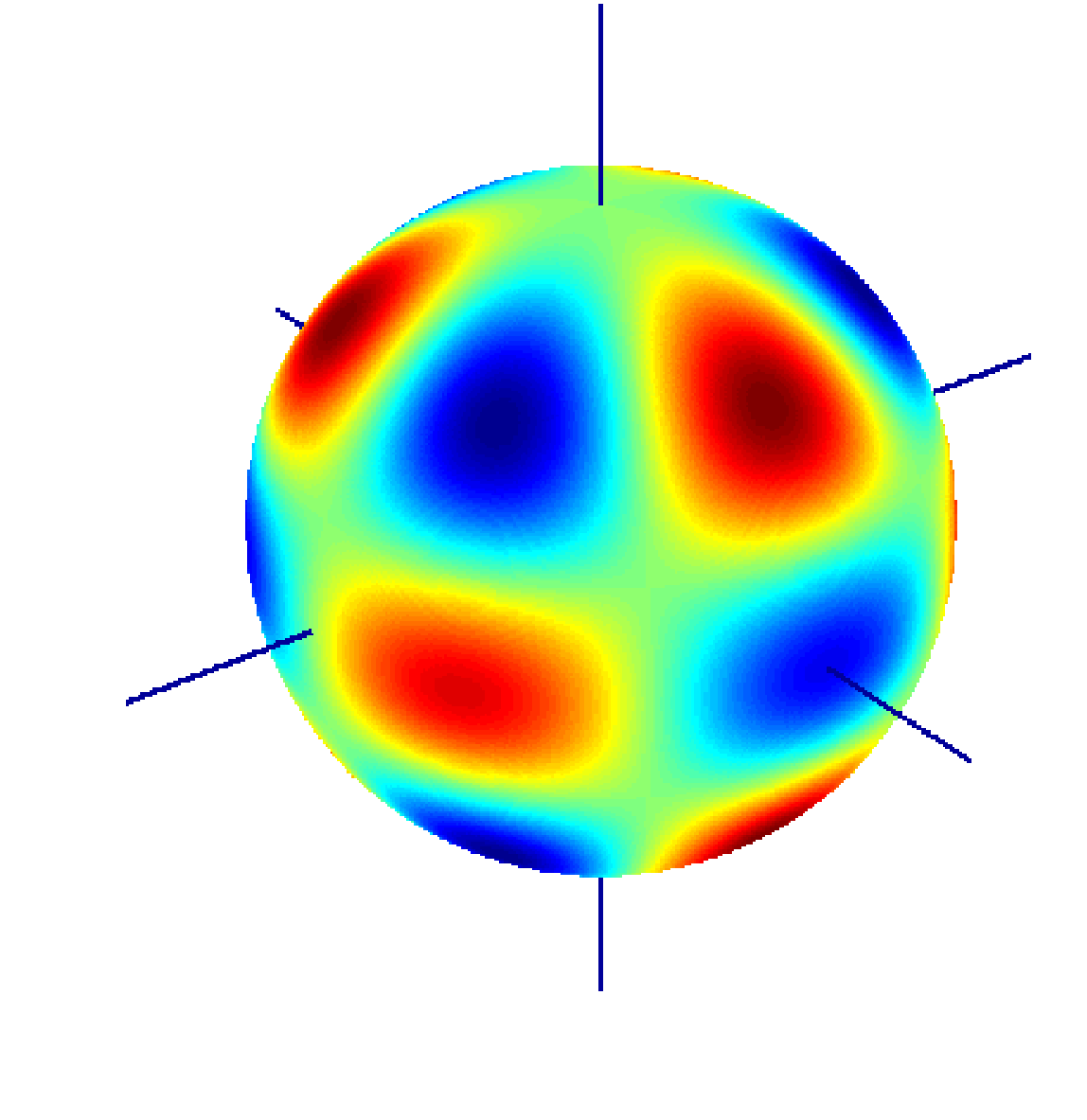}
\includegraphics[scale=0.25]{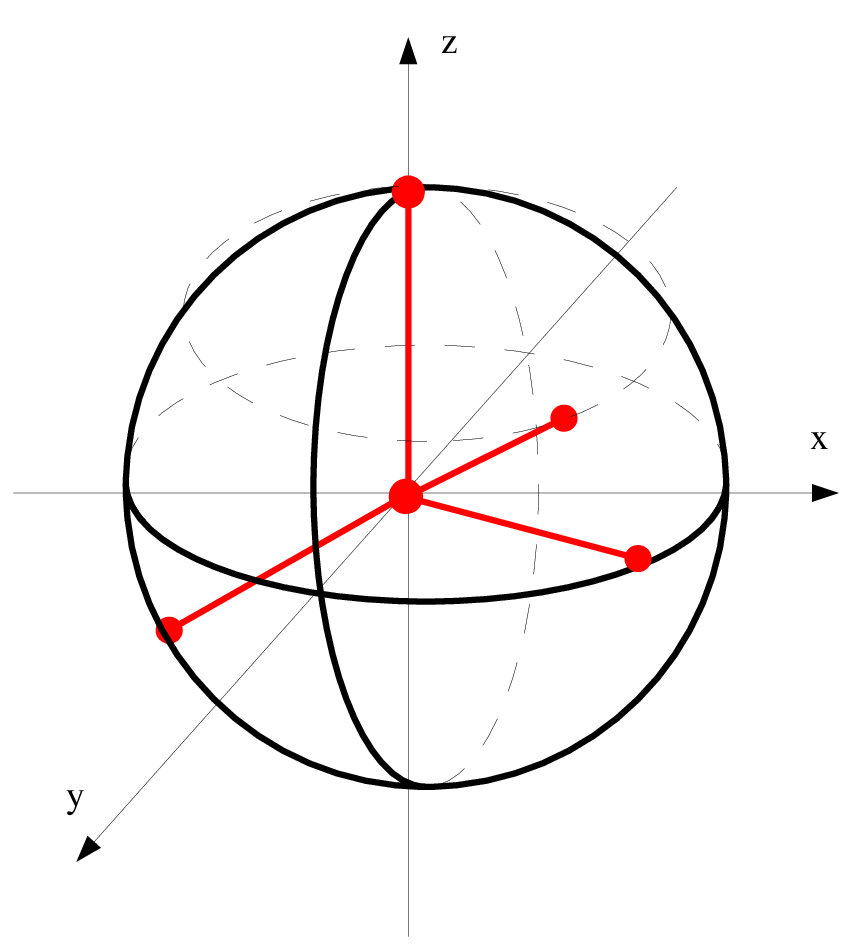}
\includegraphics[scale=0.25]{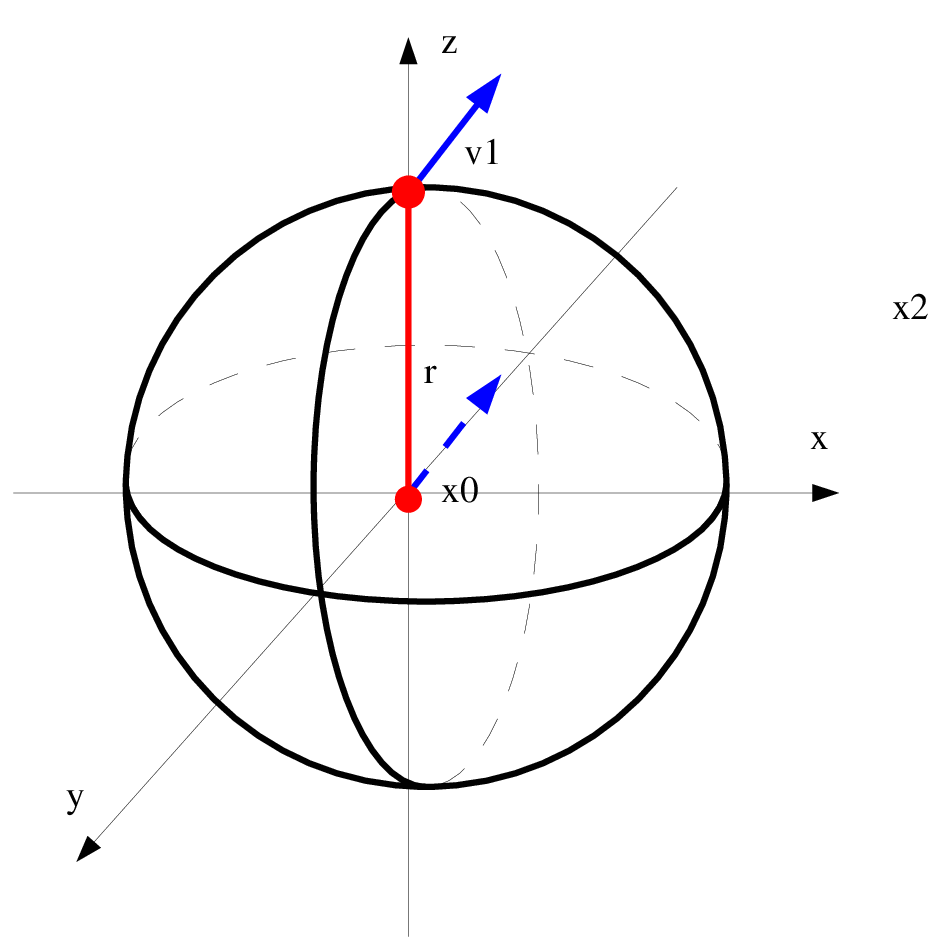}}
\author[1,2]{Janis Keuper \\ \href{mailto:keuper@imla.ai}{keuper@imla.ai}}
\affil[1]{Institute for Machine Learning and Analytics (IMLA), Offenburg University}
\affil[2]{CC-HPC, Fraunhofer ITWM, Kaiserslautern}
 
\begin{document}

\maketitle

\section*{Abstract}

The mathematical representations of data in the \textit{Spherical Harmonic} (${\cal SH}$) domain has recently regained increasing interest in the machine learning community. This technical report gives an in-depth introduction to the theoretical foundation and practical implementation of ${\cal SH}$ representations, summarizing works on rotation invariant and equivariant features, as well as convolutions and exact correlations of signals on spheres. In extension, these methods are then generalized from scalar ${\cal SH}$ representations to \textit{Vectorial Harmonics} (${\cal VH}$), providing the same capabilities for 3d vector fields on spheres.     

\paragraph{NOTE 1:} This document is a re-publication of a subset of works originally published in my PhD thesis (I changed my last name from \textit{Fehr} to \textit{Keuper}):
\begin{verbatim}
Fehr, Janis. Local invariant features for 3D image analysis. 
PhD Thesis. University of Freiburg, 2009.
\end{verbatim}

\noindent Hence, it does NOT provide any references or findings since 2009. The sole intention of this re-publication is to provide old (but still very useful) insights to an \textit{Arxiv} audience (which occasionally appears not to be aware of pre-\textit{Arxiv} works).\\ 

\noindent Please cite the thesis or the original publications:
\begin{verbatim}
Fehr, Janis. "Local rotation invariant patch descriptors for 3D vector
fields." 20th International Conference on Pattern Recognition. IEEE, 2010.

Fehr, Janis, Marco Reisert, and Hans Burkhardt. "Fast and accurate rotation 
estimation on the 2-sphere without correspondences." 10th European
Conference on Computer Vision, Marseille, France, October 12-18, 2008 

Fehr, Janis, and Hans Burkhardt. "Harmonic Shape Histograms for 3D Shape 
Classification and Retrieval." MVA. 2007.

Ronneberger, Olaf, Janis Fehr, and Hans Burkhardt. "Voxel-wise gray scale
invariants for simultaneous segmentation and classification." Joint Pattern
Recognition Symposium. Berlin, Heidelberg: Springer Berlin Heidelberg, 2005.
\end{verbatim}

\noindent when using this content for your work.

\paragraph{NOTE 2:} The original thesis and publications where all targeting multi-channel volumetric input data given by the target applications at that time. However, the actual methods in the harmonic domain directly extend to other data and applications in most cases.

\tableofcontents

\addcontentsline{toc}{chapter}{Introduction and Perquisites}
\chapter*{\label{sec:feature:intro}Introduction and Perquisites}
\index{Features}

\paragraph{Structure of of the report: }
The following report is structured as follows: in this introductory chapter we review the aspects of feature design 
in general (section \ref{sec:feature:featuredesign}), and take a closer look at local (section \ref{sec:feature:localfeatures}) and invariant
 features (section \ref{sec:feature:invariance}).\\
In chapter \ref{sec:feature:mathbg} we introduce the essential mathematical basics and derive further mathematical techniques needed for the 
formulation of our features, including correlation and convolution.\\
Chapter \ref{sec:feature:implement} discusses basic implementation issues like sampling problems or parallelization and fills the gap 
between the continuous mathematical theory and discrete implementation: for each of the following features, we first derive the theoretic
foundation in a continuous setting, and then give details on the actual discrete implementation based on these methods.\\
Then we introduce several different classes of features and their feature extraction algorithms:  
chapter \ref{sec:feature:SH} introduces the class of ${\cal SH}$-Features,
chapter \ref{sec:feature:SHHaar} derives new features based on Haar-Integration and finally the chapters \ref{sec:feature:VHfeature} and 
\ref{sec:feature:VHHaar} show how
we can compute different features on 3D vector fields. An overview of all features which are covered in this work can be found in table 
\ref{tab:feature:overview}.\\
%In chapter \ref{sec:feature:featureselect} we tackle the problem of automatic and data-driven feature selection from (weakly) labeled training
%data.\\
Finally, we evaluate and compare the introduced features on an artificial benchmark (chapter \ref{sec:feature:experiments}). 

\begin{table}[ht]
\begin{tabular}{||lr||c|c|c||}
\hline
\hline
{\bf Feature} & &{\bf Invariance} & {\bf Input domain} & {\bf Output domain}\\
\hline

${\cal SH}_{abs}$ & \ref{sec:feature:SHabs} & {\bf r} invariance \& {\bf g} robustness & scalar & band-wise scalar\\
\hline
${\cal SH}_{phase}$ & \ref{sec:feature:SHphase} & {\bf r} invariance \& {\bf g} invariance & scalar & band-wise scalar\\
\hline
${\cal SH}_{corr}$ & \ref{sec:feature:SHautocorr} & {\bf r} invariance \& {\bf g} invariance & scalar & scalar\\
\hline
${\cal SH}_{bispectrum}$ & \ref{sec:feature:SHbispectrum} & {\bf r} invariance \& {\bf g} robustness & scalar & sub-band-wise scalar\\
\hline
%\hline
%fuLBP & \ref{sec:feature:uLBP} & {\bf r} invariance \& {\bf g} invariance & scalar & scalar\\
%\hline
%fLBP& \ref{sec:feature:LBPfast3D} & {\bf r} invariance \& {\bf g} invariance & scalar & scalar\\
%\hline
\hline
2p-Haar & \ref{sec:feature:2p} & {\bf r} invariance \& {\bf g} robustness & scalar & scalar\\
\hline
3p-Haar & \ref{sec:feature:3p} & {\bf r} invariance \& {\bf g} robustness & scalar & scalar\\
\hline
np-Haar & \ref{sec:feature:np} & {\bf r} invariance \& {\bf g} invariance & scalar & scalar\\
\hline
\hline
${\cal VH}_{abs}$ & \ref{sec:feature:VHabs} & {\bf r} invariance \& {\bf g} invariance & vectorial & band-wise scalar\\
\hline
${\cal VH}_{autocorr}$ & \ref{sec:feature:VH_autocorr} & {\bf r} invariance \& {\bf g} invariance & vectorial & scalar\\
\hline
\hline
1v-Haar & \ref{sec:feature:1v} & {\bf r} invariance \& {\bf g} invariance & vectorial & scalar\\
\hline
2v-Haar & \ref{sec:feature:2v} & {\bf r} invariance \& {\bf g} invariance & vectorial & scalar\\
\hline
nv-Haar & \ref{sec:feature:nv} & {\bf r} invariance \& {\bf g} invariance & vectorial & scalar\\
\hline
\hline
\end{tabular}
\caption[Schematic Overview Of All Features]{\label{tab:feature:overview} Schematic overview of all features with their invariance properties and input/output domains ({\bf r} = rotation, {\bf g} = gray-scale).}
\end{table}

\section{\label{sec:intro:notation}Mathematical Notation}
\begin{table}[ht]
\begin{tabular}{|l|l|}
\hline
$x \in \mathbb{R}, x \in \mathbb{C}$ & real or complex scalar value\\
$\Re(x)$& real part of a complex value\\
$\Im(x)$& imaginary part of a complex value\\
$\bar{x}$ & complex conjugate\\
${\bf x} \in \mathbb{R}^n, {\bf x} \in \mathbb{C}^n$ & n-dimensional position or vector\\
$X: \mathbb{R}^n \rightarrow \mathbb{R}, X: \mathbb{Z}^n \rightarrow \mathbb{R}$ & image function representing a $n$D scalar image\\
$X[c_i]: \mathbb{R}^n \rightarrow \mathbb{R}$ & $i$-th channel of a $m$-channel $n$D 
continuous scalar image\\
${\bf X}: \mathbb{R}^n \rightarrow \mathbb{R}^m$ & image function representing a $n$D field of $m$D vectors\\
$X({\bf x}) \in \mathbb{R}$ & scalar value at position ${\bf x}$\\
${\bf X}({\bf x}) \in \mathbb{R}^n$ & vectorial value at position $\bf{x}$\\
$\kappa(x)$ & non-linear kernel function $\kappa:\mathbb{R} \rightarrow \mathbb{R}$\\
$T[X]$ and $T[\bf{X}]$ & voxel-wise feature extraction \\
${\cal S}[r]({\bf x}) := \{{\bf x_i} \in \mathbb{R}^n | \|{\bf x}-{\bf x_i}\|=r\}$ & spherical neighborhood around ${\bf x}$\\
${\cal F}(\cdot)$ & Fourier transform\\
${\cal SH}\big(X|_{{\cal S}[r]({\bf x})}\big)$ & Spherical Harmonic transform of a local neighborhood\\
${\cal SH}[r](X)$ & element-wise Spherical Harmonic transform with radius $r$\\
${\cal VH}\big({\bf X}|_{{\cal S}[r]({\bf x})}\big)$ & Vectorial Harmonic transform of a local neighborhood\\
${\cal VH}[r]({\bf X})$ & element-wise Vectorial Harmonic with radius $r$\\
$\widehat{X}:= {\cal F}(X)$ or $\widehat{X}:= {\cal SH}[r](X)$ & $X$ transformed into frequency domain\\
$A\cdot$ B& scalar or voxel-wise multiplication\\
$\cal{R}_{(\phi,\theta,\psi)}$ or just $\cal{R}$ & rotation matrix\\
$A * B$ & convolution in $\mathbb{R}^n$ or $S^2$\\
$A \# B$ & correlation in $\mathbb{R}^n$ or $S^2$\\
$\cal{C}^*$ & convolution matrix\\
$\cal{C}^\#$ & correlation matrix\\
$\cal{G}$ & mathematical group\\
${g}\in \cal{G}$ & group element\\
${\cal SO}(3)$ & rotation group in  $\mathbb{R}^3$\\
$\phi,\theta,\psi$ & parameterization angles of ${\cal SO}(3)$\\
$S^2$ & sphere\\
$\Phi, \Theta$ & parameterization angles of $S^2$\\
%$\bf{\nabla} = \left(\frac{\partial}{\partial x_1}, \frac{\partial}{\partial x_2}, \frac{\partial}{\partial x_3}\right)$& del operator, represented by the nabla symbol\\
%$\bf{\nabla}_{\!\bf{x}'} = \left(\frac{\partial}{\partial x_1'}, \frac{\partial}{\partial x_2'}, \frac{\partial}{\partial x_3'}\right)$ &del operator with respect to $\bf{x}'$\\
$\bf{\nabla}\bf{X}$& vector field containing gradients of scalar field $\bf{X}$\\
$\bf{\nabla}X(\bf{x})$& gradient at position $\bf{x}$\\
\hline
\end{tabular}
\caption[Overview of the Mathematical Notation]{\label{notation} Overview of the mathematical notation used throughout this work.}
\end{table}

\section{\label{sec:feature:featuredesign} General Feature Design}
\begin{figure}[ht]
\centering
\psfrag{R1}{$\mathbb{R}^\infty$}
\psfrag{R2}{$\mathbb{R}^n$}
\psfrag{R3}{$\mathbb{R}^p$}
\psfrag{R4}{$\mathbb{Z}$}
\psfrag{G}{$\Gamma'$}
\psfrag{smily}{``smiley''}
\includegraphics[width=1\textwidth]{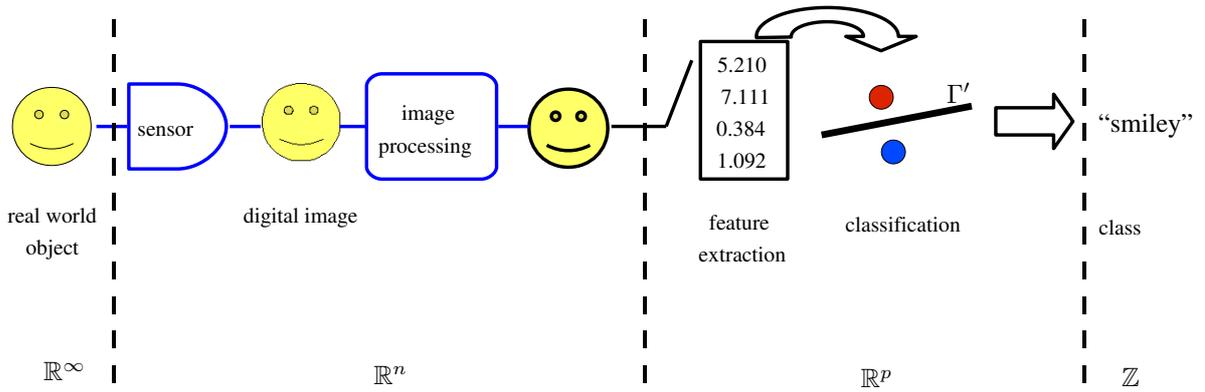}
\caption[Pattern recognition pipeline.]{\label{fig:feature:patternrecog} Idealized pattern recognition pipeline on images: in general, we try
to reduce the problem complexity from continuous real world data to an 1D categorization space. The semantic match of real world objects to
such categories is given by definitions based the human perception and thinking. }
\end{figure}

Most pattern recognition tasks can be derived from a very general and basic problem setting: given an arbitrary set of patterns 
$\{X_i| X_i\in {\cal X}\}$, we are looking for some function $\Gamma: X_i \rightarrow y_i$ which denotes each pattern with a semantic label 
$y_i \in Y$ from the category space $Y\subset \mathbb{Z}$.\\
 In general, ${\cal X}$ contains all possible patterns, which are usually defined as the digitalized signals obtained from a sensor capturing
the ``real world'' (see figure \ref{fig:feature:patternrecog}). $Y$ holds the semantic meaning (categorization) of the real world, where
each category $y_i$ defines an equivalence class.\\
The actual task of assigning the label $y_i$ is called classification and $\Gamma$ is often referred as decision function or classifier which
should hold:
\begin{equation}
X_1 \sim_y X_2 \Leftrightarrow \Gamma(X_1) = \Gamma(X_2).
\label{eq:feature:eqrel}
\end{equation}

The most crucial step towards a suitable $\Gamma$ is to find an adequate equality measure on $X$. 
Since the notion of equivalence of real world objects is given by the human perception and is often highly semantic, it is usually very  
hard to construct a measure which fulfills (\ref{eq:feature:eqrel}).\\

In practice, there are two strategies to tackle this problem: learning and feature extraction - which are usually combined.\\ 
The first approach
tries to learn $\Gamma$ from a set of training examples - we discuss this method in depth in part II of this work. 
However, most practical problems
are too complex to construct or learn $\Gamma$ directly by raw ``pattern matching''. Such a ``pattern matching'' is usually too expensive in 
terms of computational complexity, or even completely intractable in cases with a large intra class variance, e.g. if
patterns of the same equivalence class are allowed to have strong variations in their appearance.\\
The second approach tries to solve the problem by simplifying the original problem: the goal is to find a reduced representation $\widetilde{X}$
of the original pattern $X$ which still preserves the distinctive properties of $X$. A commonly used analogy for the feature concept is the
notion of ``fingerprints'' which are extracted from patterns to help to find a simpler classifier $\Gamma'$ which holds:
\begin{equation}
X_1 \sim_y X_2 \Leftrightarrow \Gamma(X_1) = \Gamma(X_2) \Leftrightarrow \Gamma'(\widetilde{X_1}) = \Gamma'(\widetilde{X_2}).
\label{eq:feature:eqre2}
\end{equation}            

\begin{figure}[ht]
\centering
\psfrag{X1}{$\widetilde{X_1}=\widetilde{X_2}$}
\psfrag{X2}{$\widetilde{X_3}$}
\psfrag{Rp}{$\mathbb{R}^n$}
\psfrag{Rn}{$\mathbb{R}^p$}
\psfrag{G}{$\Gamma'$}
\includegraphics[width=0.75\textwidth]{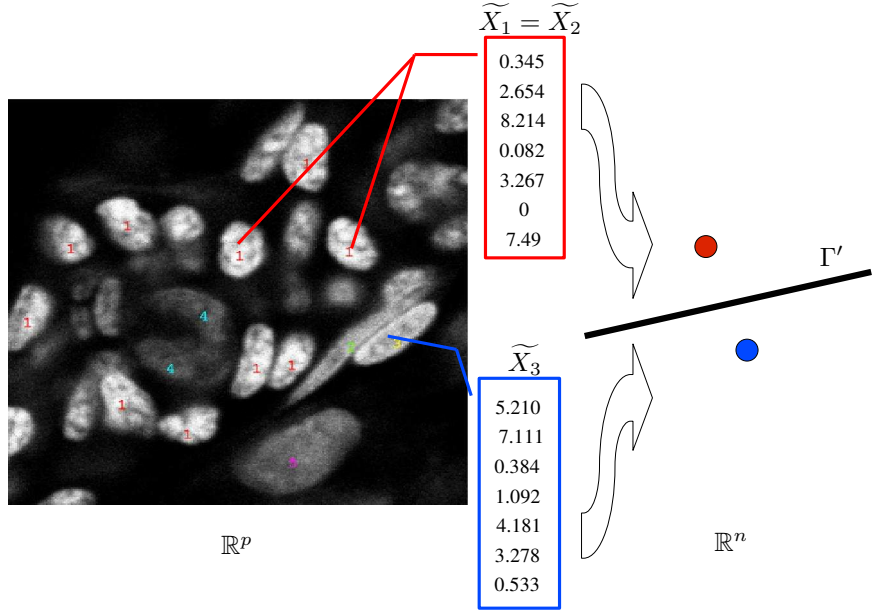}
\caption[Feature extraction duality.]{\label{fig:feature:featureduality} Feature extraction duality: instead of trying to solve the difficult 
problem in the original $\mathbb{R}^p$ space, we extract ``mathematical fingerprints'' (features) and try to find a dual classifier $\Gamma'$
in the simpler $\mathbb{R}^n$ space, where $n \ll p$ } 
\end{figure}

Either a perfect feature extraction or a perfect classifier would solve the problem completely, but in practice we have to combine both 
methods to obtain reasonable results: We use features to reduce the problem and then learn $\Gamma'$ (see figure \ref{fig:feature:featureduality}). 

\subsection{\label{sec:feature:extraction}Feature Extraction} 
\index{Features}
We formalize the feature extraction in form of some function $T(X_i)$ which maps all input signals ${X_i}$ into the so-called feature space 
$\widetilde{\cal X}$:
\begin{equation}
\widetilde{X_i} =: T(X_i).
\label{eq:feature:featuremapping}
\end{equation}
For the theoretical case of a ``perfect'' feature, $T(X_i)$ maps all input signals ${X_i}$ belonging to the same semantic class with label 
$y_i$ onto one point $\widetilde{X_i}$ in this features space:  
\begin{equation}
X_1 \sim_y X_2 \Leftrightarrow T(X_1) = T(X_2). 
\label{eq:feature:perfect_feature}
\end{equation}
As mentioned before, the nature of practical problems includes that there are intra class variations which make things more complicated.   
We model these intra class variations by transformations $h_i\in H_y$, where $H_y$ is the set of all possible transformations, which
do not change the label $y$ of the ideal class template $X_y$: 
\begin{equation}
X_i := h_iX_y.
\label{eq:feature:feature_variance}
\end{equation}
If it is impossible to construct the ``perfect'' feature for a practical application,  the goal is to find feature mappings $T(h{X_y})$ 
which at least fulfill the following properties:
\index{Feature properties}
\begin{itemize}
\item \textbf{(I) size:} the feature space should be much smaller than the pattern space: $n <<< p$ with $\widetilde{\cal X} \subset 
\mathbb{R}^n,  {\cal X}\subset \mathbb{R}^p$. 
\item \textbf{(II) continuity:} small changes in the input pattern $X_i$ should have only small effects in feature space $\widetilde{\cal X}$
\item \textbf{(III) cluster preservation:} local neighborhoods should be transfered from input to feature space
\end{itemize}
If the extracted feature $\widetilde{X_i}$ adheres to these properties, $\widetilde{\cal X}$ provides several advantages for the further construction or 
learning of $\Gamma$:
first, (I) drastically reduces the computational complexity and second, (II) and (III) make it possible to introduce a meaningful similarity
measure on $\widetilde{\cal X}$ (like a simple Euclidean-Norm), which is an essential precondition to the application of learning algorithms (see
part II).\\

Still, the question remains how to construct features which hold the properties I-III. While size property (I) is rather easy to meet, 
continuity (II) and cluster preservation (III) are more difficult to obtain. This leads us to the notions of invariance and 
robustness of features, which are central to the methods presented in this work.

\subsection{\label{sec:feature:invariance}Invariance}
\index{Invariance}\index{Invariant features}
Feature extraction methods are strongly interlaced with the concept of invariance. The basic idea of invariant features is to
construct $T(X)$ in such a way that the effect of those transformations $h_i \in H_y$ (\ref{eq:feature:feature_variance}) which are 
not affecting the semantic class label $y$ of $X$, e.g. $X \sim_y h_iX$, is canceled out by $T$: 
\begin{equation}
T(h_i{X_y}) = \widetilde{X_y}, \forall h_i\in H_y.
\end{equation}
For two signals ${X_1}$ and ${X_2}$ which are considered to be equivalent under a certain transformation $h_i\in H_y$,
${X_1}\overset{h_i}{\sim} {X_2}$,
the {\bf necessary condition} \cite{habilBu} for invariance against $h_i$ is:
\index{Invariance: necessary condition}
\begin{equation}
\label{eq:feature:Necessary}
{X_1}\overset{h_i}{\sim} {X_2}\Rightarrow T({X_1})=\widetilde{X_1}=\widetilde{X_2}= T({X_2}).
\end{equation}
In order to achieve {\bf completeness} \cite{habilBu}, $T$ has to hold:
\index{Invariance: completeness}
\begin{equation}
\label{eq:feature:sufficient}
T({X_1})=T({X_2})\Rightarrow {X_1}\overset{h_i}{\sim} {X_2}.
\end{equation}
In most cases the mathematical completeness condition is too strict, since it is not practicable to have a distinct mapping for every
theoretically possible pattern ${X_i}$. However, with only little a priori knowledge, one can
determine a sufficient subset of likely patterns ${\cal X}'$. If (\ref{eq:feature:sufficient}) holds for all ${X_i},{X_j} \in {\cal X}'$,
\textbf {separability} \cite{habilBu} can be guaranteed for the likely patterns.\\
\index{Invariance: separability}
It is straightforward to see that a feature which holds the necessary condition (\ref{eq:feature:Necessary}) and achieves at least
separability meets the properties II and III. 

\subsubsection{\label{sec:feature:grouptrans}Group Transformations}
\index{Transformation groups}
The construction of an invariant feature requires that we are able to model the allowed transformations $h_i \in H_y$ of the equivalence class
with label $y$. In general this is a hard and sometimes infeasible task, e.g. just think of arbitrary deformations. However, for the 
subset of transformations $G_y\subset H_y$, where $G_y$ forms a compact mathematical group, we have sophisticated mathematical tools to model the 
individual transformations $g_i\in G_y$.\\
Luckily, many practically relevant transformations like rotations are groups or can easily be transformed to groups, e.g. translations if we
consider cyclic translations. Overall, we can formulate translations, rotations, shrinking, shearing and even affine mappings as group
operations \cite{Burkhardt2001}.

\subsubsection{\label{sec:feature:constinvariance}General Techniques For The Construction Of Invariant Features}
In general, there are three generic ways of constructing invariant features: by normalization, derivation and integration
\cite{Burkhardt2001}. For allowed transformations 
$H_y$, the individual transformations $h \in H_y$ differ only by their associated set of parameters $\vec{\lambda}$, which cover the
degrees of freedom under $H_y$. The most popular method for invariant feature construction is to eliminate the influence of
$\vec{\lambda}$ via normalization of the class members $X_i := h_{\lambda} X_y$ with a class template $X_y$.\\
We apply normalization techniques in the following features: 
${\cal SH}_{abs}$ (chapter \ref{sec:feature:SHabs}), ${\cal SH}_{phase}$ (chapter \ref{sec:feature:SHphase}), ${\cal SH}_{bispectrum}$, 
 and 
${\cal VH}_{abs}$ (chapter \ref{sec:feature:VHabs})\\
However, it should be noted that normalization techniques in general tend to suffer in
cases of noisy or partially corrupted data and are often totally infeasible for complex data where no normalized template can be found.\\

A second possibility is the elimination of  $\vec{\lambda}$ via derivation:
\begin{equation}
{\partial T(g_{\vec{\lambda}}{X_i})\over \partial \lambda}\equiv 0.
\end{equation}
The resulting differential equations can be solved using Lie-Theory \cite{lenz} approaches, but in practice it is often very 
difficult to obtain solutions to the differential equations. \\

Finally, the approach which has been proposed by \cite{Schulz:Mirbach95} can be applied on the subset of group transformations: 
It generates invariant features via Haar-Integration
over all degrees of freedom of the transformation group $G$. We take an in-depth look at the Haar-Integration approach in chapter
\ref{sec:feature:SHHaar} and apply it in several of our features: 
2p-Haar (chapter \ref{sec:feature:2p}), 3p-Haar (chapter \ref{sec:feature:3p}), np-Haar (chapter \ref{sec:feature:np}), 1v-Haar 
(chapter \ref{sec:feature:1v}), 2v-Haar (chapter \ref{sec:feature:2v}) and nv-Haar (chapter \ref{sec:feature:nv}).\\

For many practical applications invariance can be achieved by the combination of several different approaches: we can split transformations $h$
into a combination of several independent transformations $h := h_1 \circ h_2 \circ \dots$, where $h_1$ might be a group transformation
like i.e. rotation and $h_2$ a non-group transformation like gray-scale changes.\\

The concept of invariance provides us with a powerful tool for the construction of features which is suitable for a wide range of problems.
However, there are still many practically relevant cases where some of the underlying transformations $h_i$ cannot be
sufficiently modelled, or are even partially unknown. Then it becomes very hard or impossible to construct invariant features. In these
cases we have to fall back to the sub-optimal strategy to construct robust instead of invariant features. \\

\subsection{\label{sec:feature:robustness}Robustness}
\index{Robustness}
Robustness is a weaker version of invariance: if we are not able to cancel out the effect of the transformations $h_i$ like in 
(\ref{eq:feature:Necessary}), we can at least try to minimize the impact of these intra class variations.\\ 
Given ${X_1}\overset{h}{\sim} {X_2}, {X_1}, {X_2} \in {\cal X}$, 
we are looking for a feature $T$ which maps ${X_1},{X_2}$ in such a way that the intra class variance
in $\widetilde{\cal X}$ is smaller than the extra class distances given some distance measure $d$ in $\widetilde{\cal X}$:
\begin{equation}
{X_1}\overset{h_i}{\sim} {X_2}\Rightarrow  d\big(T({X_1}), T({X_2})\big) < d\big(T({X_{1,2}}),T({X'})\big), \text{\quad}\forall X'\in 
{\cal X}: X'\neg\overset{h_i}{\sim} X_{1,2}.
\label{eq:feature:robustness}
\end{equation}
It is obvious that the robustness property (\ref{eq:feature:robustness}) directly realizes the feature properties II and III. In practice,
robustness is often achieved by simplified approximations of complex intraclass variations, e.g. linear approximations of actually 
non-linear transformations $h_i$. In theses cases, we often use an even weaker definition of robustness and demand that 
(\ref{eq:feature:robustness}) has only to hold for most but not all  $X' \in {\cal X}$.

\subsection{\label{sec:feature:equivariance}Equivariance}
\index{Equivariance}
For some applications it is desirable to explicitly transfer the variations to the feature space:  
\begin{equation}
\label{eq:feature:equivariance}
{X_1}\overset{h_i}{\sim} {X_2}\Rightarrow T({X_1})= h_iT({X_2}).
\end{equation}
These features are called equivariant, and are often used to compute the parameters of known transformations $h_i$.

\section{\label{sec:feature:localfeatures}Local Features}
\index{Features}\index{Local features}
The feature definition in the last section (\ref{sec:feature:extraction}) considered only the extraction of so-called ``global'' features, i.e.
features are extracted as descriptors $\widetilde{X_i} = T(X_i)$ (or ``Fingerprints'') of the entire pattern $X_i$. This global approach
is suitable for many pattern recognition problems, especially when the patterns are taken from prior segmented objects (see part III).
In other cases, it can be favorable to describe a global pattern as an ensemble of locally constrained sub-patterns. Such a local
approach is suitable for object retrieval, object detection in unsegmented data, or data segmentation itself (see part III).

\subsection{\label{sec:feature:localfeaturesvolume}Local Features on 3D Volume Data}
Throughout the rest of this work we deal with 3D volume data or 3D vector fields. In general we derive the theoretical background of the
local features 
in settings of continuous 3D volumes, which we define as functions $X: \mathbb{R}^3 \rightarrow \mathbb{R}^m$ with 
values $X({\bf x}) \in \mathbb{R}^m$ at evaluation coordinates  ${\bf x} \in \mathbb{R}^3$.  
We then transfer the feature algorithms to operate on the practical relevant discrete 3D volume grids: $X:\mathbb{Z}^3 
\rightarrow \mathbb{R}^m $, where we often refer to the position ${\bf x}$ as a ``{\bf voxel}''.

Given 3D volume data, we capture the locality of the features extracted from $X$ in 
terms of a spatial constraining of the underlying sub-pattern. More precisely, we define a sub-pattern as ``local neighborhood'' around a 
data point at ${\bf x}$ with the associated local feature $\widetilde{X({\bf x})}$.\\
Further, we parameterize the local ``neighborhood'' in concentric spheres with radii $r$ around ${\bf x}$. 
This has several advantages over a rectangular 
definition of the ``local neighborhood'':\\
First, we can easily define the elements of the sub-pattern by a single parameter $r$ using the following notation for the sub-pattern
around ${\bf x}$: 
\begin{equation}
\label{eq:feature:subpatterndef}
{\cal S}[r]\left({\bf x}\right) := \{{\bf x}_i \in \mathbb{R}^3 | \|{\bf x}-{\bf x}_i\|_2 = r\}. 
\end{equation}
Second, we can address all points in ${\cal S}[r]\left({\bf x}\right)$ via the parameterization in radius $r$ and the spherical 
coordinates $(\Phi, \Theta$) - see section \ref{sec:feature:shrot} for more details on the parameterization. And finally, we can rely on
a well known and sound mathematical theory to handle signals (patterns) in spherical coordinates which provides us with very useful
tools to handle common transformations such as rotations.\\
We give an in-depth introduction and further extensions to this mathematical basis for our local features in chapter \ref{sec:feature:mathbg}.

\subsubsection{Gray-Scale Data}
\index{Gray-scale data}\index{Voxel}
In cases where the 3D volume data is scalar $X: \mathbb{R}^3 \rightarrow \mathbb{R}$, 
we can directly apply the locality definition (\ref{eq:feature:subpatterndef}). Note,
that we usually refer to scalar data as ``gray-scale'' data, this term is derived from the usual data visualization as gray-scale images - 
even though the scalar values might encode arbitrary information. 
Analogous to this, we denote intensity changes as gray-scale changes.\\
For many pattern recognition tasks on scalar 3D volume data we like to obtain gray-scale and rotation invariant local features in order to
cancel out the dominant transformations which act locally. Other transformations of the data do not act locally, like translations, or
are very hard to model like arbitrary deformations. In these cases we try to obtain local robustness, which is usually easier to obtain than
global robustness since the local affect of complex global transformations is limited in most cases.

\subsubsection{Multi-Channel Data}
\begin{figure}[ht]
\centering
\includegraphics[width=0.3\textwidth]{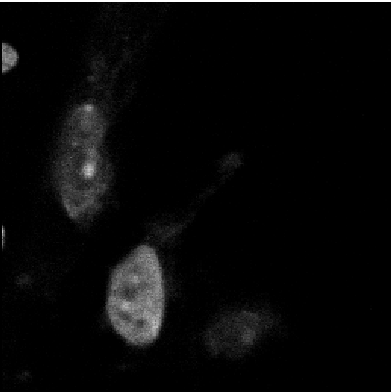}
\includegraphics[width=0.3\textwidth]{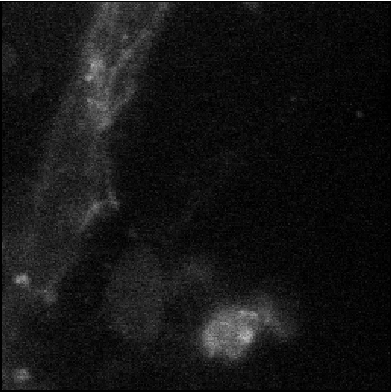}
\includegraphics[width=0.3\textwidth]{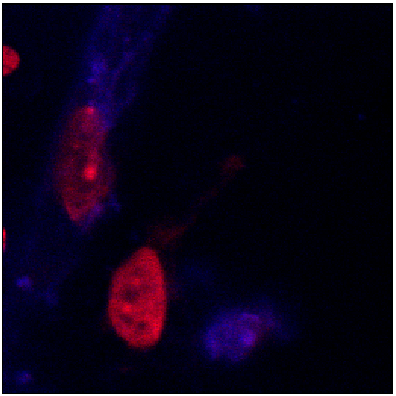}
\caption[Local Features on 3D Multi-Channel Volume Data.]{\label{fig:feature:localmultichannelfeatures} Example of multi-channel volume data:
xy-slice of volume data recorded by a Laser Scanning Microscope (LSM): {\bf Left:} channel with {\it YoPro} staining. {\bf Center:} channel
with {\it SMA} staining. {\bf Right:} pseudo coloration of the combined channels.}
\end{figure}

\index{Multi-channel data}
In many cases we face volumes with data which holds more than a single scalar value at each position ${\bf x}$.
Then we define $X: \mathbb{R}^3 \rightarrow \mathbb{R}^m$ for data with $m$ scalar values per position.
The classic example
could be a RGB color coding at each voxel, but we might also have other multi-modal data with an arbitrary number of scalar values.\\
We refer to these volumes as multi-channel data, where we address the individual channels $c_i$ by $X[c_i]({\bf x}) \in \mathbb{R}$. 
Figure \ref{fig:feature:localmultichannelfeatures} shows an example of such multi-channel data.

It is obvious that we also need features which operate on multiple channels - this is an important aspect we have to take into account 
for the feature design. 

\subsection{Local Features on 3D Vector Fields}
\begin{figure}[ht]
\centering
\includegraphics[width=0.08\textwidth]{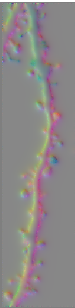}
\includegraphics[width=0.5\textwidth]{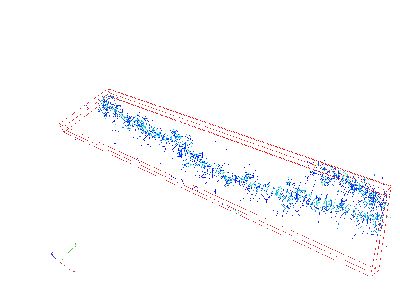}
\caption[Local Features on 3D Vector Fields.]{\label{fig:feature:localvectorfeatures} Example of vectorial data: xy-slice of the thresholded 
gradient vector field of volume data recorded by a Laser Scanning Microscope (LSM): {\bf Left:} pseudo coloring of the gradient directions.
{\bf Right:} 3D vector field reconstruction of the same data.}
\end{figure}

\index{Vector fields}
Besides local features for scalar gray-scale and multi-channel scalar volumes, we further investigate and derive features which
operate on 3D vector fields $\bf X:{\mathbb R}^3 \rightarrow {\mathbb R}^3$. 
Usually these vector fields are directly obtained by the extraction of gradient information from 
scalar volumes (see figure \ref{fig:feature:localvectorfeatures}).\\
In contrast to multi-channel data, the elements of the vectors in the field are not independent and change according to transformations,
e.g. under rotation. This makes the feature design a lot more complicated.

\section{\label{sec:feature:featurerelatedwork}Related Work}
The number of publications on feature extraction methods and their applications is countless. Hence, we restrict our review of related work
to methods which provide local rotation invariant features for 3D volume data or 3D vector fields. This restriction reduces the number of
methods we have to consider to a manageable size. 
Since we provide an in-depth discussion of most of the suitable methods in the next chapters (see table \ref{tab:feature:overview}), we 
are left with those few methods we are aware of, but which are not further considered throughout the rest of this work:\\

\begin{description}
\item The first class of rotational invariant features which operate on spherical signals are based on the so-called ``Spherical Wavelets''
\cite{Swavelet1} which form the analog to standard wavelets on the 2-sphere. These methods have mostly been used for 3D shape analysis,
but also for the characterization of 3D textures \cite{Swavelet2}. 

\item Second, we have to mention methods based on 3D Zernike moments. For shape retrieval (also see Part III), 3D Zernike moments have been
successfully applied as 3D shape descriptors, i.e. by \cite{saupemoments} and \cite{Novotni033dzernike}. In both cases, only the absolute
value of the Zernike coefficients were used to obtain rotation invariance which leads to rather weakly discriminative features just as in the 
case of the ${\cal SH}_{abs}$ features \ref{sec:feature:SHabs}.\\
\cite{Canterakis} introduced a  set of complete affine invariant 3D Zernike moments which overcome these problems. However, just as for the 
${\cal SH}_{bispectrum}$ features \ref{sec:feature:SHbispectrum}, the completeness comes at the price of very high complexity.

\item Finally, we were not able to find much significant prior work on rotation invariant features operating on 3D vector fields.
Mentionable is the work in \cite{schulz}, which uses a generalized Hough approach \cite{hough} to detect spherical structures in a 3D gradient
vector field. This method is closely related to our 1v-Haar feature \ref{sec:feature:1v}.

\end{description}

%\section{Contributions}
%Our original contributions in this part are manifold: we have introduced and derived %all of the features (see table \ref{tab:feature:overview})
%with the exception of ${\cal SH}_{abs}$ (which is our reference feature) and ${\cal SH}_{bispectrum}$, which has been used before for 2D 
%feature extraction - but not for 3D volume data.\\
%We have also introduced the $np$- and $vp$-Feature selection algorithm (section \ref{sec:feature:featureselect})  and several novel 
%mathematical operations in the
%Spherical Harmonic domain (e.g. the fast correlation (section \ref{sec:feature:shcorr}) and convolution (section \ref{sec:feature:shconvolve})).\\
%Finally, we have provided the benchmark with a 3D texture generation method for the evaluation of the local 3D feature methods (section 
%\ref{sec:feature:experiments}). 

\cleardoublepage
\chapter{\label{sec:feature:mathbg}Mathematical Background}
In this chapter we introduce and review the mathematical background of important methods we use later on. First we exploit
and formulate the basics of mathematical operations on the 2-sphere, which are essential to derive our features. The theoretical
foundation of these methods has been adapted for our purposes from angular momentum theory \cite{angular}, which plays an 
important role in Quantum Mechanics. Hence, we can rely on a well established and sound theoretical basis when we extend existing and 
derive novel operations in the second part of this chapter.\\
The reader may refer to \cite{angular}\cite{angular2}\cite{angular3} and \cite{sh} for a detailed introduction to angular momentum theory. 

\section{\label{sec:feature:sh}Spherical Harmonics}
\begin{figure}[htbp]
\index{Base functions}\index{Spherical Harmonics}
\begin{flushright}
\includegraphics[width=0.1\textwidth]{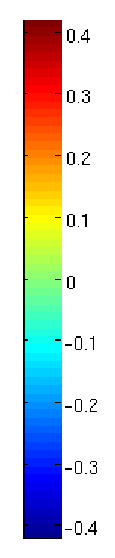}
\begin{tabular}{c|ccccc}
$m$&$l=1$ & $l=2$ & $l=3$ & $l=4$ & $l=5$\\
\hline
$-5$&
&
&
&
&
\includegraphics[width=0.12\textwidth]{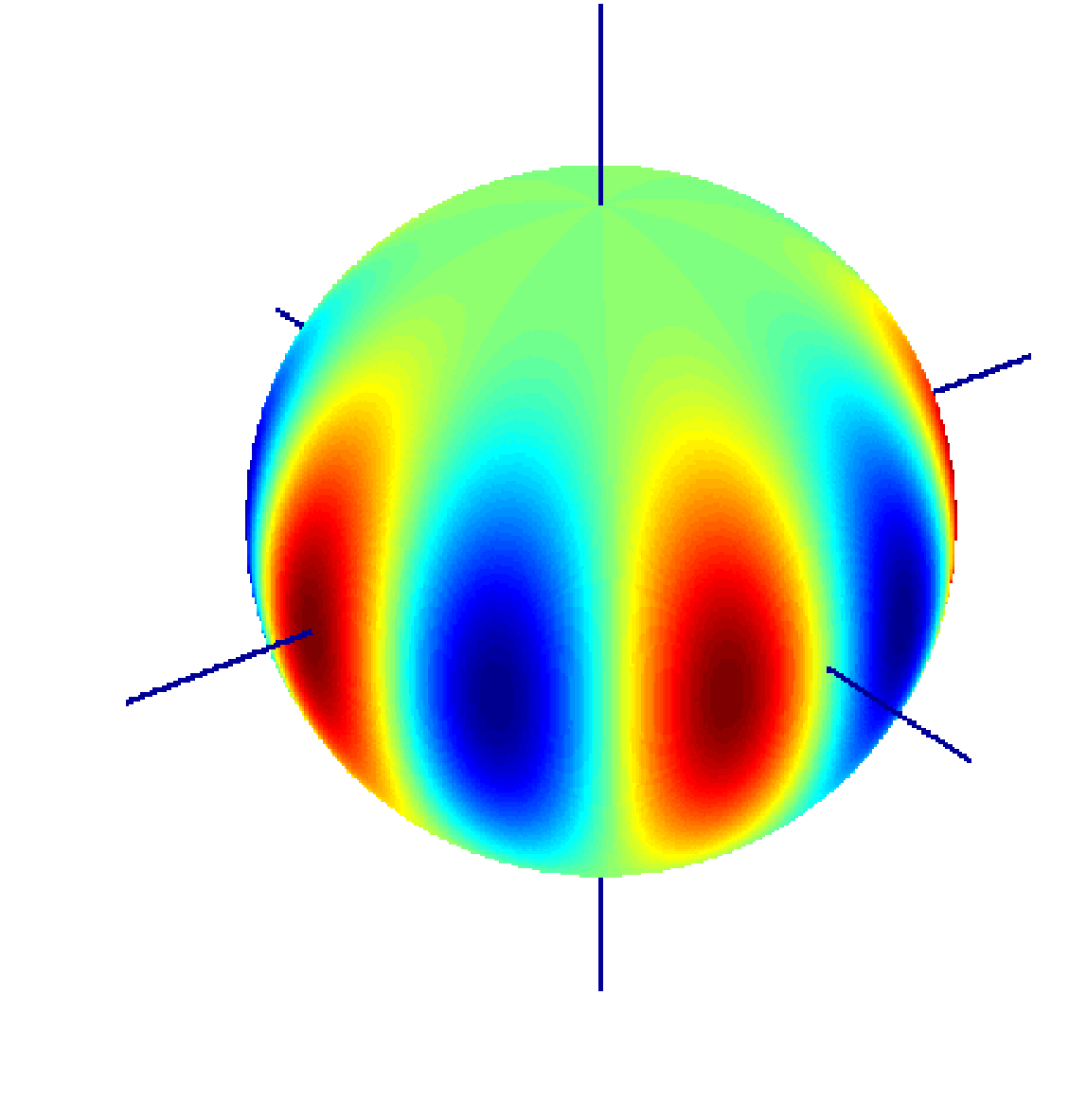}\\
$-4$&
&
&
&
\includegraphics[width=0.12\textwidth]{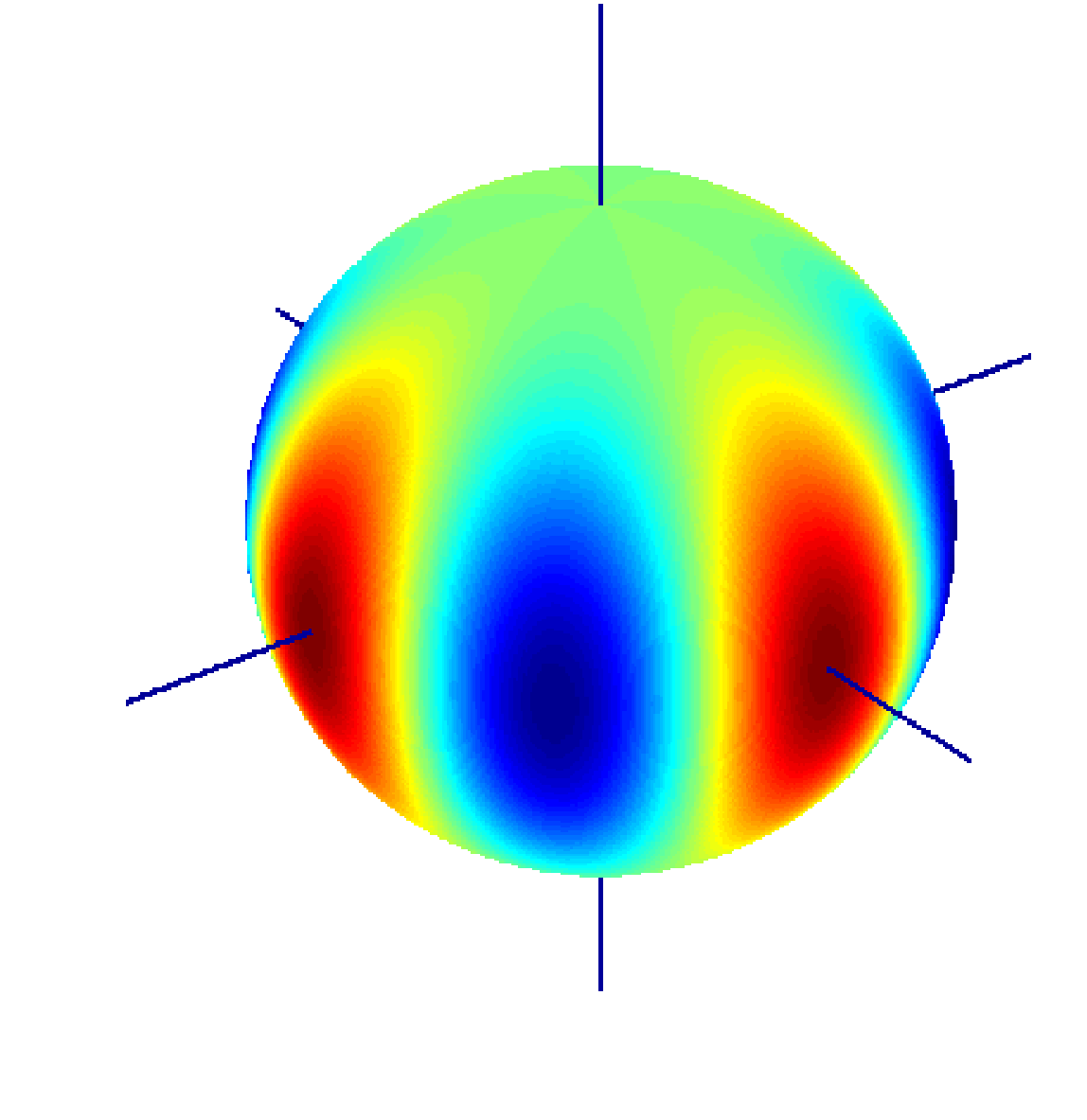}&
\includegraphics[width=0.12\textwidth]{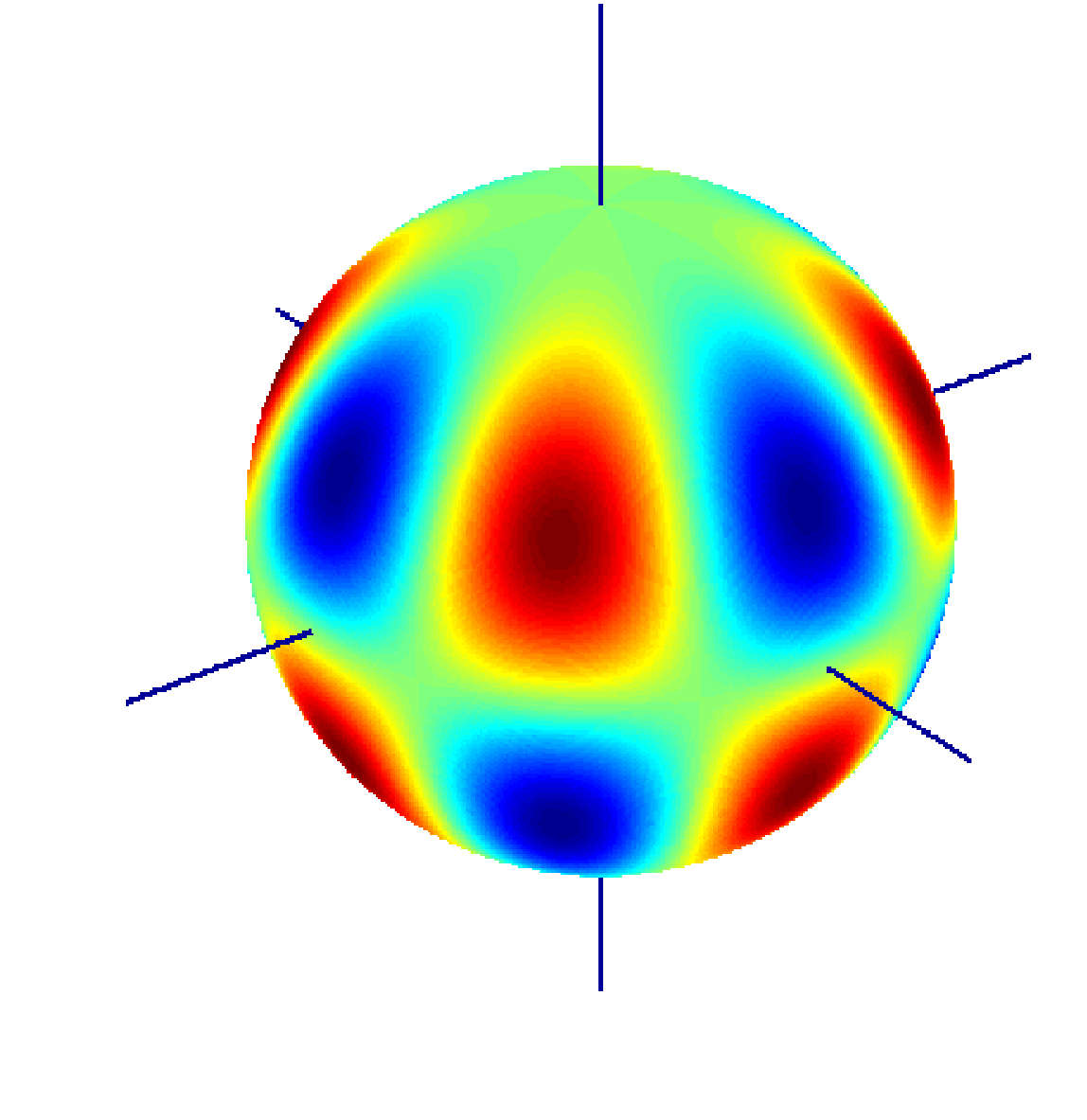}\\
$-3$&
&
&
\includegraphics[width=0.12\textwidth]{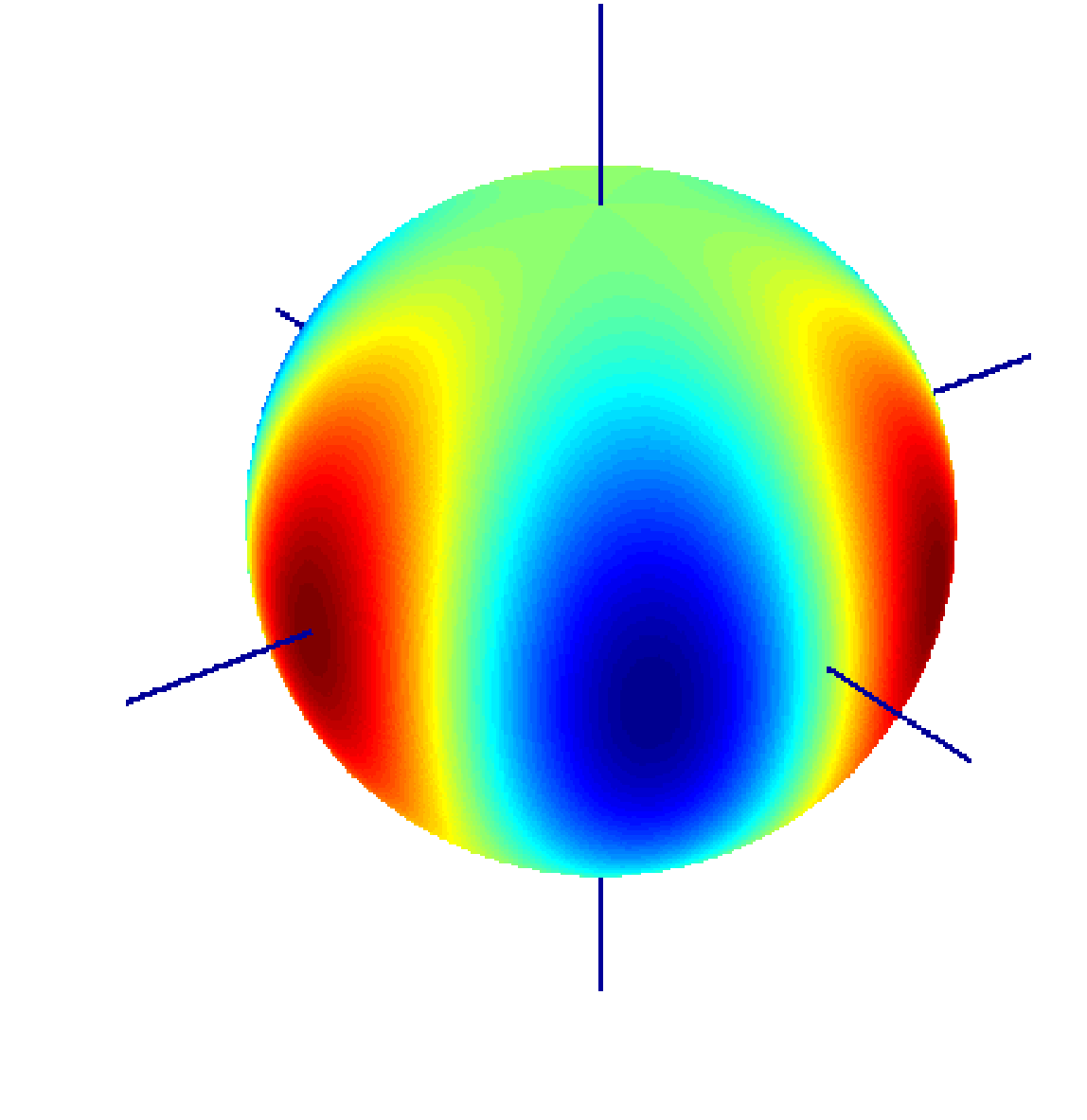}&
\includegraphics[width=0.12\textwidth]{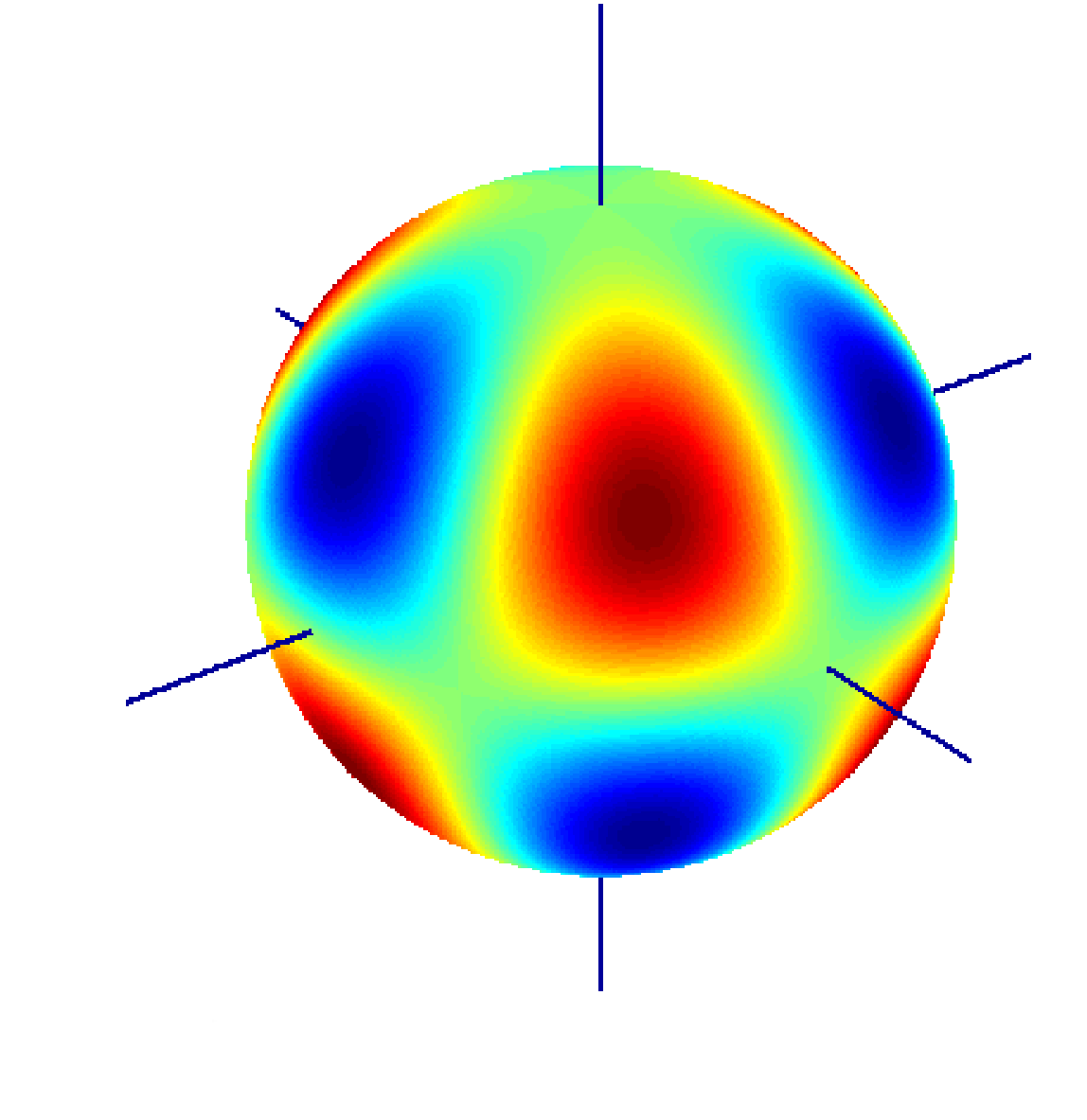}&
\includegraphics[width=0.12\textwidth]{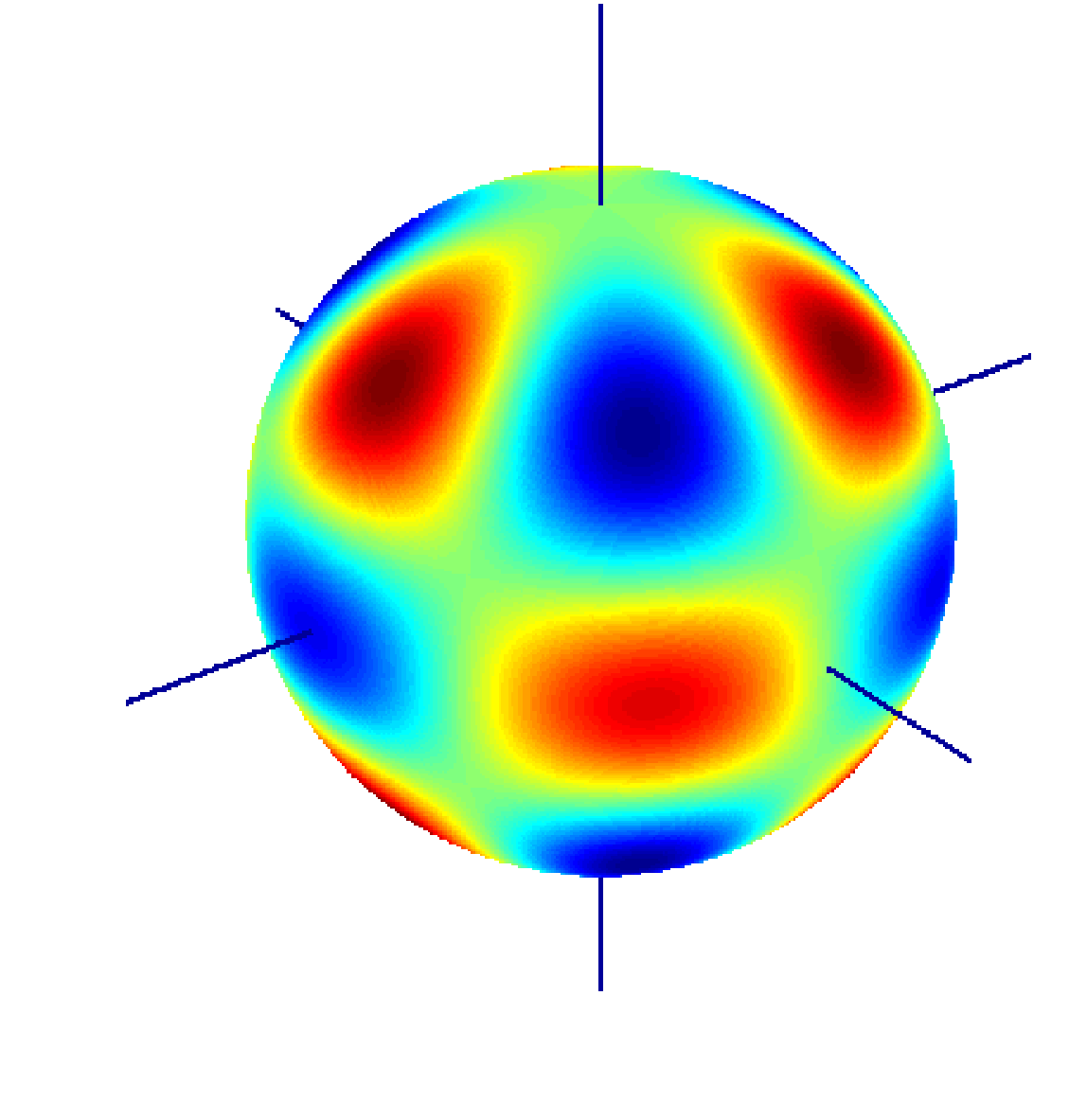}\\
$-2$
&
&
\includegraphics[width=0.12\textwidth]{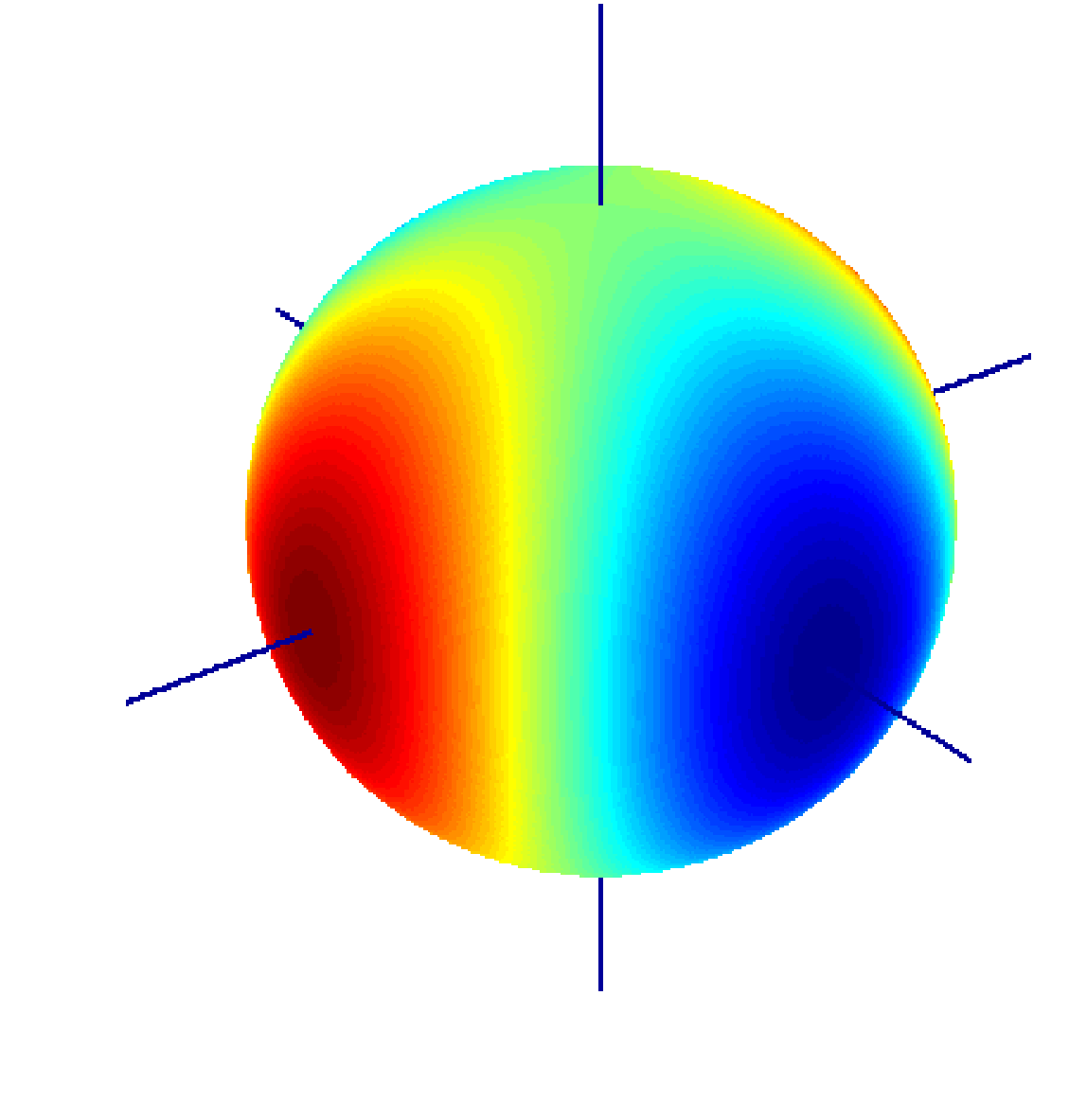}&
\includegraphics[width=0.12\textwidth]{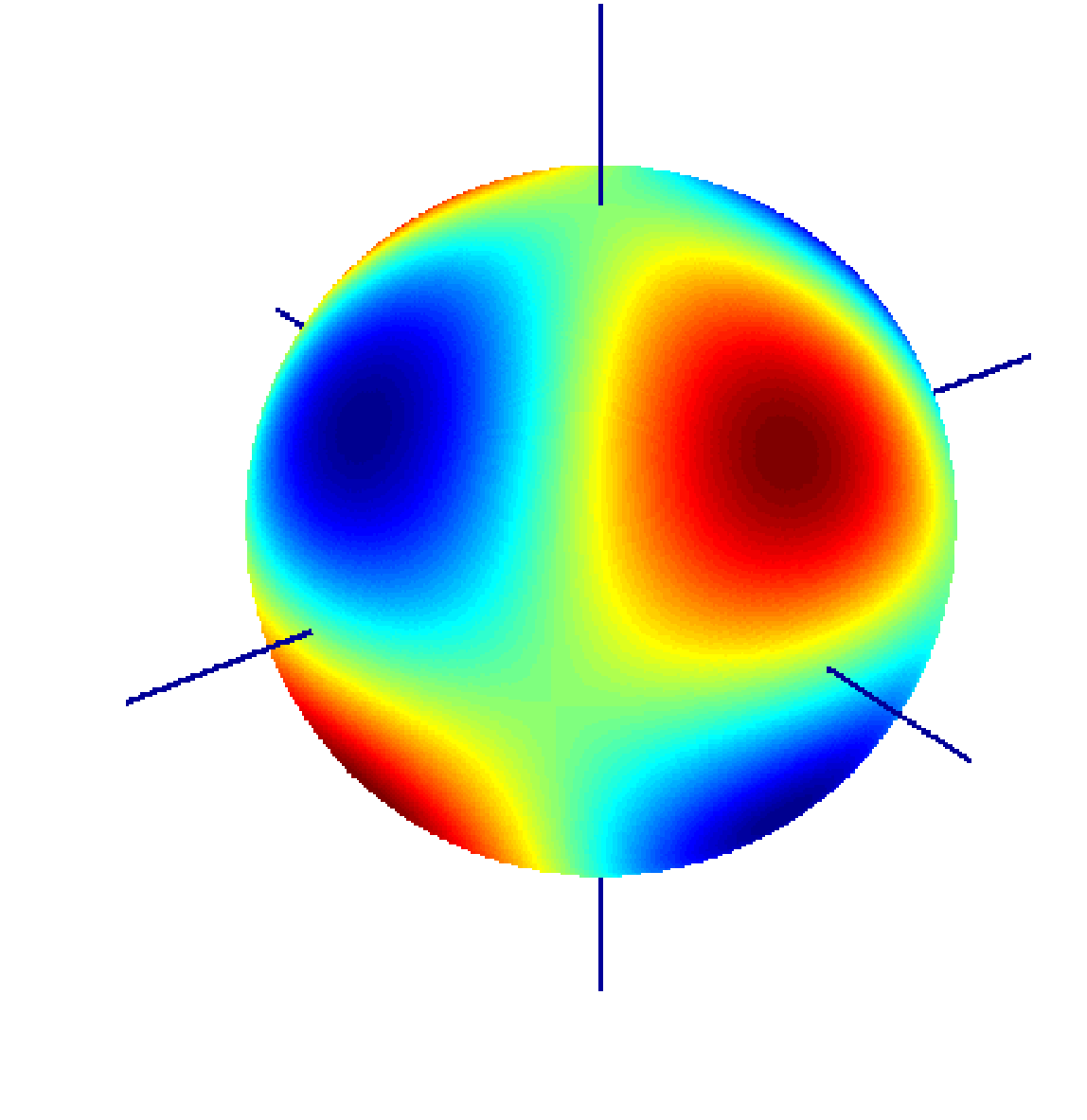}&
\includegraphics[width=0.12\textwidth]{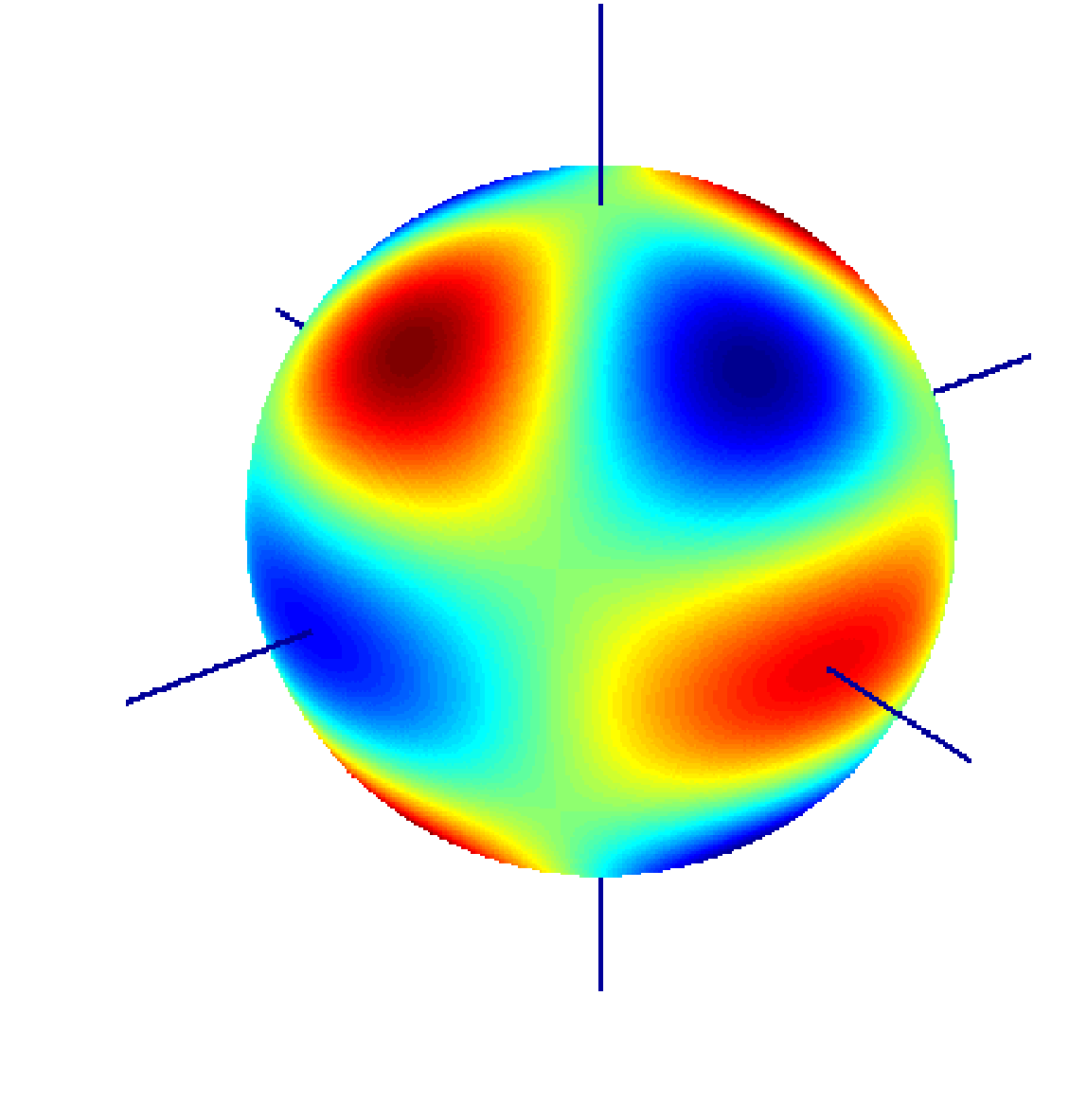}&
\includegraphics[width=0.12\textwidth]{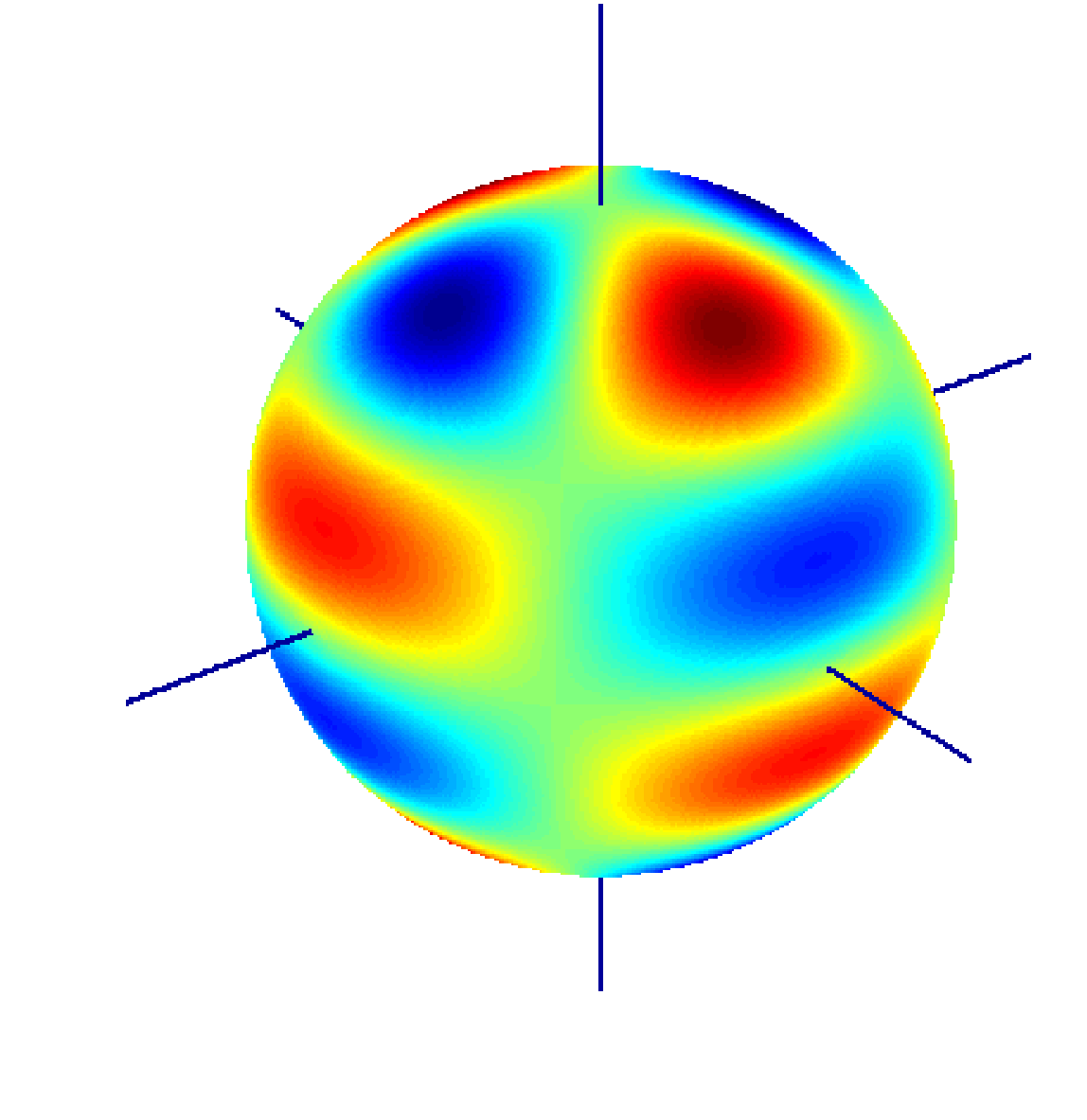}\\
$-1$
&
\includegraphics[width=0.12\textwidth]{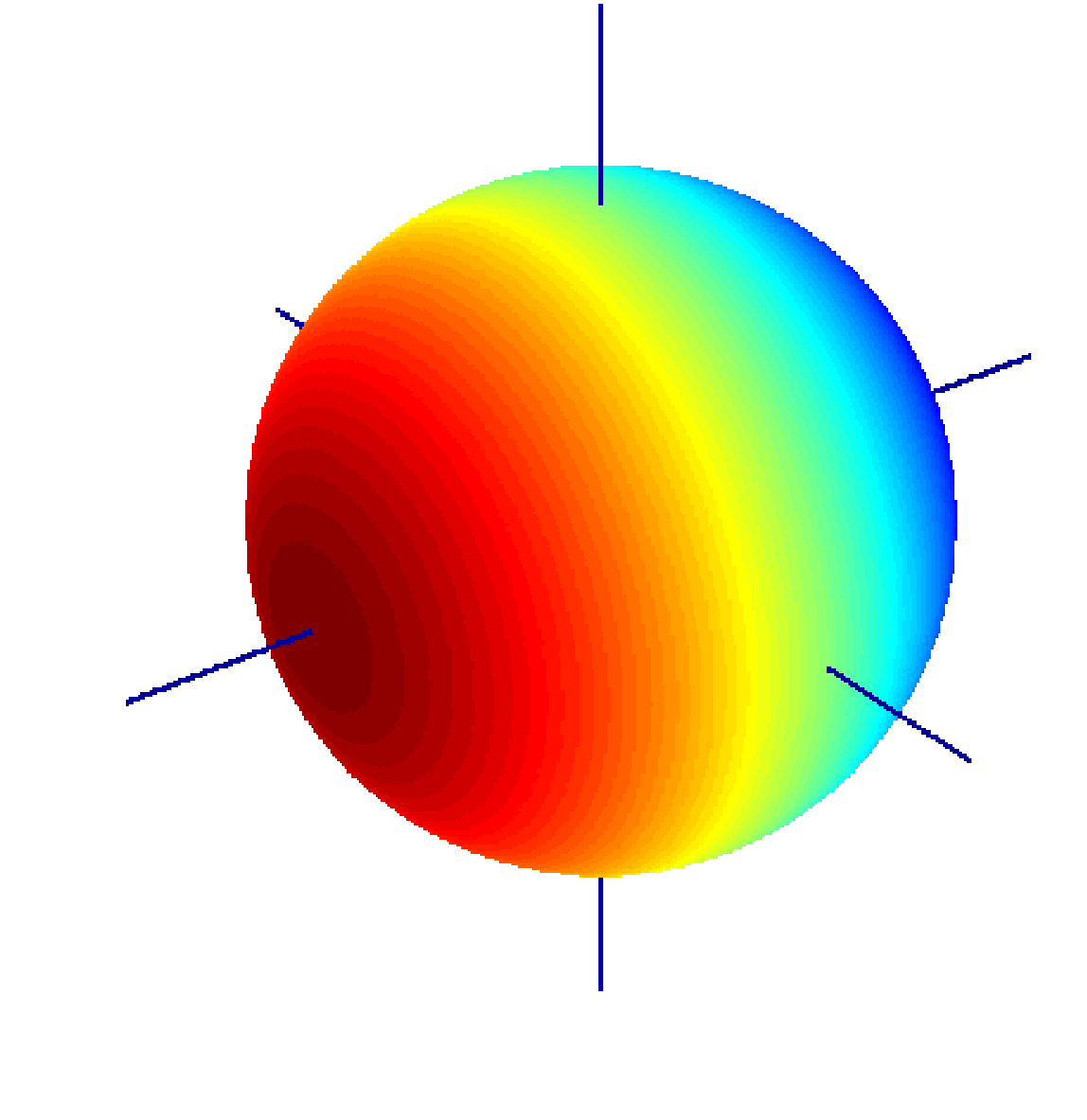}&
\includegraphics[width=0.12\textwidth]{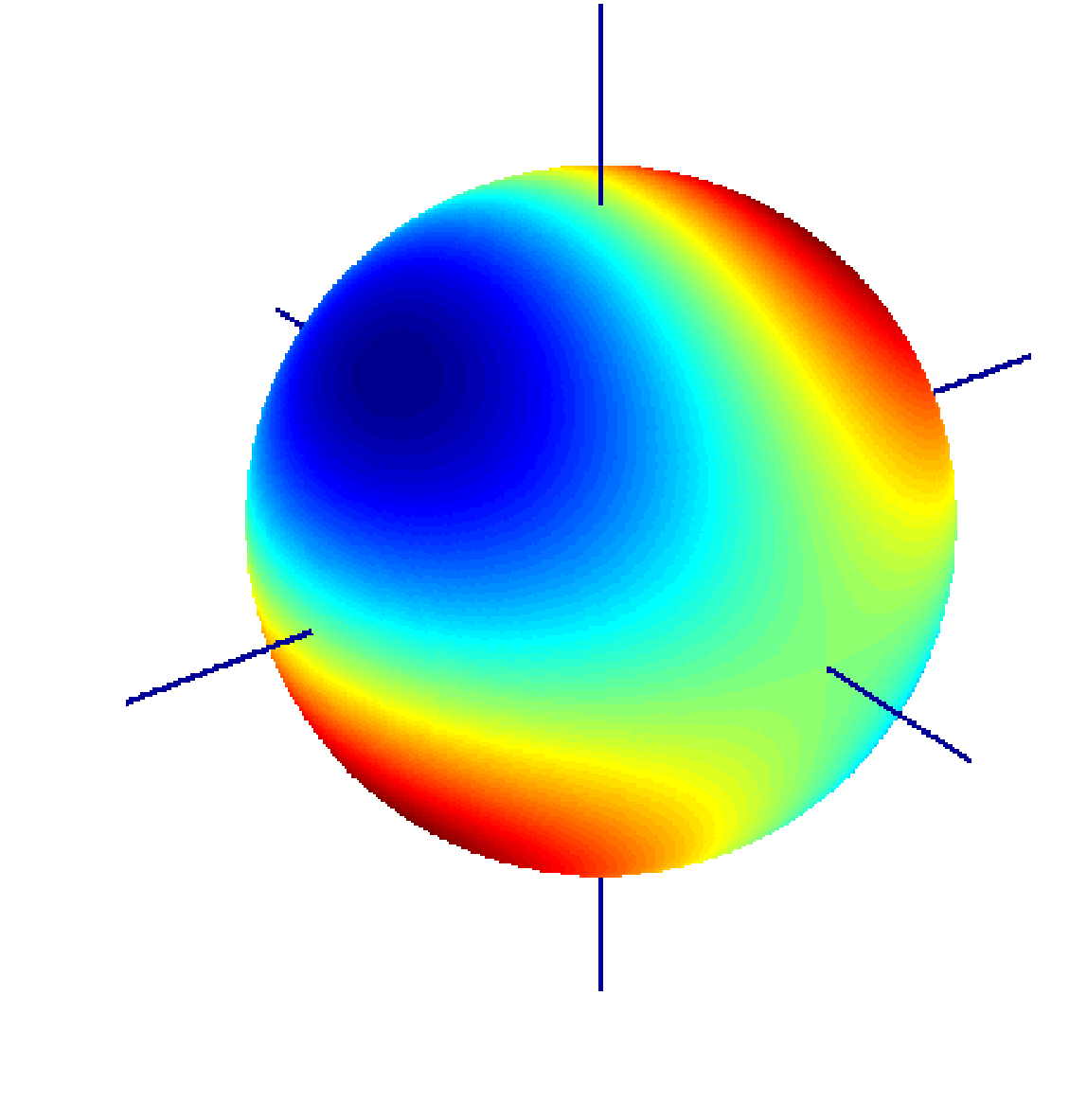}&
\includegraphics[width=0.12\textwidth]{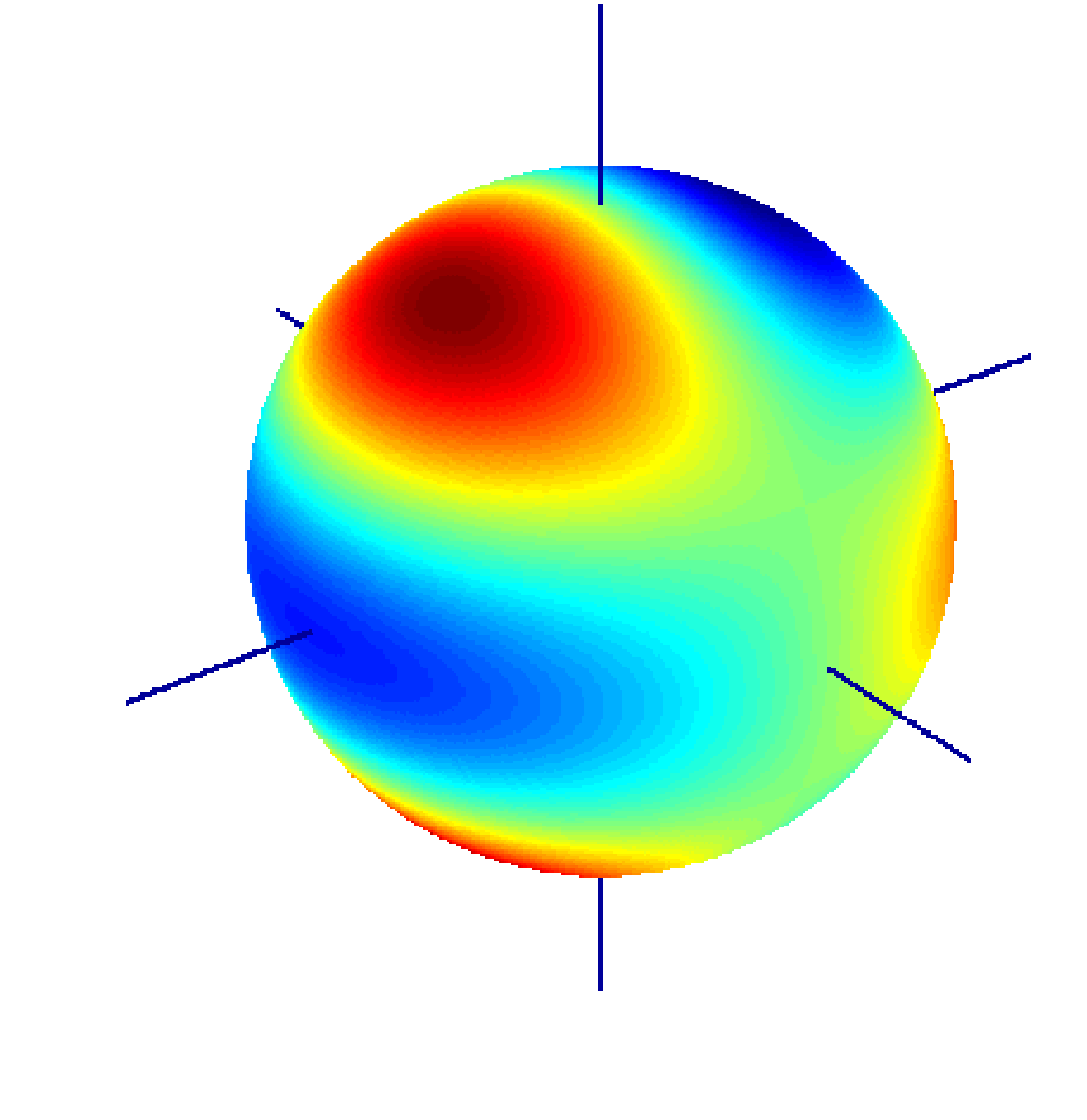}&
\includegraphics[width=0.12\textwidth]{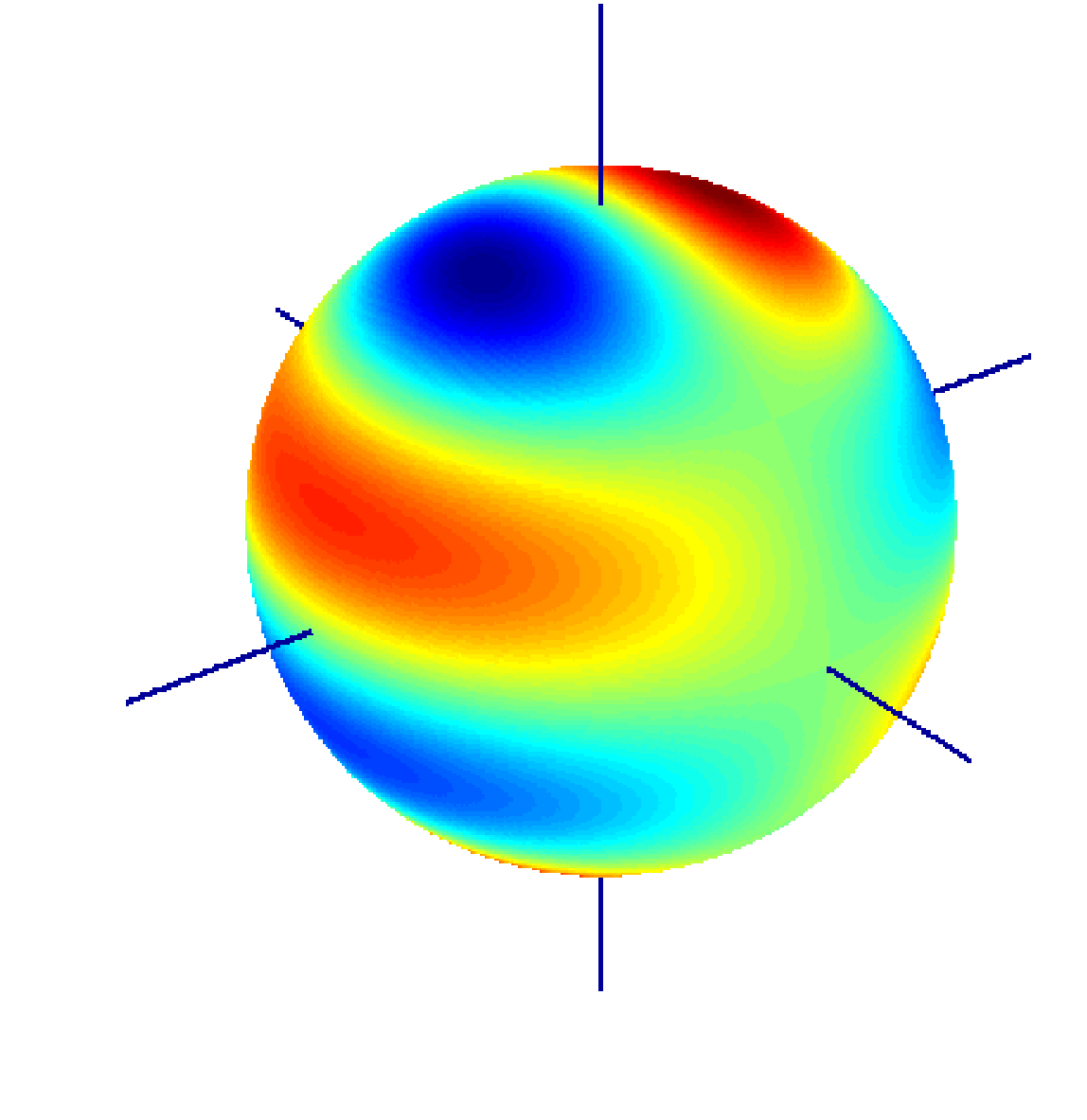}&
\includegraphics[width=0.12\textwidth]{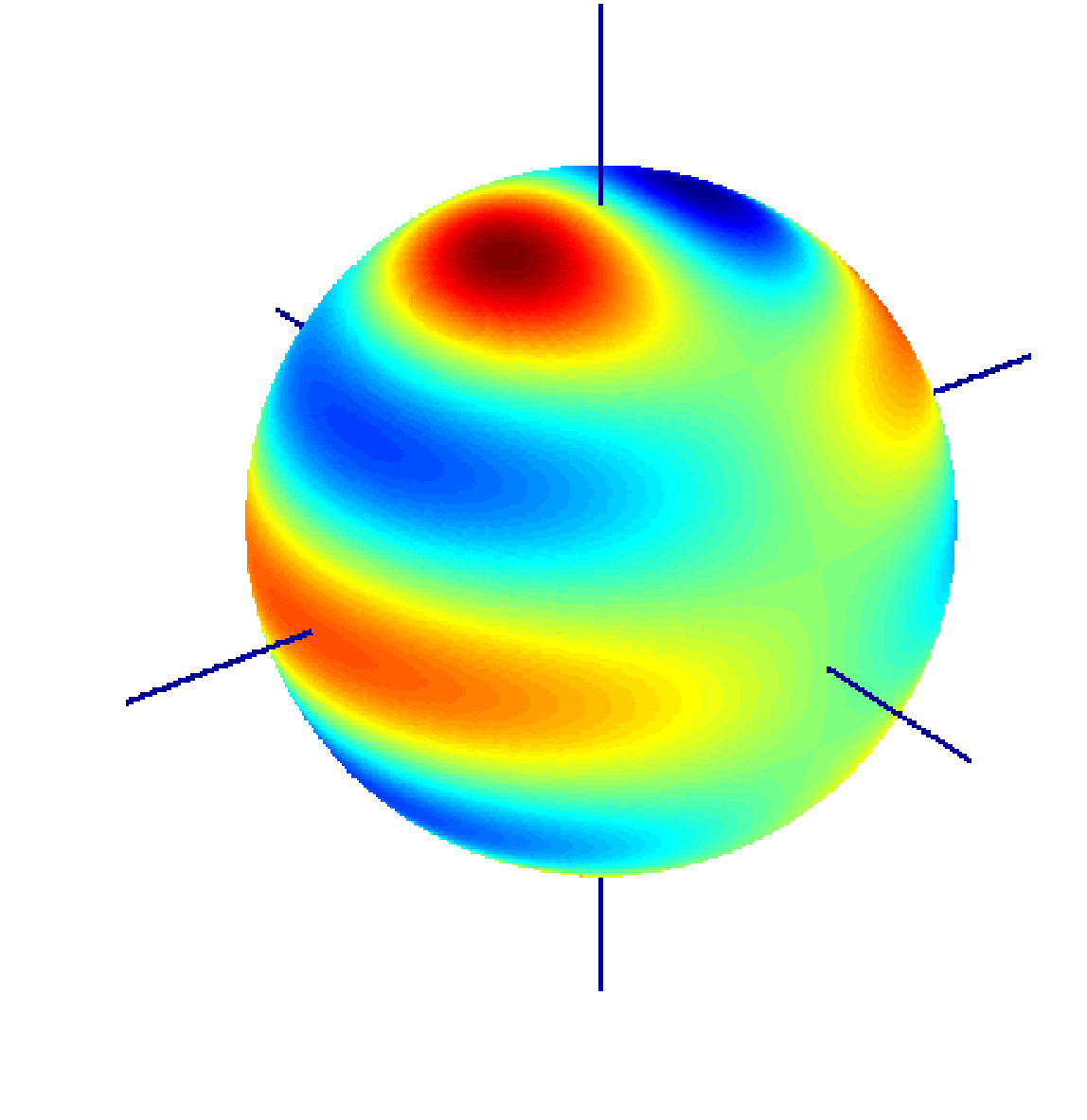}\\
$0$&
\includegraphics[width=0.12\textwidth]{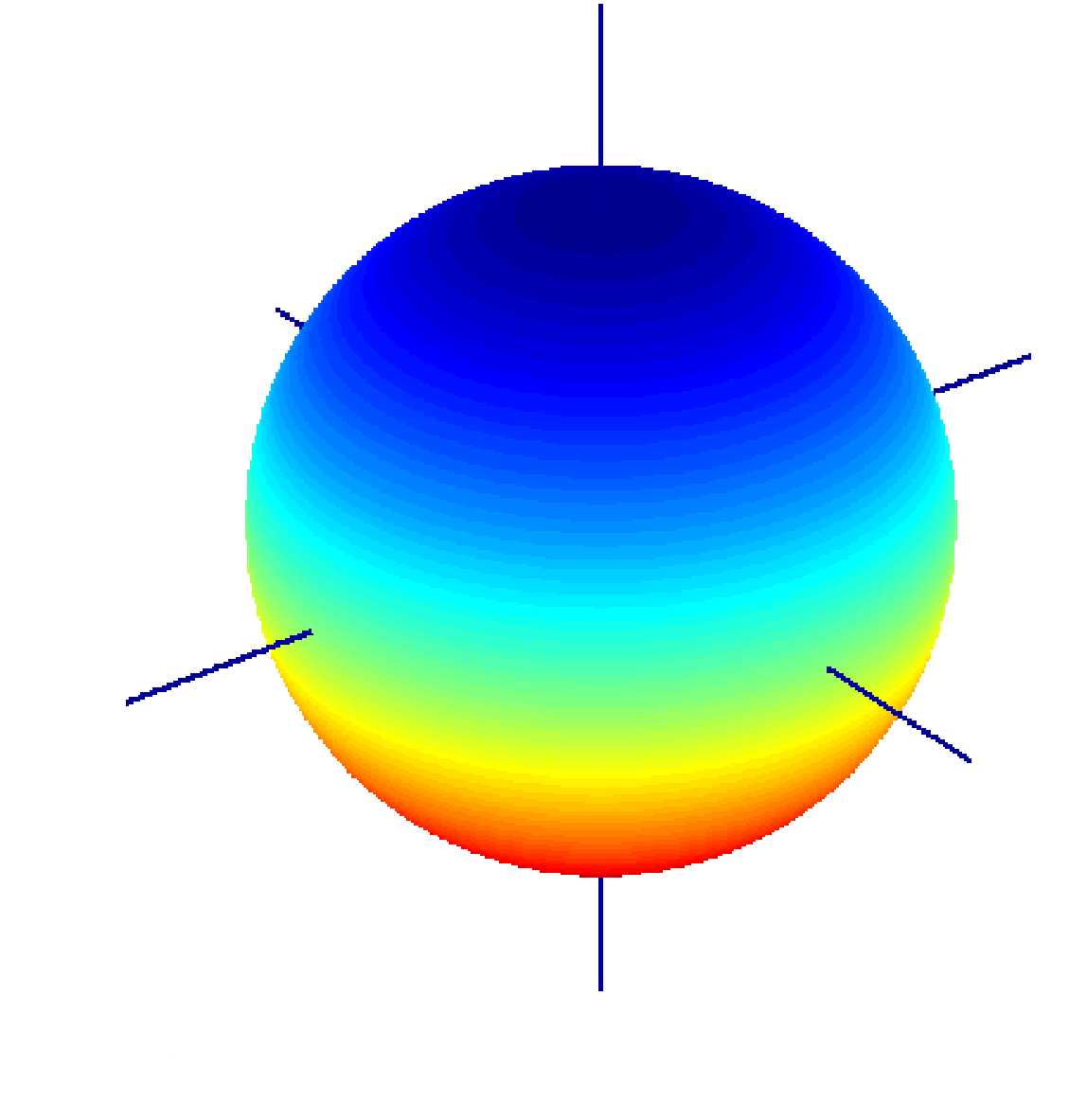}&
\includegraphics[width=0.12\textwidth]{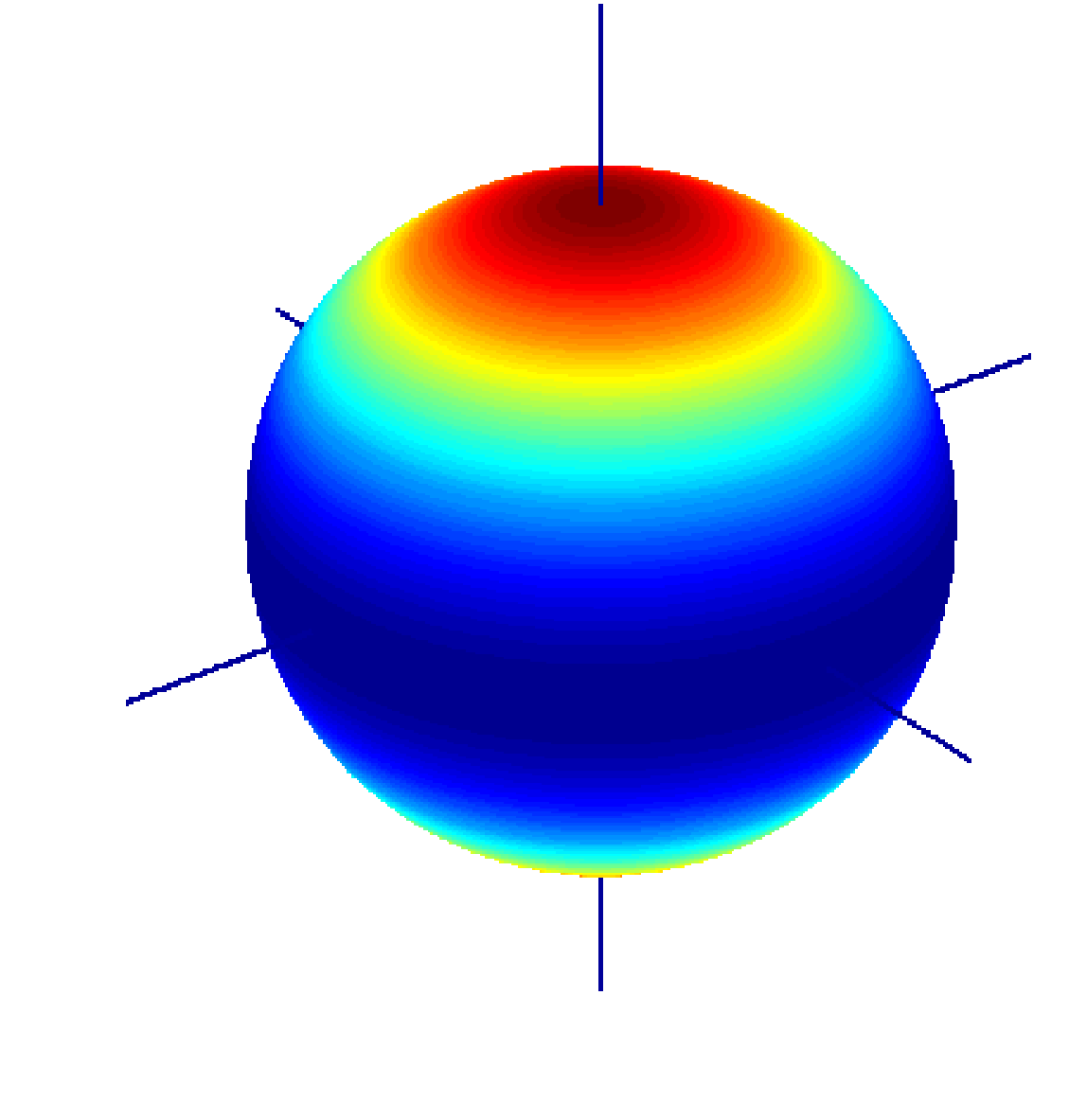}&
\includegraphics[width=0.12\textwidth]{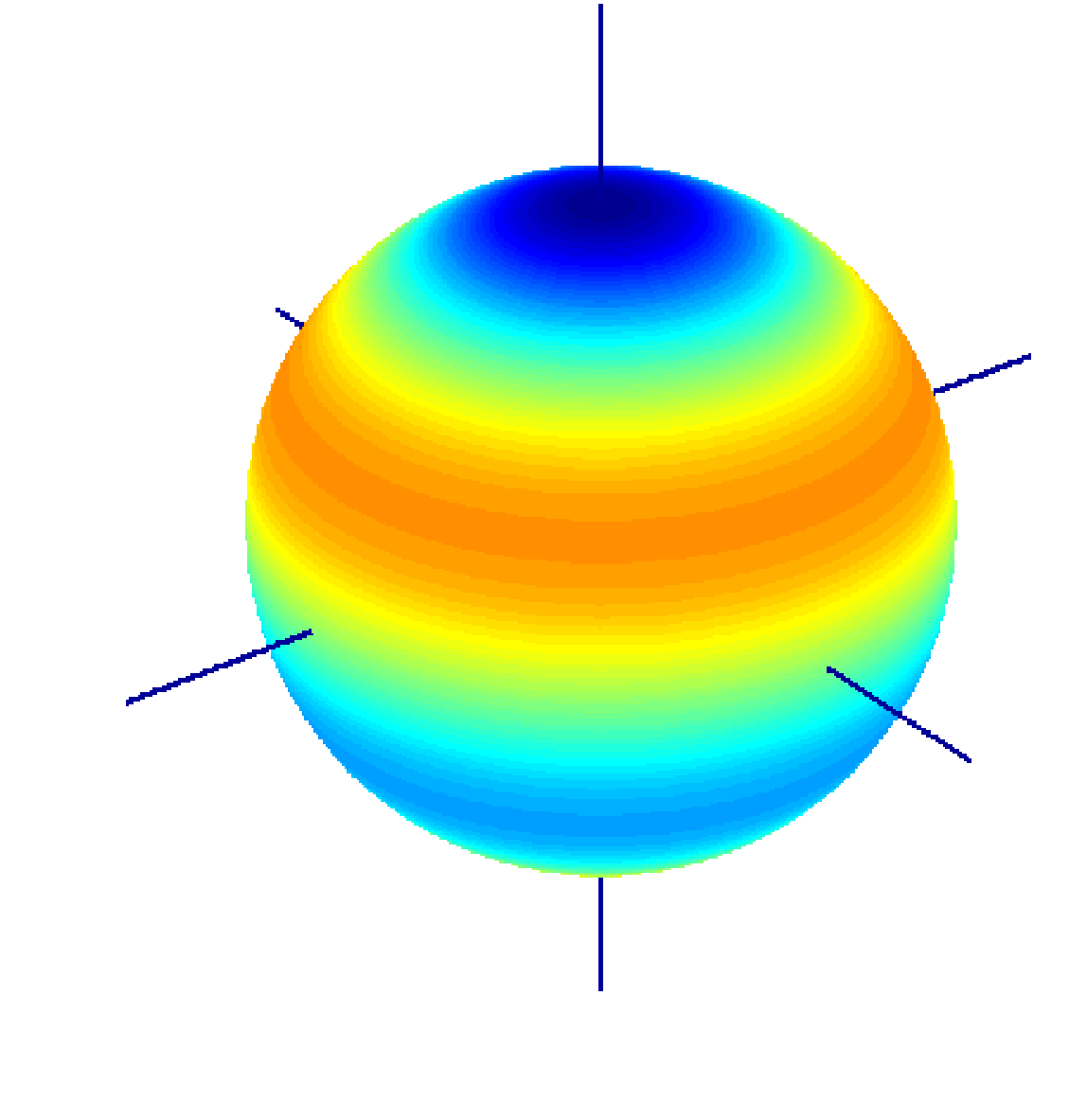}&
\includegraphics[width=0.12\textwidth]{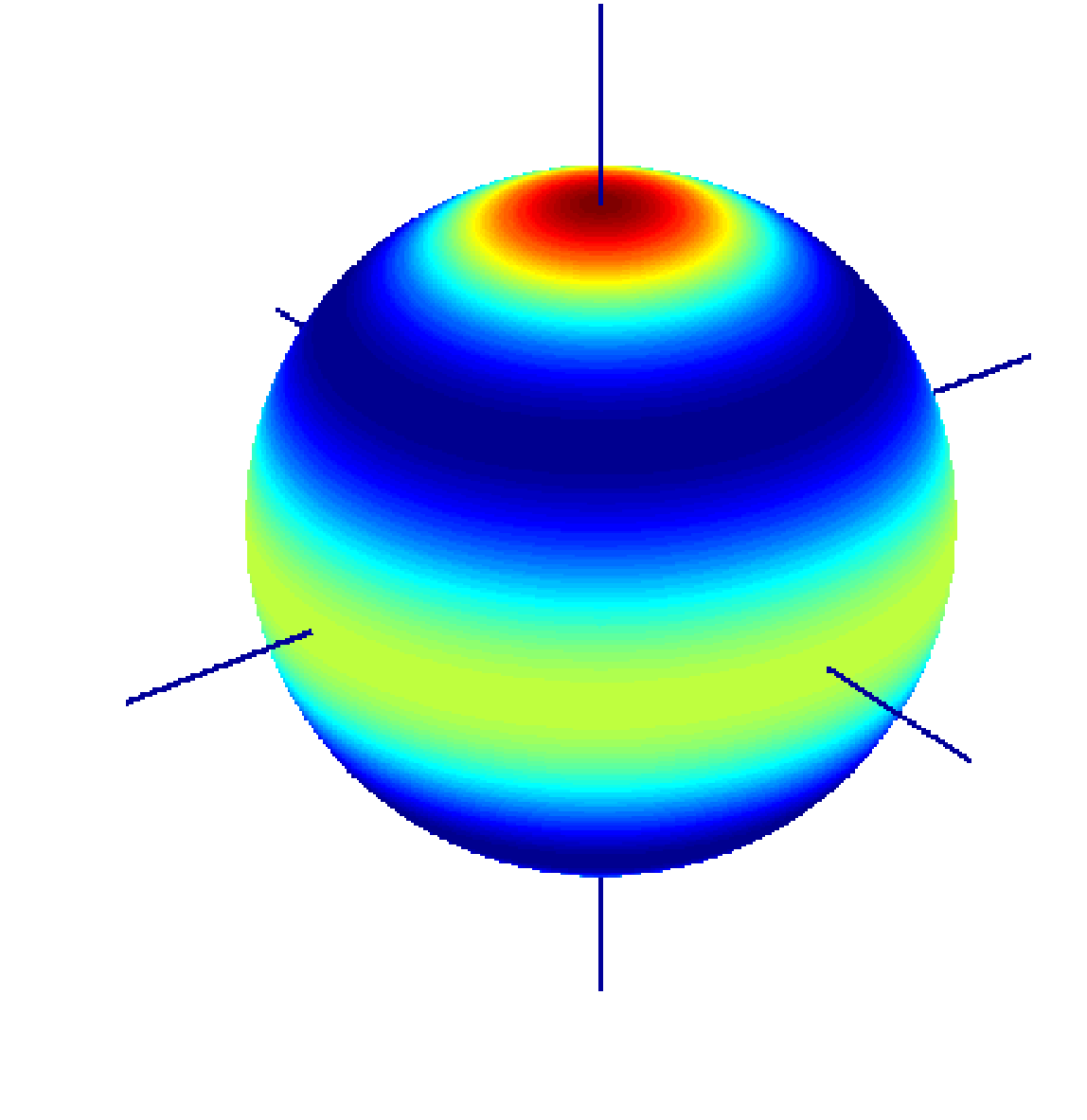}&
\includegraphics[width=0.12\textwidth]{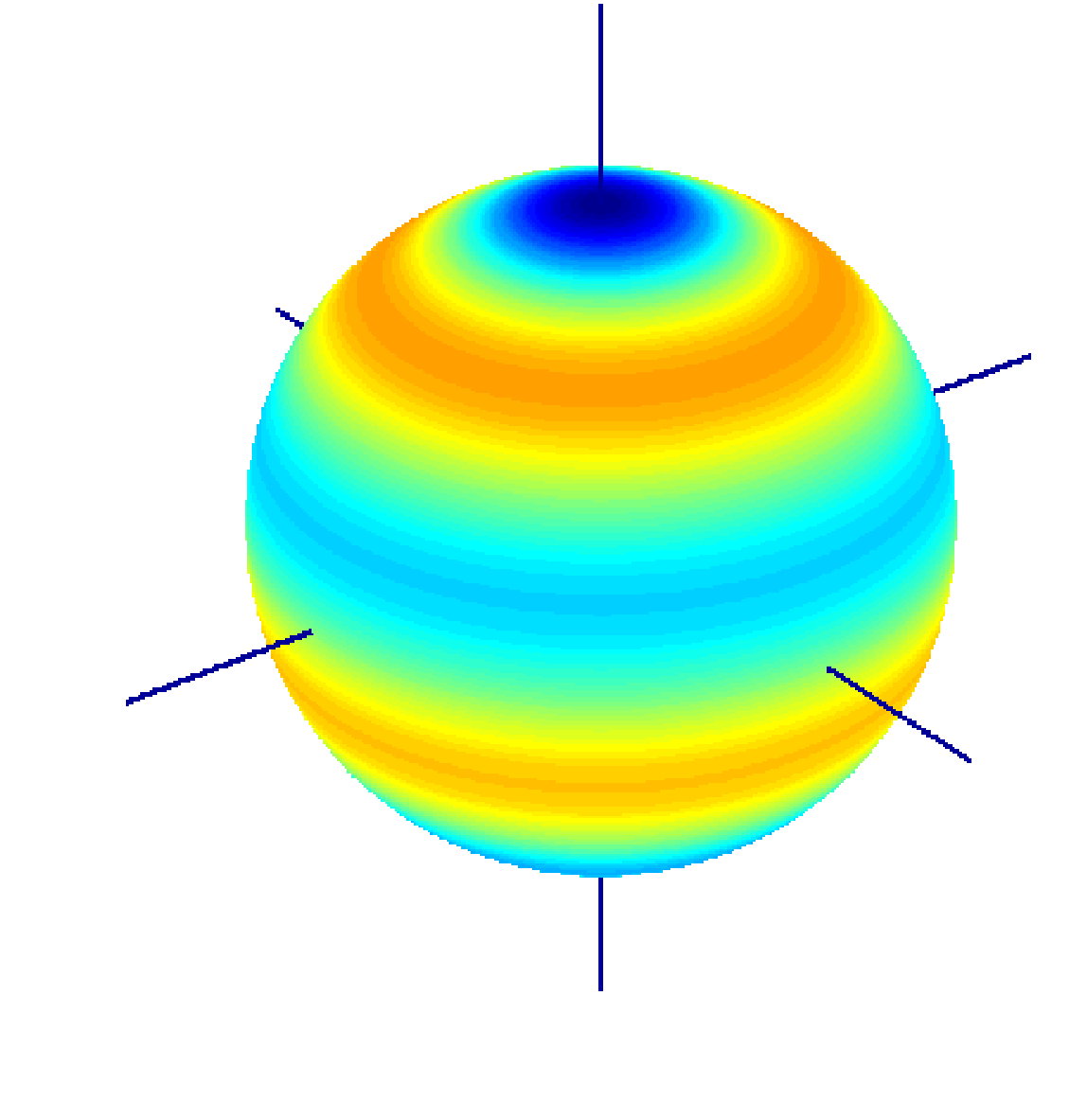}\\
$1$
&
\includegraphics[width=0.12\textwidth]{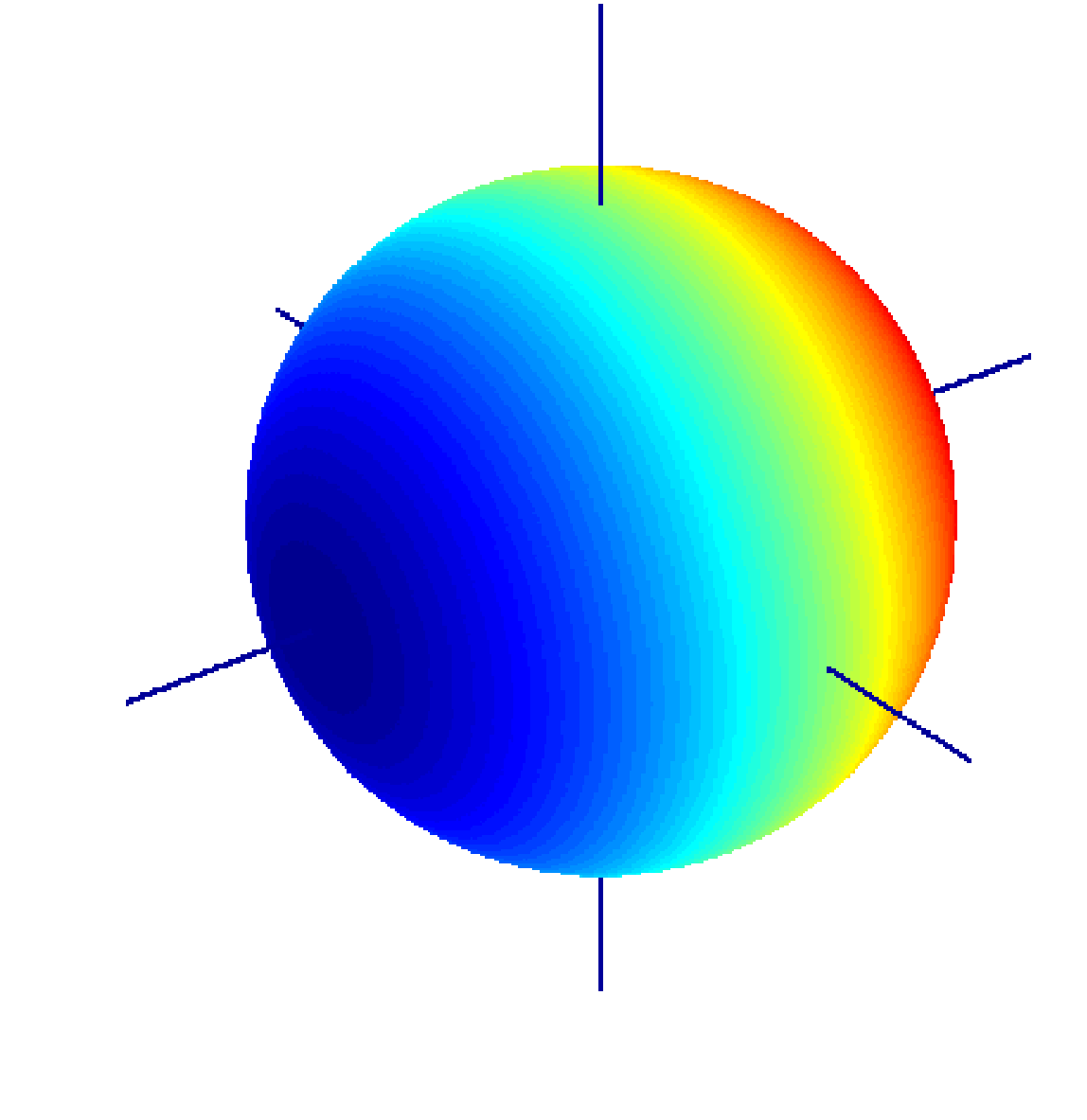}&
\includegraphics[width=0.12\textwidth]{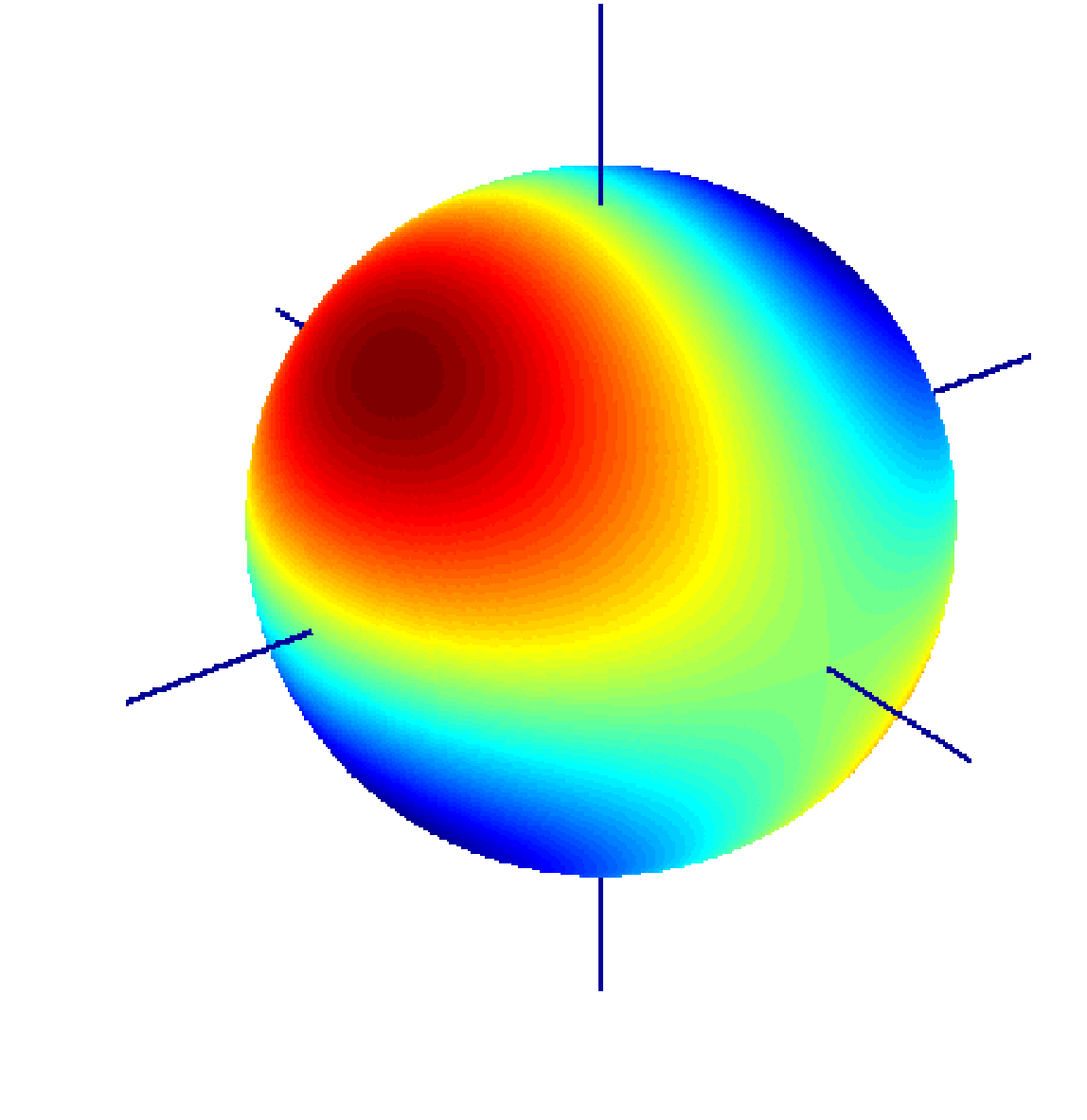}&
\includegraphics[width=0.12\textwidth]{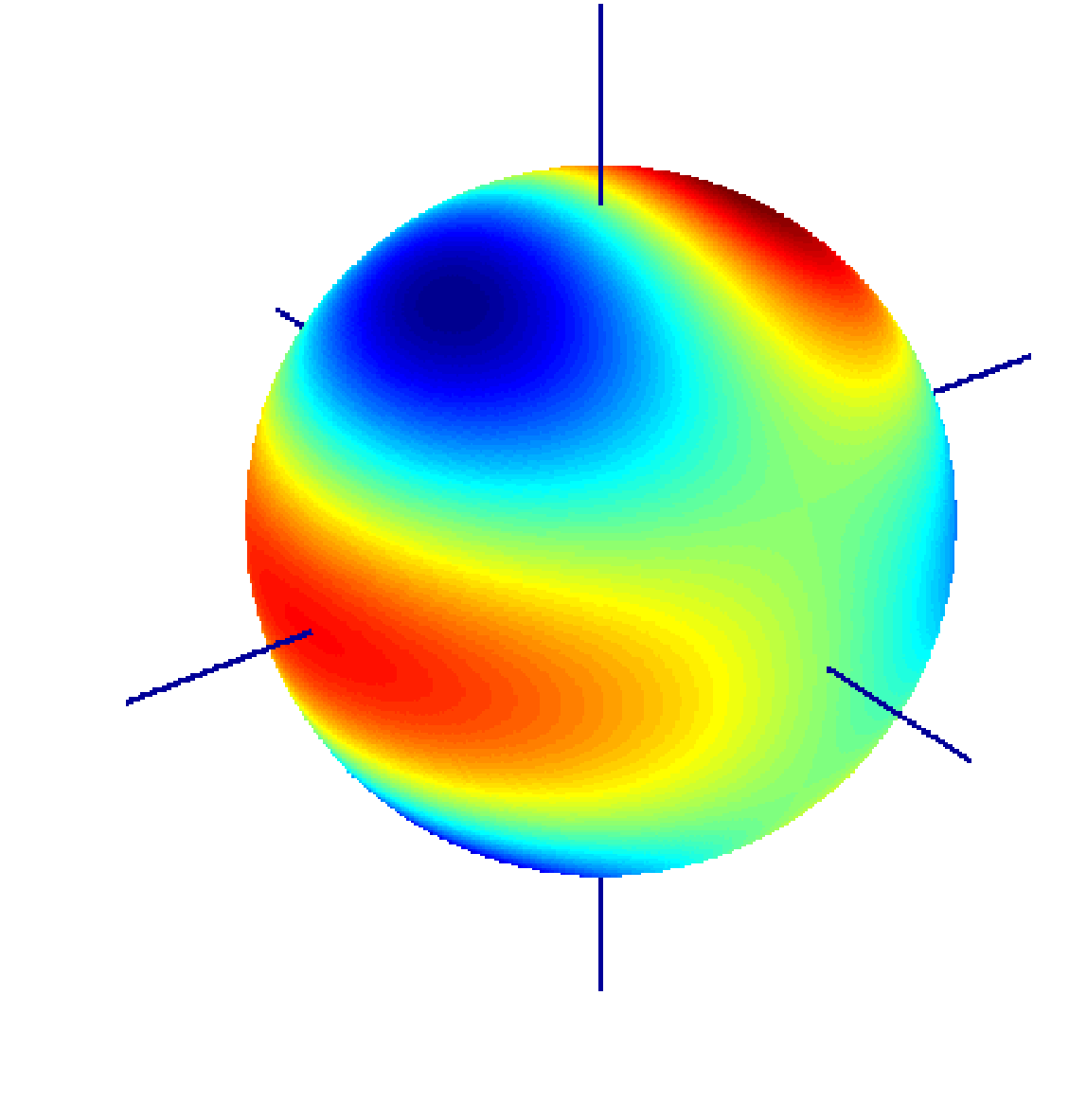}&
\includegraphics[width=0.12\textwidth]{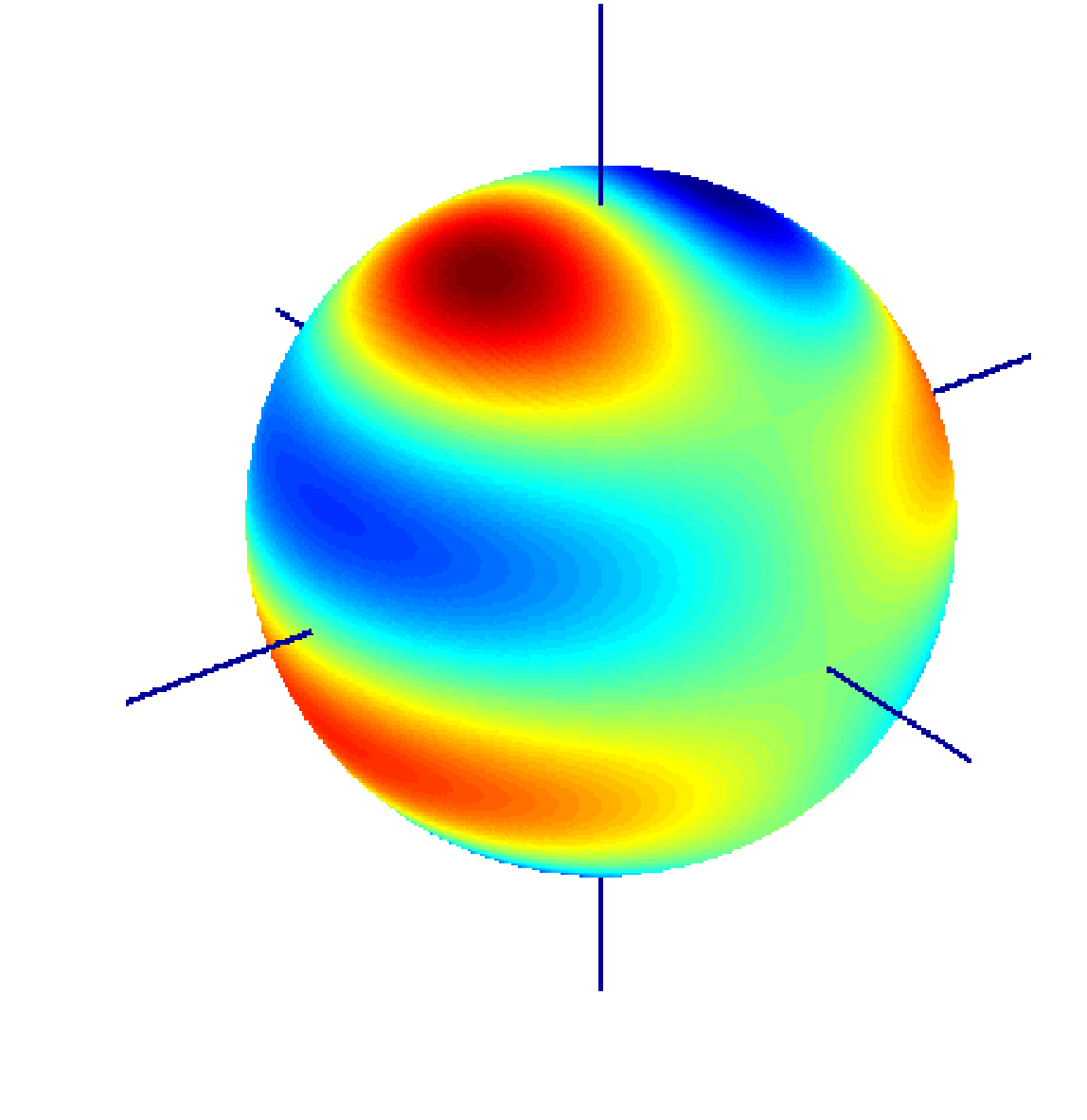}&
\includegraphics[width=0.12\textwidth]{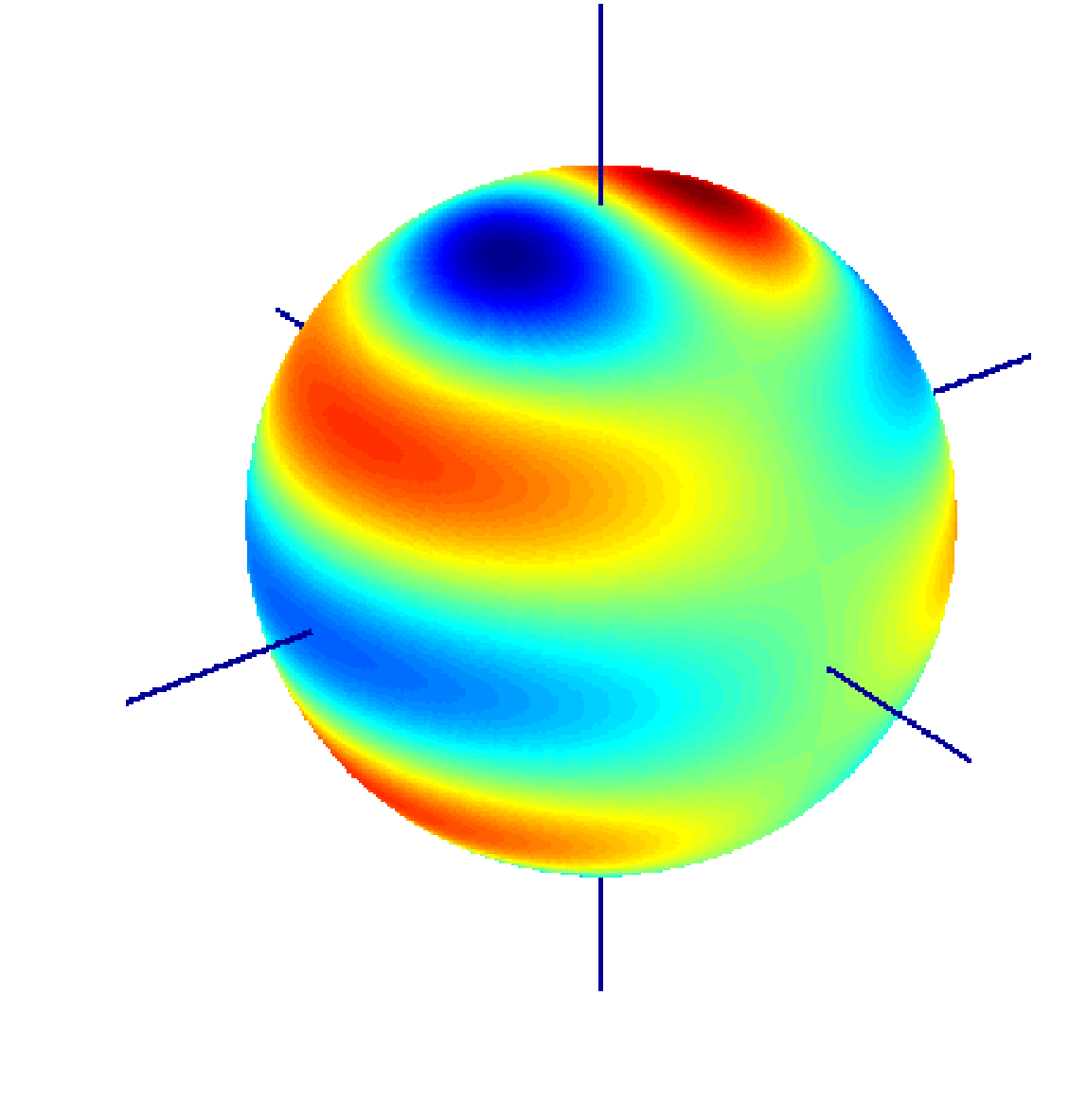}\\
$2$
&
&
\includegraphics[width=0.12\textwidth]{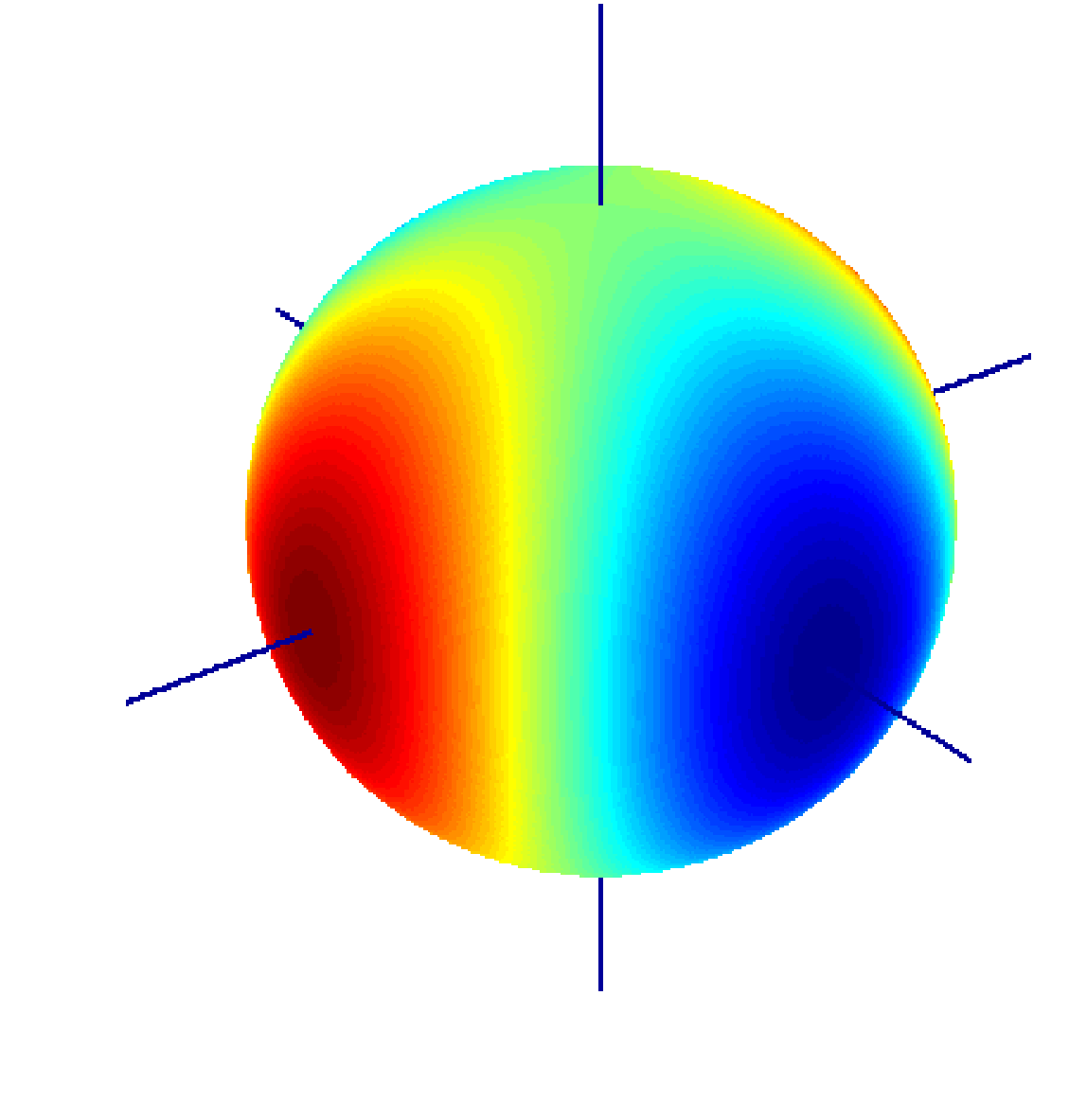}&
\includegraphics[width=0.12\textwidth]{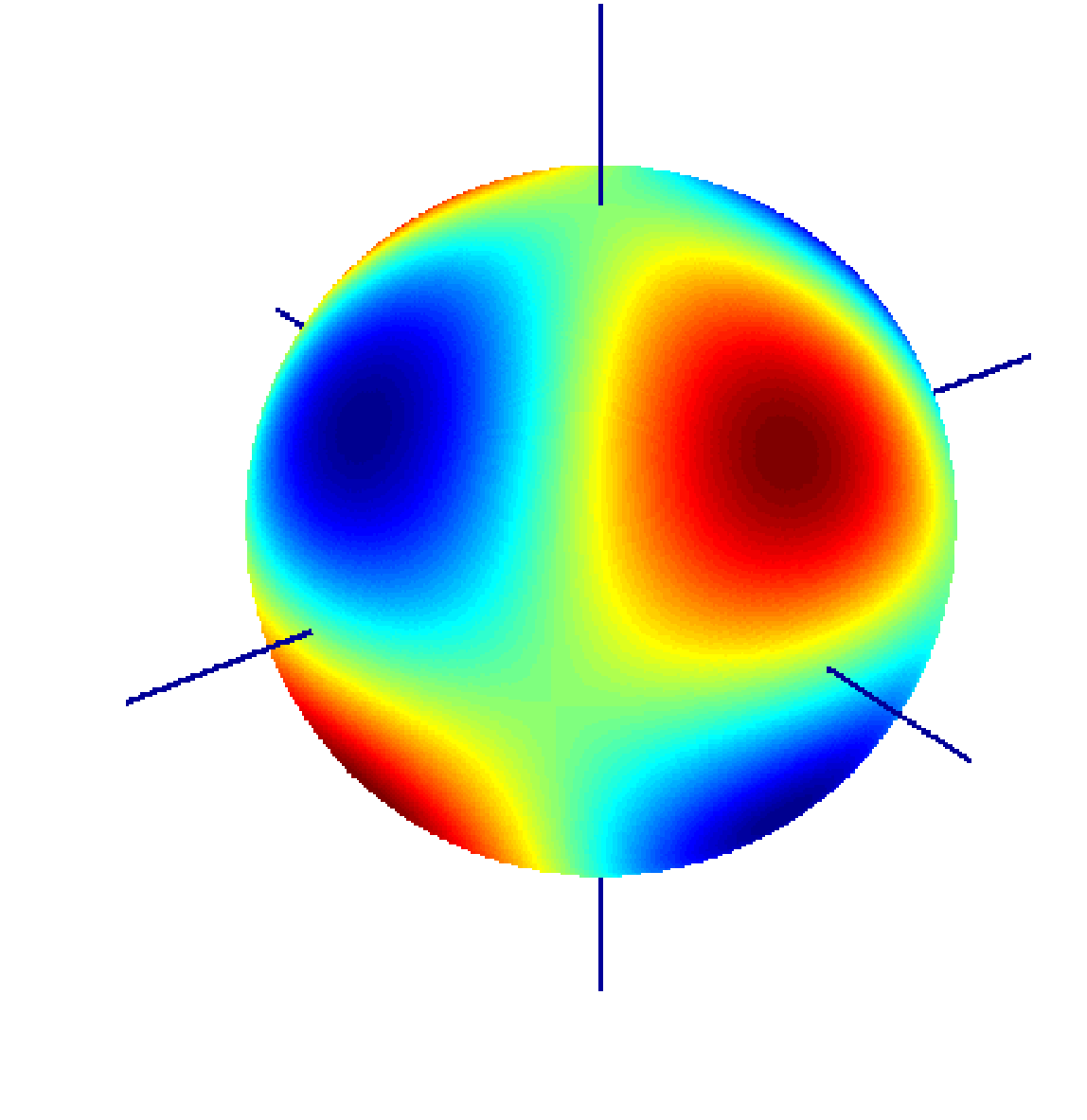}&
\includegraphics[width=0.12\textwidth]{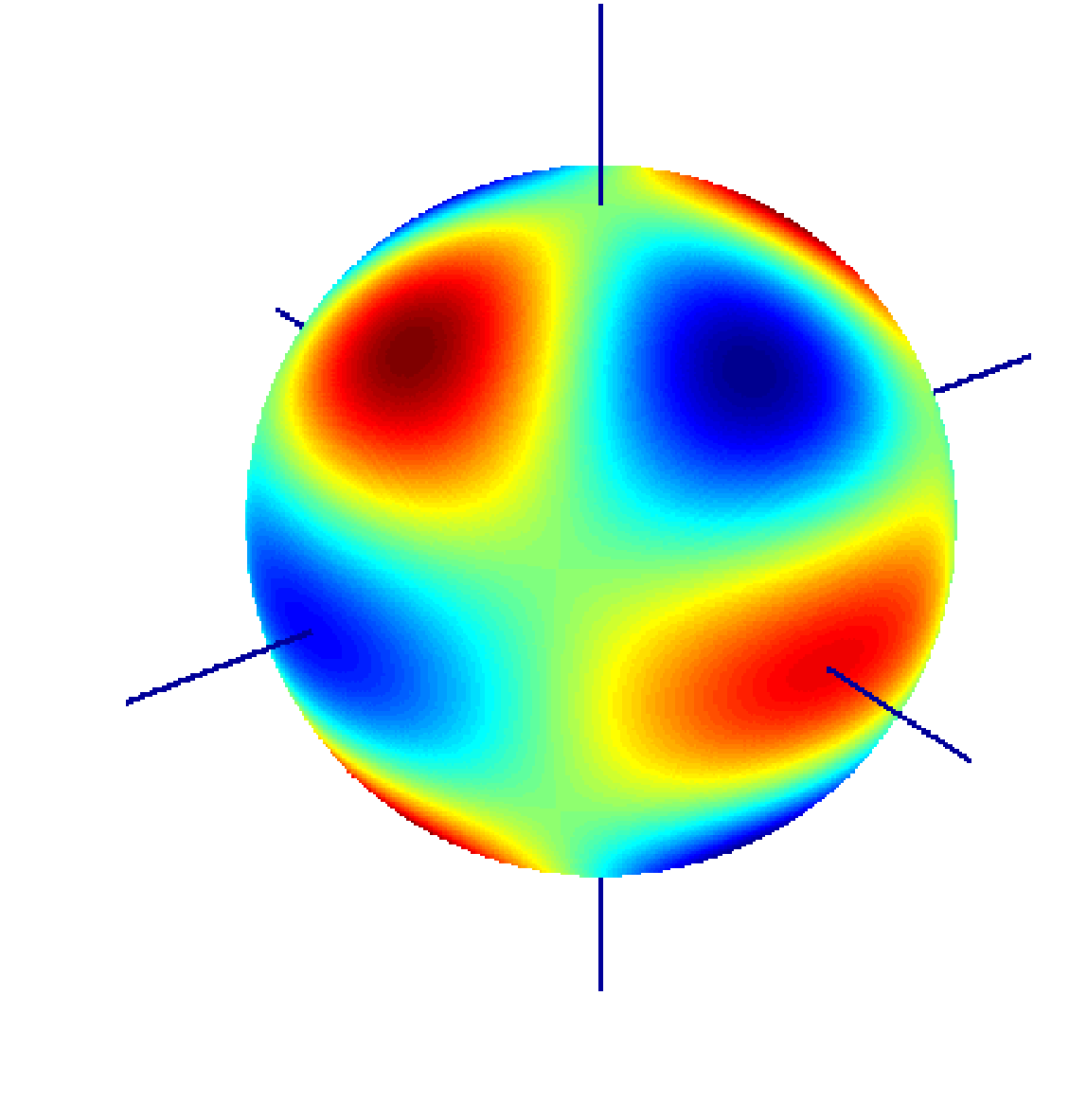}&
\includegraphics[width=0.12\textwidth]{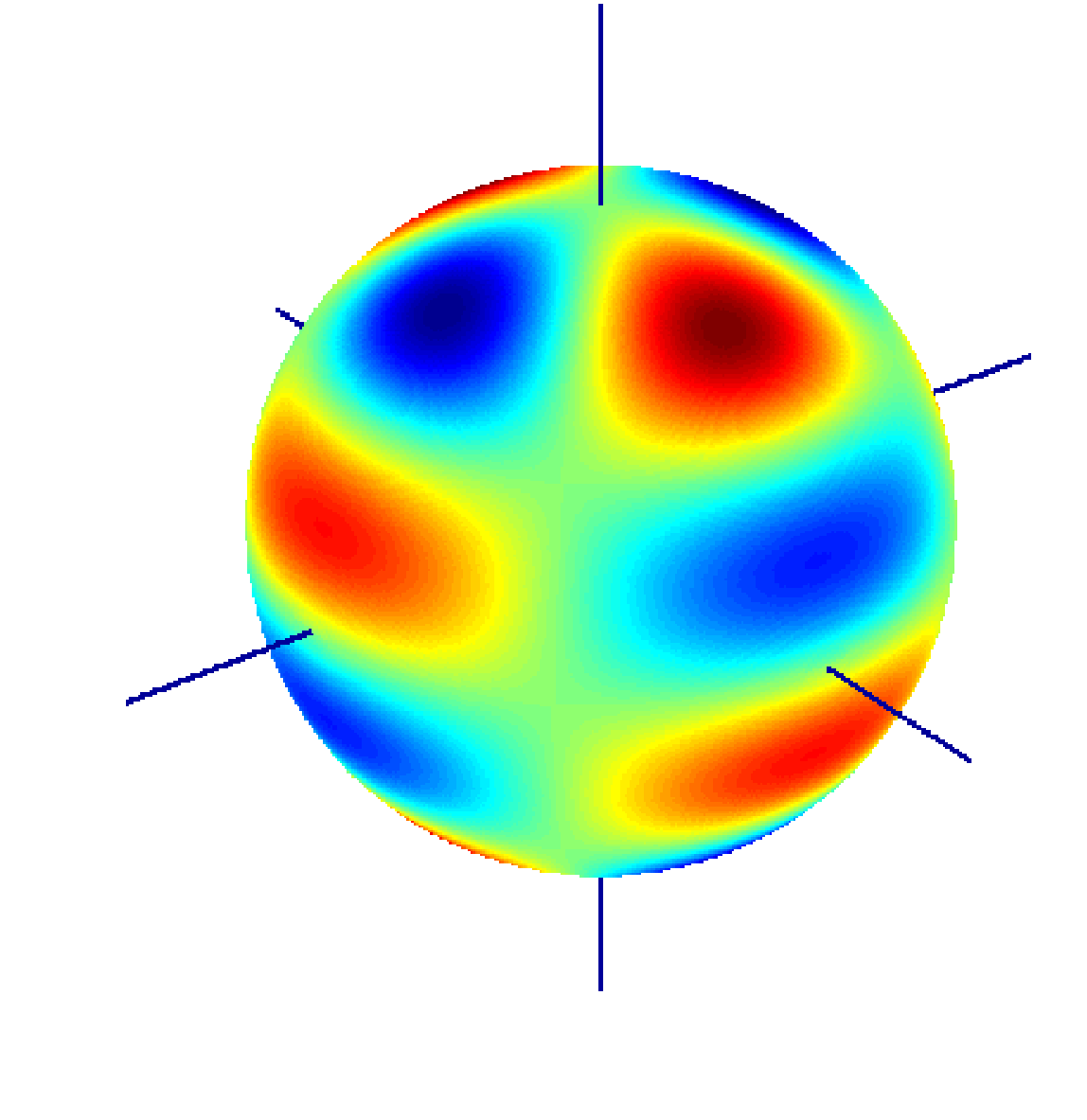}\\
$3$
&
&
&
\includegraphics[width=0.12\textwidth]{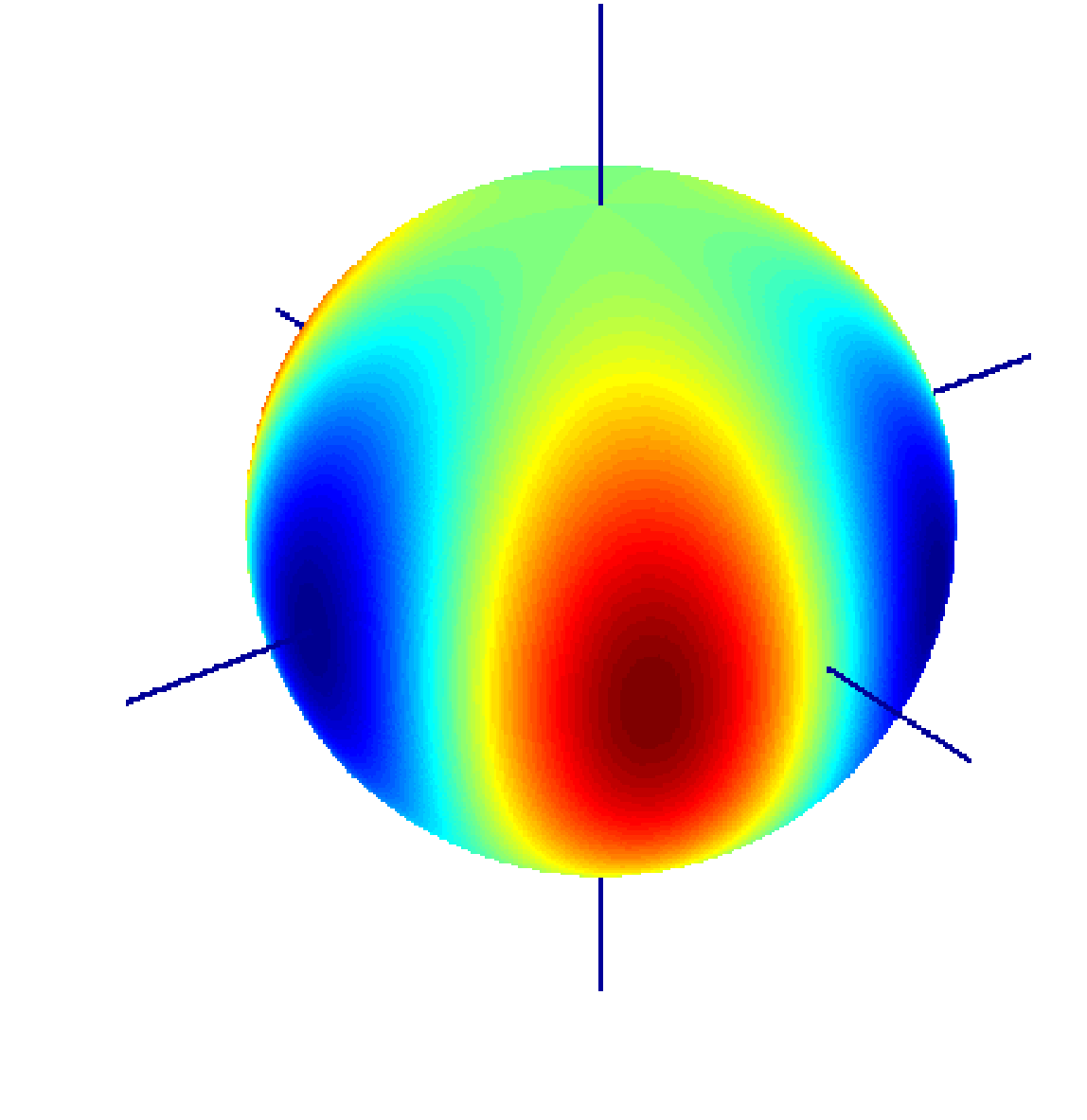}&
\includegraphics[width=0.12\textwidth]{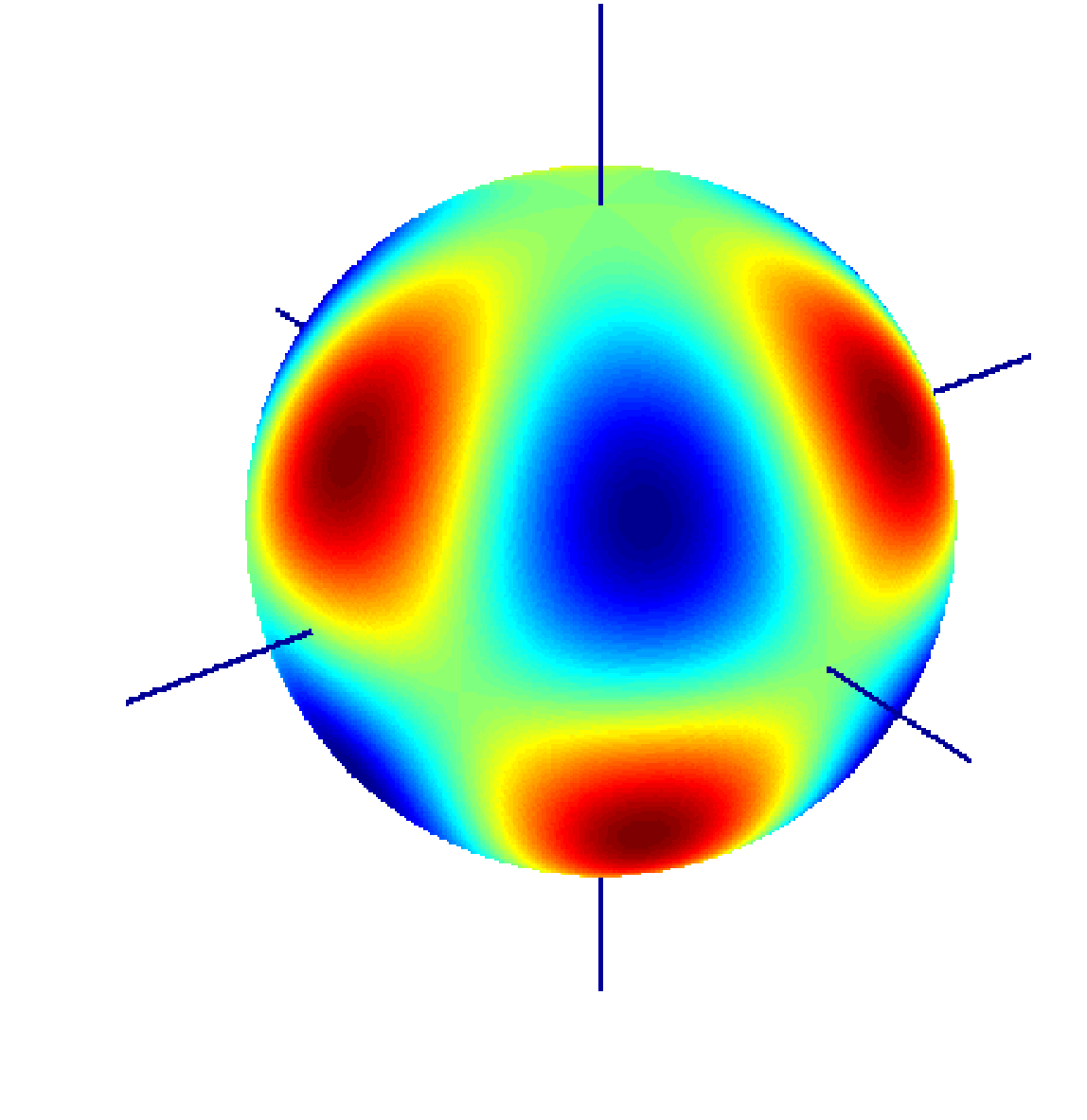}&
\includegraphics[width=0.12\textwidth]{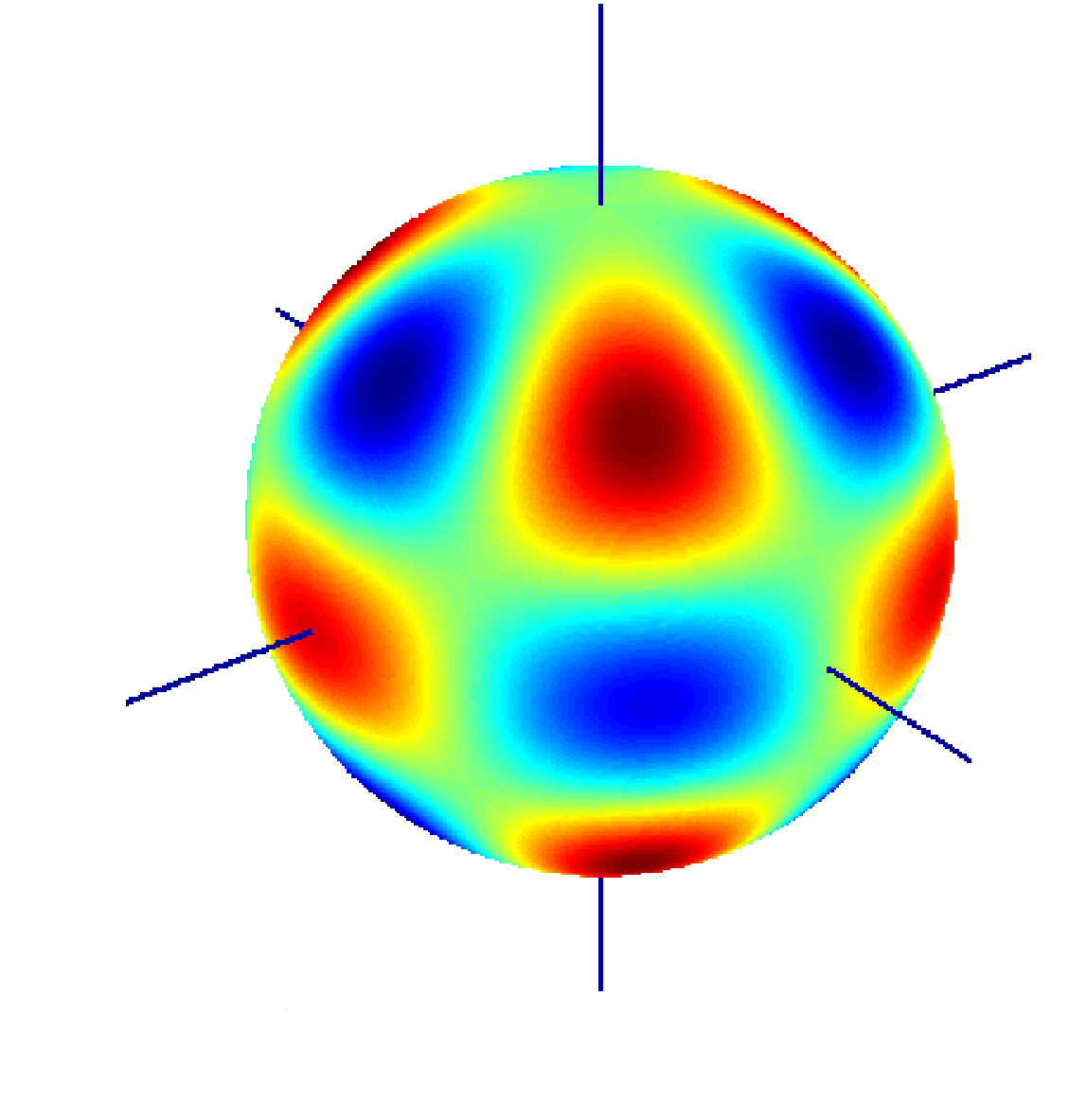}\\
$4$
&
&
&
&
\includegraphics[width=0.12\textwidth]{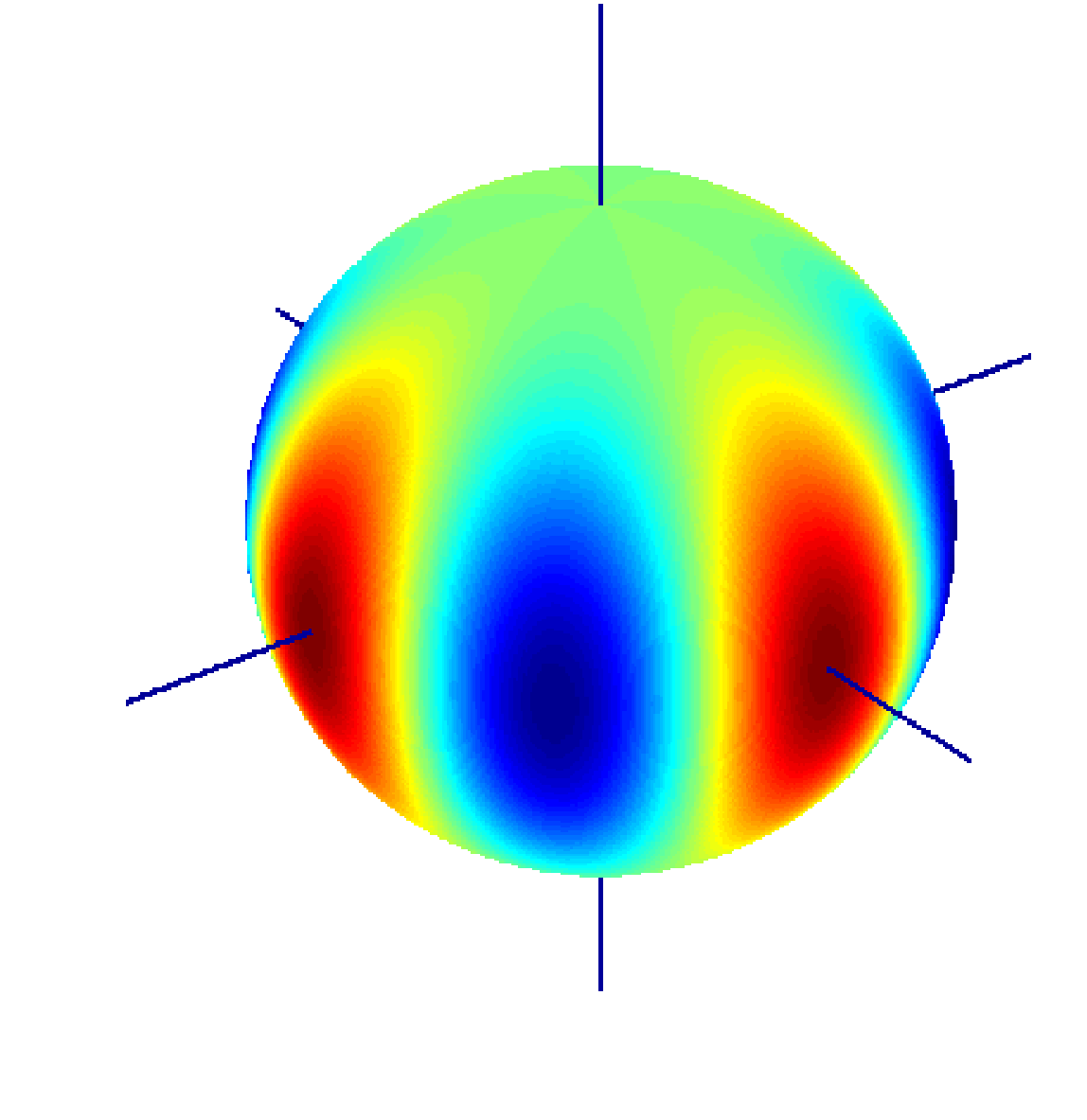}&
\includegraphics[width=0.12\textwidth]{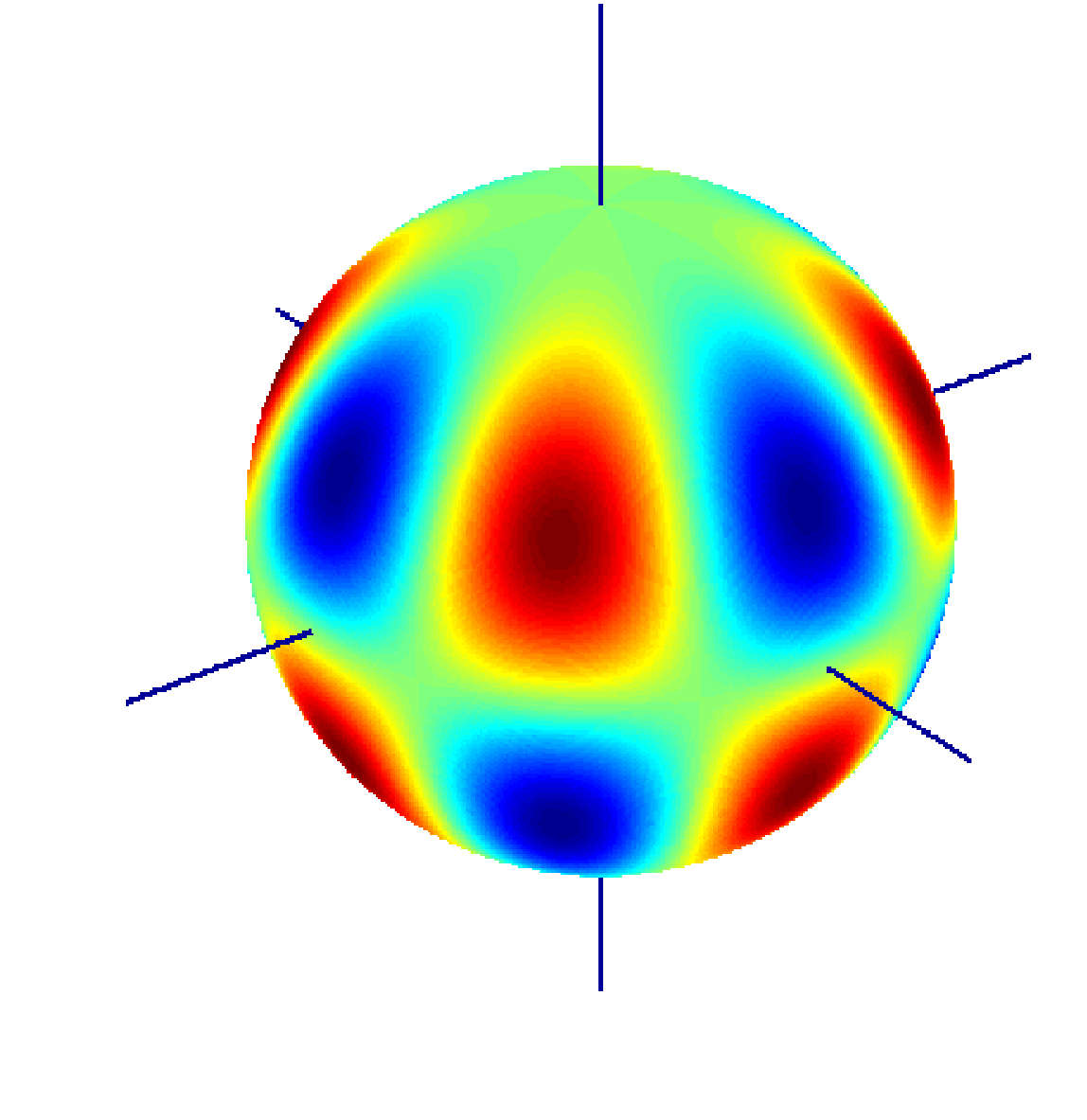}\\
$5$
&
&
&
&
&
\includegraphics[width=0.12\textwidth]{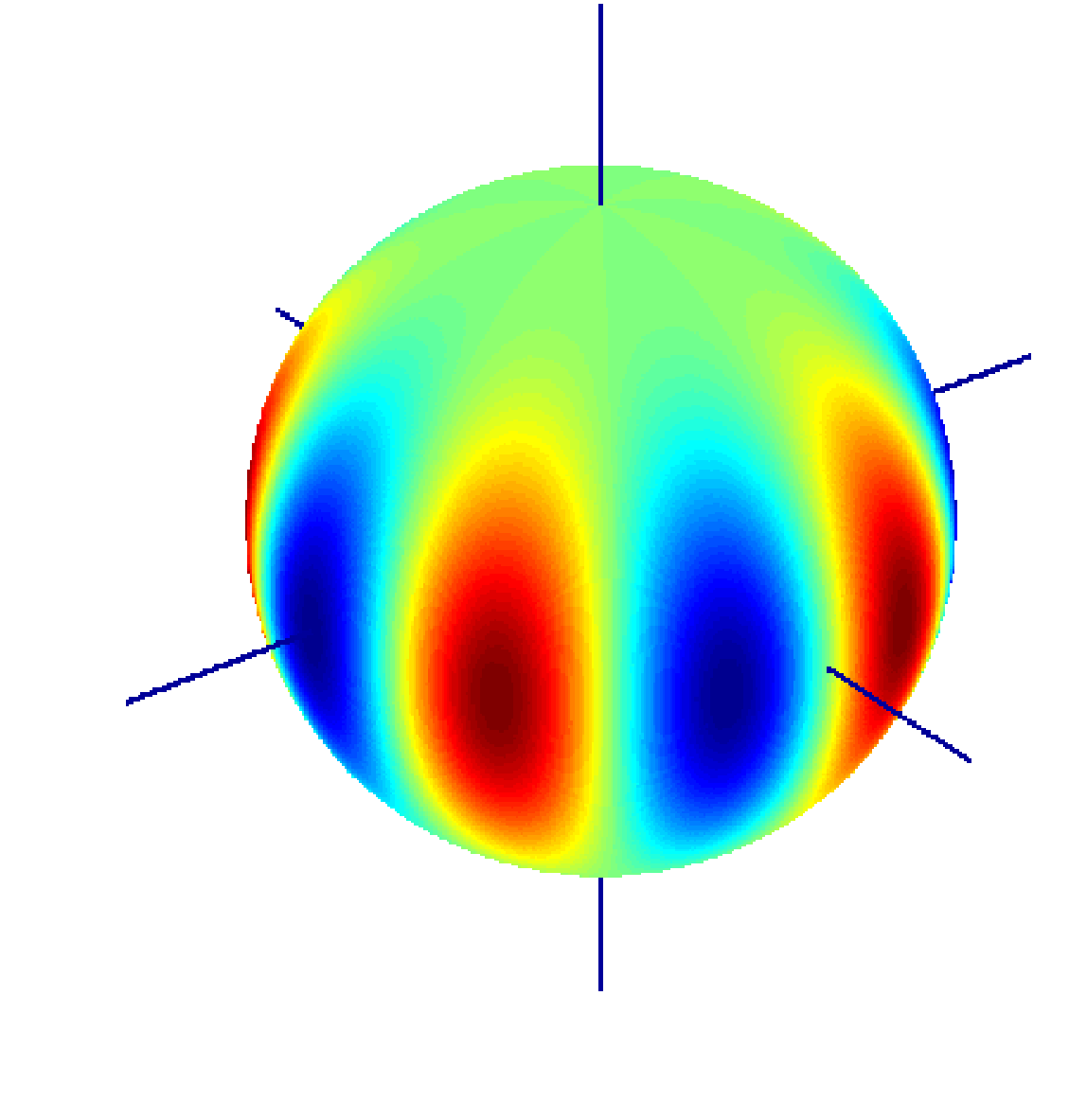}
\end{tabular}
\end{flushright}
\caption[Spherical Harmonic base functions (real part)]{\label{fig:feature:SHbase} Real part of the first 5 bands of the complex Spherical 
Harmonic base functions\footnotemark.}
\end{figure}

\index{Spherical Harmonics}\index{Harmonic expansion}
Spherical Harmonics (${\cal SH}$) \cite{sh} form an orthonormal base on the 2-sphere $S^2$. Analogical to the Fourier Transform,
any given real or complex valued, integrable function $f$ in some Hilbert space on a sphere with its parameterization over the angles 
$\Theta \in [0,\pi[$ and $\Phi \in [0,2\pi[$ (latitude and longitude of the sphere) can be represented by an expansion in its 
harmonic coefficients by: 
\begin{equation}
f(\Phi,\Theta)=\sum\limits_{l=0}^\infty\sum\limits^{m=l}_{m=-l} \widehat{f}^l_m Y_m^l(\Phi,\Theta),
\label{eq:feature:SHcoeff}
\end{equation}
where $l$ denotes the band of expansion, $m$ the order for the $l$-th band and $\widehat{f}^l_m$ the harmonic coefficients.
The harmonic base functions $Y_m^l(\Theta,\Phi)$ are calculated (using the standard normalized \cite{angular} formalization) as follows:
\begin{equation}
Y_m^l(\Phi,\Theta) = \sqrt{ \frac{2l+1}{4\pi}\frac{(l-m)!}{(l+m)!}} \cdot P_m^l(\cos \Theta)\mathrm{e}^{im\Phi},
\label{eq:feature:SHbase}
\end{equation}
where $P_m^l$ is the associated Legendre polynomial (see \ref{sec:feature:legendrepoly}). Fig. \ref{fig:feature:SHbase} illustrates
the $Y_m^l$ base functions of the first few bands.\\
The harmonic expansion of a function $f$ will be denoted by $\widehat{f}$ with corresponding coefficients $\widehat{f}^l_m$.
We define the forward Spherical Harmonic transformation as:
\begin{equation}
{\cal SH}(f) := \widehat{f}, \text{\quad with \quad} \widehat{f}^l_m = \int\limits_{\Phi,\Theta} \overline{Y_m^l}(\Phi,\Theta)f(\Phi,\Theta)
\sin{\Theta}d\Phi d\Theta,
\label{eq:feature:SHforward}
\end{equation}
where $\widehat{x}$ denotes the complex conjugate, and the backward transformation accordingly:
\begin{equation}
{\cal SH}^{-1}(\widehat{f})(\Phi,\Theta) := \sum\limits_{l=0}^\infty\sum\limits^{m=l}_{m=-l} \widehat{f}^l_m Y_m^l(\Phi,\Theta).  
\label{eq:feature:SHbackward}
\end{equation}
\subsection{\label{sec:feature:legendrepoly}Associated Legendre Polynomials}
\begin{figure}[htbp]
\centering
\psfrag{data1}{\footnotesize$P^0_0(x)$}
\psfrag{data2}{\footnotesize$P^1_0(x)$}
\psfrag{data3}{\footnotesize$P^2_0(x)$}
\psfrag{data4}{\footnotesize$P^3_0(x)$}
\psfrag{data5}{\footnotesize$P^4_0(x)$}
\psfrag{data6}{\footnotesize$P^5_0(x)$}
\includegraphics[width=0.8\textwidth]{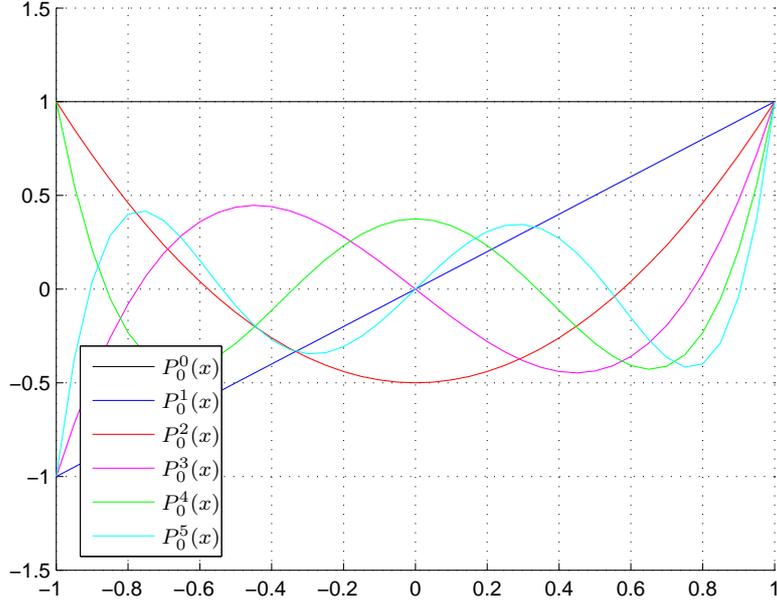}
\caption[Associated Legendre polynomials]{\label{fig:feature:legendre} A few sample Associated Legendre polynomials of the 0th order which are equal to the General Legendre polynomials.}
\end{figure}

\index{Associated Legendre polynomial}
Associated Legendre polynomials $P^l_m(x)$ are derived as the canonical solution of the General Legendre differential 
equation \cite{angular}: 
\begin{equation}
\left((1-x^2)y'\right)+\left(l(l+1) - \frac{(m^2)}{1-x^2}\right)y = 0,
\end{equation}
which plays an important role for the solution of many well known problems such as the Laplace equation \cite{angular} in our case.
For integer values of $-l \leq m \leq l$,  
\begin{equation}
P^l_m(x) = \frac{(-1)^m}{2^ll!}(1-x^2)^{m/2}\frac{d^{l+m}}{dx^{l+m}}(x^2-1)^l
\label{eq:feature:asslegendere}
\end{equation}
has non-singular solutions in $[-1,1]$. The Associated Legendre polynomials are linked to the General Legendre polynomials by:
\begin{equation}
P^l_m(x) = (-1)^m(1-x^2)^{m/2}\frac{d^{m}}{dx^{m}}(P^l(x)),
\label{eq:feature:legendere}
\end{equation}
which implies that $P^l_0(x) = P^l(x)$ - as shown in Fig. \ref{fig:feature:legendre}. 
\paragraph{Properties:} 
Two main properties of the Associated Legendre polynomials in context of this work are the orthogonality of the $P^l_m(x)$ \cite{angular} 
as well as the symmetry property:
\begin{equation}
P^l_{-m} = (-1)^m\frac{(l-m)!)}{(l+m)!} P^l_{m}.
\end{equation}
Another notable fact is that in contrast to its name, the $P^l_m(x)$ are actually only polynomials if $m$ has a even integer value.

\subsection{\label{sec:feature:deriveSH}Deriving Spherical Harmonics}
We give a brief sketch of how Spherical Harmonics have been derived in literature \cite{angular}\cite{sh} focusing on some aspects which 
are useful for our purposes. For more details please refer to \cite{angular} or \cite{sh}.\\
Given a function $f$ parameterized in $\Phi, \Theta$ on $S^2$, its  Laplacian is: 
\begin{equation}
\nabla^2\Phi = \frac{\partial^2 f}{\partial\Theta^2}+\cot\Theta\frac{\partial f}{\partial\Theta}+\csc^2\Theta\frac{\partial^2 f}{\partial\Phi^2}.  
\end{equation}
A solution to the partial differential equation
\begin{equation}
\frac{\partial^2 f}{\partial\Theta^2}+\cot\Theta\frac{\partial f}{\partial\Theta}+\csc^2\Theta\frac{\partial^2 f}{\partial\Phi^2} + \lambda f = 0
\end{equation}
can be obtained \cite{sh} by separation into $\Phi$-dependent parts
\begin{eqnarray}
\sin(m\Phi)&& \text{\quad for \quad} m < 0 \nonumber\\
\cos(m\Phi)&& \text{\quad else}
\end{eqnarray}
and $\Theta$-dependent parts
\begin{equation}
\frac{d^2y}{d\Theta^2}+\cot\Theta\frac{dy}{d\Theta}+\left(\lambda-\frac{m^2}{\sin^2\Theta}\right) y = 0,
\end{equation}
with solutions given by $P^l_m(\cos(\Theta))$ (section \ref{sec:feature:legendrepoly}) for the integer valued $m\geq 0$ and $\lambda=l(l+1)$.
Rewriting the $\Phi$-dependent parts in exponential notation and adding the normalization to $\sum|Y^l_m|^2=1$ \cite{angular}, 
we obtain the Spherical 
Harmonics:
\begin{equation}
Y_m^l(\Phi,\Theta) := \sqrt{ \frac{2l+1}{4\pi}\frac{(l-m)!}{(l+m)!}} \cdot P_m^l(\cos \Theta)\mathrm{e}^{im\Phi}.
\label{eq:feature:SHbasederive}
\end{equation}

\footnotetext{Thanks to O. Ronneberger for the MATLAB visualization}

\begin{figure}[htbp]
\index{Base functions}\index{Spherical Harmonics}
\begin{flushleft}
\begin{tabular}{ccccc|c}
$l=5$ & $l=4$ & $l=3$ & $l=2$ & $l=1$&$m$\\
\hline
\includegraphics[width=0.12\textwidth]{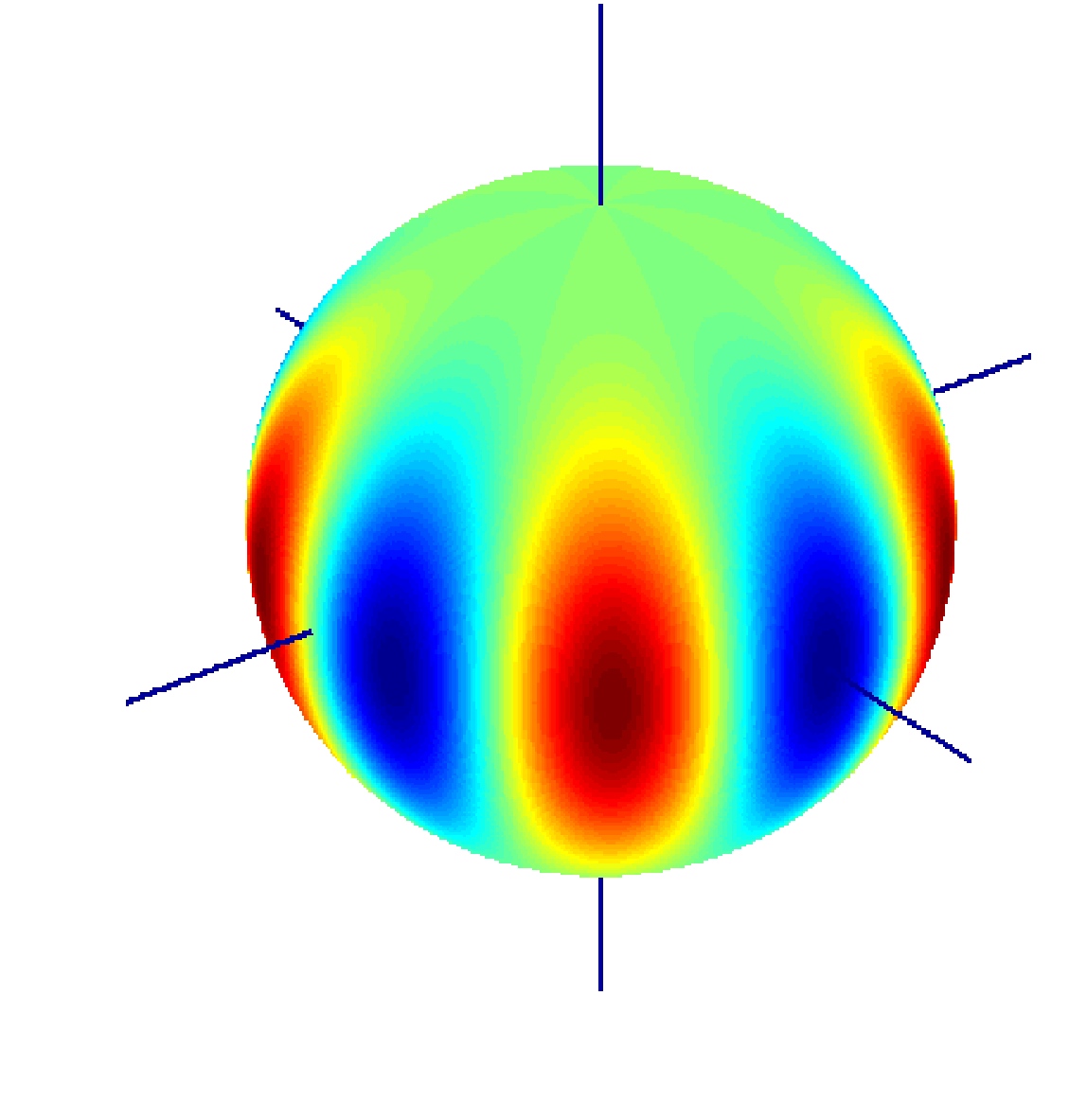}&
&
&
&
&
$-5$\\
\includegraphics[width=0.12\textwidth]{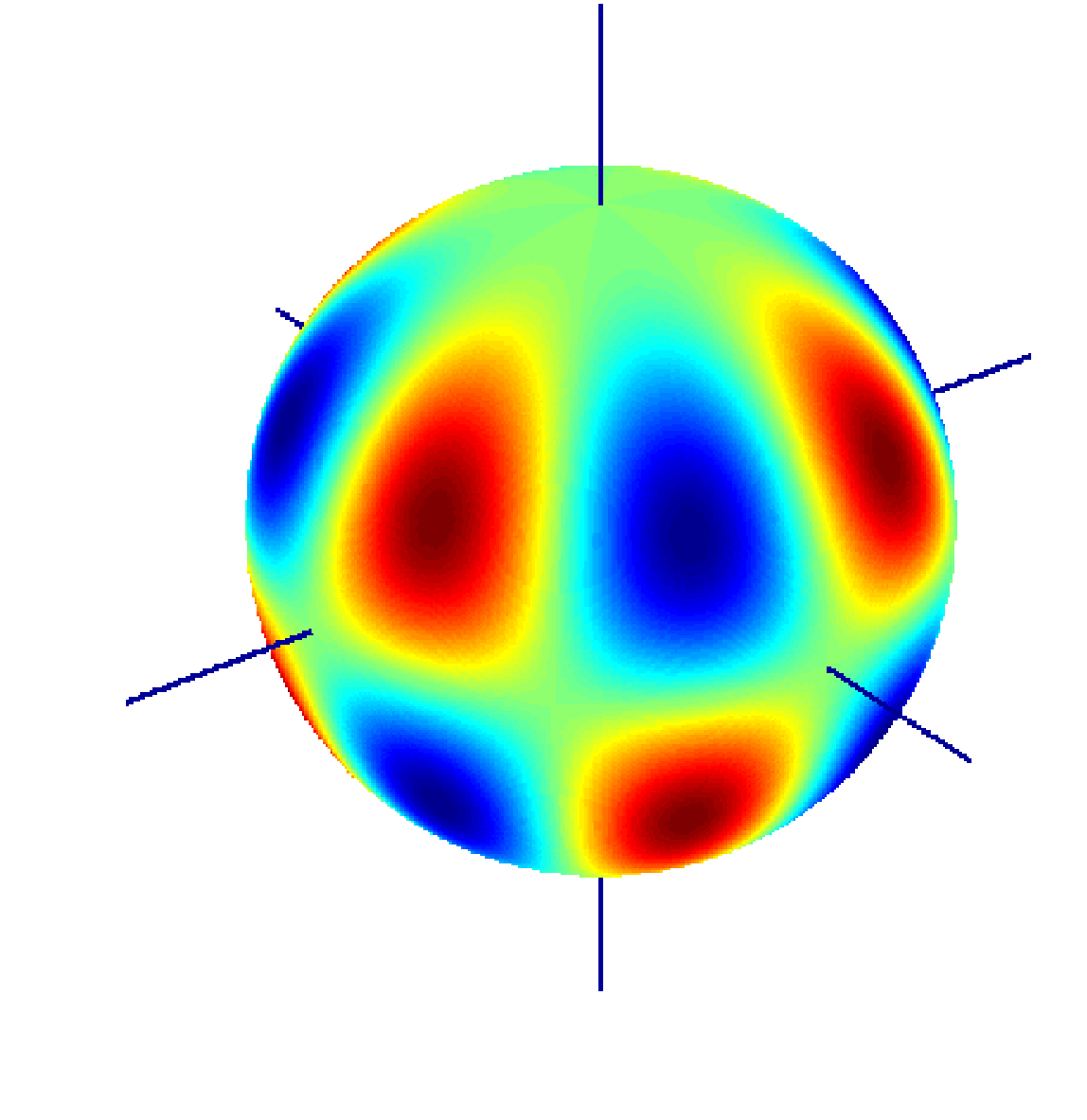}&
\includegraphics[width=0.12\textwidth]{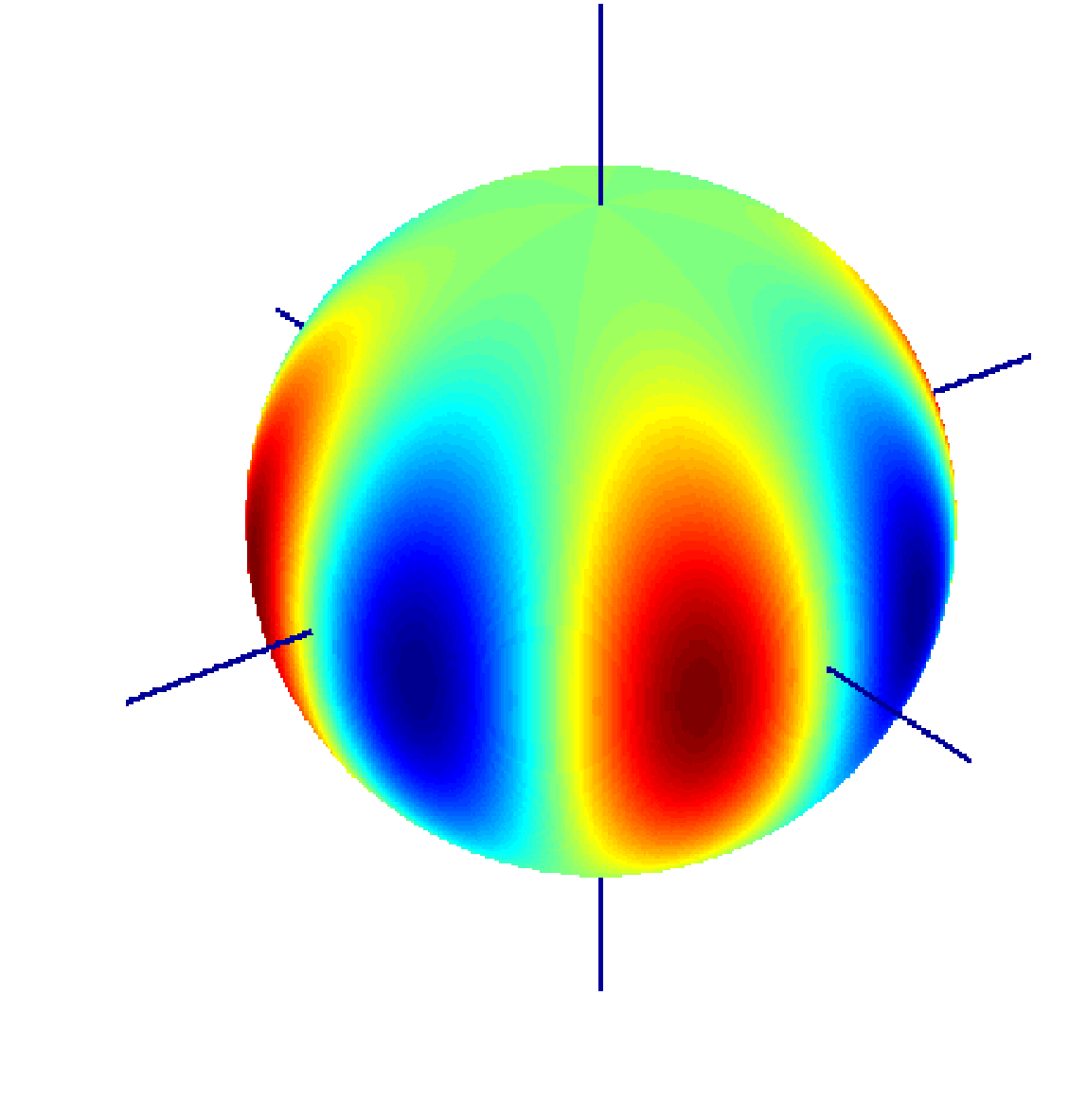}&
&
&
&
$-4$\\
\includegraphics[width=0.12\textwidth]{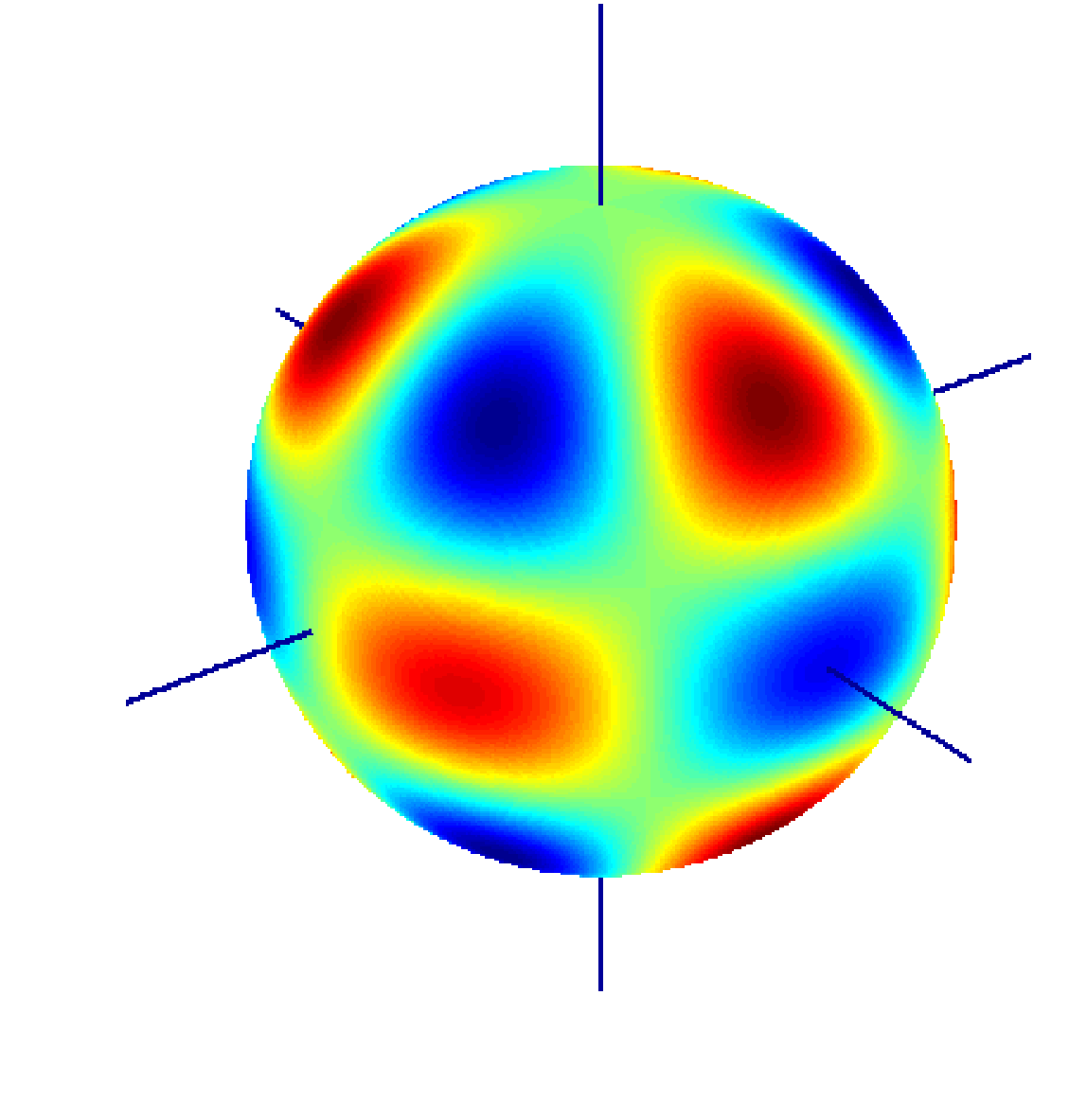}&
\includegraphics[width=0.12\textwidth]{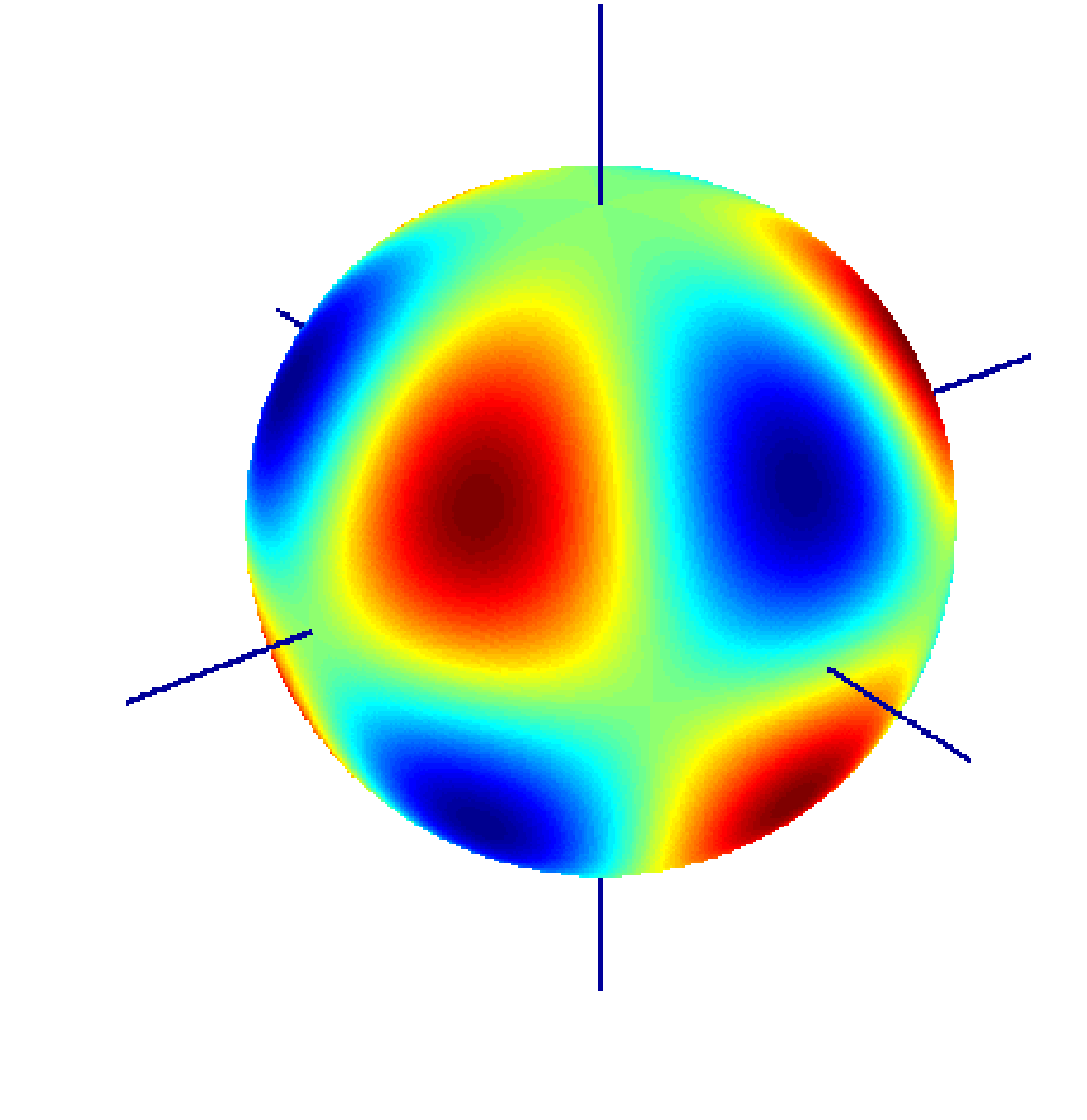}&
\includegraphics[width=0.12\textwidth]{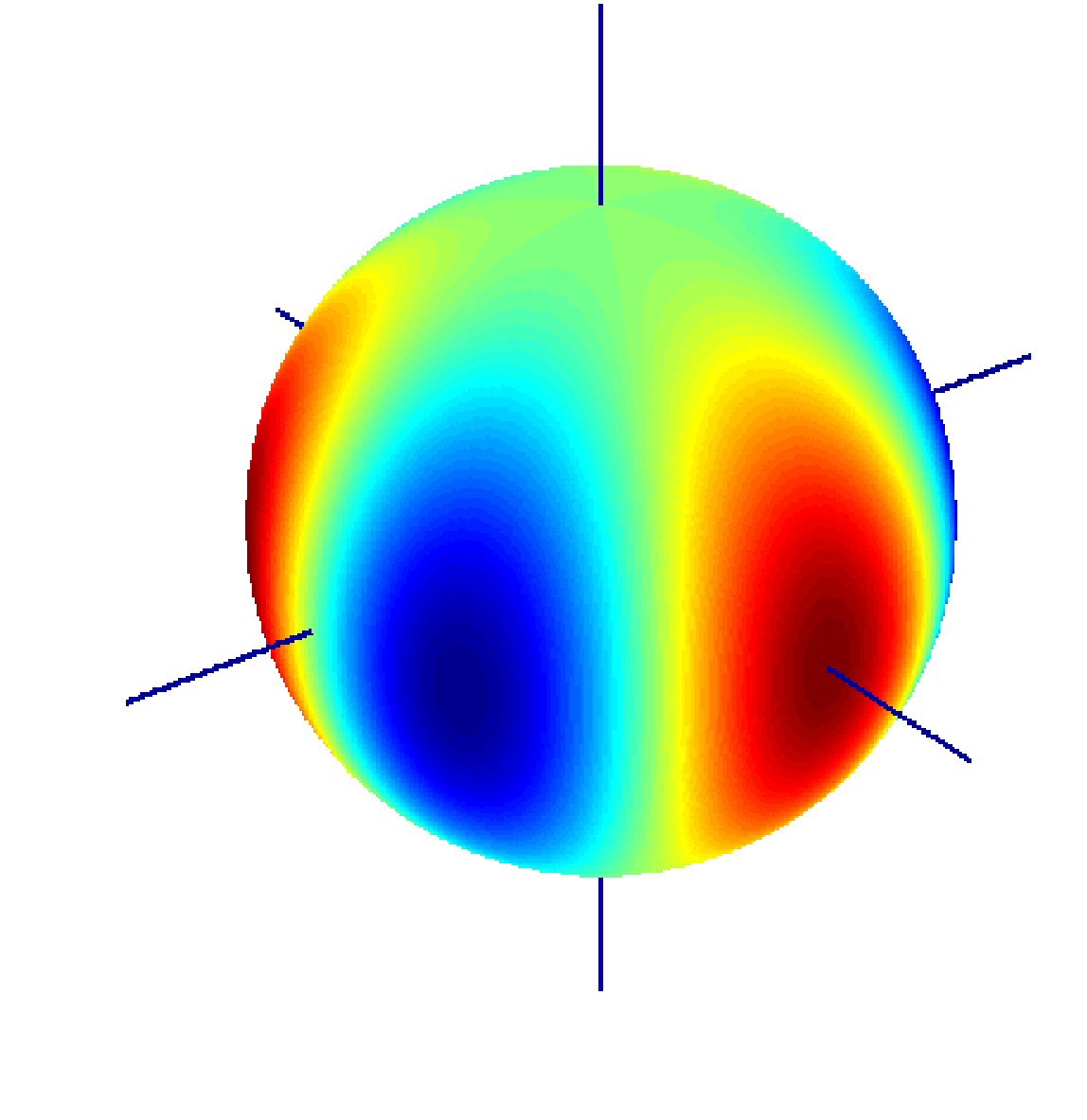}&
&
&
$-3$\\
\includegraphics[width=0.12\textwidth]{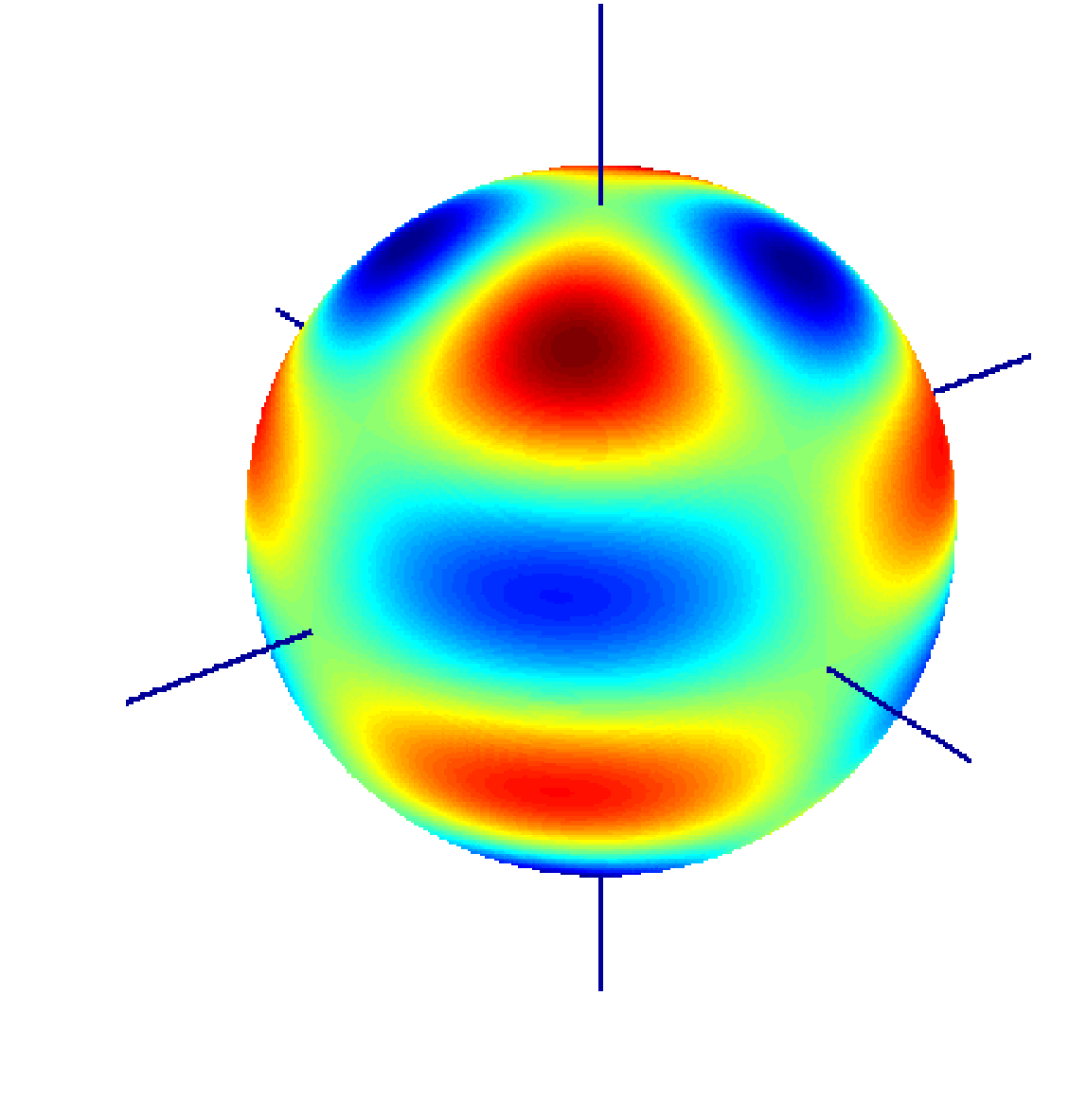}&
\includegraphics[width=0.12\textwidth]{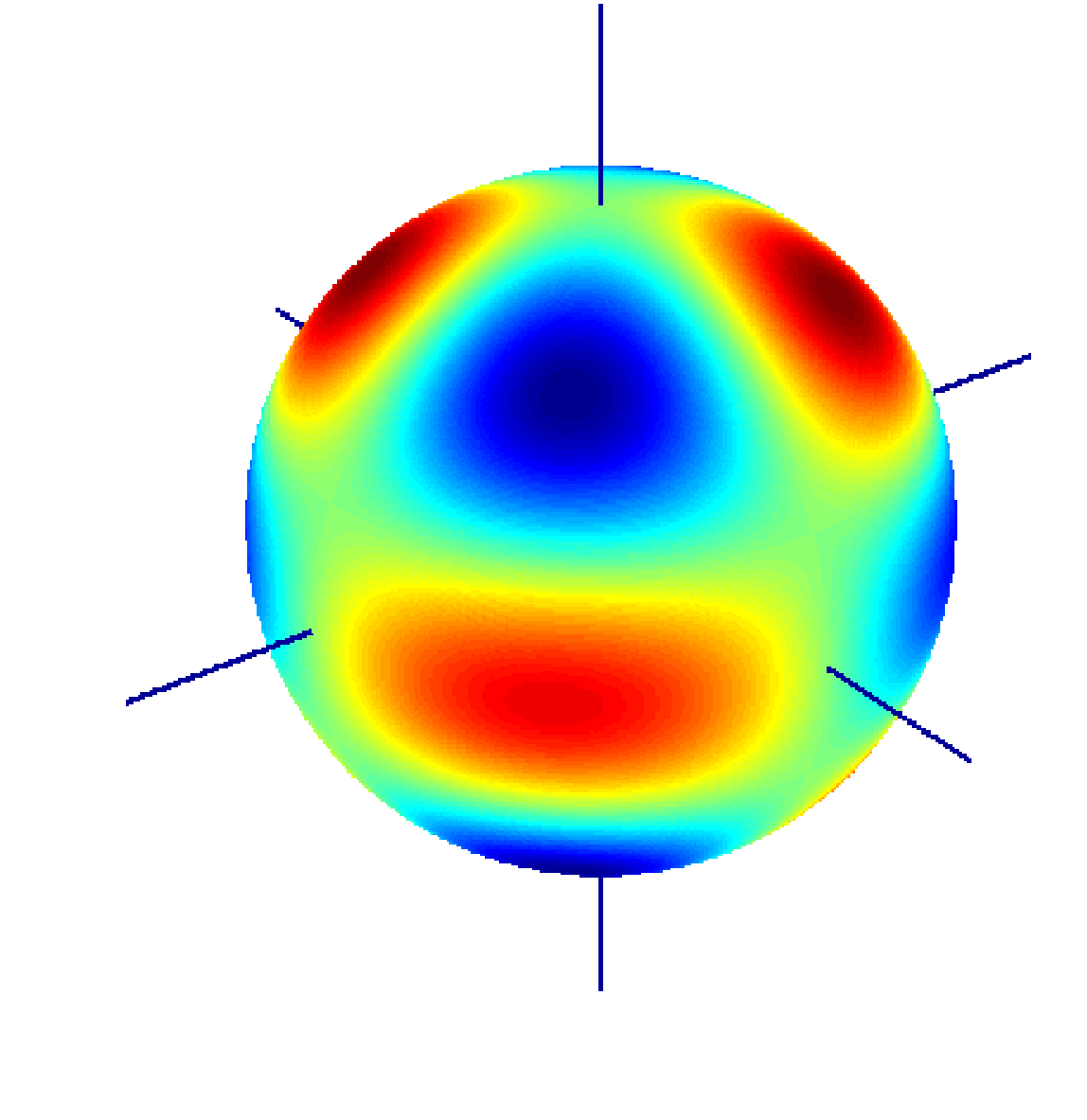}&
\includegraphics[width=0.12\textwidth]{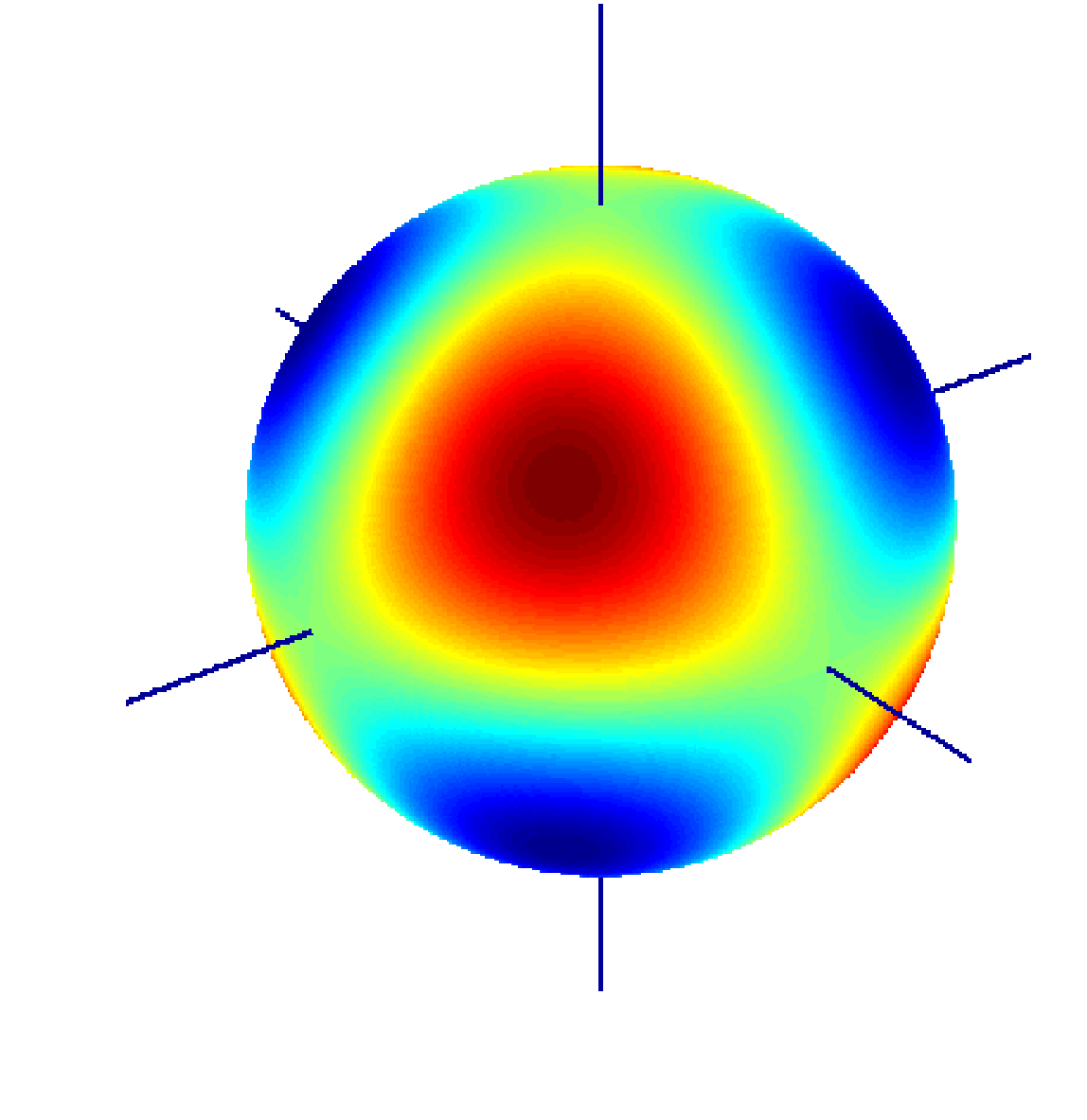}&
\includegraphics[width=0.12\textwidth]{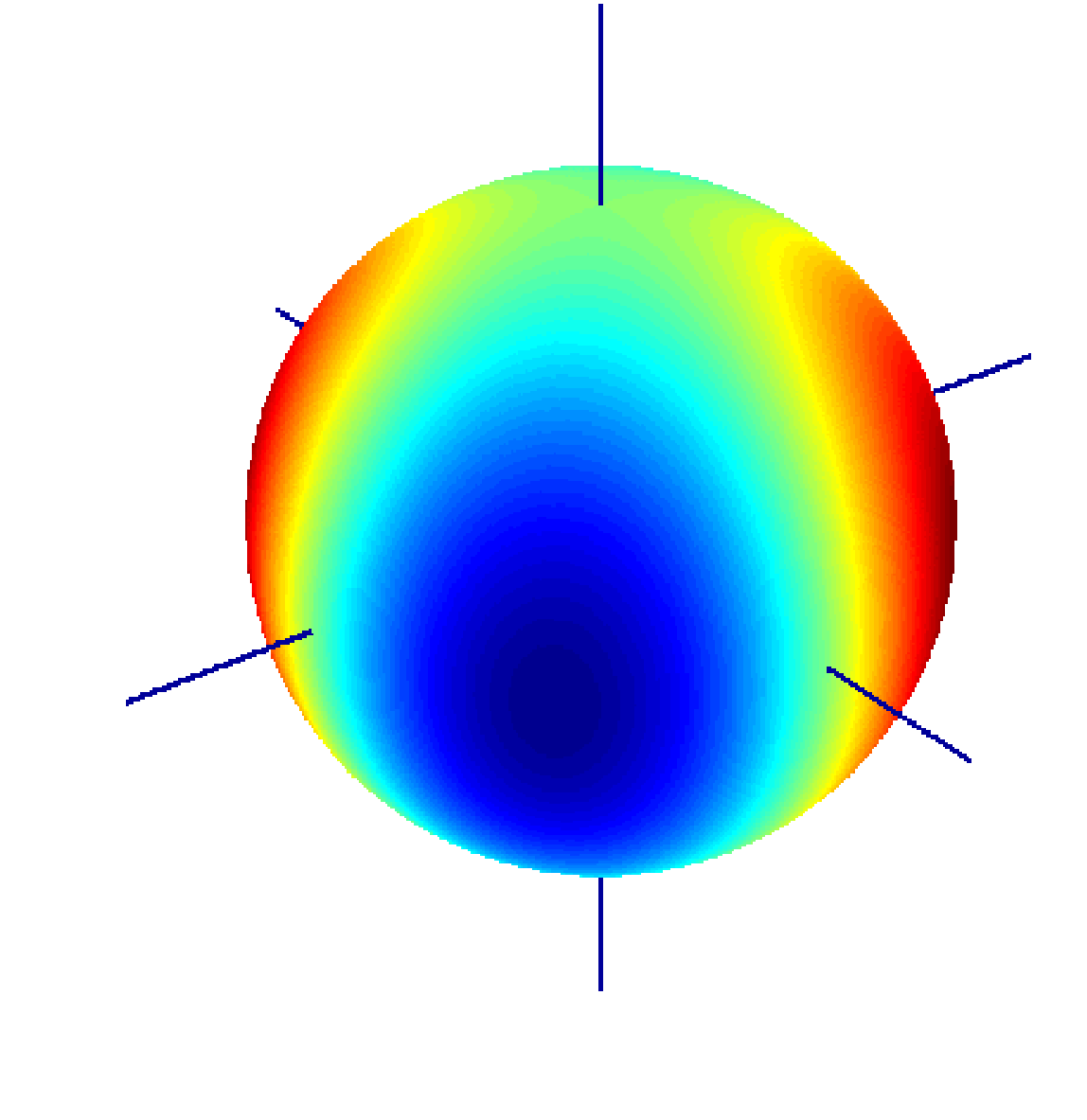}&
&
$-2$\\
\includegraphics[width=0.12\textwidth]{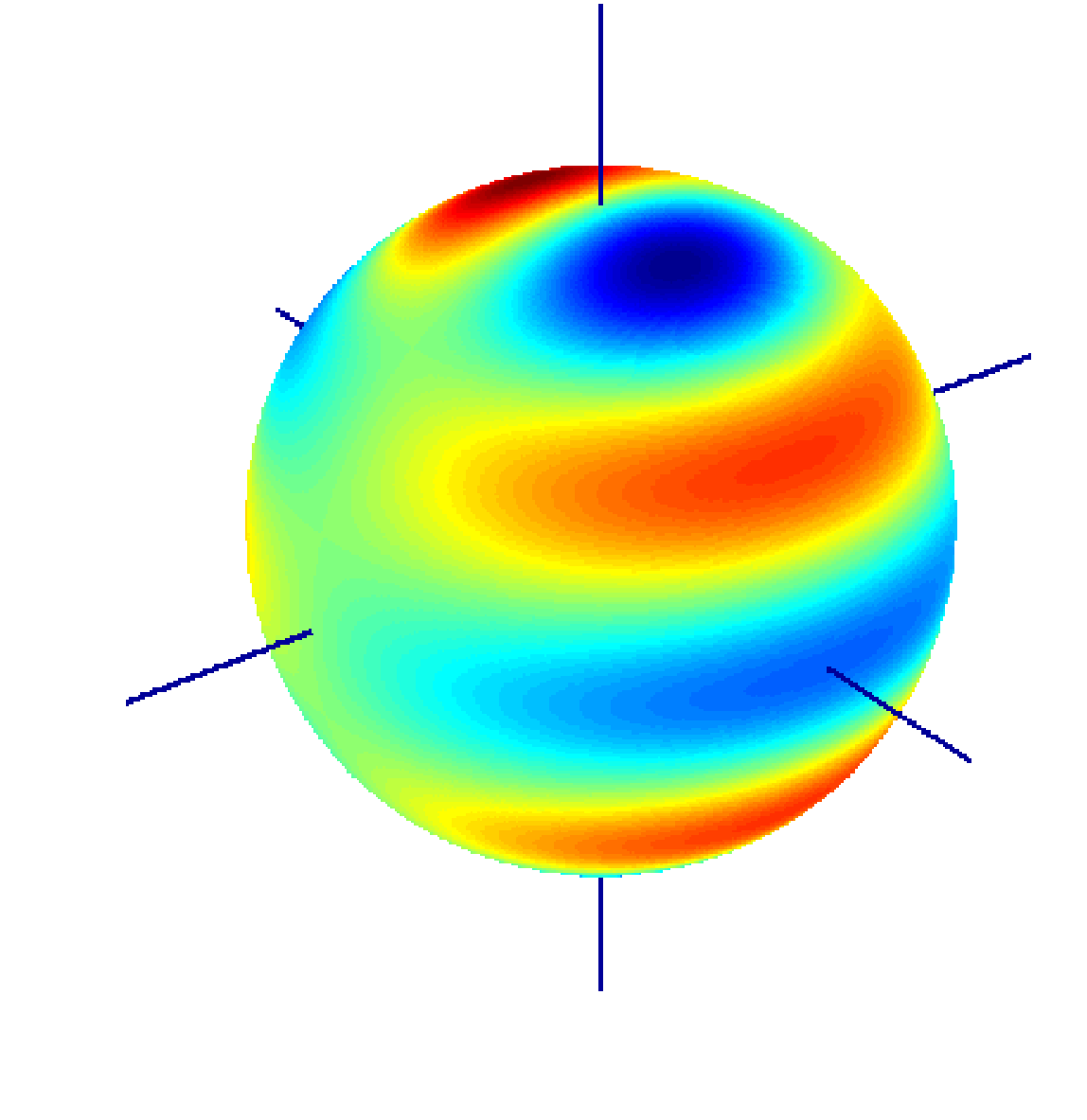}&
\includegraphics[width=0.12\textwidth]{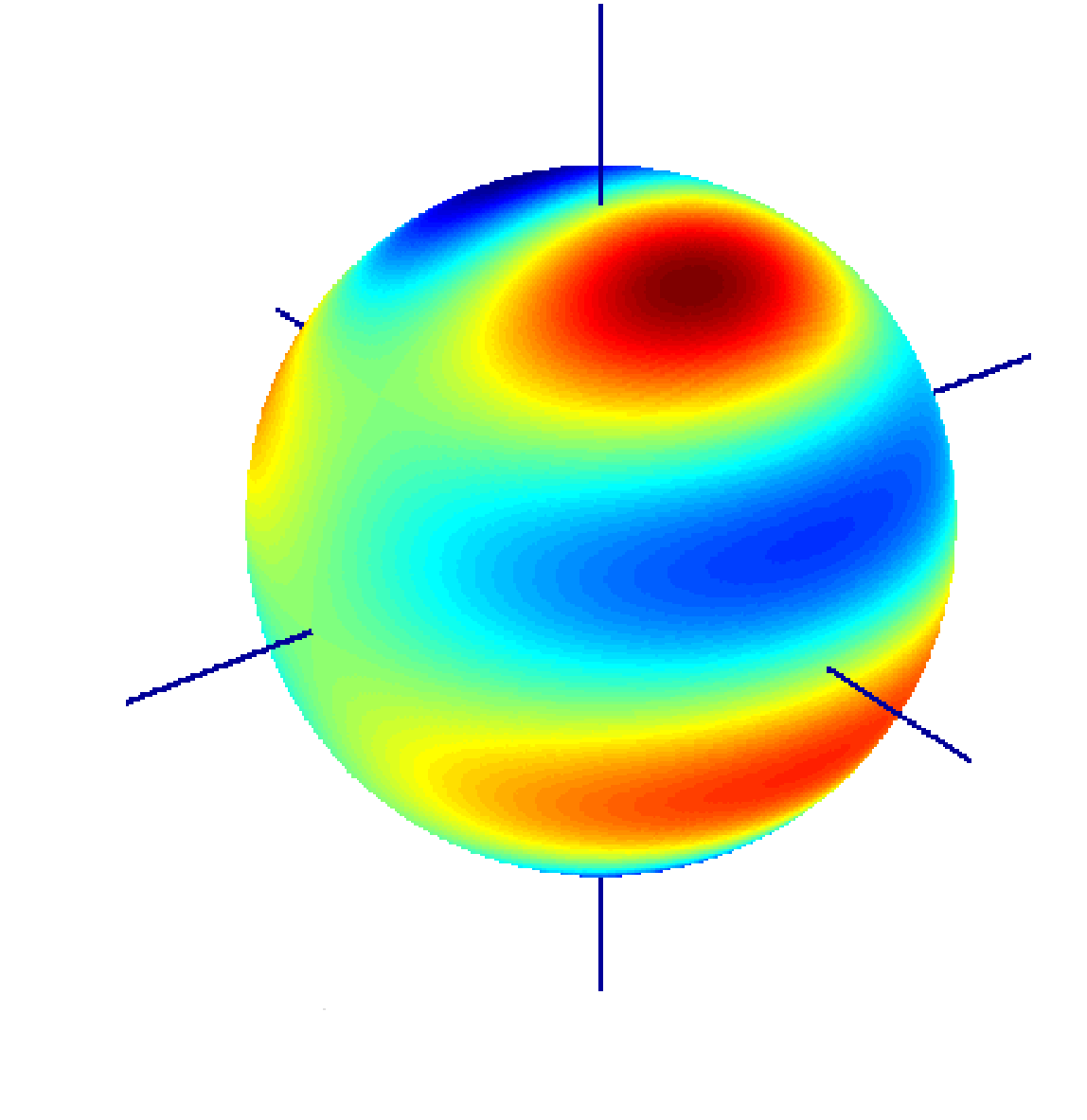}&
\includegraphics[width=0.12\textwidth]{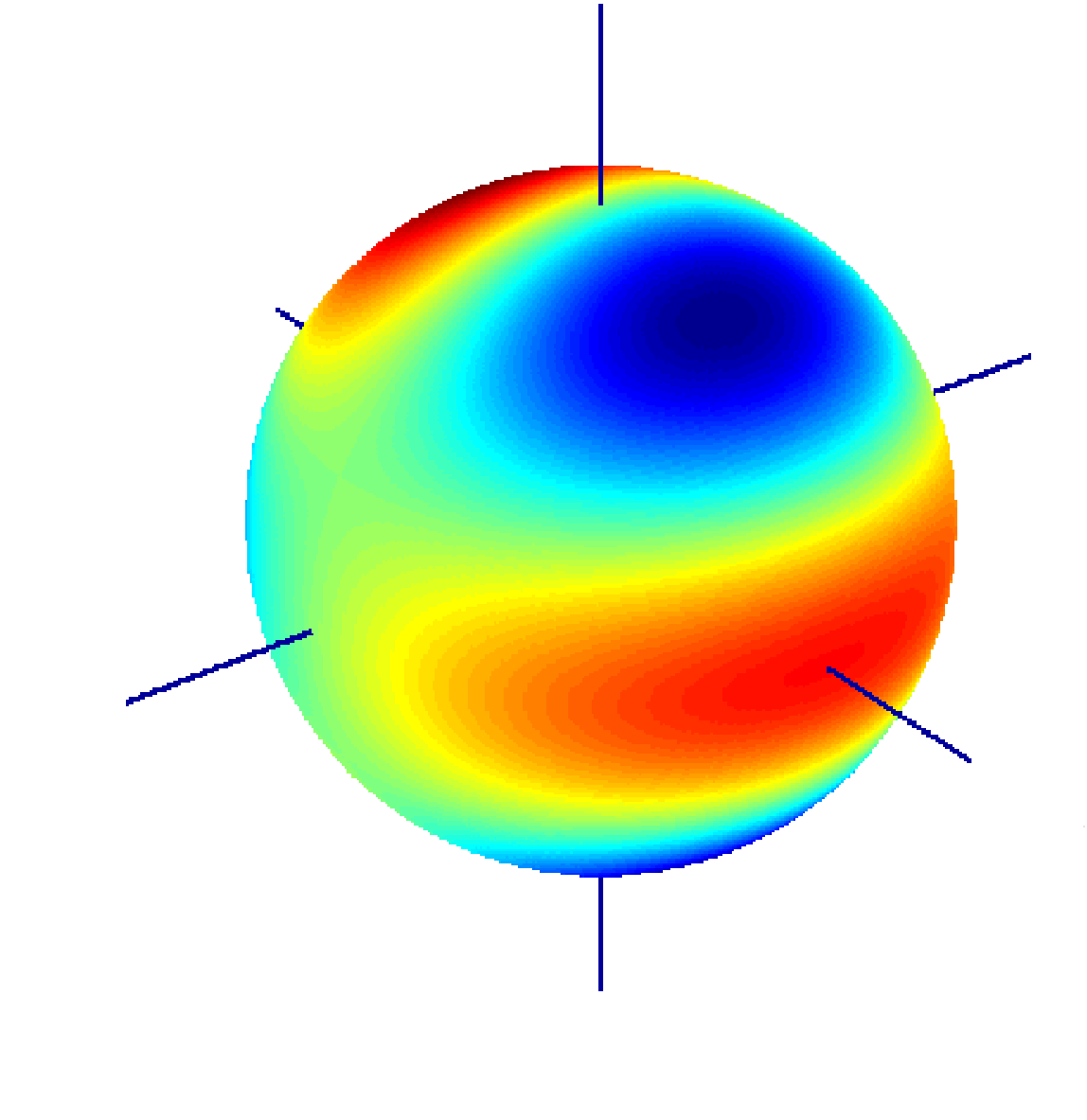}&
\includegraphics[width=0.12\textwidth]{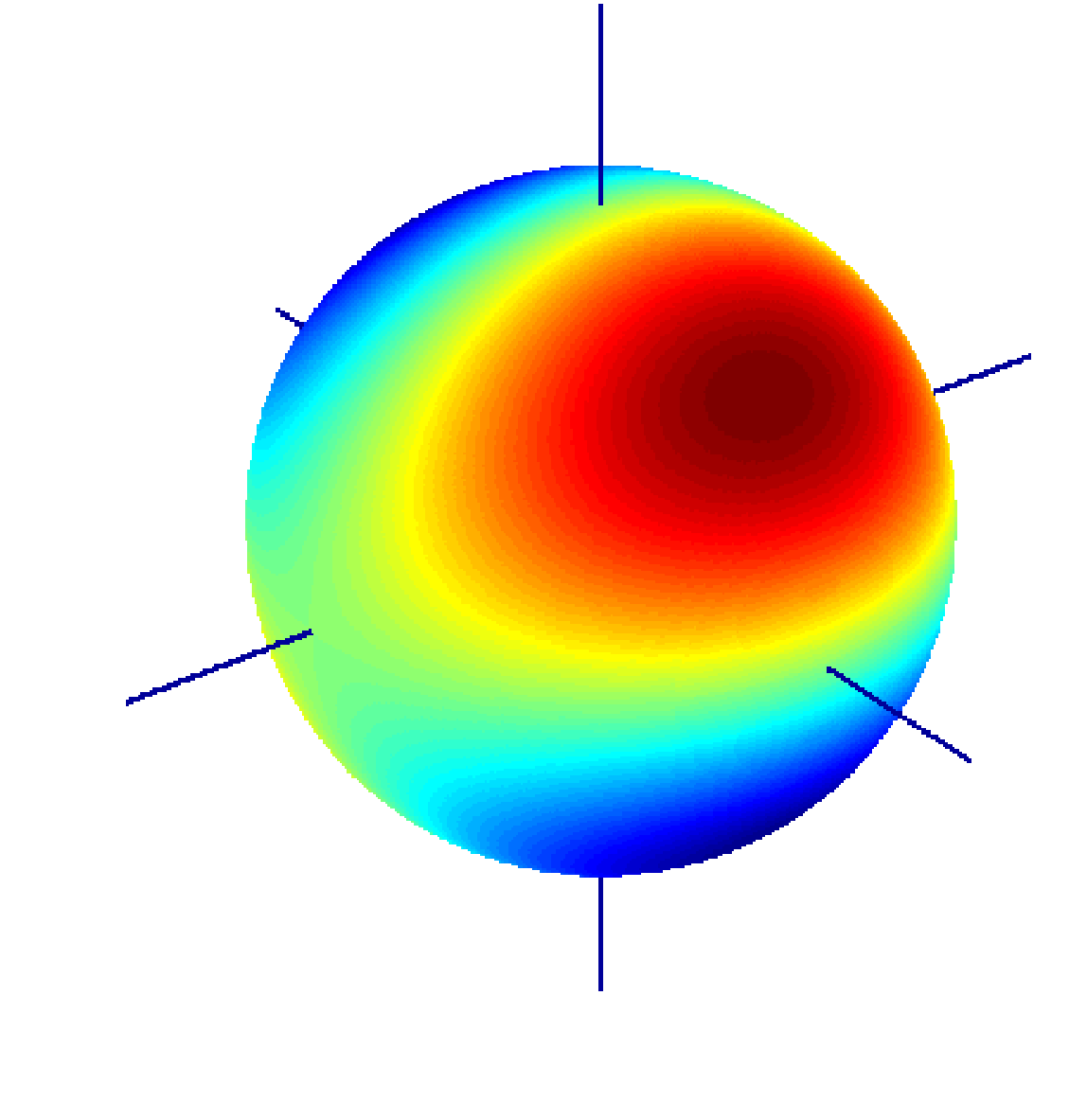}&
\includegraphics[width=0.12\textwidth]{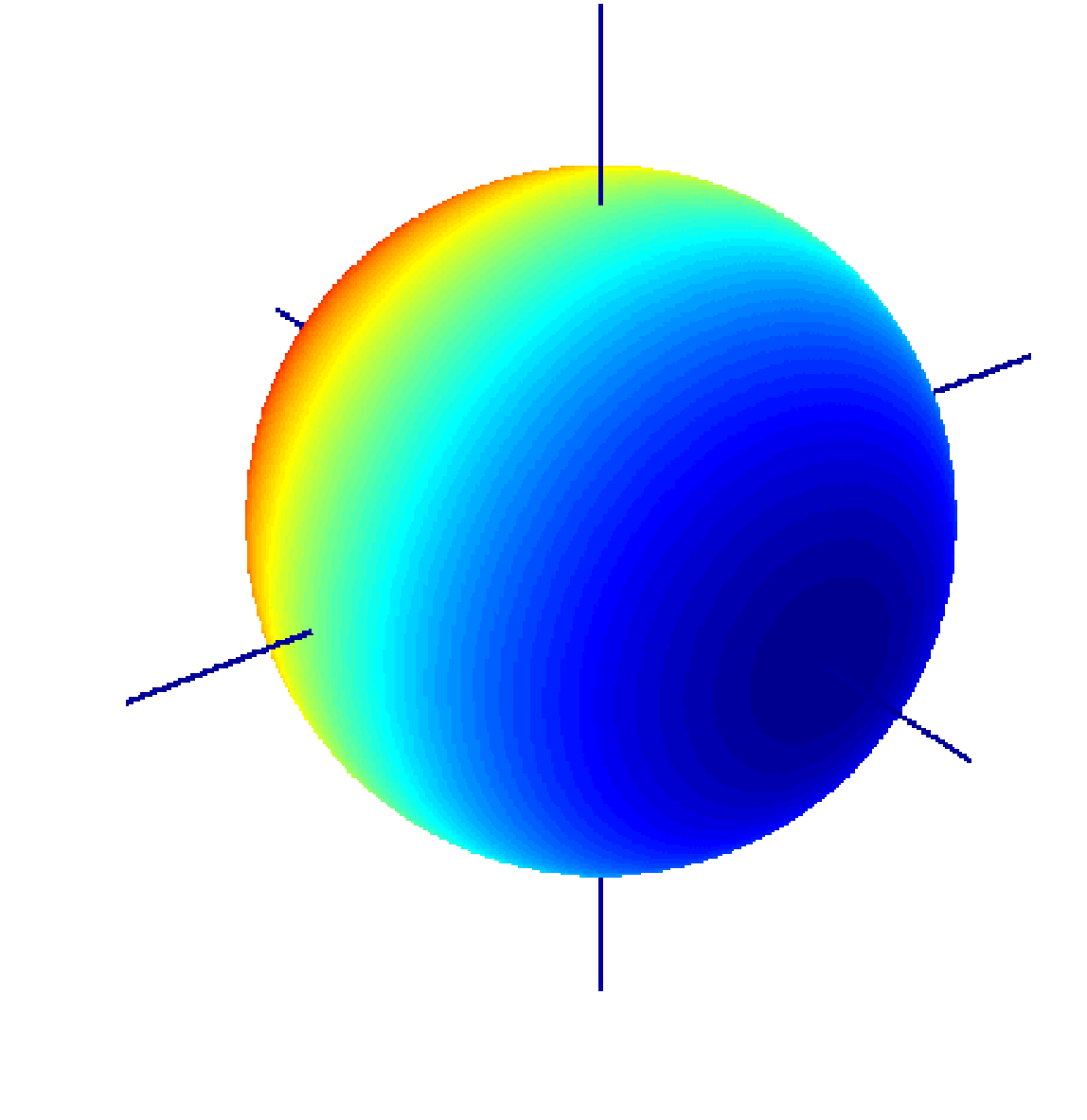}&
$-1$\\
&
&
&
&
\includegraphics[width=0.12\textwidth]{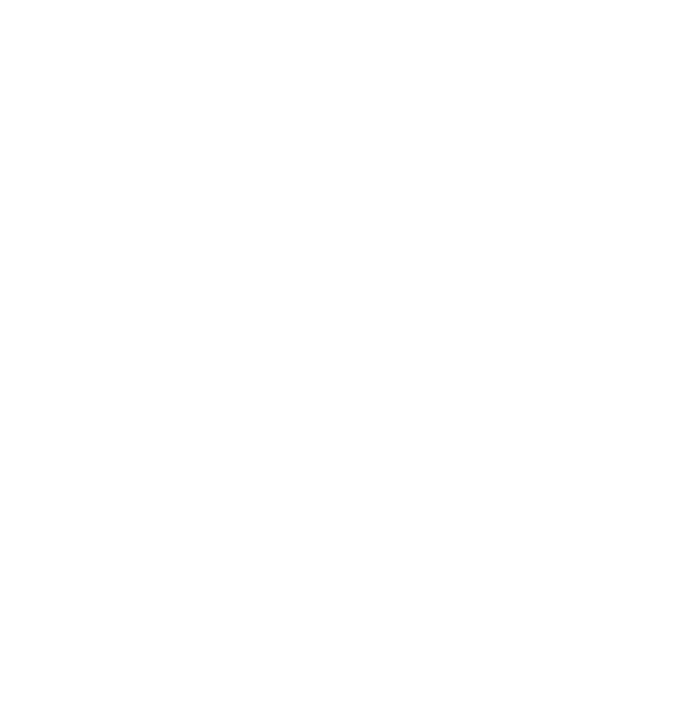}&
$0$\\
\includegraphics[width=0.12\textwidth]{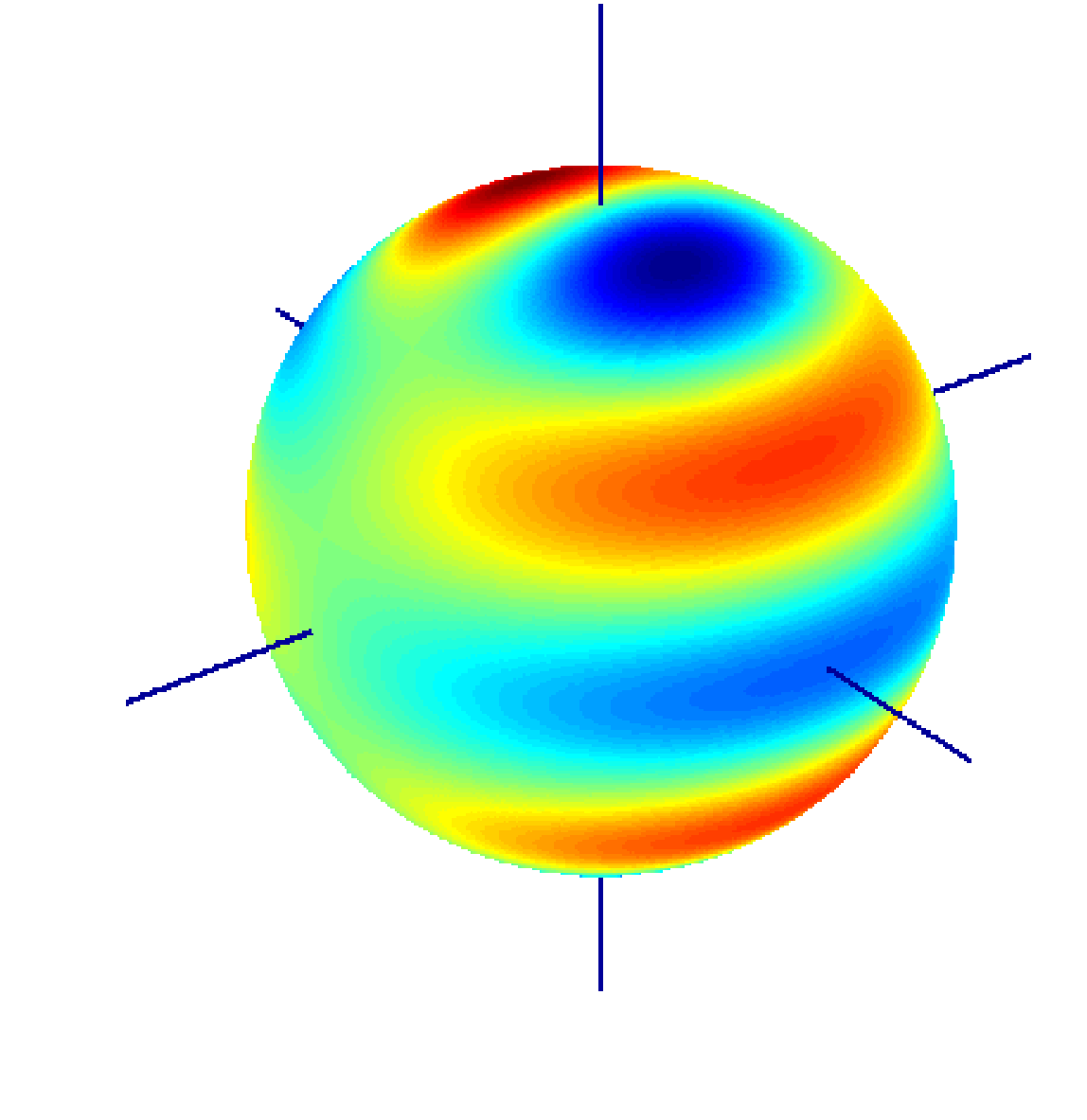}&
\includegraphics[width=0.12\textwidth]{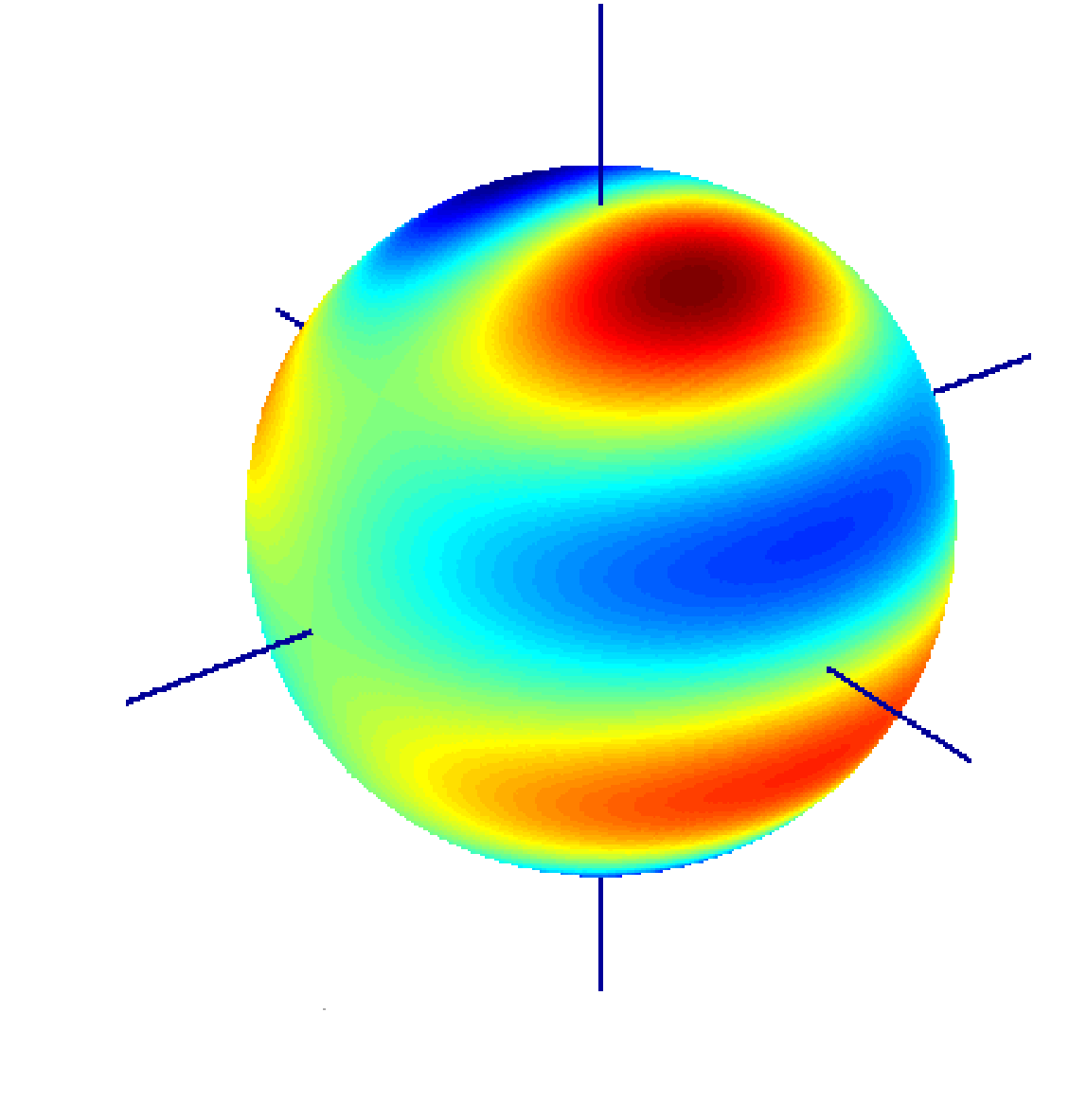}&
\includegraphics[width=0.12\textwidth]{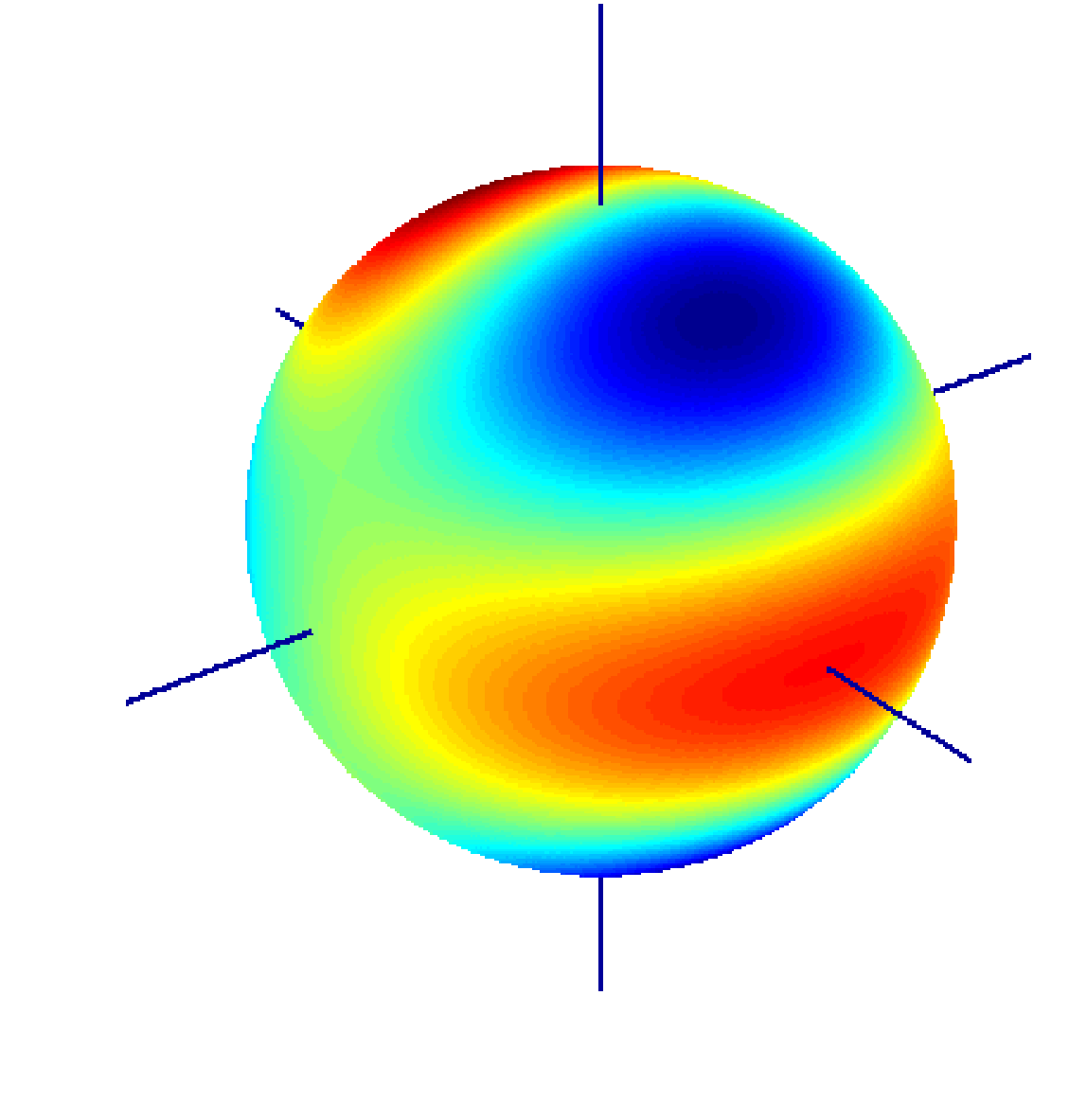}&
\includegraphics[width=0.12\textwidth]{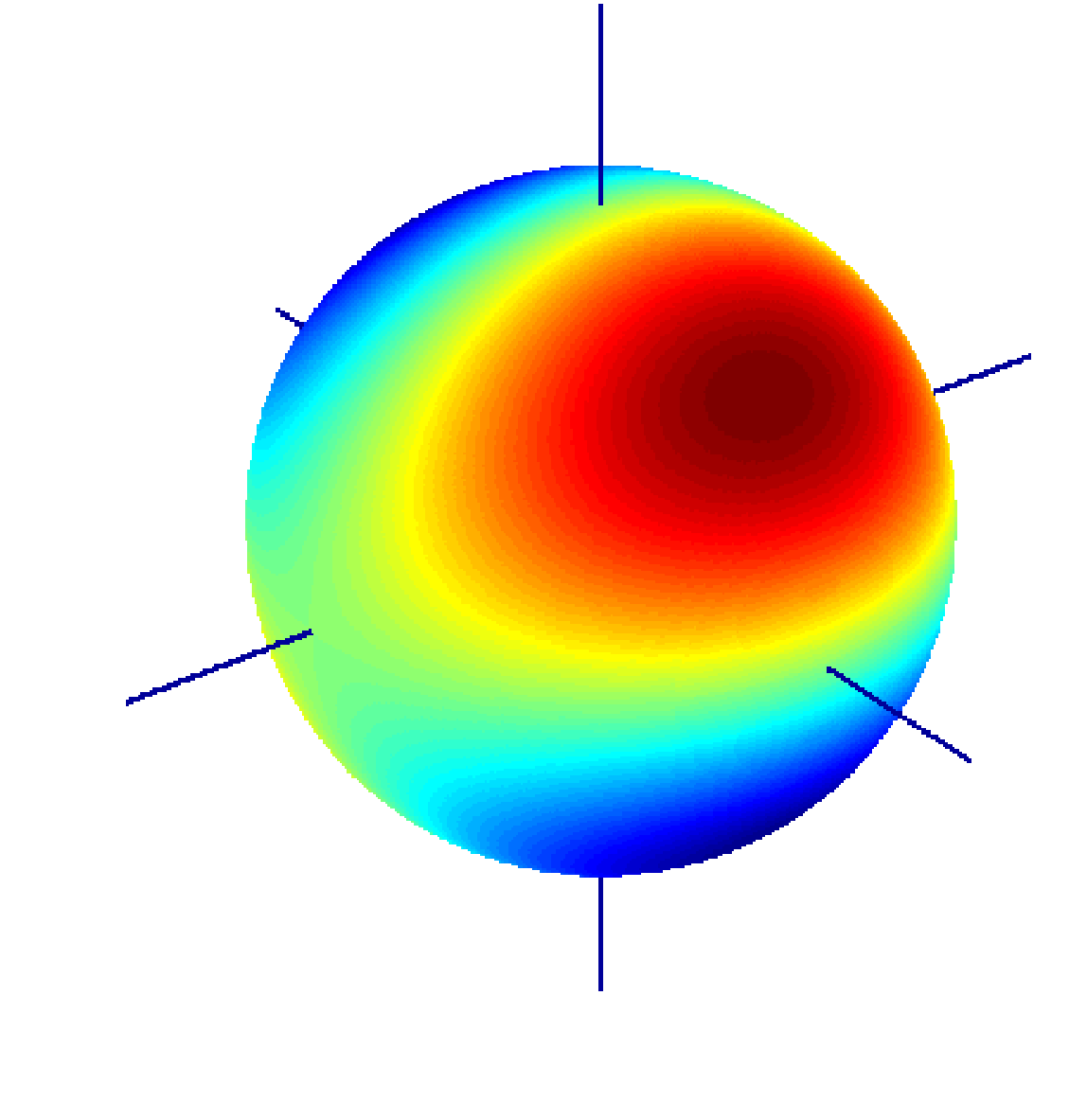}&
\includegraphics[width=0.12\textwidth]{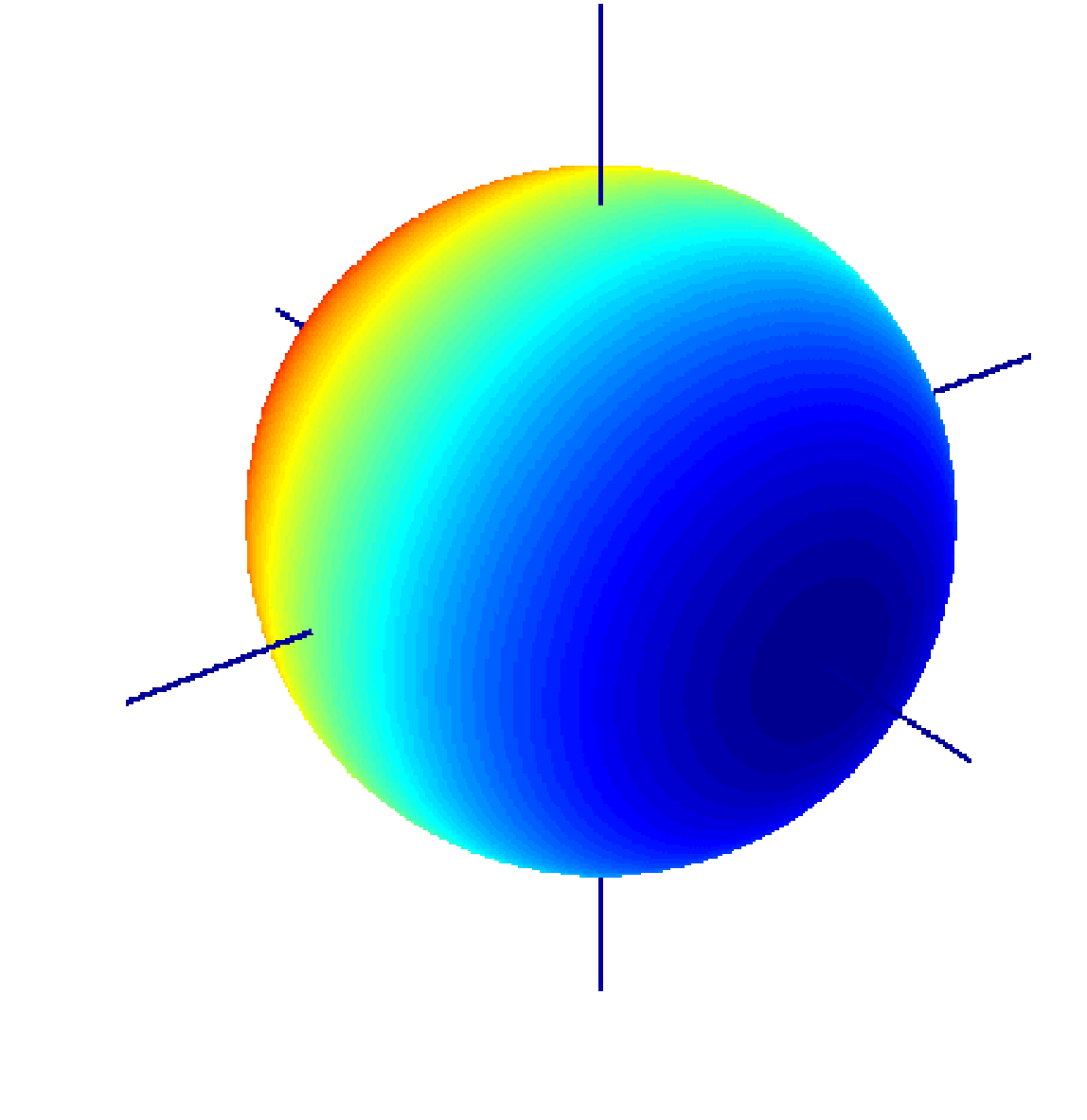}&
$1$\\
\includegraphics[width=0.12\textwidth]{FIGURES/comp-Y_2_5.eps}&
\includegraphics[width=0.12\textwidth]{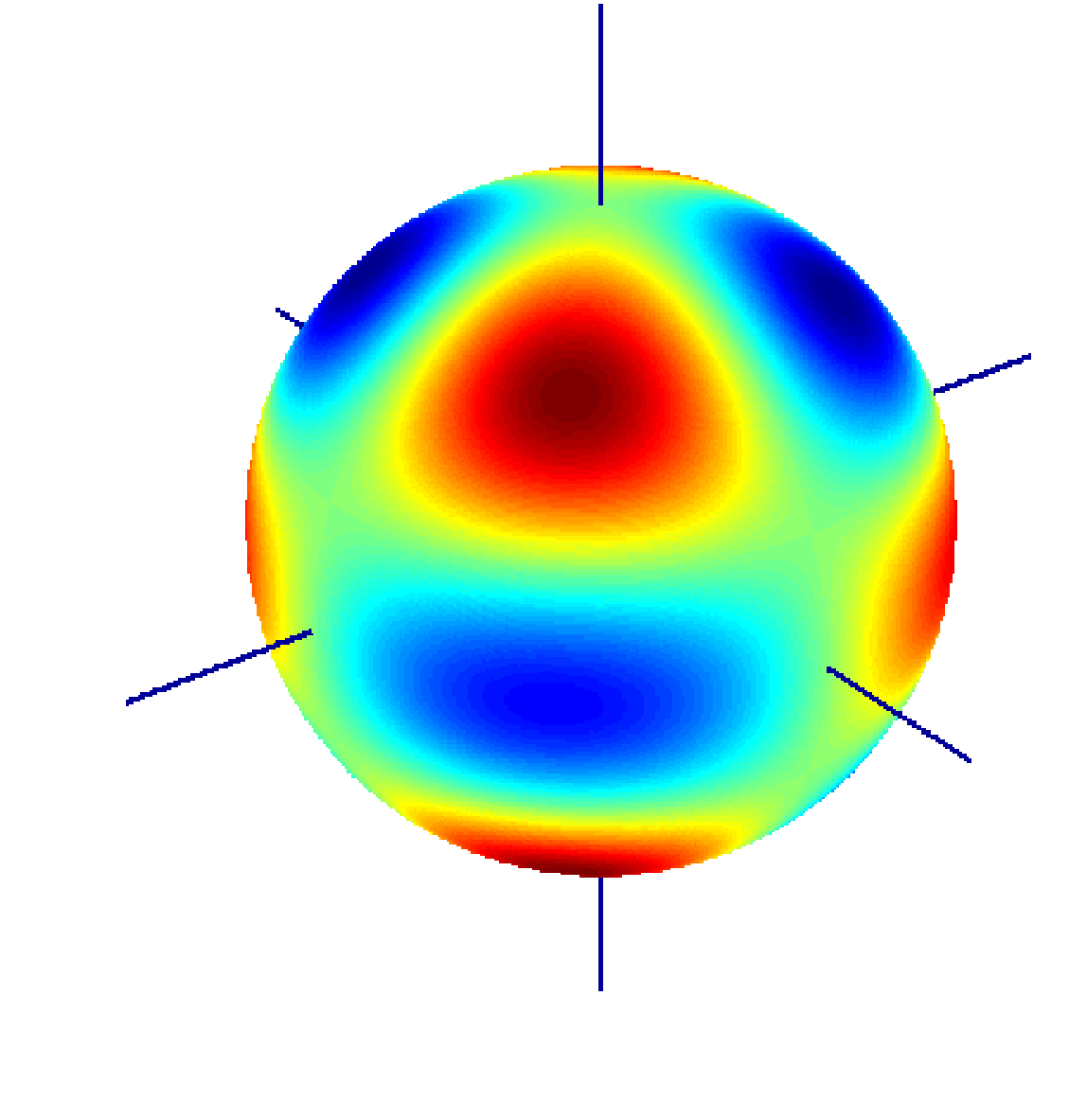}&
\includegraphics[width=0.12\textwidth]{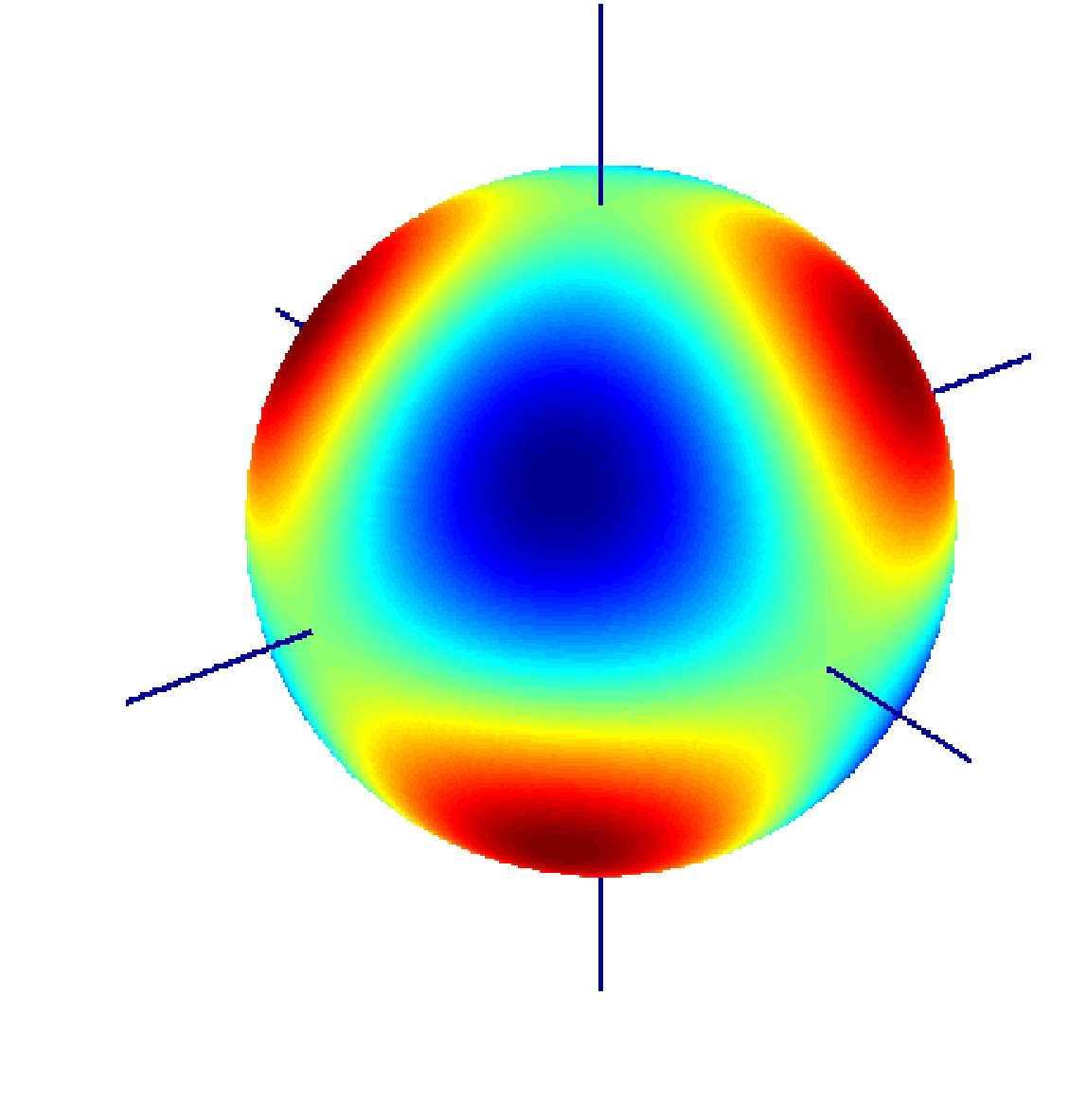}&
\includegraphics[width=0.12\textwidth]{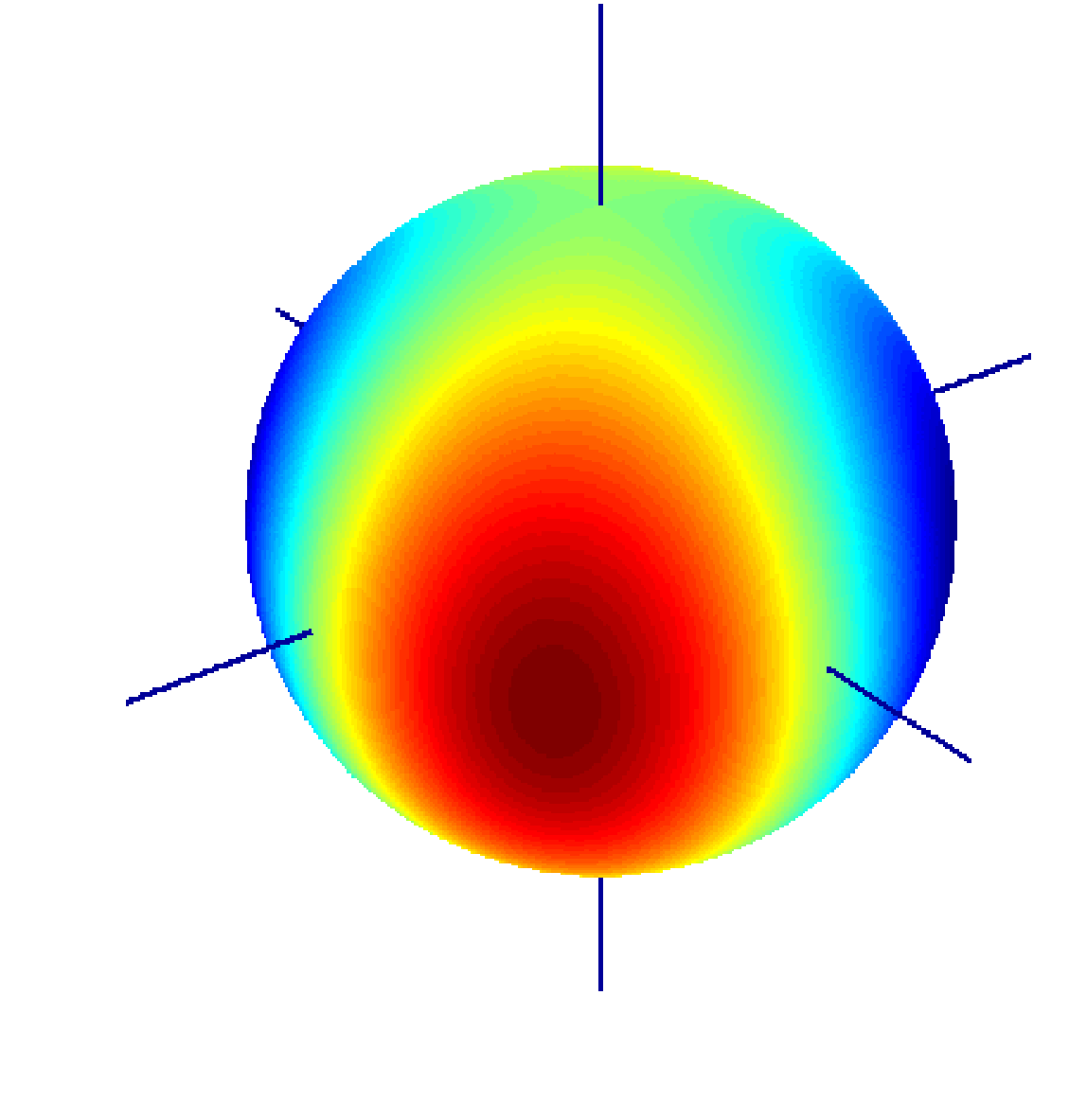}&
&
$2$\\
\includegraphics[width=0.12\textwidth]{FIGURES/comp-Y_3_5.eps}&
\includegraphics[width=0.12\textwidth]{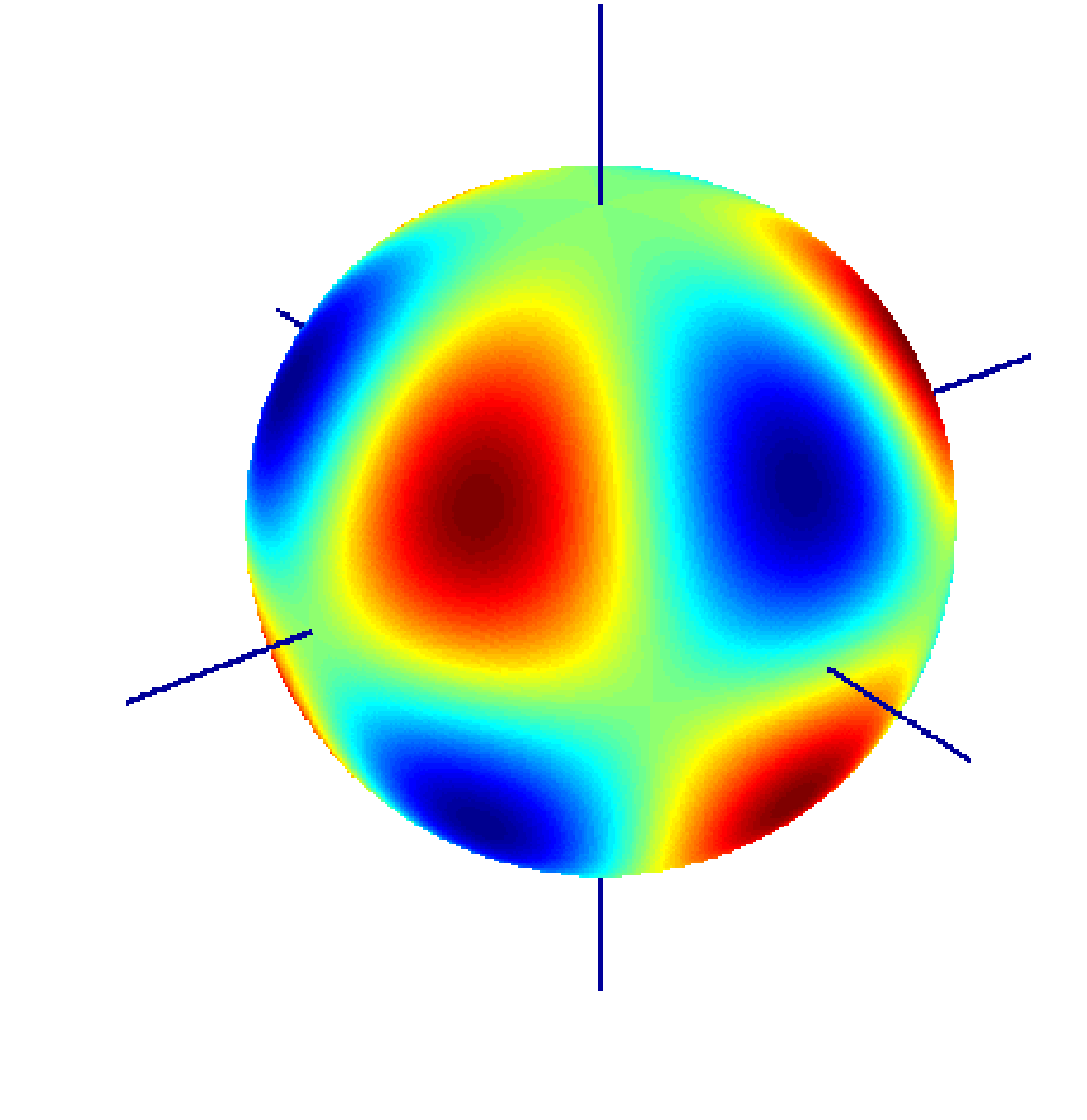}&
\includegraphics[width=0.12\textwidth]{FIGURES/comp-Y_3_3.eps}&
&
&
$3$\\
\includegraphics[width=0.12\textwidth]{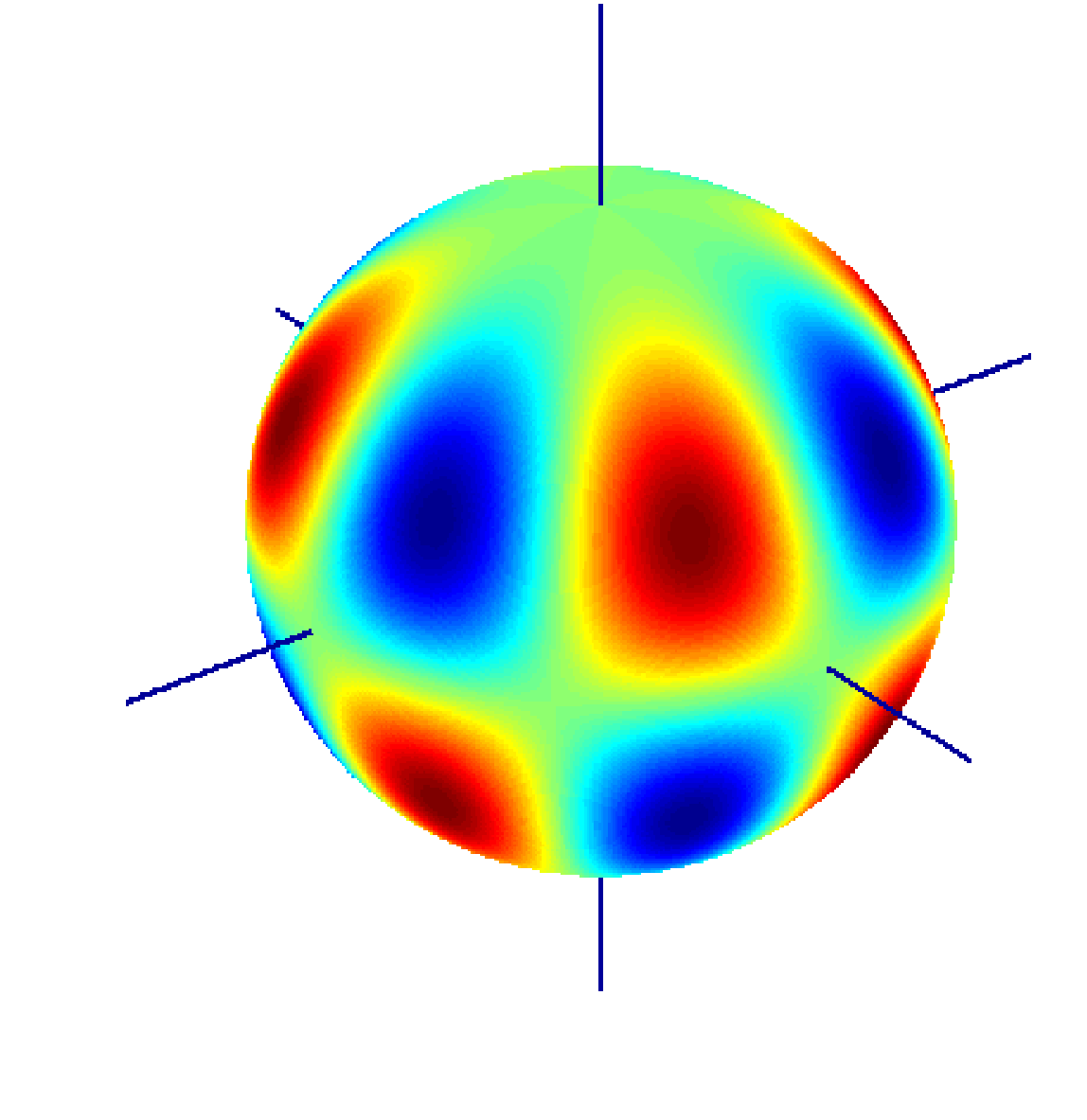}&
\includegraphics[width=0.12\textwidth]{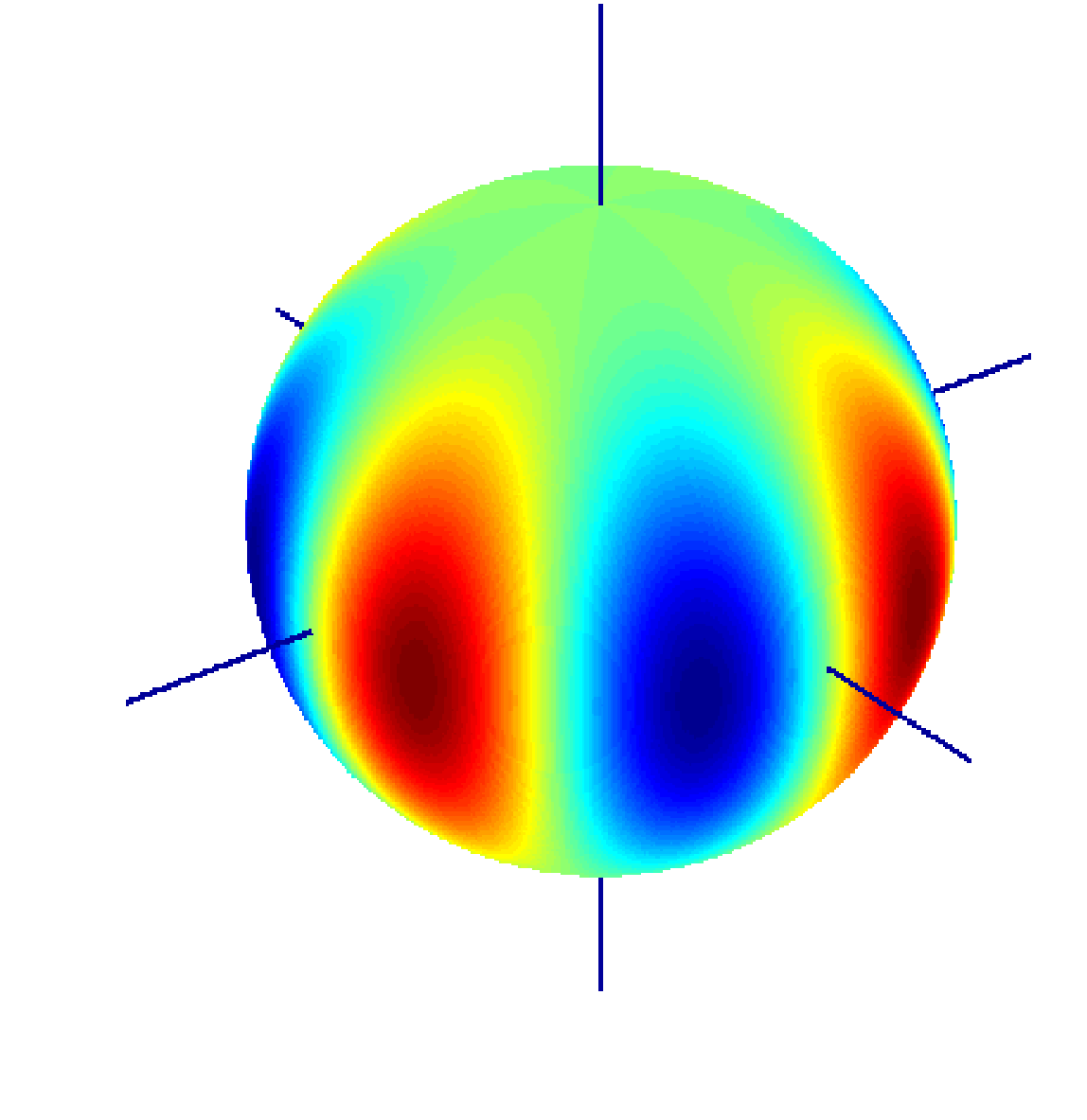}&
&
&
&
$4$\\
\includegraphics[width=0.12\textwidth]{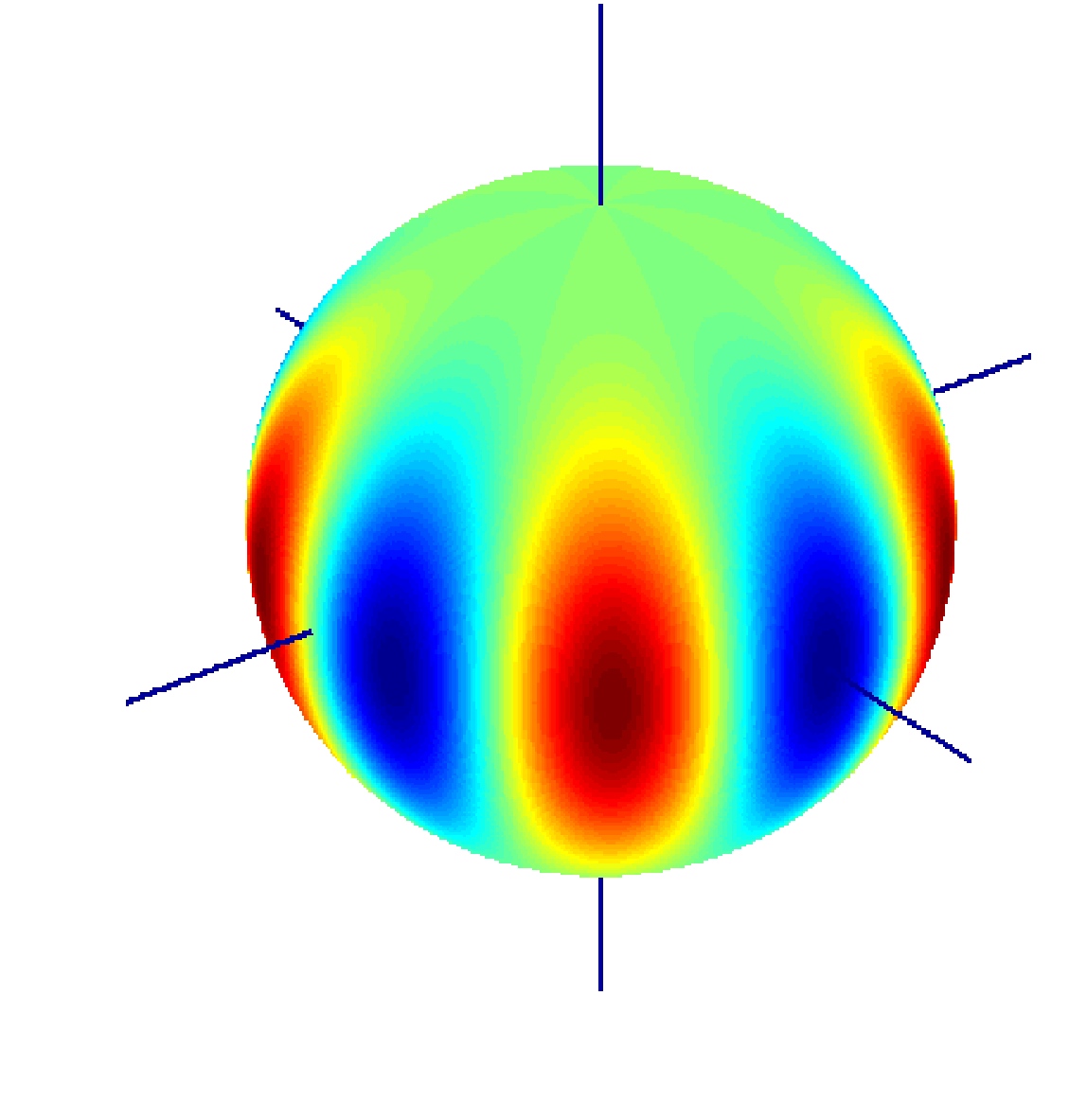}&
&
&
&
&
$5$
\end{tabular}
\end{flushleft}
\caption[Spherical Harmonic base functions (complex part)]{\label{fig:feature:SHbase_comp} Complex part of the first 5 bands of the complex 
Spherical Harmonic base functions. Note that all $Y^l_0$ are only real-valued.}
\end{figure}

\subsection{\label{sec:feature:shpoper}Useful Properties of Spherical Harmonics}
\index{Spherical Harmonics}
We give some of the useful properties of Spherical Harmonics which we exploit later. All presented properties are valid for the use
of normalized base functions.
\index{Orthonormal}
\paragraph{Orthonormality:} As mentioned before, the key property is that the base functions $Y^l_m$ are orthonormal:
\begin{equation}
\int\limits_{\Theta\Phi}Y^l_m\overline{Y^{l'}_{m'}} \sin{\Theta}d\Theta d\Phi = \delta_{ll'}\delta_{mm'},
\label{eq:feature:SHpropOrtho}
\end{equation}
with the Kronecker symbol $\delta$.
\paragraph{Symmetry:} Symmetry of the Spherical Harmonic base functions can be nicely observed in Fig. \ref{fig:feature:SHbase_comp} and
is given by: 
\index{Symmetry}
\begin{eqnarray}
\overline{Y^l_m} &=& (-1)^m Y^l_{-m}.\\
\label{eq:feature:SHpropSym}
\end{eqnarray}
\paragraph{Addition Theorem:}
\index{Addition Theorem}
For $\gamma$ given by
$$ \cos(\gamma) = \cos(\Theta)\cos(\Theta')+\sin(\Theta)\sin(\Theta')\cos(\Phi-\Phi')$$
the Addition Theorem \cite{angular} states that $P^l(\cos(\gamma))$ can be obtained by:
\begin{equation}
P^l(\cos(\gamma)) = \frac{4\pi}{2l+1}\sum\limits_m \overline{Y^{l}_{m}}(\Theta,\Phi)Y^{l}_{m}(\Theta',\Phi'),
\label{eq:feature:SHadditiontheorem}
\end{equation}
which also implies the property \cite{angular}
\begin{equation}
Y^l_0 = \left(\frac{2l+1}{4\pi}\right)^{1/2} P^l(\cos(\Theta)).
\end{equation}

\section{\label{sec:feature:shrot}Rotations in ${\cal SH}$}
\index{Rotation}
\begin{figure}[ht]
\centering
\psfrag{phi}{$\phi$}
\psfrag{theta}{$\theta$}
\psfrag{psi}{$\psi$}
\includegraphics[width=0.25\textwidth]{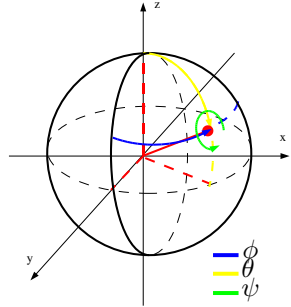}
\caption[Parameterization of rotations in Euler angles]{\label{fig:feature:eulerschema} Rotations in Euler angles using the $zyz'$ 
convention. First we rotate $\phi$ around the $z$-Axis, then $\theta$
around $y$ and finally $\psi$ around the new $z$-Axis $z'$.}
\end{figure}
Throughout the rest of this work we will use the Euler notation in $zyz'$-convention (see Fig. \ref{fig:feature:eulerschema}) denoted 
by the angles
$\phi,\theta,\psi$ with $\phi,\psi \in [0, 2\pi[$ and $\theta \in [0, \pi[$ to parameterize the rotations ${\cal R}\in {\cal SO}(3)$
(abbreviated for ${\cal R}(\phi,\theta,\psi)\in {\cal SO}(3)$).\\
Rotations ${\cal R}(\phi,\theta,\psi)$ in the Euclidean space find their equivalent representation in the harmonic domain in terms
of the so called Wigner D-Matrices, which form an irreducible representation of the rotation group ${\cal SO}(3)$ \cite{angular}.
For each band $l$, $D^l(\phi,\theta,\psi)$ (or abbreviated $D^l({\cal R})$) defines a band-wise rotation in the ${\cal SH}$ coefficients.
A rotation of $f$ by ${\cal R}$ in the Euclidean space can be computed in the harmonic domain by:
\begin{equation}
\label{eq:feature:shrot}
{\cal R} f = \sum\limits_{l=0}^\infty\sum\limits_{m=-l}^l\sum\limits_{n=-l}^l D^l_{mn}({\cal R}) \widehat{f}^l_n Y^l_m.
\end{equation}
Hence, we rotate $\widehat{f}^l_m$ by ${\cal R}(\phi,\theta,\psi)$ via band-wise multiplications:
\begin{equation}
f'={\cal R}(\phi,\theta,\psi)f \Rightarrow \widehat{f'}^l_m = \sum\limits_{n=-l}^{l} D^l_{mn}(\phi,\theta,\psi) \widehat{f}^l_n.
\label{eq:feature:SHrotForward}
\end{equation}
Due to the use of the $zyz'$-convention, we have to handle inverse rotations with some care:
\begin{equation}
f'={\cal R}^{-1}(\phi,\theta,\psi)f \Rightarrow \widehat{f'}^l_m = \sum\limits_{n=-l}^{l} D^l_{mn}(-\psi,-\theta,-\phi) \widehat{f}^l_n.
\label{eq:feature:SHrotBack}
\end{equation}

\subsection{\label{sec:feature:dwig}Computation of Wigner d-Matrices}
\index{Wigner d-Matrix}
The actual computation of the Wigner d-Matrices is a bit tricky.
In a direct approach, the d-Matrices can be computed by the sum
\begin{eqnarray}
\label{eq:feature:dwig1}
d^l_{mn}(\theta) = &&\sum\limits_t (-1)^t {{\sqrt{(l+m)!(l-m)!(l+n)!(l-n)!}}\over{(l+m-t)!(l-n-t)!t!(t+n-m)!}}\nonumber\\
&\cdot& \cos(\theta/2)^{2l+m-n-2t} \cdot \sin(\theta/2)^{2t+n-m}
\end{eqnarray}
over all $t$ which lead to non-negative factorials \cite{angular}.
It is easy to see that the constraints on $t$ are causing the computational complexity to grow
with the band of expansion. To overcome this problem, \cite{dmatrix} introduced a recursive method for the d-Matrix computation. We
are applying a closely related approach inspired by \cite{reisert}, where we retrieve d-Matrices from  recursively computed D-Matrices.\\

\subsubsection{\label{sec:feature:Dwig}Recursive Computation of Wigner D-Matrices}
\index{Wigner D-Matrix}
Given $D^l$ for the first two bands $l=0$ and $l=1$,
\begin{eqnarray}
\label{eq:feature:dwig2}
D^0(\phi,\theta,\psi) &:=& 1\nonumber\\
D^1(\phi,\theta,\psi) &:=&
\left(
\begin{matrix}
\mathrm{e}^{ -i\psi} {{1+\cos(\theta)}\over 2} \mathrm{e}^{ -i \phi}\text{\quad} &
{{-\sin(\theta)}\over{\sqrt{2}}} \mathrm{e}^{ -i \phi}\text{\quad} &
\mathrm{e}^{ i\psi} {{1-\cos(\theta)}\over 2} \mathrm{e}^{ -i \phi}\\
\mathrm{e}^{ -i \psi} {\sin(\theta)\over\sqrt{2}} \text{\quad} &
\cos(\theta) \text{\quad} &
-\mathrm{e}^{ i \psi} {\sin(\theta)\over\sqrt{2}}\\
\mathrm{e}^{ -i \psi} {{1-\cos(\theta)}\over 2} \mathrm{e}^{ i \phi} \text{\quad} &
{\sin(\theta)\over\sqrt{2}}  \mathrm{e}^{ i \phi} \text{\quad} &
\mathrm{e}^{ i \psi} {{1+\cos(\theta)}\over 2} \mathrm{e}^{ i \phi}
\end{matrix}
\right)\nonumber
\end{eqnarray}
we can compute $D^l$ via band-wise recursion:
\begin{eqnarray}
\label{eq:feature:dwig3}
D^l_{mn} = &&\sum\limits_{m,m',n,n'=-l}^l D^1_{m'n'} D^{l-1}_{(m-m')(n-n')}\nonumber\\
&\cdot& \langle (l-1)m|1m',l(m-m') \rangle\nonumber\\
&\cdot& \langle (l-1)n|1n',l(n-n') \rangle
\end{eqnarray}
where $\langle lm|l'm',l''m'' \rangle$ denotes Clebsch-Gordan coefficients (see section \ref{sec:feature:clebschgordan}) known from angular momentum 
theory.
Using (\ref{eq:feature:substD}), we finally obtain:
\begin{equation}
d^l_{mn}(\theta) = D^l_{mn}(0,\theta,0).
\label{eq:feature:dwig4}
\end{equation}

\subsection{\label{sec:feature:dwigProp}Properties of Wigner Matrices}

\paragraph{Orthogonality:}
The Wigner D-matrix elements form a complete set of orthogonal functions over the Euler angles \cite{angular2}:
\begin{equation}
\int\limits_{\phi,\theta,\psi} D^l_{mn}(\phi,\theta,\psi) \overline{D^{l'}_{m'n'}}(\phi,\theta,\psi) \sin{\Theta}d\phi d\theta d\psi = \frac{8\pi^2}{2l+1}
\delta_{ll'}\delta_{mm'}\delta_{nn'},
\label{eq:feature:dwigOrthogonal}
\end{equation}
with Kronecker symbol $\delta$.
\paragraph{Symmetry:}
\begin{equation}
D^l_{mn}(\phi,\theta,\psi) = \overline{D^l_{-m-n}}(\phi,\theta,\psi).
\label{eq:feature:dwigSym}
\end{equation}
\paragraph{Relations to Spherical Harmonics:}
The D-Matrix elements with second index equal to zero, are proportional to Spherical Harmonic base functions \cite{angular3}:
\begin{equation}
\overline{D^l_{m0}}(\phi,\theta,\psi) = \sqrt{\frac{4\pi}{2l+1}}Y^l_m(\phi,\theta).
\label{eq:feature:dwigRelationSH}
\end{equation}
\paragraph{Relations to Legendre Polynomials:}
The Wigner small d-Matrix elements with both indices set to zero are related to Legendre polynomials \cite{angular2}:
\begin{equation}
d^l_{00}(\theta) = P^l(\cos(\theta)).
\label{eq:feature:dwigRelationLegendre}
\end{equation}

\section{\label{sec:feature:clebschgordan}Clebsch-Gordan Coefficients}
\index{Clebsch-Gordan Coefficients}\index{Angular Coupling}
Clebsch-Gordan Coefficients (CG) of the form
$$ \langle lm|l_1 m_1, l_2 m_2 \rangle $$
are commonly used for the representation of direct sum decompositions of ${\cal SO}(3)$ tensor couplings \cite{angular}.
The CG define the selection criteria for couplings and are by definition only unequal to zero if the constraints 
\begin{equation}
m=m_1+m_2 \text{ and } |l_1-l_2|\leq l\leq l_1 +l_2\nonumber
\end{equation}
hold. In most cases non-zero  Clebsch-Gordan Coefficients are not directly evaluated, we rather utilize their orthogonality
and symmetry properties to reduce and simplify coupling formulations. The quite complex closed form for the computation of CG can be 
found in \cite{angular2}.

\subsection{Properties of Clebsch-Gordan Coefficients}
Some useful properties of Clebsch-Gordan Coefficients \cite{angular}:
\paragraph{Exceptions:}
For $l=0$ the CG are:
\begin{equation}
\langle 00|l_1m_1,l_2m_2 \rangle = \delta_{l_1,l_2}\delta_{m_1,-m_2}\frac{(-1)^{l_1-m_1}}{\sqrt{2l_2+1}}
\label{eq:feature:CGexept1}
\end{equation}
and for $l=(l_1+l_2)$ and $m_1=l_1, m_2=l_2$: 
\begin{equation}
\langle (l_1+l_2)(l_1+l_2)|l_1l_1,l_2l_2 \rangle = 1.
\label{eq:feature:CGexept2}
\end{equation}
\index{Orthogonality}
\paragraph{Orthogonality:}
\begin{eqnarray}
\sum\limits_{l=|l_1-l_2|}^{l_1+l_2}\sum\limits_{m=-l}^{l}\langle lm|l_1m_1,l_2m_2\rangle\langle lm|l_1m_1',l_2m_2'\rangle &=& \delta_{m_1,m_1'}\delta_{m_2,m_2'}\\
\sum\limits_{m_1m_2}\langle lm|l_1m_1,l_2m_2\rangle\langle l'm'|l_1m_1,l_2m_2\rangle &=& \delta_{l,l'}\delta_{m,m'}.
\end{eqnarray}
\index{Symmetry}
\paragraph{Symmetry:} Some symmetry properties of CG. There are even more symmetries \cite{angular2}, but we only provide those 
which we will use later on:
\begin{eqnarray}
\langle lm|l_1m_1,l_2m_2\rangle &=& (-1)^{l_1+l_2-l}\langle l(-m)|l_1(-m_1),l_2(-m_2)\rangle\\
&=& (-1)^{l_1+l_2-l} \langle lm|l_2m_2,l_1m_1 \rangle\\
&=& (-1)^{l_1-m_1} \sqrt{\frac{2l+1}{2l_2+1}}\langle l_2(-m_2)|l_1m_1,lm\rangle\\
&=& (-1)^{l_2+m_2} \sqrt{\frac{2l+1}{2l_1+1}}\langle l_1(-m_1)|l(-m),l_2m_2\rangle.
\end{eqnarray}

\section{\label{sec:feature:shcorr}Fast and Accurate Correlation in ${\cal SH}$}
\index{Correlation}\index{Rotation}
So far we have introduced many basic properties of the Spherical Harmonic domain, which we are using now to derive more complex operations.
In analogy to the Fourier domain, where the Convolution Theorem enables us to compute a fast convolution and correlation of
signals in the frequency domain, we now derive fast convolution and correlation for the Spherical Harmonic domain which
we introduced in \cite{CorrECCV}.\\
Since some important features and feature selection
methods have been derived from the key ideas of this approach, we review this method in detail:
\paragraph{\label{sec:feature:spericalcorridea} Correlation on the 2-Sphere: }
The full correlation function ${\cal C}^{\#}:{\cal SO}(3)\rightarrow \mathbb{R}$ of two signals $f$ and $g$ under the rotation ${\cal R}\in
{\cal SO}(3)$ on a 2-sphere is
given as:
\begin{equation}
{\cal SH}_{corr}({\cal R}) := \int\limits_{S^2} f  ({\cal R} g) \text{\quad} \sin{\Theta}d\Phi d\Theta. 
\label{eq:feature:SHcorr0}
\end{equation}
Obviously, the computational cost of a direct evaluation approach - over all possible rotations ${\cal R}$ - is way too high. Especially
when we are considering arbitrary resolutions of the rotation parameters. To cope with this problem, we derive a fast but accurate
method for the computation of the correlation in the harmonic domain.\\
Besides the obvious usage of the (cross)-correlation as similarity measure, the correlation on the 2-sphere can also be used
to perform a rotation estimation of similar signals on a sphere. 
\paragraph{\label{sec:feature:rotestimateidea}Rotation Estimation:} 
given any two real valued signals $f_1$ and $f_2$ on a 2-sphere which are considered to be equal
or at least similar under some rotational invariant measure $(\sim_{\cal R})$:
\begin{equation}
\label{eq:feature:SHcorr_problem}
f_1 \sim_{\cal R} f_2, {\cal R}\in SO(3),
\end{equation}
the goal is to estimate the parameters of an arbitrary rotation ${\cal R}$ as accurate as possible without any additional information other than $f_1, f_2$ and considering arbitrary resolutions of the rotation parameters.
\paragraph{Related Approaches:}
Recently, there have been proposals for several different methods which try to overcome the direct matching problem. Here, we are only
considering methods which provide full rotational estimates (there are many methods covering only rotations around the z-axis) without
correspondences.\\
A direct nonlinear estimation (DNE) which is able to retrieve the parameters for small rotations via
iterative minimization techniques was introduced in \cite{Rot2}. However, this method fails for larger rotations and was proposed only for
``fine tuning'' of pre-aligned rotations.
Most other methods use representations in the Spherical Harmonic domain to solve the problem.\\
The possibility to recover the rotation parameters utilizing the spherical harmonic shift theorem (SHIFT) \cite{angular}
has been shown in \cite{shift}. This approach also uses an iterative minimization and was later refined by \cite{Rot3}.
Again, the estimation accuracy is limited to small rotations.
\paragraph{Rotation Estimation via Correlation:}
The basis of our method was first suggested by \cite{first}, presenting a fast correlation in two angles followed by a correlation in the 
third Euler angle in an iterative way (known as FCOR).
This method
was later extended to a full correlation in all three angles by \cite{correlation}. This approach allows the direct computation of the 
correlation from the harmonic coefficients via FFT, but was actually not intended to be used to recover the rotation parameters.
Its angular resolution directly depends on the range of the harmonic expansion - making high angular resolutions rather expensive.
But FCOR was used by \cite{Rot2} to initialize the DNE and SHIFT ``fine tuning'' algorithms. The same authors
used a variation of FCOR (using inverse Spherical Fourier Transform \cite{SOFT} in stead of FFT) in combination with
SHIFT \cite{Rot1} to recover robot positions from omni-directional images  via rotation parameter estimation.
\subsection{\label{sec:feature:SHcorrbase}Basic ${\cal SH}$-Correlation Algorithm}
Starting from the full correlation function (\ref{eq:feature:SHcorr0})
we use  the Convolution Theorem and substitute $f$ and $g$ with their ${\cal SH}$ expansions
(\ref{eq:feature:shrot}, \ref{eq:feature:SHcoeff})
, which leads to
\begin{equation}
\label{eq:feature:SHcorr1}
{\cal SH}_{corr}({\cal R}) = \sum\limits_{l=0}^\infty\sum\limits_{m=-l}^l\sum\limits_{n=-l}^l \overline{D^{l}_{mn}(\cal R)}  \widehat{f}^l_m \overline{\widehat{g}^l_n}.
\end{equation}
The actual ``trick'' to obtain the fast correlation is to factorize the original rotation ${\cal R}(\phi,\theta,\psi)$ into
${\cal R} = {\cal R}_1 \cdot {\cal R}_2$, choosing
${\cal R}_1(\xi,\pi/2,0)$ and ${\cal R}_2(\eta,\pi/2,\omega)$ with $\xi = \phi-\pi/2, \eta = \pi - \theta, \omega = \psi-\pi/2$.\\

Using the fact that
\begin{equation}
D^l_{mn}(\phi,\theta,\psi) = \mathrm{e}^{-im\phi}d^l_{mn}(\theta)\mathrm{e}^{-in\psi},
\label{eq:feature:SHcorr2a}
\end{equation}
where $d^l$ is a real valued ``Wigner (small) d-matrix'' (see (\ref{sec:feature:dwig})), and
\begin{equation}
D^l_{mn}({\cal R}_1\cdot {\cal R}_2) = \sum\limits_{h=-l}^l D^l_{nh}({\cal R}_1) D^l_{hm}({\cal R}_2),
\label{eq:feature:SHcorr3}
\end{equation}
we can rewrite
\begin{equation}
\label{eq:feature:substD}
{D^{l}_{mn}(\cal R)} = \sum\limits_{h=-l}^l d^l_{nh}(\pi/2) d^l_{hm}(\pi/2) \mathrm{e}^{-i(n\xi + h\eta + m\omega)}.
\end{equation}
Substituting (\ref{eq:feature:substD}) into (\ref{eq:feature:SHcorr1}) provides the final formulation for the correlation function 
regarding
the new angles $\xi, \eta$ and $\omega$:
\begin{equation}
\label{eq:feature:SHcorr2}
{\cal SH}_{corr}(\xi, \eta, \omega) = \sum\limits_{l=0}^\infty\sum\limits_{m=-l}^l\sum\limits_{h=-l}^l\sum\limits_{m'=-l}^l d^l_{mh}(\pi/2) d^l_{hm'}(\pi/2) \widehat{f}^l_m \overline{\widehat{g}^l_{m'}} \mathrm{e}^{-i(m\xi + h\eta + m'\omega)}.
\end{equation}
The direct evaluation of this correlation function is of course not possible - but it is rather straightforward to obtain the Fourier
transform of (\ref{eq:feature:SHcorr2}), hence eliminating the missing angle parameters:\index{FFT}
\begin{equation}
\label{eq:feature:SHcorrFT}
\widehat{{\cal SH}_{corr}}(m, h, m') = \sum\limits_{l=0}^\infty d^l_{mh}(\pi/2) d^l_{hm'}(\pi/2) \widehat{f}^l_m \overline{\widehat{g}^l_{m'}}.
\end{equation}
Finally, the correlation ${\cal SH}_{corr}(\xi, \eta, \omega)$ can be retrieved via inverse Fourier transform of $\widehat{{\cal SH}_{corr}}$,
\begin{equation}
\label{eq:feature:SHcorrFinal}
{\cal SH}_{corr}(\xi, \eta, \omega) = {\cal F}^{-1}(\widehat{{\cal SH}_{corr}}(m, h, m')),
\end{equation}
revealing the correlation values in a three dimensional ${\cal C}^{\#}(\xi, \eta, \omega)$-space.\\

\begin{figure}[th]
\centering
\includegraphics[width=0.25\textwidth]{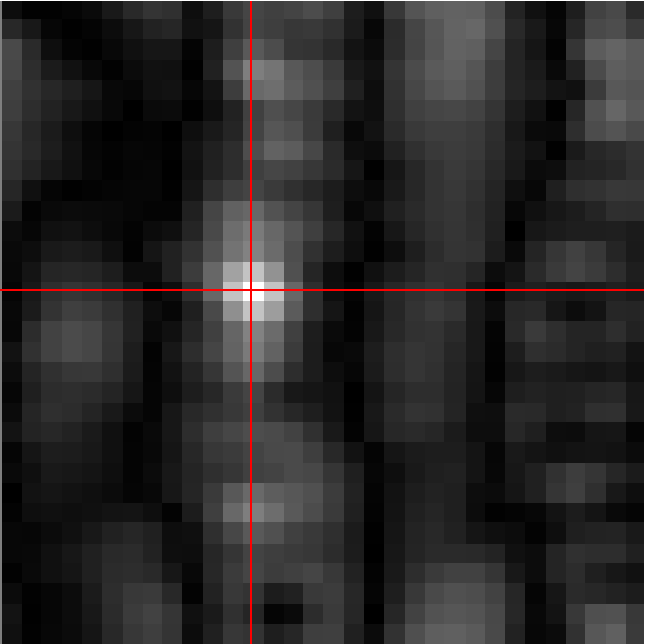}
\includegraphics[width=0.25\textwidth]{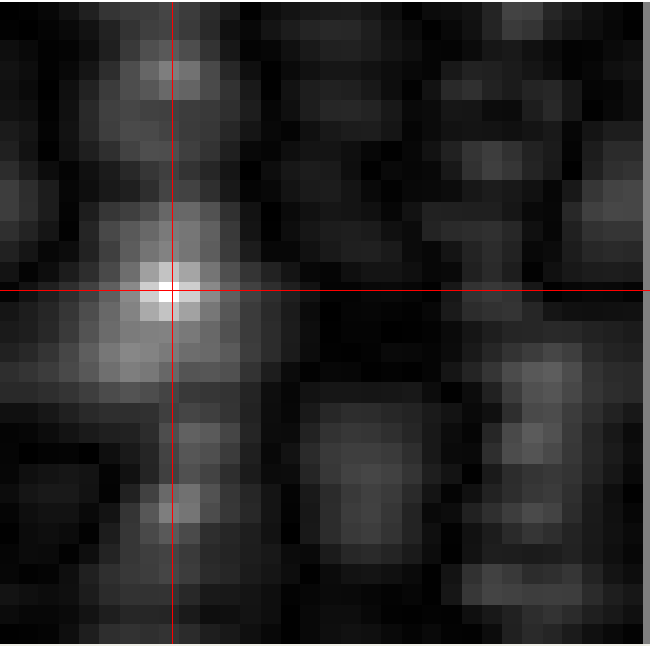}
\includegraphics[width=0.25\textwidth]{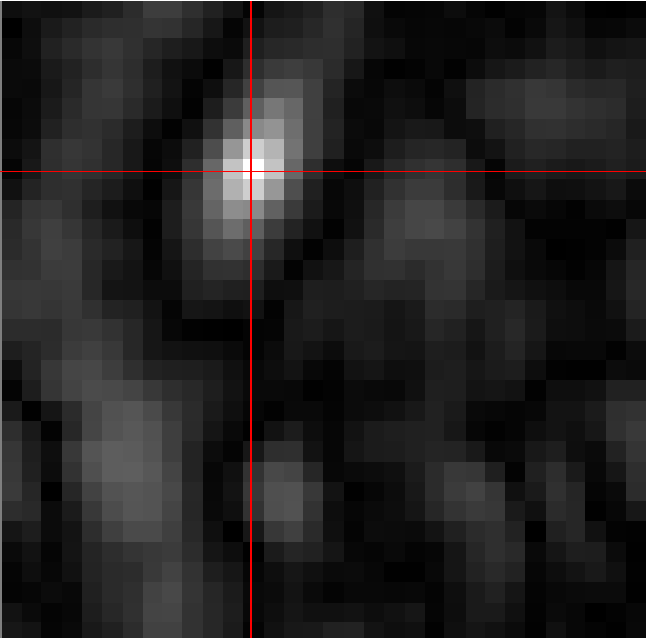}
\caption[${\cal SH}$ correlation matrix]{\label{fig:feature:Fcorrgrid}Orthoview of a resulting 3D correlation grid in the ${\cal C}^{\#}(\xi, \eta, 
\omega)$
-space with a maximum spherical harmonic expansion to the 16th band, $\phi=\pi/4, \theta=\pi/8, \psi=\pi/2$. From left to right: xy-plane, 
zy-plane, xz-plane.}
\end{figure}

\subsection{\label{sec:feature:shift}Euler Ambiguities}
The final obstacle towards
the recovery of the rotation parameters inherits from the Euler parameterization used in the correlation function. Unfortunately, Euler
angle formulations cause various ambiguities and cyclic shift problems.\\
One minor problem is caused by the fact that our parameter grid range is from $0, \dots, 2\pi$ in all dimensions, while the
angle $\theta$ is only defined $\theta\in [0,\pi[$. This causes two correlation peaks at $\theta = \beta$ and
$\theta = 2\pi - \beta$ for an actual rotation of $\theta = \beta$. We avoid this problem by restricting the maximum search to
$\theta\in [0,\pi[$, hence neglecting half of the correlation space.\\
The formulation of the correlation function also causes further cyclic shifts in the grid representation of the Euler angles.
This way, the zero rotation ${\cal R}(\phi=0, \theta=0, \psi=0)$ does not have its peak at the zero position $C^\#(0,0,0)$ of the parameter 
grid as
one would expect. For a more intuitive handling of the parameter extraction from the grid, such that the $(0,0,0)$ position in the grid
corresponds to no rotation,
we extend the original formulation of
(\ref{eq:feature:SHcorrFT}) and use a shift in the frequency space in order to normalize the mapping of ${\cal R}(\pi,0,\pi)$ to $C^\#(0,0,0)$:
\begin{equation}
\label{eq:feature:SHcorrShift}
\widehat{C^\#}(m, h, m') = \sum\limits_{l=0}^{\infty} d^l_{mh}(\pi/2) d^l_{hm'}(\pi/2) \widehat{f}_{lm}
\overline{\widehat{g}_{lm'}} \cdot i^{m+2h+m'}.
\end{equation}

\subsection{\label{sec::feature:SHcooranglarres}Increasing the Angular Resolution}
For real world applications, where the harmonic expansion is limited to some maximum expansion band $b_{\max}$:
\begin{equation}
\label{eq:feature:SHcorrFFT}
\widehat{{\cal C}^{\#}}(m, h, m') = \sum\limits_{l=0}^{b_{\max}} d^l_{mh}(\pi/2) d^l_{hm'}(\pi/2) \widehat{f}^l_m \overline{\widehat{g}^l_{m'}}
\cdot i^{m+2h+m'},
\end{equation}
the resulting $(\xi, \eta, \omega)$ space turns into a sparse and discrete space. Unfortunately, this directly affects the angular 
resolution of the correlation.
Let us take a closer look at figure (\ref{fig:feature:Fcorrgrid}): first of all, it appears (and our experiments in 
section \ref{sec:feature:Shcorrexperiment}) clearly support this 
assumption) that
the fast correlation function has a clear and stable maximum in a point on the grid. This is a very nice property, and we could simply
recover the corresponding rotation parameters which are associated with this maximum position. But there are still some major problems:
The image in Figure (\ref{fig:feature:Fcorrgrid}) appears to be quite coarse - and in fact, the parameter grids for expansions up to 
the 16th band ($b_{\max}=16$) have
the size of $33\times 33\times 33$ since the parameters $m,m',h$ in (\ref{eq:feature:SHcorrFT}) are running from $-b_{\max},\dots,b_{\max}$.
Given rotations up to $360^{\circ}$, this leaves us in the worst case with an overall estimation
accuracy of less than $15^{\circ}$.\\
In general, even if our fast correlation function (\ref{eq:feature:SHcorrFinal}) would perfectly estimate the maximum position in all cases,
we would have to expect a worst case accuracy of
\begin{equation}
\label{eq:feature:Corr_err}
Err_{corr} = 2\cdot{180^{\circ}\over 2b_{\max}} + {90^{\circ} \over 2b_{\max}},
\end{equation}
accumulated over all three angles.
Hence, if we would like to achieve an accuracy of $1^{\circ}$, we would have to take the harmonic expansion roughly beyond the 180th band.
This
would be computationally expensive. Even worse, since we are considering discrete data, the signals on the sphere are band-limited.
So for smaller radii, higher bands of the expansion are actually not carrying any valuable information.\\
Due to this resolution problem, the fast correlation has so far only been used to initialize iterative algorithms \cite{Rot1}\cite{Rot2}.\\

\subsubsection{\label{sec:feature:sincinterpol}Sinc Interpolation.}
\index{Sinc interpolation}
Now, instead of increasing the sampling rate of our input signal by expanding the harmonic transform, we have found an alternative way to
increase the correlation accuracy: interpolation in the frequency domain.\\
In general, considering the Sampling Theorem and given appropriate discrete samples $a_n$ with step size $\Delta_x$ of some continuous 1D
signal $a(x)$, we can reconstruct the original
signal via sinc interpolation \cite{sinc1}:
\begin{equation}
\label{eq:feature:sinc}
a(x) = \sum\limits_{n=-\infty}^{\infty} a_n \mathrm{sinc}(\pi(x/\Delta_x -n)),
\end{equation}
with
\begin{equation}
\mathrm{sinc}(x) = { \sin(x)\over x}.
\end{equation}
For a finite number of samples, (\ref{eq:feature:sinc}) changes to:
\begin{equation}
\label{eq:feature:dsinc}
a(x) = \sum\limits_{k=0}^{N} a_k {{\sin(\pi(x/\Delta_x-k))}\over {N \sin(\pi(x/\Delta_x-k)/N)}}.
\end{equation}
This sinc interpolation features two nice properties \cite{sinc1}: it entirely avoids aliasing errors and it can easily be applied in
the discrete Fourier space. Given the DFT coefficients $\alpha_n, n=0,1, \dots, N-1$ of the discrete signal $a_n, n=0,1, \dots, N-1$, 
the sinc interpolation is implemented by adding a zero padding between $\alpha_{(N/2)-1}$ and $\alpha_{(N/2)}$.\\

\begin{figure}[th]
\includegraphics[width=0.19\textwidth]{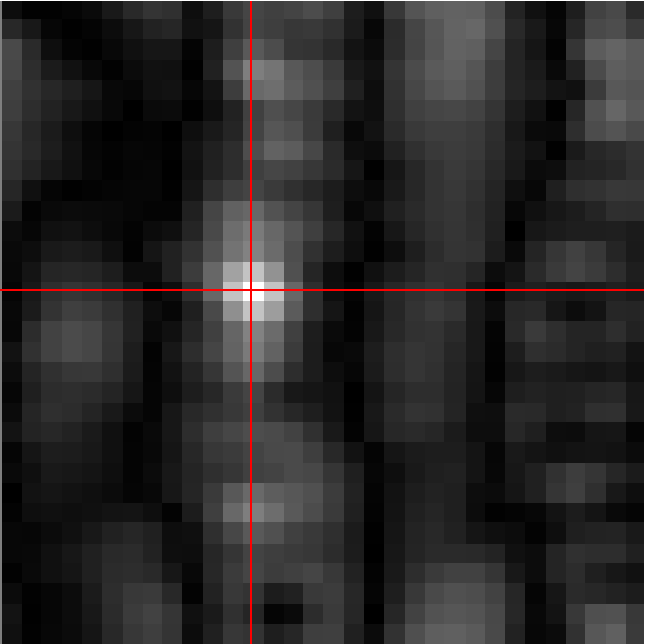}
\includegraphics[width=0.19\textwidth]{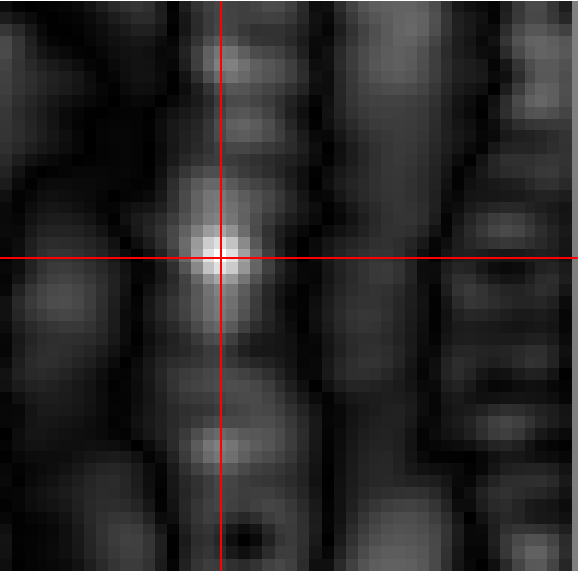}
\includegraphics[width=0.19\textwidth]{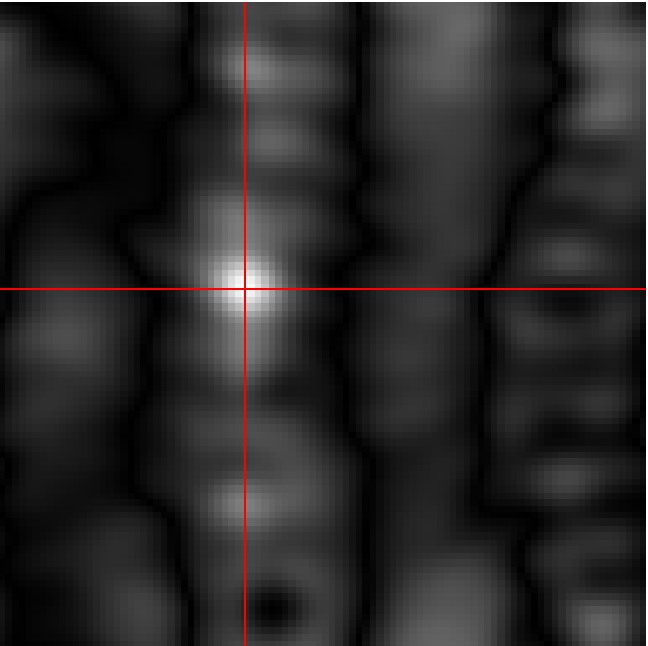}
\includegraphics[width=0.19\textwidth]{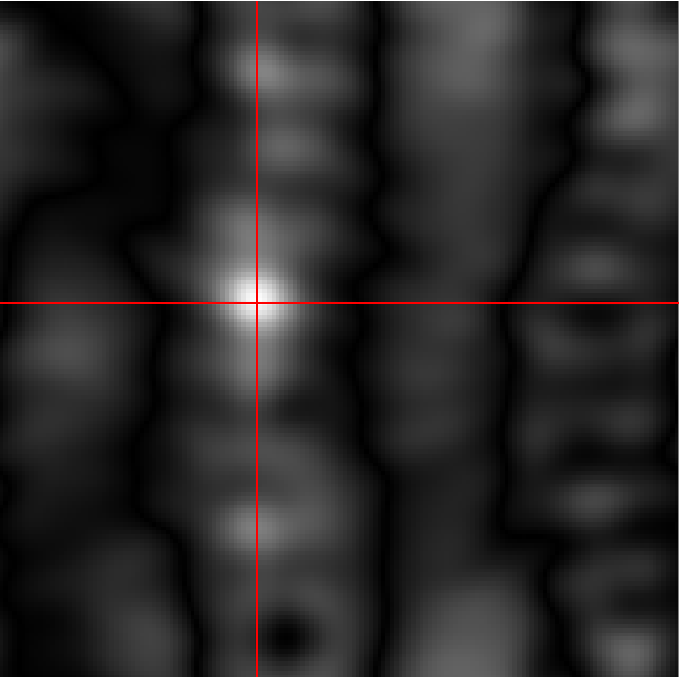}
\includegraphics[width=0.19\textwidth]{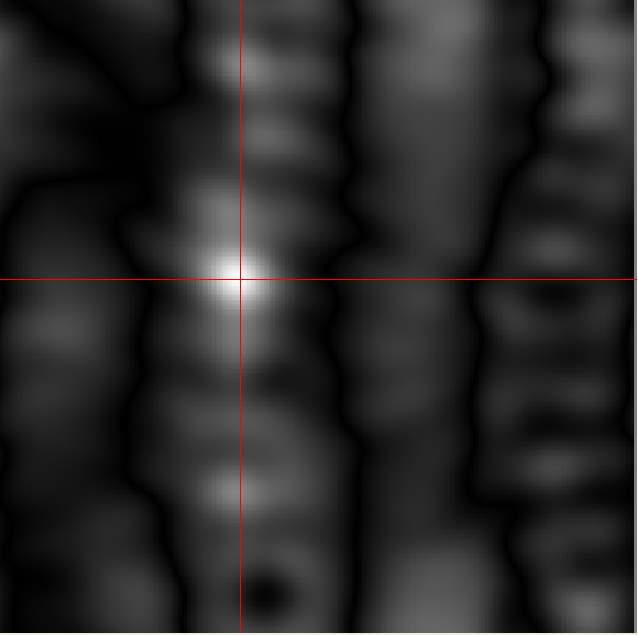}
\caption[Sinc interpolation]{\label{fig:feature:Fpad}Same experiment as in Fig. (\ref{fig:feature:Fcorrgrid}) but with increasing 
size of the sinc interpolation padding. From left to right: $p=0, p=16, p=64, p=128, p=256$ }
\end{figure}

Returning to our original correlation problem, it is easy to see that the $(m, h, m')$-space in (\ref{eq:feature:SHcorrFT}) 
is actually nothing else
but a discrete 3D Fourier spectrum. So we can directly apply the 3D extension of (\ref{eq:feature:dsinc}) and add a zero padding into the
$(m, h, m')$-space. This way, we are able to drastically increase the resolution of our correlation function at very low additional
cost for implementation issues as well as suitable pad sizes). Figure (\ref{fig:feature:Fpad}) 
shows the effect of
the interpolation on the correlation matrix for different pad sizes $p$.

It has to be noted that even though the sinc interpolation implies some smoothing characteristics to the correlation matrix,
the maxima remain fixed to singular positions in the grid.\\
Theoretically, we are now finally able to reduce the worst case accuracy to arbitrarily small angles for any given band:
\begin{equation}
\label{eq:feature:Corr_err_pad}
Err_{corr}^{pad} = 2\cdot {180^{\circ}\over 2b_{\max} + p} + {90^{\circ} \over 2b_{\max} + p}.
\end{equation}
Of course, the padding approach has practical limitations - inverse FFTs are becoming computationally expensive at some point. But
as our experiments in \ref{sec:feature:Shcorrexperiment} show, resolutions below one degree are possible even for very low expansions.
\index{FFT}
\paragraph{Implementation:} The implementation of the inverse FFT in (\ref{eq:feature:SHcorrFinal}) combined with the frequency space
padding requires some care: we need an inverse complex to real FFT with an in-place mapping (the grid in the frequency space has the same
size as the resulting grid in $\mathbb{R}^3$). Most FFT implementations are not providing such an operation. Due to the symmetries
in the frequency space not all complex coefficients need to be stored, hence most implementations are using reduced grid sizes.
We can avoid the tedious construction of such a reduced grid from $\widehat{C^\#}$ by using an inverse complex to complex FFT
and taking only the real part of the result.
In this case, we only have to shuffle the coefficients  of $\widehat{C^\#}$, which can be done via simple modulo operations while simultaneously
applying the padding. We rewrite (\ref{eq:feature:SHcorrShift}) to:
\begin{equation}
\label{eq:feature:SHcorrPad}
\widehat{C^\#}(a, b, c) = \sum\limits_{l=0}^{b_{\max}} d^l_{mh}(\pi/2) d^l_{hm'}(\pi/2) \widehat{f}_{lm} \overline{\hat{g}_{lm'}} \cdot i^{m+2h+m'},
\end{equation}
where
\begin{equation}
s:=2bp, \text{\quad}  a:=(m + s + 1)\text{mod }s, \text{\quad} b:=(h + s + 1)\text{mod }s, \text{\quad} c:=(m' + s + 1)\text{mod }s. \nonumber
\end{equation}
Concerning the pad size: due to the nature of the FFT, most implementations achieve notable speed-ups for certain grid sizes. So it
is very useful to choose the padding in such a way that the overall grid size has, e.g., prime factor decompositions of mostly small primes
\cite{FFTW}.

\subsection{\label{sec:feature:recrotparams}Rotation Parameters} Finally, we are able to retrieve the original rotation parameters. 
For a given correlation peak
at the grid position $c(x,y,z)$, with maximum harmonic expansion $b$ and padding $p$ the rotation angles are:
\begin{eqnarray}
\label{eq:feature:angles}
\phi &=&
\left\{
\begin{matrix}
\pi +(2\pi-x\Delta) & \text{\quad\quad for } x\Delta>\pi\\
\pi - x\Delta  & \text{\quad\quad otherwise}
\end{matrix}
\right.\\
\theta &=&
\left\{
\begin{matrix}
(2\pi-y\Delta) & \text{\quad\quad\quad  for } y\Delta>\pi\\
y\Delta  & \text{\quad\quad\quad  otherwise}
\end{matrix}
\right.\\
\psi &=&
\left\{
\begin{matrix}
\pi +(2\pi-z\Delta) & \text{\quad\quad for } z\Delta>\pi\\
\pi - z\Delta  & \text{\quad\quad otherwise}
\end{matrix}
\right.\\
\nonumber\\
\text{with\quad} \Delta &=& 2\pi/(b+p).\nonumber
\end{eqnarray}
\begin{figure}[ht]
\centering
\includegraphics[width=0.2\textwidth]{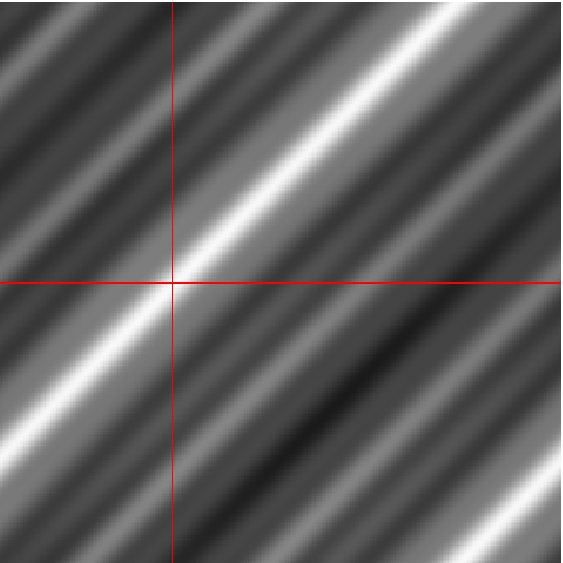}
\caption[Padded Correlation Matrix]{\label{fig:feature:corr_theta_0} $\phi\psi$-plane for of the correlation matrix with $\theta=0$. }
\end{figure}
The resulting rotation estimates return very precise and unique parameter sets. Only one ambiguous
setting has to be noted: for $\theta=0,\pi$ all $zyz'$-Euler formulations which hold $\phi+\psi=2\pi$ encode the very same
rotation (see Figure (\ref{fig:feature:corr_theta_0})). This is actually not a problem for our rotation estimation task,
but it might be quite confusing especially in the case of numerical evaluation of the estimation accuracy.

\subsection{\label{sec:feature:ncrosscorr}Normalized Cross-Correlation}
\index{Cross-Correlation}
In many cases, especially when one tries to estimate the rotation parameters between non-identical objects,
it is favorable to normalize the (cross-)correlation results. We follow an approach which is widely known from
the normalized cross-correlation of 2D images: First, we subtract the mean from both functions prior to the
correlation and then divide the results by the variances:
\begin{equation}
{\cal SH}_{corr-norm}({\cal R}) := \int\limits_{S^2} \frac{(f-\overline{f}) ({\cal R} (g-\overline{g}))}{\sigma_f\sigma_g} \text{\quad} \sin{\Theta}d\Phi d\Theta. 
\label{eq:feature:SHcorrnormFinal}
\end{equation}
Analogous to Fourier transform, we obtain the expected values $\overline{f}$ and $\overline{g}$ directly from the 0th
${\cal SH}$ coefficient. The variances $\sigma_f$ and $\sigma_g$ can be estimated from the band-wise energies:
\begin{equation}
\sigma_f \approx \sqrt{\sum\limits_l |\widehat{f_l}|^2}.
\label{eq:feature:SHcorrnormFinal2}
\end{equation}

\subsection{\label{sec:feature:shcorrradii}Simultaneous Correlation of Signals on Concentric Spheres}
In many applications we consider local signals which are spread over the surfaces of several concentric spheres with different radii.
Instead of computing the correlation for each surface separately, we can simply extend (\ref{eq:feature:SHcorrFinal}) to compute
the correlation over all signals at once.\\
This can be achieved by the use of a single correlation matrix $C^{\#}$. We simply add the $\widehat{{\cal SH}_{corr}}(m, h, m')$ 
(\ref{eq:feature:SHcorrFT}) for all radii and retrieve the combined correlation matrix $C^{\#}$ via inverse FFT as before.

\subsection{\label{sec:feature:shcorrcomplex}Complexity}
Following the implementation given in section \ref{sec:feature:implement}, we obtain the harmonic expansion to band $b_{\max}$ at each
point of a volume with $m$ voxels in $O(m(b_{\max})^2 + (m \log m))$. Building the correlation matrix $\widehat{C^{\#}}$ at each point 
takes $O((2b_{\max})^4)$ plus the inverse FFT in $O((b_{\max}+p)^3 \log (b_{\max}+p)^3)$.  

\paragraph{Parallelization:} Further speed-up can be achieved by parallelization (see section \ref{sec:feature:implement}): the 
transformation into the harmonic domain can be parallelized as well as the point-wise computation of $\widehat{C^{\#}}$.

\section{\label{sec:feature:shconvolve}Convolution in ${\cal SH}$}
\index{Convolution}
After the fast correlation has been introduced, it is obvious to also take a look at the convolution in the harmonic domain. If we are only
interested in the result of the convolution of two signals at a given fixed rotation, we can apply the so-called ``left''-convolution.
\subsection{\label{sec:feature:shleftconvolve}``Left''-Convolution}
\index{Convolution}
We define the ``left''-convolution of two spherical functions $f$ and $g$ in the harmonic domain as $\widehat{f} * \widehat{g}$.
Following the Convolution Theorem this convolution is given as:
\begin{equation}
\left(\widehat{f}* \widehat{g}\right)^l_m = 2\pi\sqrt{\frac{4\pi}{2l+1}}\widehat{f}^l_m\cdot \widehat{g}^l_0.
\label{eq:feature:shleftconvolve}
\end{equation}
Note that this definition is asymmetric and performs an averaging over the translations (rotations) of the ``left'' signal.\\

The ``left''-convolution is quite useful, but for our methods we typically encounter situations like in the case of the fast correlation,
where we need to evaluate the convolution at all possible rotations of two spherical functions.  

\subsection{\label{sec:feature:fastshconvolve}Fast Convolution over all Angles}
\index{Convolution}
Following the approach used for the fast correlation, we introduce a method for the fast computation of full convolutions over all angles on
the sphere in a very similar way:\\
Again, the full convolution function ${\cal SH}_{conv}:{\cal SO}(3)\rightarrow \mathbb{R}$ of two signals $f$ and $g$ under the rotation ${\cal R}\in {\cal SO}(3)$ on a 2-sphere
is
given as:
\begin{equation}
{\cal SH}_{conv}({\cal R}) := \int\limits_{S^2} f  ({\cal R} \overline{g}) \sin{\Theta}d\Phi d\Theta. 
\label{eq:feature:shleftconvolve2}
\end{equation}
Applying the same steps as in the case of the correlation, we obtain a convolution matrix:
\begin{equation}
\label{eq:feature:SHconvFT}
\widehat{{\cal C}^*}(m, h, m') = \sum\limits_{l=0}^\infty d^l_{mh}(\pi/2) d^l_{hm'}(\pi/2) \widehat{f}^l_m \widehat{g}^l_{m'}.
\end{equation}

Analog to equation (\ref{eq:feature:SHcorrFinal}),
\begin{equation}
\label{eq:feature:SHconvFinal}
{\cal C}^*(\xi, \eta, \omega) = {\cal F}^{-1}(\widehat{C^*}),
\end{equation}
an inverse Fourier transform reveals the convolution $f * g$ for each possible rotation in the three dimensional ${\cal C}^{*}(\xi, \eta, \omega)$-space.\\

Regarding computational complexity and angular resolution, this convolution method shares all the properties of the fast correlation
(see sections \ref{sec::feature:SHcooranglarres} to \ref{sec:feature:Shcorrexperiment}). 

\section{\label{sec:feature:VH}Vectorial Harmonics}
\index{Vectorial Harmonics}
So far, we have exploited and utilized the nice properties of the harmonic expansion of scalar valued functions on $S^2$ in Spherical
Harmonics to derive powerful methods like the fast correlation. These methods can be operated on single scalar input in form of gray-scale 
volumes, which is one of the most common data types in 3D image analysis. But there are two equally important data types: multi-channel 
scalar input (e.g. RGB colored volumes) and 3D vector fields (e.g. from gradient data).\\
In the first case, a harmonic expansion of multi-channel scalar input is straightforward: since the channels are not affected 
independently, one can simply combine the Spherical Harmonic expansions of each individual channel (e.g. see section \ref{sec:feature:SH}).\\
For 3D vector fields, the harmonic expansion turns out to be less trivial, i.e. if we rotate the field, we are not only changing the 
position of the individual vectors, but we also have to change the vector values accordingly. This dependency can be modeled by the use
of Vectorial Harmonics (${\cal VH}$).\\

\begin{figure}[htbp]
\centering
\begin{tabular}[t]{l|cccc|c}
$l$ & $m=0$ & $m=1$ & $m=2$ & $m=3$ & k\\
\hline
\multirow{1}{*}{0}
 & \includegraphics[width=0.22\textwidth]{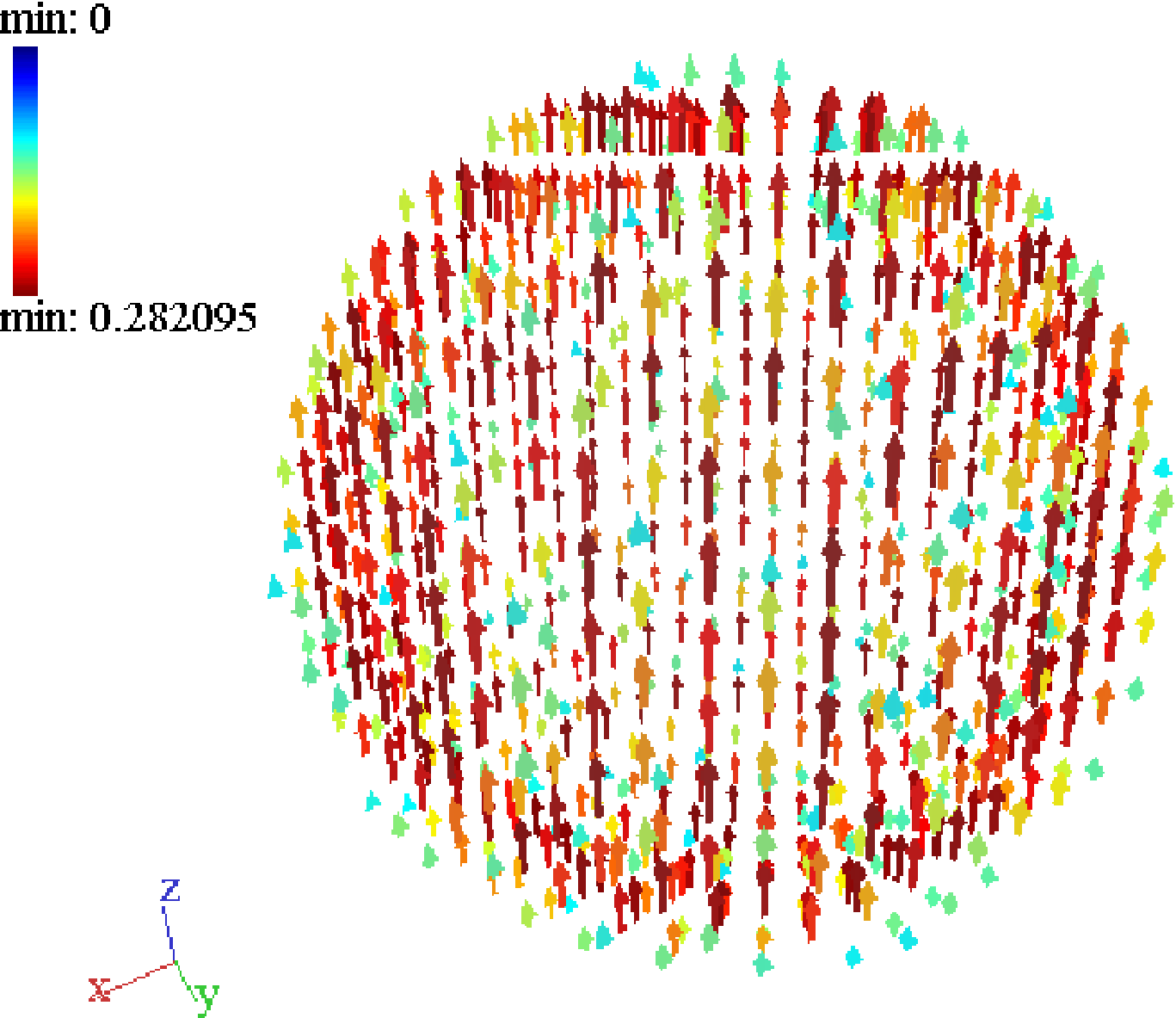}
 & \includegraphics[width=0.22\textwidth]{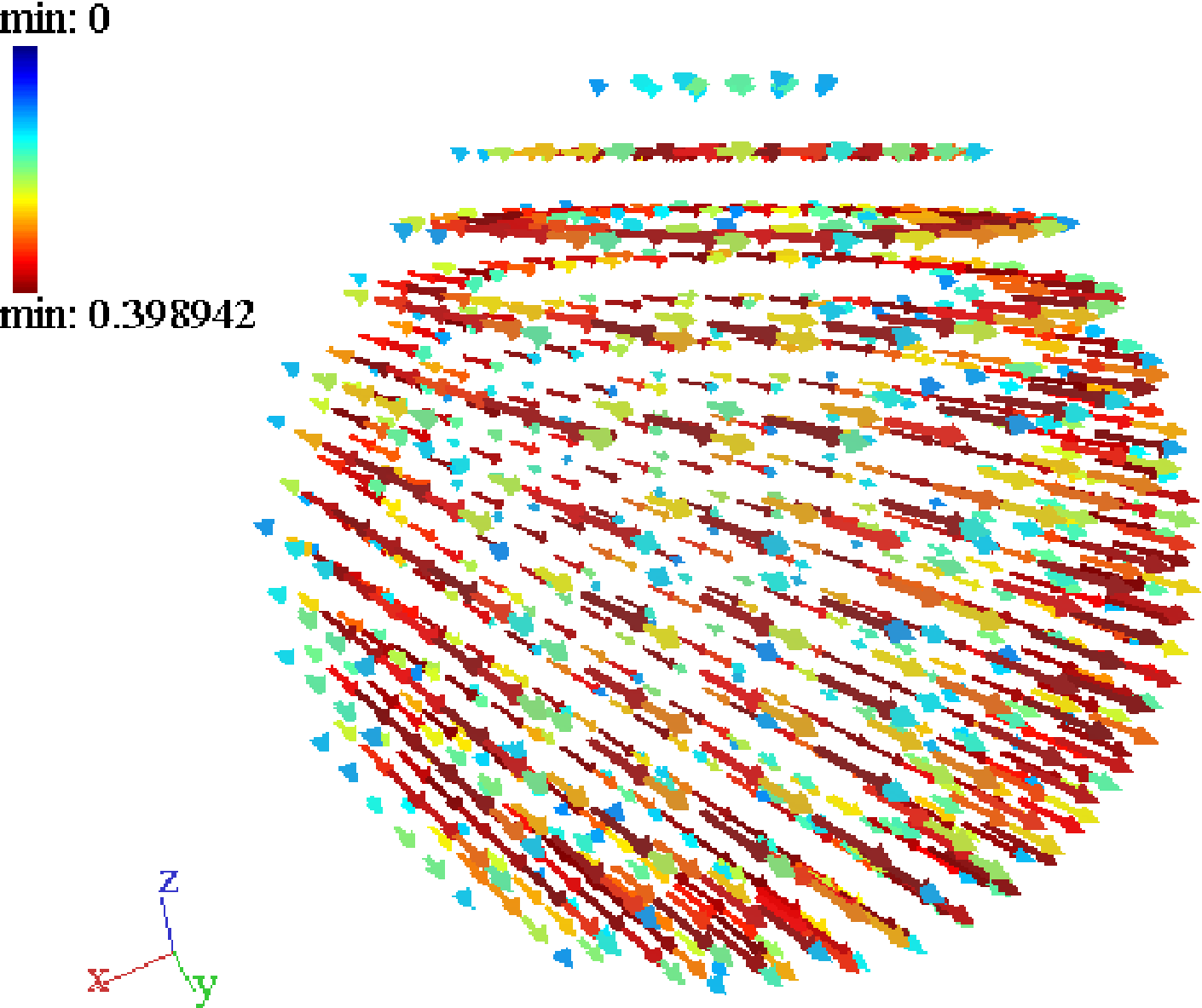}
 &
 &
 & $1$
 \\
 \hline
\multirow{3}{*}{1}
 & \includegraphics[width=0.22\textwidth]{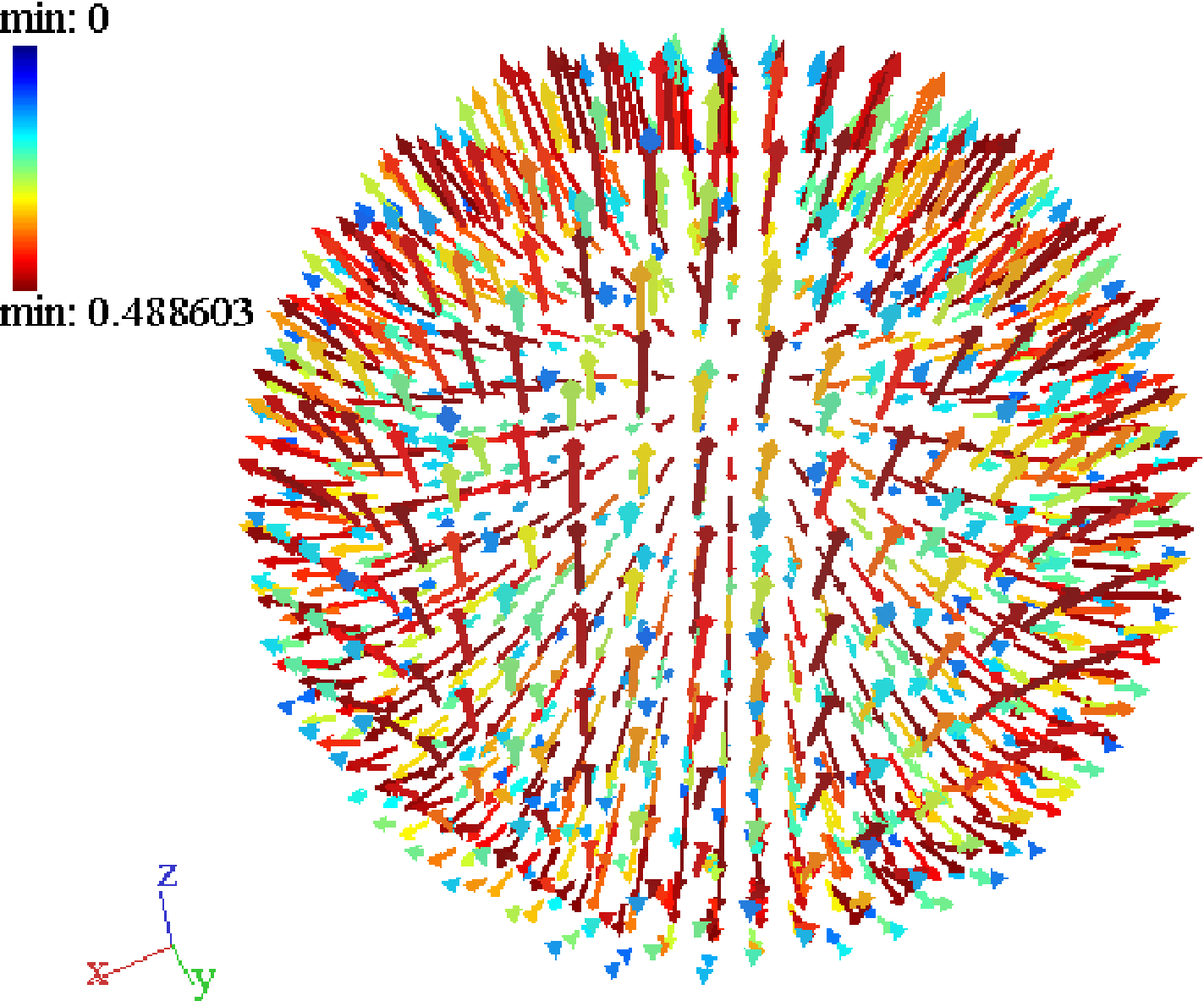}
 &
 &
 &
 & $-1$
 \\
 & \includegraphics[width=0.22\textwidth]{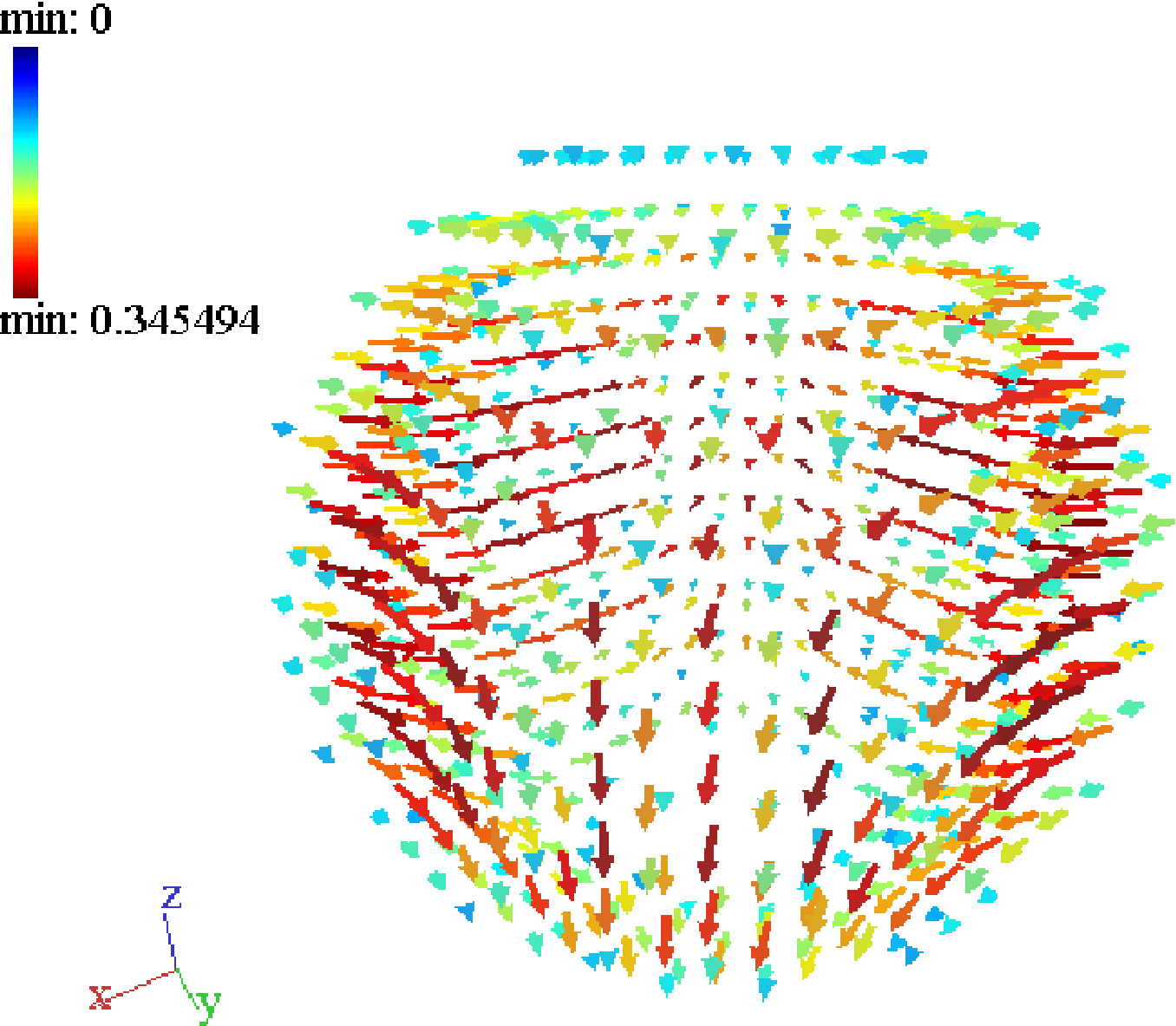}
 & \includegraphics[width=0.22\textwidth]{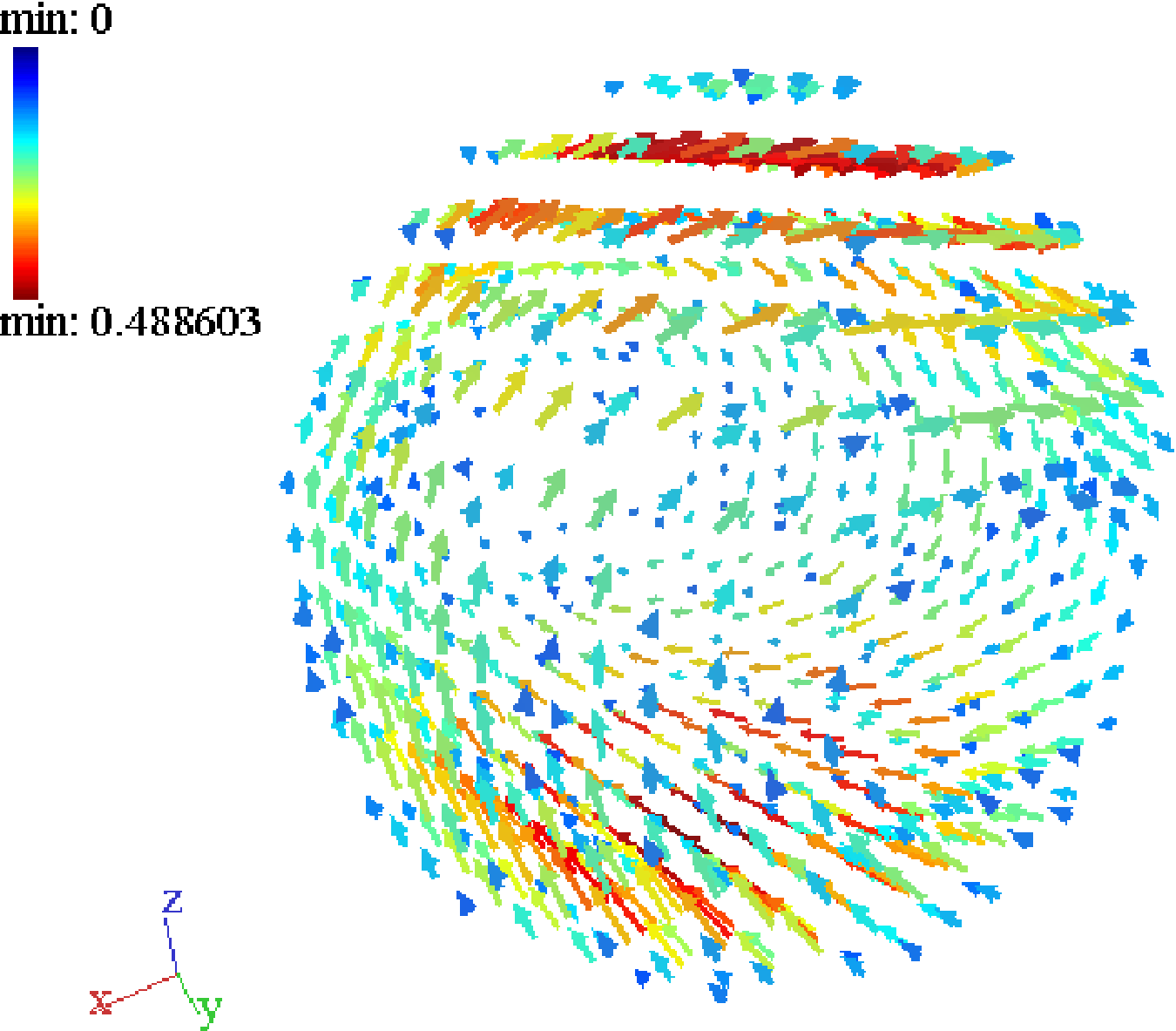}
 &
 &
 & $0$
\\
 & \includegraphics[width=0.22\textwidth]{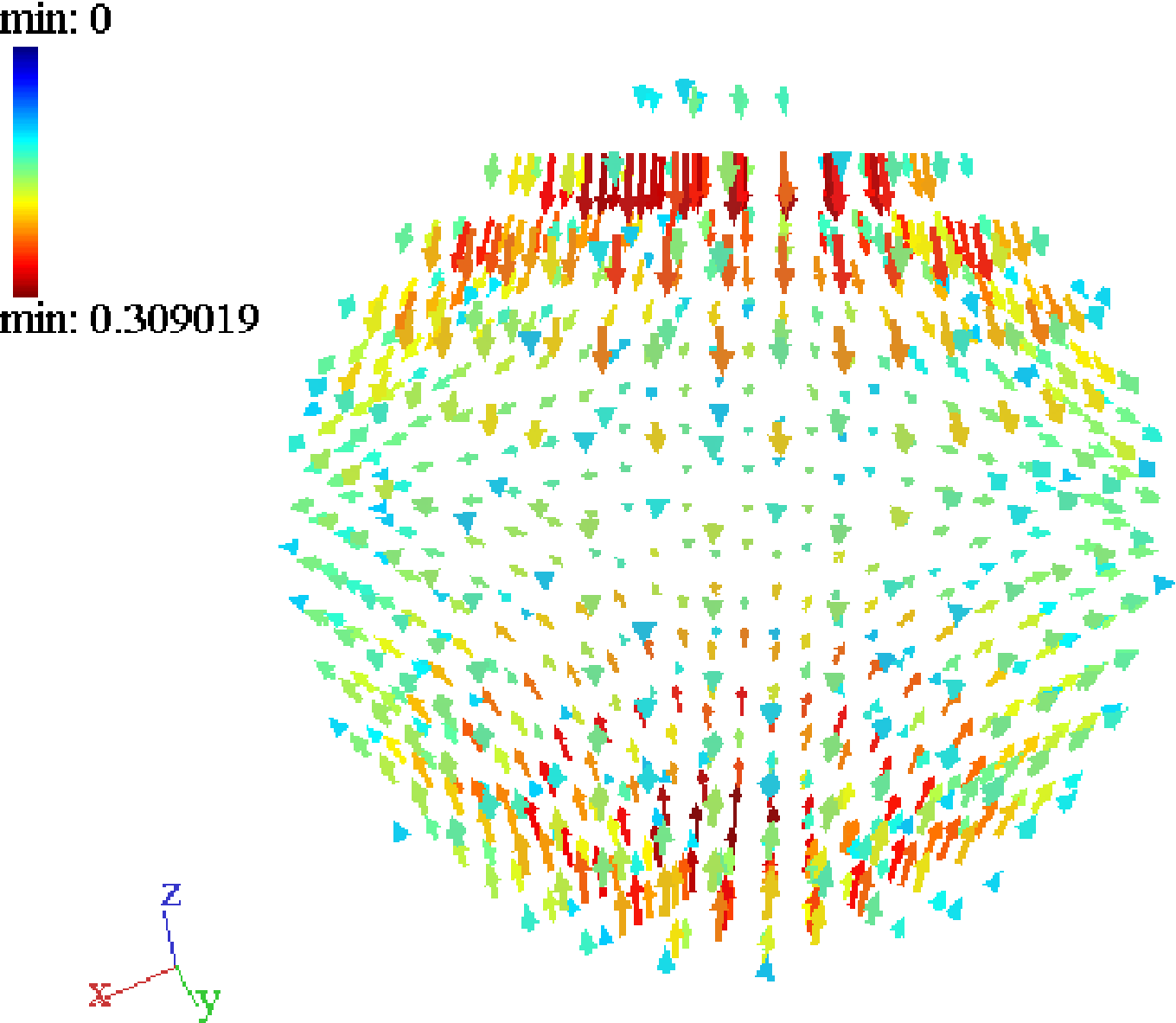}
 & \includegraphics[width=0.22\textwidth]{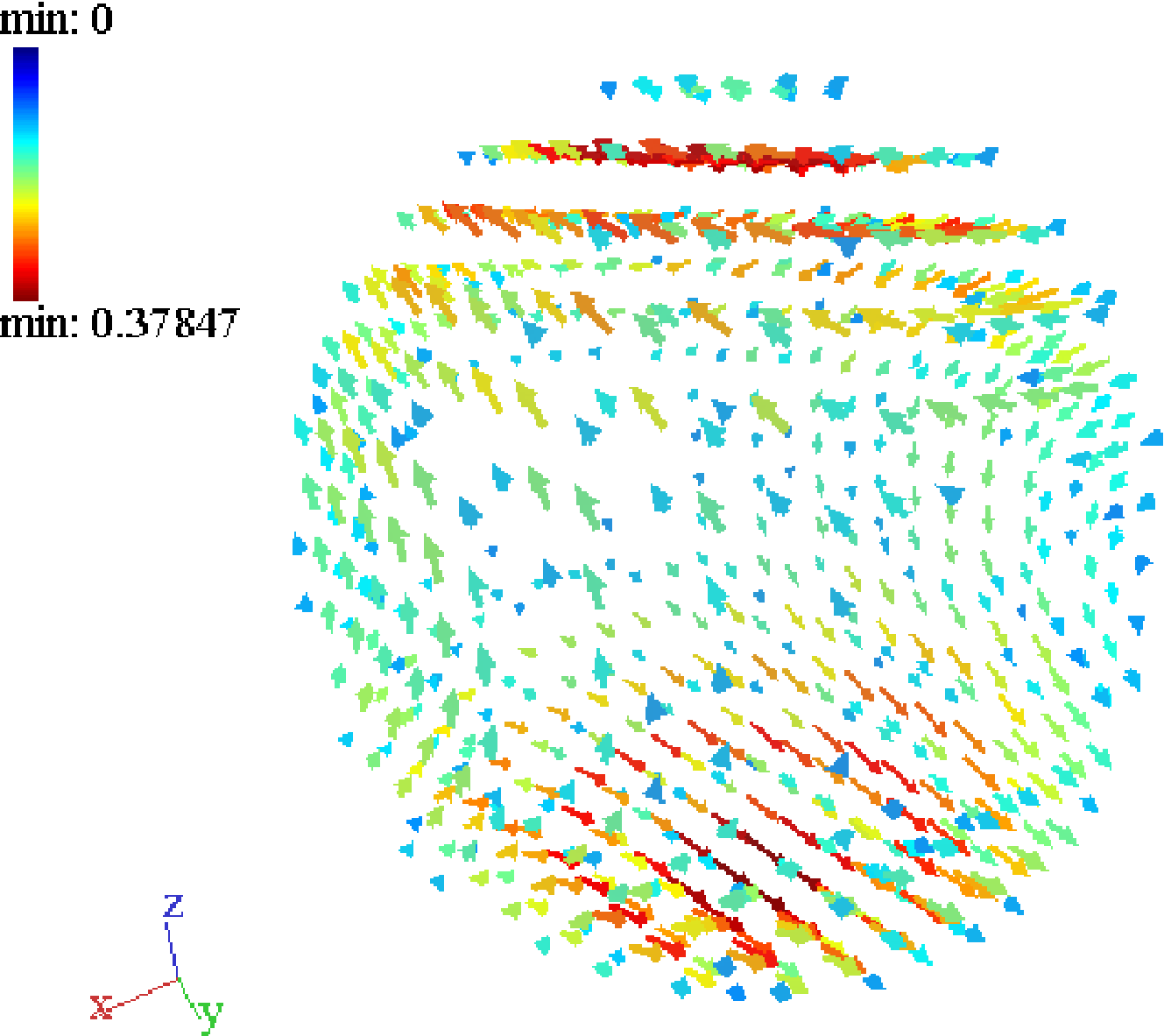}
 & \includegraphics[width=0.22\textwidth]{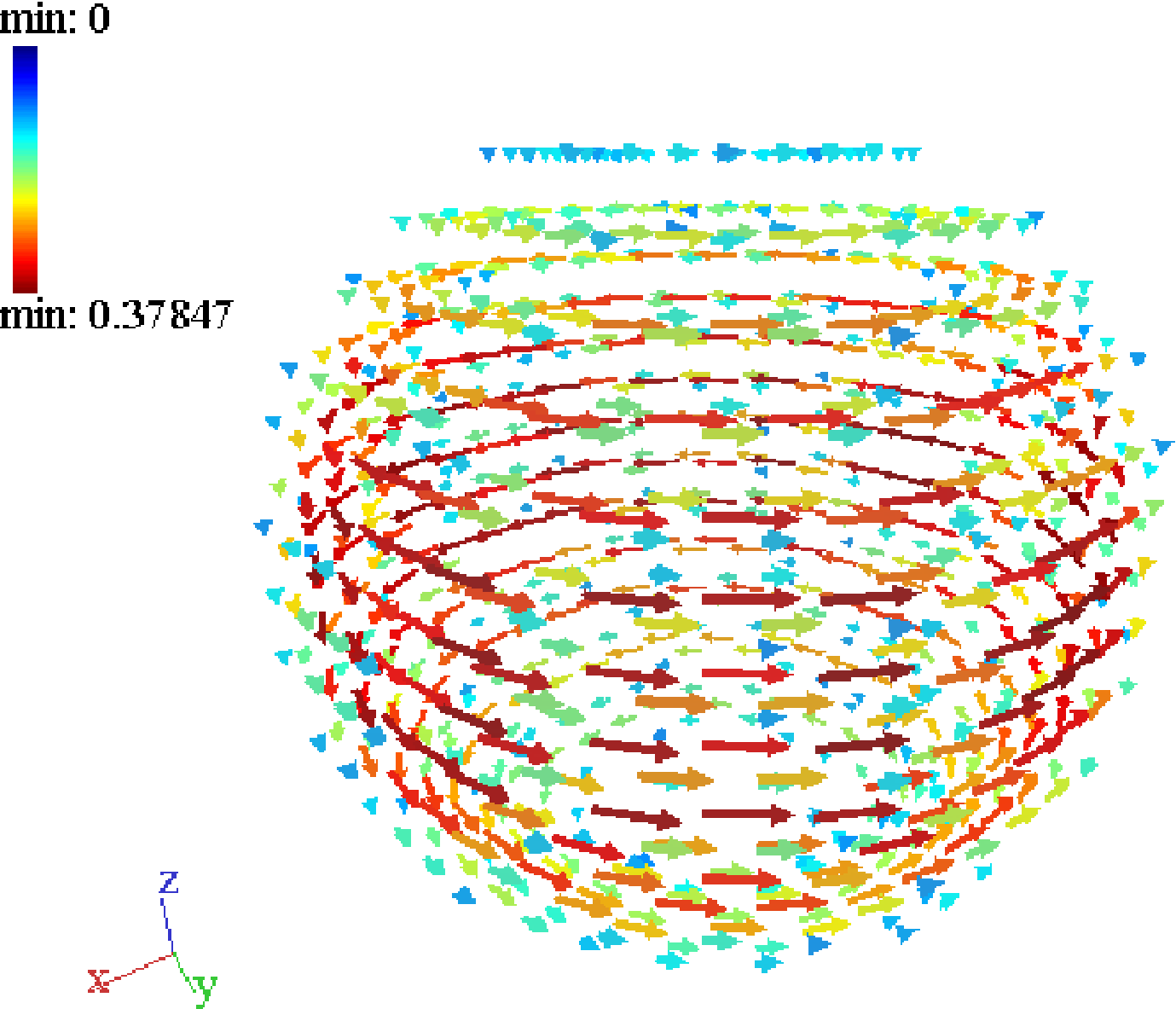}
 &
 & $1$
\\
 \hline
\multirow{3}{*}{2}
 & \includegraphics[width=0.22\textwidth]{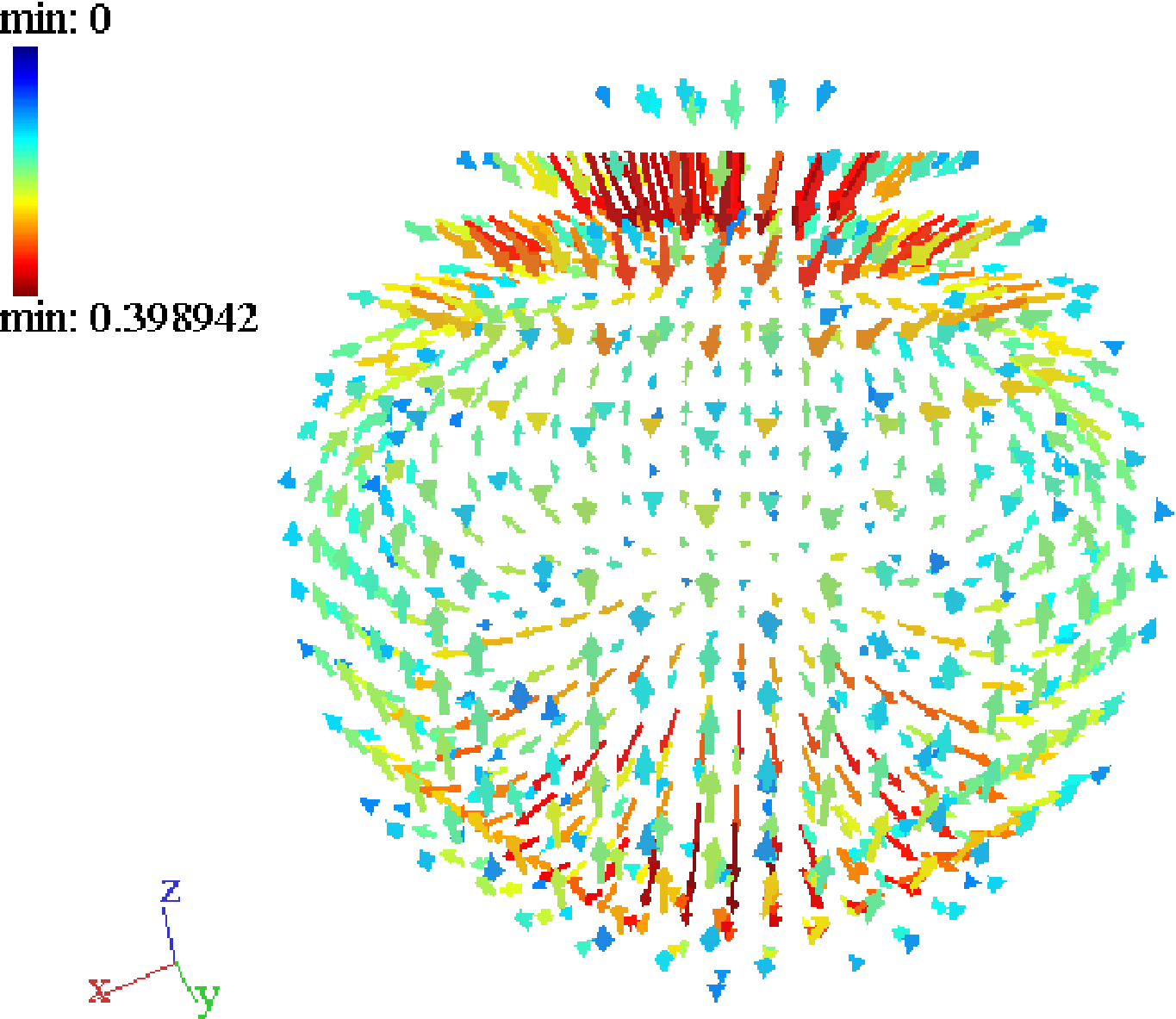}
 & \includegraphics[width=0.22\textwidth]{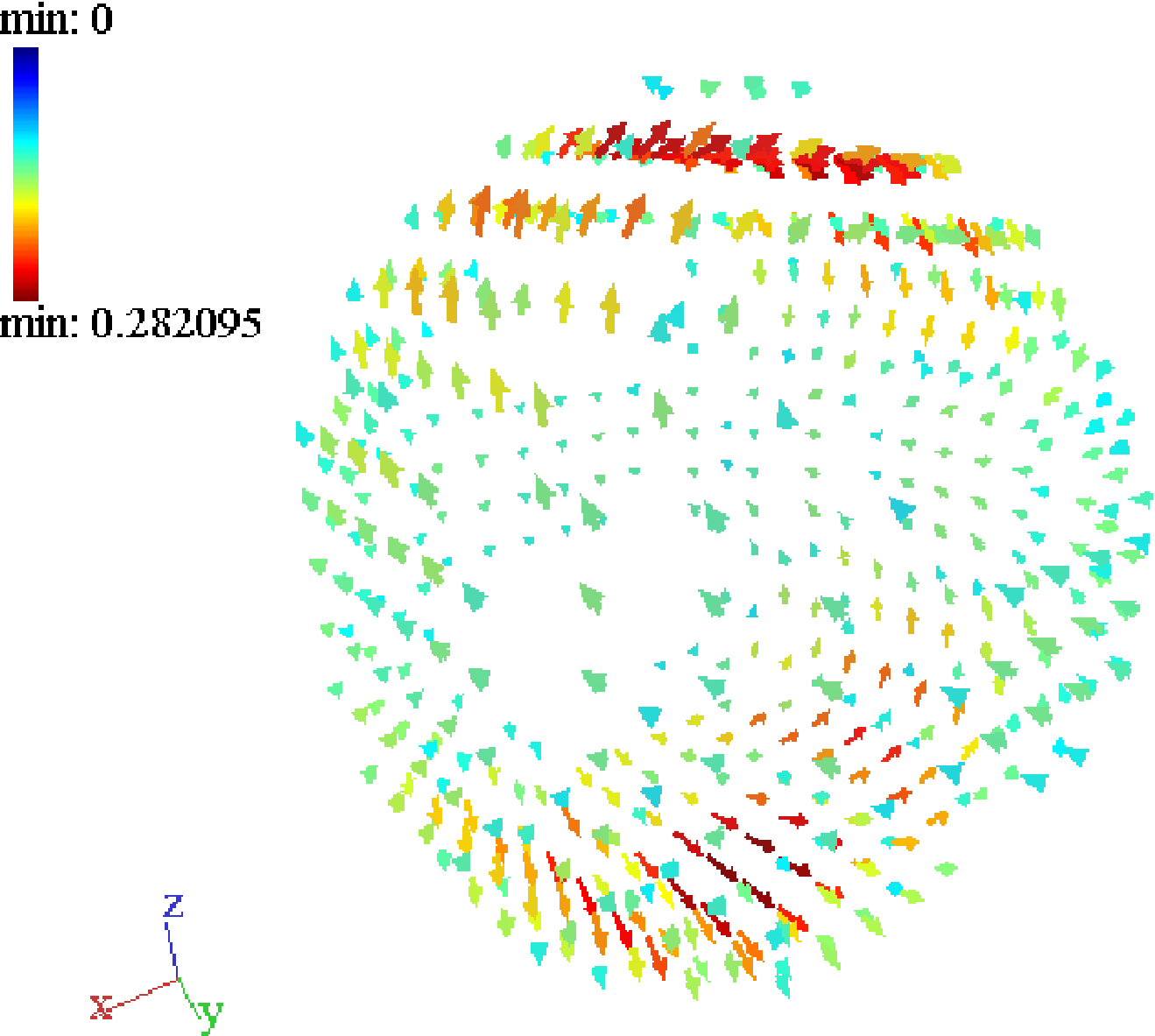}
 &
 &
 & $-1$
 \\
 & \includegraphics[width=0.22\textwidth]{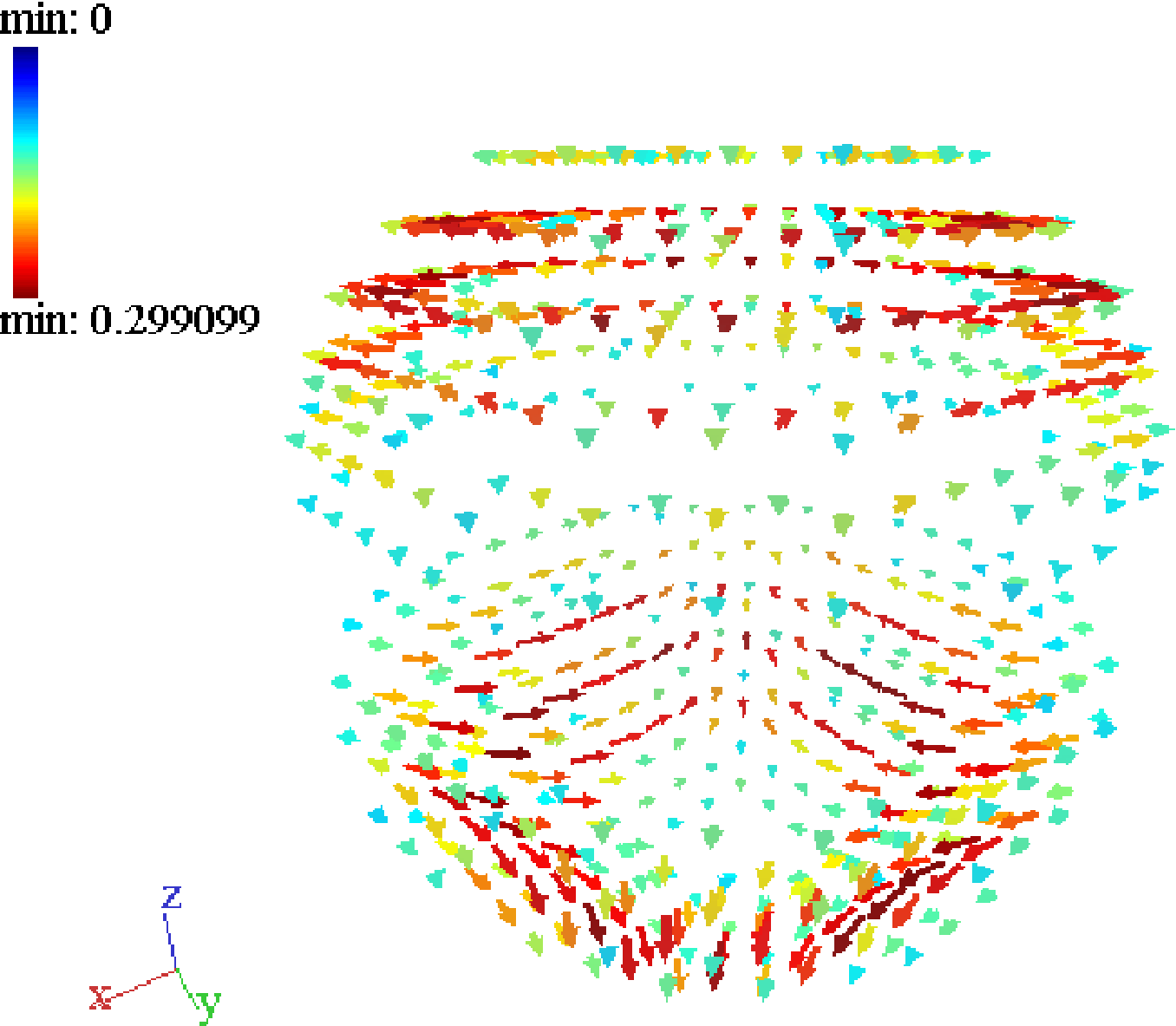}
 & \includegraphics[width=0.22\textwidth]{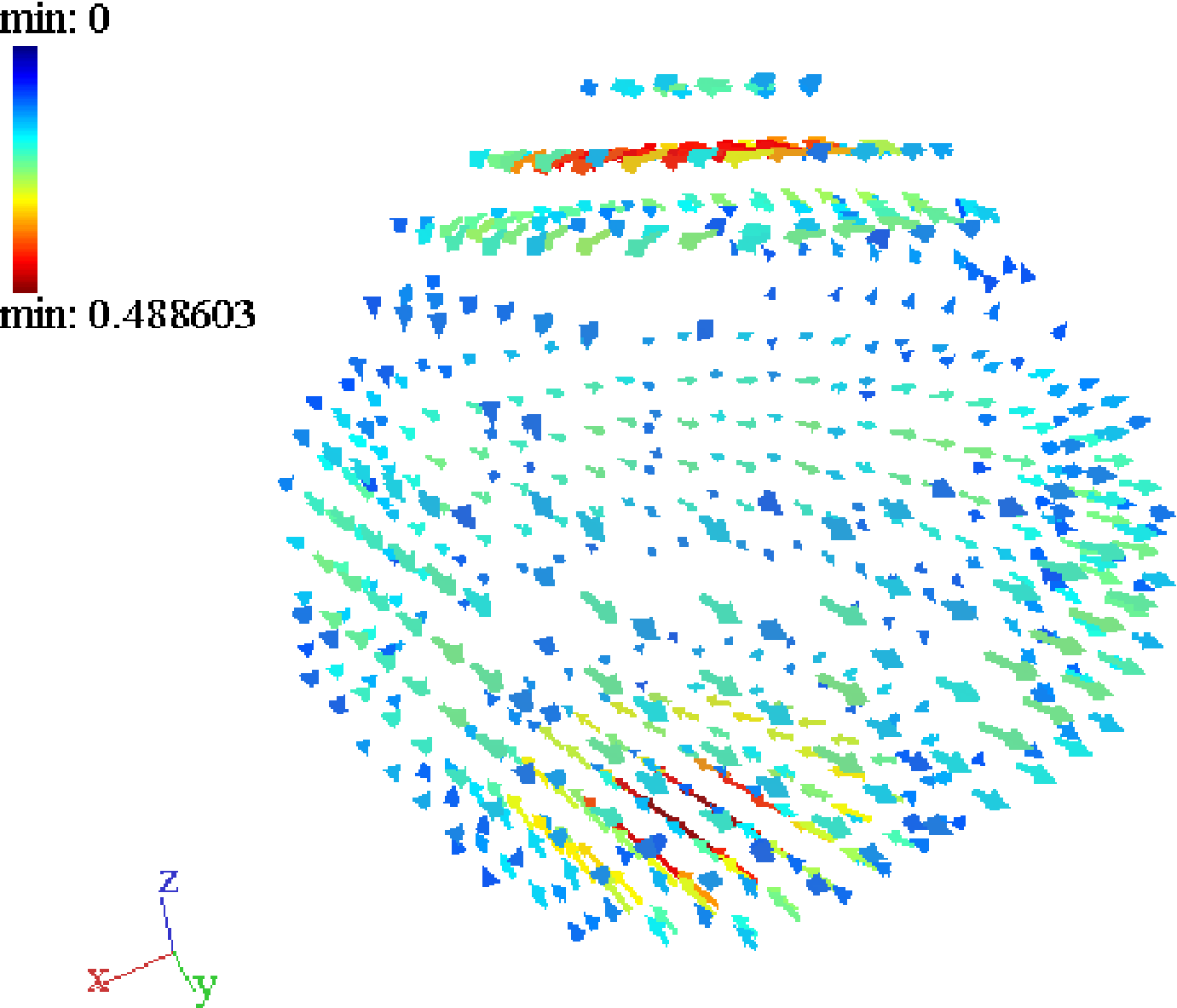}
 & \includegraphics[width=0.22\textwidth]{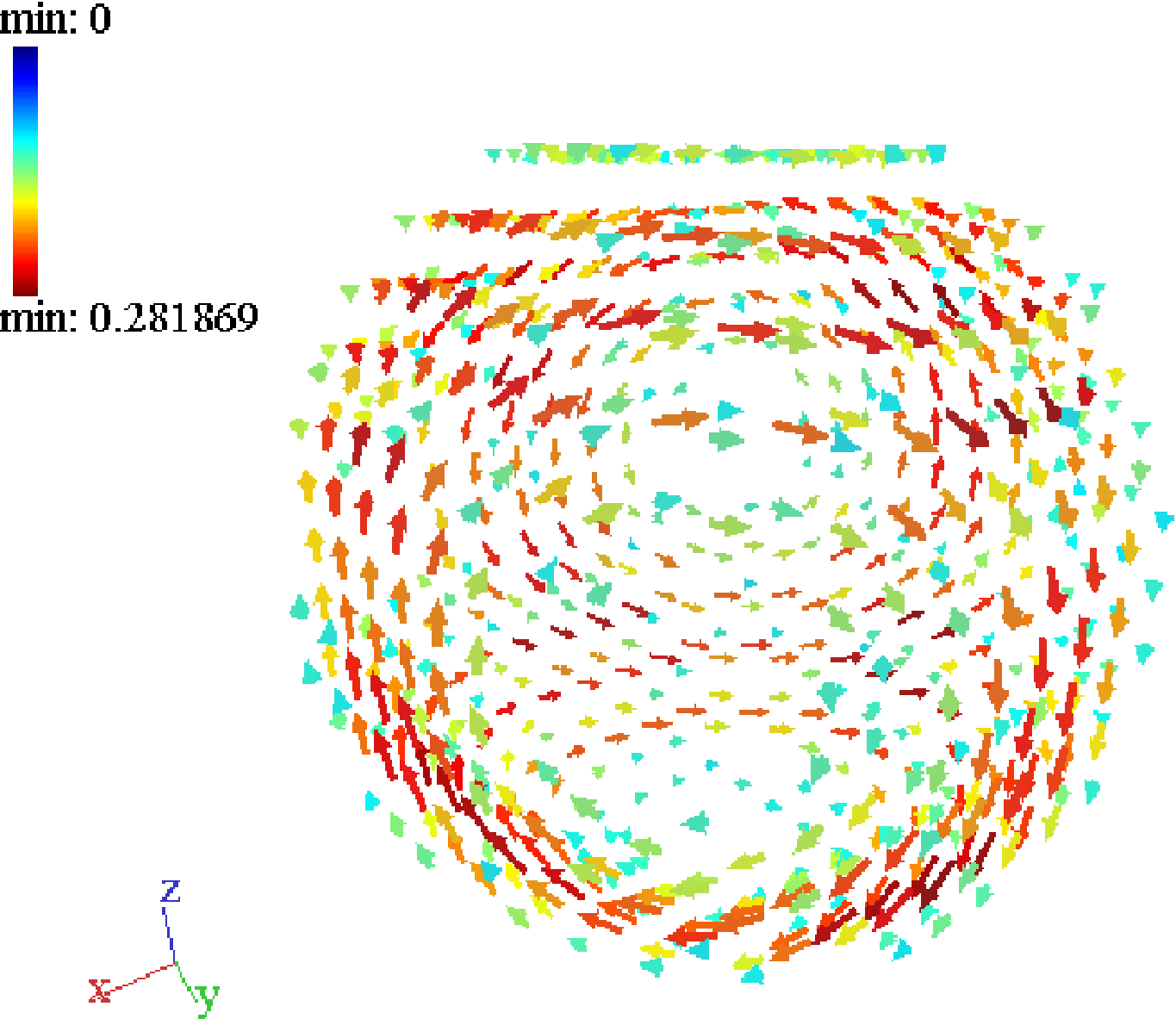}
 &
 & $0$
\\
 & \includegraphics[width=0.22\textwidth]{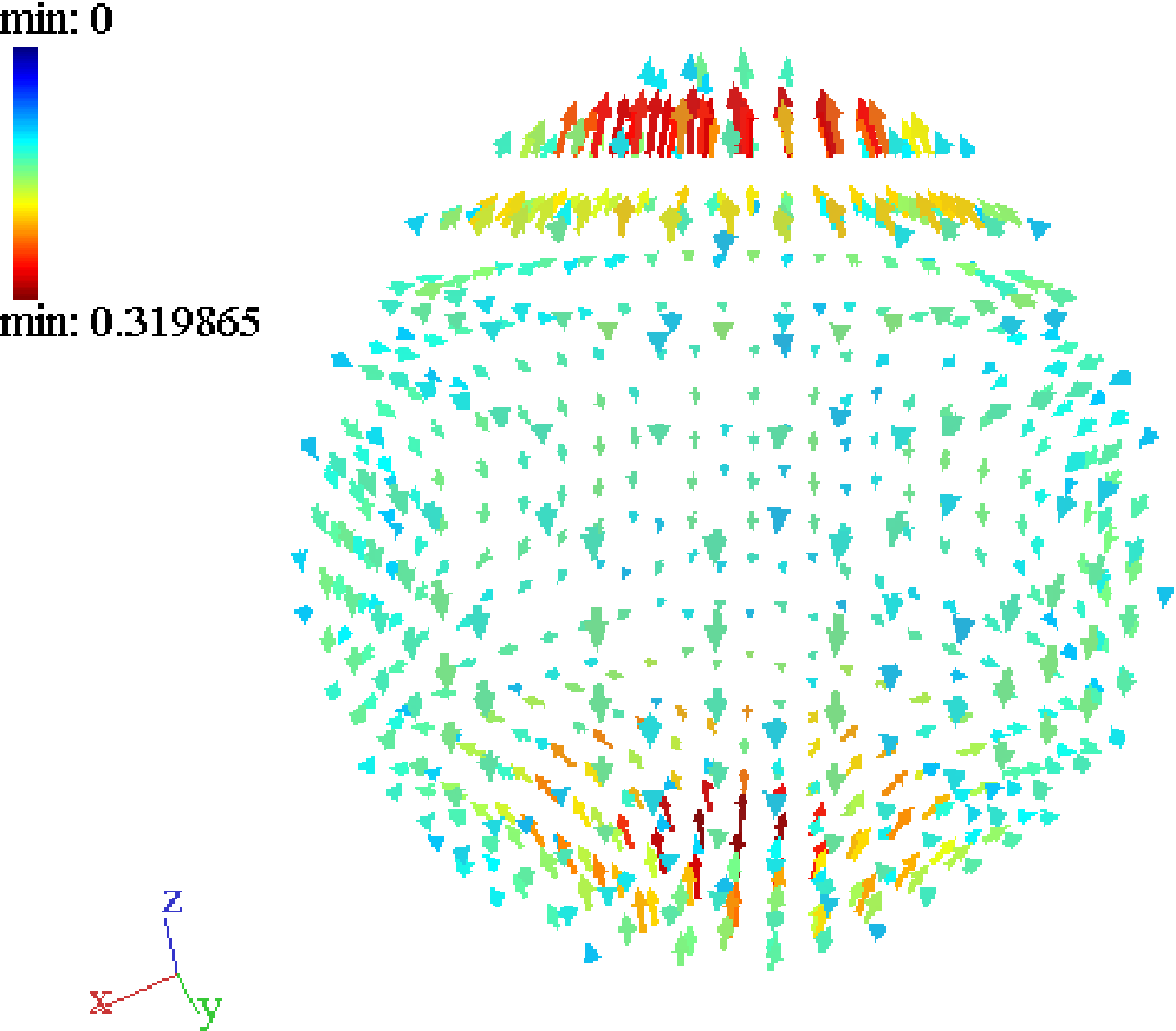}
 & \includegraphics[width=0.22\textwidth]{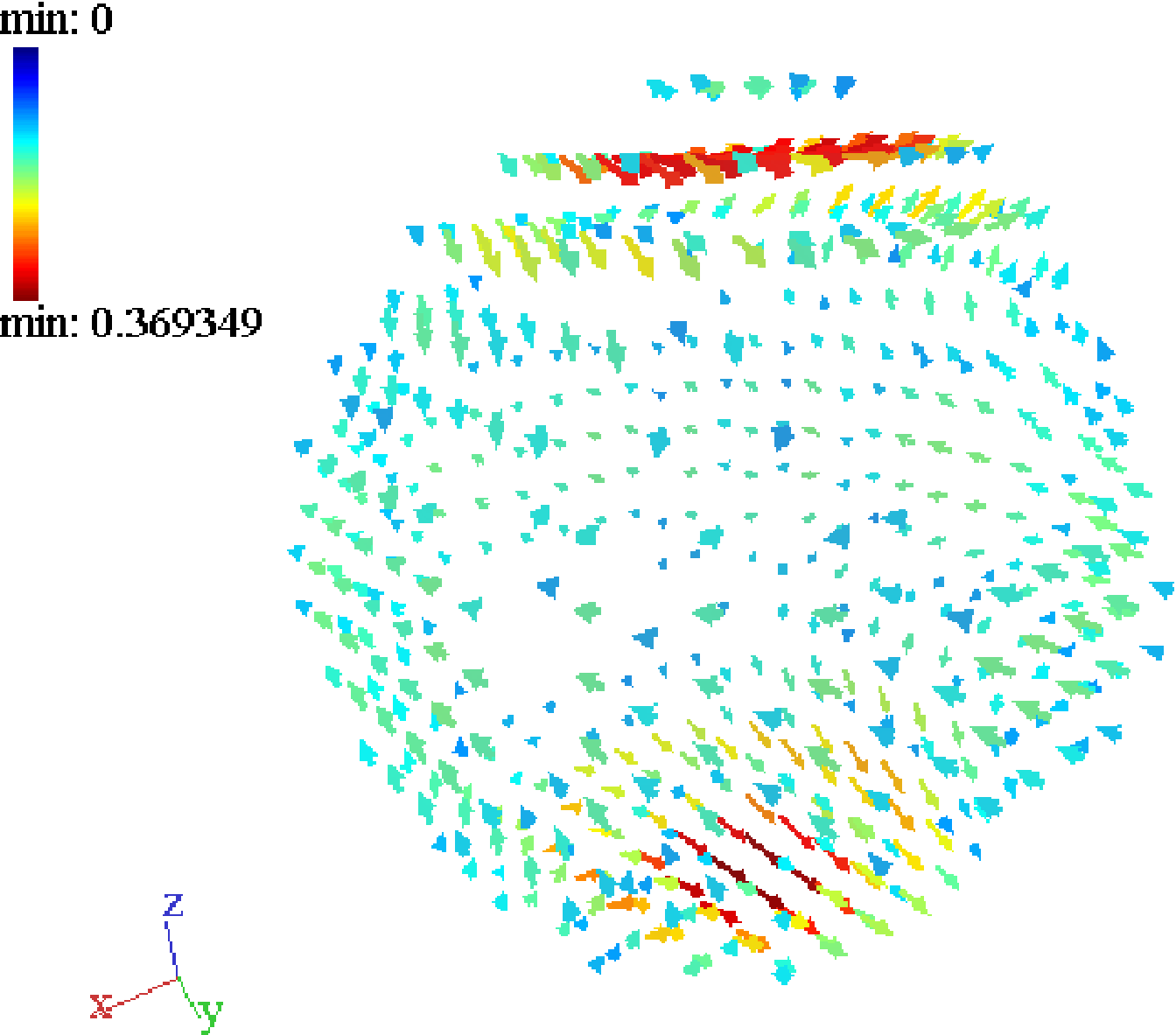}
 & \includegraphics[width=0.22\textwidth]{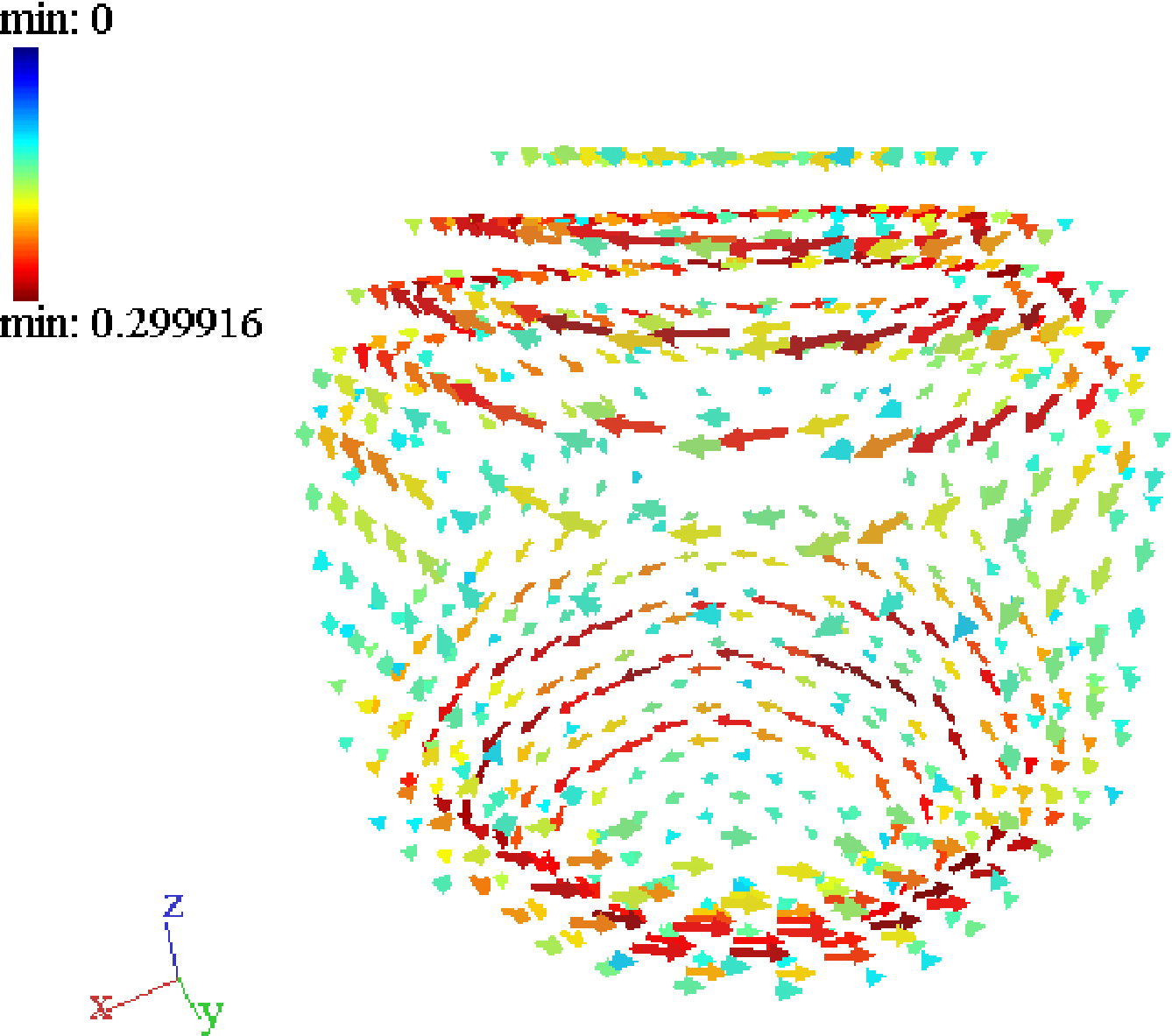}
 & \includegraphics[width=0.22\textwidth]{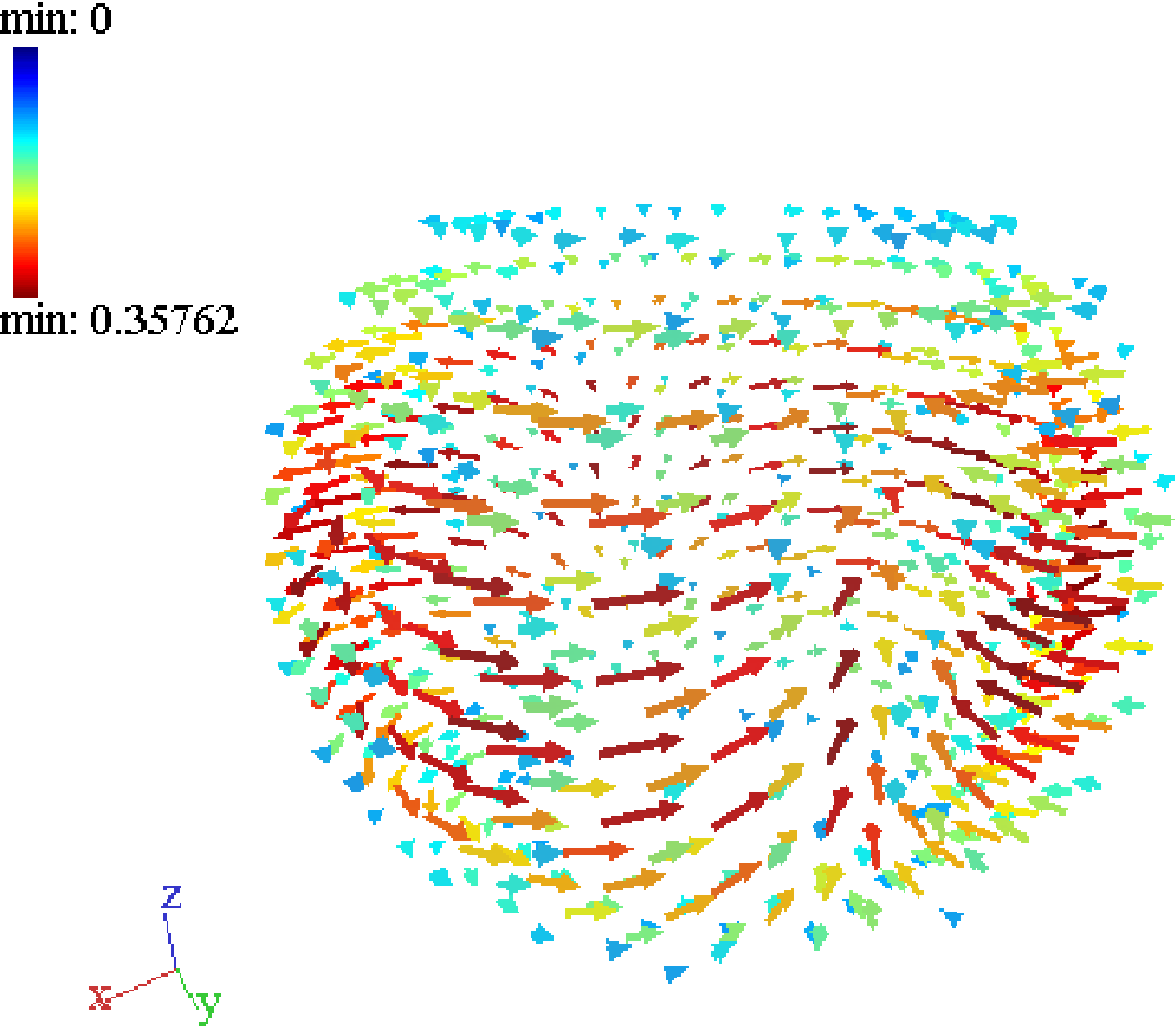}
 & $1$
\\
 \hline
\end{tabular}
\caption[Vectorial Harmonic base functions]{\label{fig:feature:VHbase} The first Vectorial Harmonic base functions to $l=2$. Due to space
limitations we only visualize functions with $0\leq m \leq l+k$ instead of the actual $-(l+k)\leq m \leq l+k$.}
\end{figure}

Given a vector valued function ${\bf f}:S^2\rightarrow \mathbb{R}^3$ with three vectorial components $[x,y,z] = {\bf f}(\Phi,\Theta)$ and parameterized in
Euler angles (Fig. \ref{fig:feature:eulerschema}) $\phi,\theta,\psi$, we can expand ${\bf f}$ in Vectorial Harmonics:
\begin{equation}
{\bf f}(\Phi,\Theta) = \sum_{l=0}^\infty\sum_{k=-1}^{1}\sum_{m=-(l+k)}^{(l+k)} \widehat{{\bf f}^l_{km}} {\bf Z }^l_{km}(\Phi,\Theta),
\label{eq:feature:VecHarmonics}
\end{equation}
with scalar harmonic coefficients $ \widehat{\bf f}^l_{km}$ and the orthonormal base functions:
\begin{equation}
\bf{Z}^l_{km} =
\left(
\begin{tabular}{rll}
$\langle1$ \ $1 $&$| l+k$ \ $m, l$ \ $1-m\rangle $&$Y^l_{1-m}$\\
$\langle1$ \ $0 $&$| l+k$ \ $m, l$ \ $-m\rangle $&$Y^l_{-m}$\\
$\langle1$ \ $-1 $&$| l+k$ \ $m, l$ \ $-1-m\rangle $&$Y^l_{-1-m}$
\end{tabular}
\right)^T.
\label{eq:feature:VecHarmonicBase}
\end{equation}
Figure \ref{fig:feature:VHbase} visualizes the first two bands of these base functions as vector fields on a sphere. We define the 
forward Vectorial Harmonic transformation as
\begin{eqnarray}
\label{eq:feature:VHforward}
{\cal VH}({\bf f}) := \widehat{{\bf f}}, \text{\quad with \quad} \widehat{{\bf f}^l_{km}} &=& 
\int\limits_{\Phi,\Theta} \overline{\bf Z}^l_{(-1)m}(\Phi,\Theta) {\bf f}[-1](\Phi,\Theta) \sin{\Theta}d\Phi d\Theta \nonumber \\
&+& \int\limits_{\Phi,\Theta} \overline{\bf Z}^l_{(0)m}(\Phi,\Theta) {\bf f}[0](\Phi,\Theta) \sin{\Theta}d\Phi d\Theta \nonumber  \\
&+& \int\limits_{\Phi,\Theta} \overline{\bf Z}^l_{(1)m}(\Phi,\Theta) {\bf f}[1](\Phi,\Theta) \sin{\Theta}d\Phi d\Theta, 
\end{eqnarray}
where ${\bf f}[-1]$ returns the scalar function on $S^2$ which is defined by the complex transformation (\ref{eq:feature:complexvector}) 
of the $z$ component of the vector-valued ${\bf f}$.
The backward transformation in is defined as:
\begin{equation}
{\cal VH}^{-1}\left(\widehat{{\bf f}}(\Phi,\Theta)\right) := \sum\limits_{l=0}^\infty\sum\limits_{k=-1}^1\sum\limits_{m=(l+k)}^{(l+k)} 
\widehat{{\bf f}^l_{km}} {\bf Z}^l_{km}(\Phi,\Theta).
\label{eq:feature:VHbackward}
\end{equation}
In our case, the Vectorial Harmonics are defined to operate on vector fields with complex vector coordinates.
For fields of real valued vectors ${\bf r}(x,y,z)\in \mathbb{R}^3$, we need to transform the vector coordinates to $\mathbb{C}^3$
according to the Spherical Harmonic relation:
\begin{equation}
\label{eq:feature:complexvector}
{\bf u} \in \mathbb{C}^3:
{\bf u} :=
\left(
\begin{tabular}{l}
$\frac{x-iy}{\sqrt{2}}$\\
$z$\\
$\frac{x+iy}{\sqrt{2}}$
\end{tabular}
\right).
\end{equation}

\subsection{Deriving Vectorial Harmonics}
There have been several different approaches towards Vectorial Harmonics, like \cite{VHref1} or \cite{VHref2}. All use a slightly different 
setting and notation. For our purposes, we derive our methods from a very general theory of Tensorial Harmonics \cite{TensoHarmonics}, 
which provides expansions for arbitrary real valued tensor 
functions ${\bf f}$ on the 2-sphere: 
\begin{equation}
{\bf f}(\Phi,\Theta) := \sum_{l=0}^\infty\sum_{k=-d}^{d}\sum_{m=-(l+k)}^{(l+k)} \widehat{\bf f^l_{km}} {\bf Z}^l_{km}(\Phi,\Theta),
\label{TensorialHarmonics}
\end{equation}
where $\widehat{\bf f}^l_{km}$ is the expansion coefficient of the $l$-th band of tensor order $d$ and harmonic order $m$.
The orthonormal Tensorial Harmonic base functions ${\bf Z}^l_{km}$ are given as:
\begin{equation}
{\bf Z}^l_{km} := {\bf e}^{(l+k)}_m \circ_1 Y^l,
\label{eq:feature:VHbasecomp}
\end{equation}
with the Spherical Harmonic bands $Y^l$. The ${\bf e}^{l}_m$ are elements of the standard Euclidean base of
${\mathbb C}^{2d+1}$, and $\circ_l$ denotes a bilinear form connecting tensors $V_{l_1}$ and $V_{l_2}$ of different ranks:
\begin{equation}
\circ_d: V_{l_1} \times V_{l_2} \rightarrow {\mathbb C}^{2d+1},
\end{equation}
where $l_1,l_2 \in {\mathbb N}$ have to hold $|l_1-l_2|\leq l \leq l_1+l_2$. $\circ_l$ is computed as follows:
\begin{equation}
({\bf e}^{l}_m)^T ({\bf v} \circ_l {\bf u}) := \sum\limits_{m=m_1+m_2} \langle lm|l_1m_1,l_2m_2 \rangle v_{m_1} u_{m_2}.
\label{eq:feature:VHbilinearform}
\end{equation}
See \cite{TensoHarmonics2} for details and proofs.\\

If we limit the general form to tensors of order one ($d:=1$) and use \ref{eq:feature:VHbilinearform} for the computation of the
base functions \ref{eq:feature:VHbasecomp},  we directly obtain Vectorial Harmonic expansions as in 
\ref{eq:feature:VecHarmonics}.

\subsection{Useful Properties of Vectorial Harmonics}
Vectorial Harmonics inherit most of the favorable properties of the underlying Spherical Harmonics, such as orthonormality.
\paragraph{Orthonormality:}
\begin{equation}
\int\limits_{\Phi,\Theta} \left({\bf Z}^l_{km}(\Phi,\Theta)\right)^T {\bf Z}^{l'}_{k'm'}(\Phi,\Theta) \sin{\Theta}d\Phi d\Theta =
\frac{4\pi}{(1/3)(2l+1)(2(l+k)+1)}\delta_{l,l'}\delta_{k,k'}\delta_{m,m'}.
\end{equation}

\section{\label{sec:feature:vhrot}Rotations in Vectorial Harmonics}
\index{Rotation}\index{Vectorial Harmonics}
The analogy of Vectorial Harmonics to Spherical Harmonics continues also in the case of rotation in the harmonic domain.
Complex 3D vector valued signals $\bf{f}$ with Vectorial Harmonic coefficients $\widehat{\bf f}$ are rotated 
\cite{TensoHarmonics} by:
\begin{equation}
\label{eq:feature:VHrot}
{\cal R} {\bf f} = \sum\limits_{l=0}^\infty \sum_{k=-1}^{k=1}\sum\limits_{m=-(l+k)}^{l+k}\sum\limits_{n=-(l+k)}^{l+k} D^{l+k}_{mn}({\cal R}) 
\widehat{\bf f}^l_{km} {\bf Z}^l_{kn},
\end{equation}
which is a straightforward extension of (\ref{eq:feature:shrot}). One notable aspect is that we need to combine Wigner-D matrices
of the upper $l+1$ and lower $l-1$ bands in order to compute the still band-wise rotation of $\widehat{\bf f}^l_{km}$.
Hence, we rotate $\widehat{\bf f}^l_{km}$ by ${\cal R}(\phi,\theta,\psi)$ via band-wise multiplications:
\begin{equation}
{\bf f}'={\cal R}(\phi,\theta,\psi){\bf f} \Rightarrow \widehat{\bf f'}^l_{km} = \sum\limits_{n=-(l+k)}^{l+k} D^{l+k}_{mn}(\phi,\theta,\psi) \widehat{\bf f}^l_{km}.
\label{eq:feature:VHrotForward}
\end{equation}
Due to the use of the $zyz'$-convention, we have to handle inverse rotations with some care:
\begin{equation}
{\bf f}'={\cal R}^{-1}(\phi,\theta,\psi){\bf f} \Rightarrow \widehat{\bf f'}^l_{km} =  \sum\limits_{n=-(l+k)}^{l+k} D^{l+k}_{mn}(-\psi,-\theta,-\phi) \widehat{\bf f}^l_{km}.
\label{eq:feature:VHrotBack}
\end{equation}

\section{\label{sec:feature:vhcorr}Fast Correlation in Vectorial Harmonics}
\index{Correlation}
We use local dot-products of vectors to define the correlation under a given rotation $\cal R$ in Euler angles $\phi, \theta, \psi$ as:
\begin{equation}
({\bf f} \# {\bf g})({\cal R}) := \int\limits_{\Phi,\Theta}  \langle{\bf f}(\Phi,\Theta),{\cal R} {\bf g}(\Phi,\Theta)\rangle \text{\quad} \sin{\Theta}d\Phi d\Theta.
\label{eq:feature:autocorr0}
\end{equation}
Using the rotational properties (\ref{eq:feature:VHrot}) of the Vectorial Harmonics, we can extend the fast correlation approach (see section
\ref{sec:feature:shcorr}) from ${\cal SH}$ to ${\cal VH}$. Starting from (\ref{eq:feature:SHcorr0}) we insert (\ref{eq:feature:VHrot}) into (\ref{eq:feature:SHcorr1}) and obtain:
\begin{equation}
\label{eq:feature:VHcorr1}
{\cal VH}_{corr}({\cal R}) = \sum_{l=0}^{l=\infty}\sum_{k=-1}^{k=1}\sum_{m,n=-(l+k)}^{(l+k)} \overline{D^{l+k}_{mn}(\cal R)}  
\widehat{\bf f}^l_{km} \overline{\widehat{\bf g}^l_{kn}}.
\end{equation}
Analogous to (\ref{eq:feature:SHcorr2}), substituting (\ref{eq:feature:substD}) into (\ref{eq:feature:VHcorr1}) provides the final 
formulation for the correlation function regarding the new angles $\xi, \eta$ and $\omega$:
\begin{eqnarray}
\label{eq:feature:VHcorr2}
{\cal VH}_{corr}(\xi, \eta, \omega) = \sum_{l=0}^{l=\infty}\sum_{k=-1}^{k=1}\sum_{m,h,m'=-(l+k)}^{m,h,m'=(l+k)} d^{l+k}_{mh}(\pi/2)
d^{l+k}_{hm'}(\pi/2)\widehat{\bf f}^l_{km} \overline{\widehat{\bf g}^l_{km'}} \mathrm{e}^{-i(m\xi + h\eta + m'\omega)}.
\end{eqnarray}
Following (\ref{eq:feature:SHcorrFT}) we obtain the Fourier transform of the correlation matrix ${\cal C}^{\#}$ (\ref{eq:feature:VHcorr2}) 
to eliminate the missing angle parameters:
\begin{equation}
\label{eq:feature:VHcorrFT}
\widehat{{\cal C}^{\#}}(m, h, m') = \sum_{l=0}^{l=\infty}\sum_{k=-1}^{k=1} d^{l+k}_{mh}(\pi/2) d^{l+k}_{hm'}(\pi/2) \widehat{\bf f}^l_{km}
\overline{\widehat{\bf g}^l_{km'}}.
\end{equation}
Again, the correlation matrix ${\cal C}^{\#}(\xi, \eta, \omega)$ can be retrieved via inverse Fourier transform of $\widehat{{\cal C}^{\#}}$:
\begin{equation}
\label{eq:feature:VHcorrFinal}
{\cal C}^{\#}(\xi, \eta, \omega) = {\cal F}^{-1}(\hat{{\cal C}^{\#}}(m, h, m')),
\end{equation}
revealing the correlation values in a three dimensional $(\xi, \eta, \omega)$-space.

\section{\label{sec:feature:VH_conv}Fast Convolution in Vectorial Harmonics}
\index{Convolution}
The fast convolution ${\cal C}^*$ in Vectorial Harmonics can be directly derived from sections \ref{sec:feature:vhcorr} and 
\ref{sec:feature:shconvolve}:
\begin{equation}
\label{eq:feature:VHconvFT}
\widehat{{\cal C}^*}(m, h, m') = \sum_{l=0}^{\infty}\sum_{-1}^{k=1} d^{l+k}_{mh}(\pi/2) d^{l+k}_{hm'}(\pi/2) \widehat{\bf f}^l_{km} 
\widehat{\bf g}^l_{km'}.
\end{equation}
Analog to equ. (\ref{eq:feature:VHcorrFinal}), we reconstruct ${\cal C}^{*}(\xi, \eta, \omega)$ from (\ref{eq:feature:VHconvFT}) via 
inverse Fourier transform:
\begin{equation}
\label{eq:feature:VHconvFinal}
{\cal C}^*(\xi, \eta, \omega) = {\cal F}^{-1}(\widehat{\cal C}^*(m, h, m')).
\end{equation}

\chapter{\label{sec:feature:implement}Implementation}
\index{Implementation}
So far, we derived the mathematical foundations for the computation of local features with a parameterization on the 2-sphere (see chapter
\ref{sec:feature:mathbg}) in a setting with strong continuous preconditions:  the input data in form of functions on 3D volumes 
$X: \mathbb{R}^3 \rightarrow \mathbb{R}$ is
continuous, and the harmonic frequency spaces of the transformed neighborhoods  ${\cal S}[r]\left({\bf x}\right)$ are infinitely large 
because we assume to have no band limitations. This setting enables us to nicely derive sound and easy to handle methods, 
however, it is obvious that these preconditions cannot be met in the case of real world applications where we have to deal with discrete 
input data
on a sparse volume grid ($X: \mathbb{Z}^3 \rightarrow \mathbb{R}$) and we have to limit the harmonic transformations to an 
upper frequency (band-limitation to $b_{\max}$). Hence, we some how have to close this gap, when applying the theoretically derived feature 
algorithms to real problems.\\

In general, we try to make this transition to the continuous setting as early as possible so that we can avoid discrete operations which
are usually causing additional problems, i.e. the need to interpolate. Since we derive all of our feature algorithms (chapters 
\ref{sec:feature:SHabs} - \ref{sec:feature:VHHaar}) in the locally expanded harmonic domain, we actually only have to worry about the 
the transition of the local neighborhoods ${\cal S}[r]\left({\bf x}\right)$ in $X$ by ${\cal SH}\left(X|_{S[r]({\bf x})}\right)$ 
(see section \ref{sec:feature:sh}) and ${\cal VH}\left({\bf X}|_{S[r]({\bf x})}\right)$ (see section \ref{sec:feature:VH}).\\
Hence, we need sound Spherical and Vectorial Harmonic transformations for discrete input data which handle the arising sampling problems
and the needed band limitation. We derive these transformations in the next sections \ref{sec:feature:SHimplement}, 
\ref{sec:feature:VHimplement} and discuss some relevant properties like complexity.\\

Another issue we frequently have to face in the context of an actual implementation of algorithms is the question of parallelization.
We tackle the basics of parallelization in section \ref{sec:feature:parallelization}.\\

The introduction of the actual features in the next chapters always follows the same structure: first, we derive the theoretic
foundation of the feature in a continuous setting, and then we give details on the actual discrete implementation based on
the methods we derive in this chapter.

\section{\label{sec:feature:SHimplement}Discrete Spherical Harmonic Transform}
\index{Spherical Harmonics}
\index{Implementation}
We are looking for discrete version of the Spherical Harmonic transform, e.g. we want to obtain the frequency decomposition of local 
discrete spherical neighborhoods ${{\cal S}[r]({\bf x})}$ (\ref{eq:feature:subpatterndef}) in $X:\mathbb{Z}^3 \rightarrow 
\mathbb{R}$.\\
If we disregard the sampling issues for a moment, the discrete implementation is rather straightforward: first, we pre-compute
discrete approximations of the orthonormal harmonic base functions $Y^l_m[r,{\bf x}]$ (\ref{eq:feature:SHcoeff}) which are centered in $\bf x$.
In their discrete version, the $Y^l_m$ are parameterized in Euclidean coordinates ${\bf x} \in \mathbb{Z}^3$ rather then Euler angles:
\begin{equation}
Y^l_m: \mathbb{Z}^3 \rightarrow \mathbb{C}. 
\end{equation}
Next, we obtain the 
transformation coefficients ${\cal SH}\big(X|_{{\cal S}[r]({\bf x})}\big)^l_m$ via the discrete dot-product:
\begin{equation}
{\cal SH}\big(X|_{{\cal S}[r]({\bf x})}\big)^l_m := \sum\limits_{ {\bf x_i} \in {\cal S}[r]({\bf x})} X({\bf x_i})
Y^l_m[r,{\bf x}]({\bf x_i}).
\label{eq:feature:discreteSH}
\end{equation}

For most practical applications we have to compute the harmonic transformation of the neighborhoods around each voxel ${\bf x}$,
which can be computed very efficiently: since (\ref{eq:feature:discreteSH}) is actually determined via convolution, we can apply the 
standard convolution theorem ``trick'' and perform a fast convolution via FFT to obtain ${\cal SH}^l_m(X): \mathbb{R}^3 
\rightarrow \mathbb{C}^b$ (with $b = b_{\max}(b_{\max}-1)$): 
\begin{equation}
 {\cal SH}[r]\left(X\right)^l_m = X * Y^l_m[r].
\label{eq:feature:discreteSHfull}
\end{equation}

This leaves us with the problems to construct correct base function templates $Y^l_m[r]$, which is essentially a sampling issue, and to find an appropriate $b_{\max}$. 

\subsection{\label{correct_sampling}Correct Sampling}
The key problem of obtaining discrete approximations of continuous signals is to avoid biased results due to false sampling. In the case of
the discrete harmonic transformations we have to handle two different sampling steps: first, the discretization of the input data, and 
second the construction of the base function templates $Y^l_m[r]$. In both cases, we can rely on the Sampling Theorem \cite{sampling}
\cite{samplingtheorem} to obtain correct discretizations:
\begin{quotation}
\it
If a function x(t) contains no frequencies higher than $B$ cycles per second\footnotemark, it is completely determined by giving its ordinates at a series of points spaced $1/(2B)$ seconds apart \cite{samplingtheorem}
\end{quotation}
\footnotetext{equivalent to modern unit hertz}

The sampling rate during the discretization of the input data is usually bound by the imaging device. While most modern microscope systems
obey the sampling theorem (see part III), other data sources might be more problematic. Hence, we are forced to introduce an artificial 
band-limitation, i.e. apply a low pass filtering on the input data whenever we face insufficient sampling.\\

\begin{figure}[ht]
\centering
\includegraphics[width=0.45\textwidth]{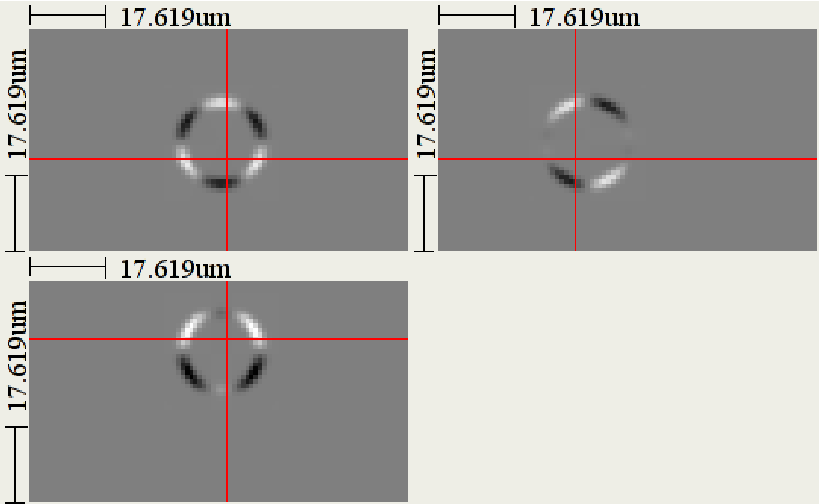}
\includegraphics[width=0.45\textwidth]{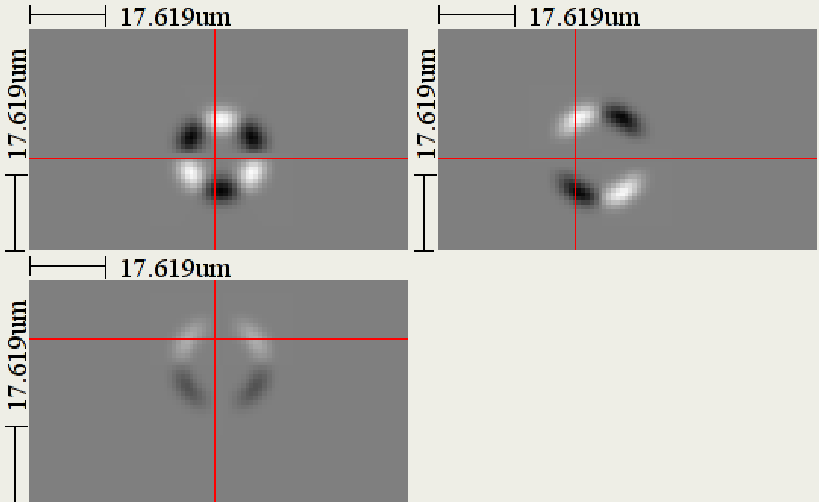}
\caption[Construction of correct discrete base function templates.]{\label{fig:feature:SHtemplate} Construction of a correct discrete base 
function template. Orthoview of the example $Y^4_3$ at $r=10$ and with an Gaussian smoothing of $\sigma=2$ ({\bf left}) and $\sigma=4$ 
({\bf right}). }
\end{figure}

\begin{figure}[ht]
\centering
\includegraphics[width=0.45\textwidth]{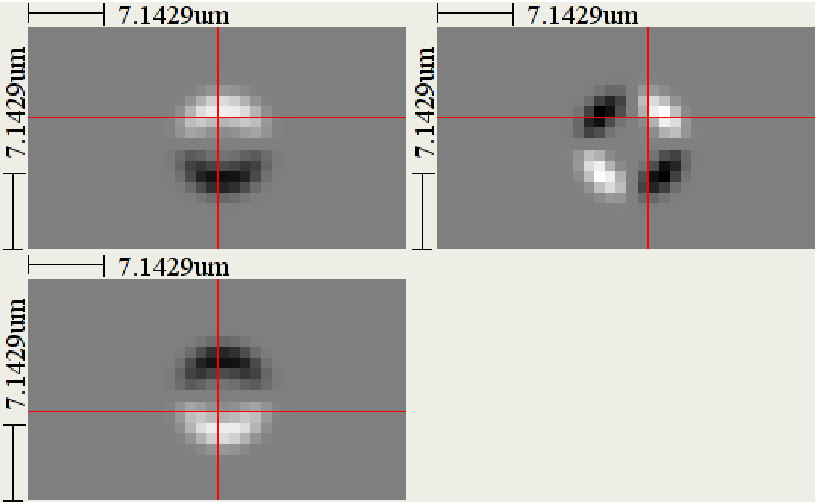}
\includegraphics[width=0.45\textwidth]{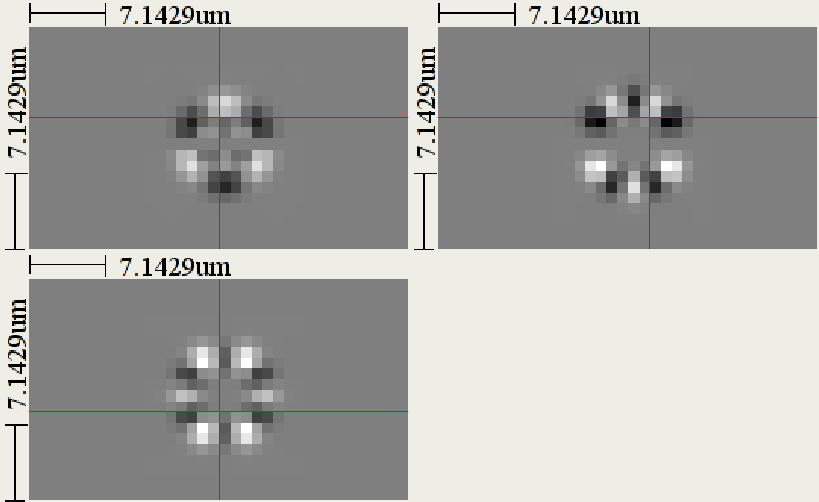}
\includegraphics[width=0.45\textwidth]{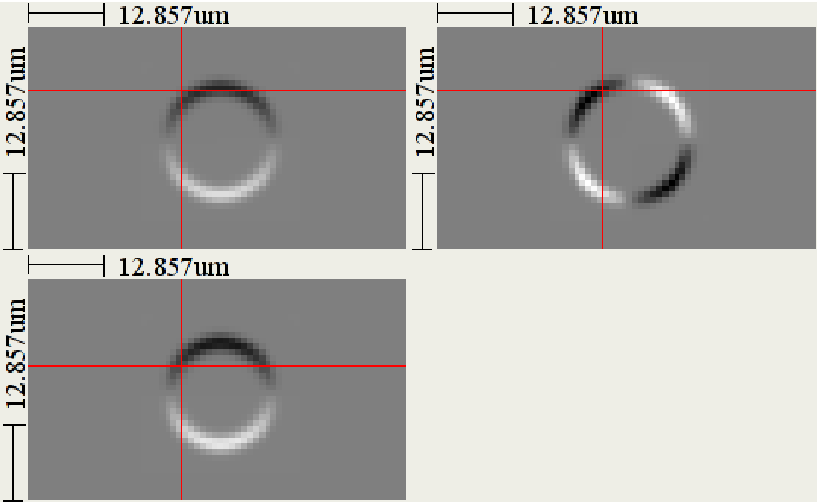}
\includegraphics[width=0.45\textwidth]{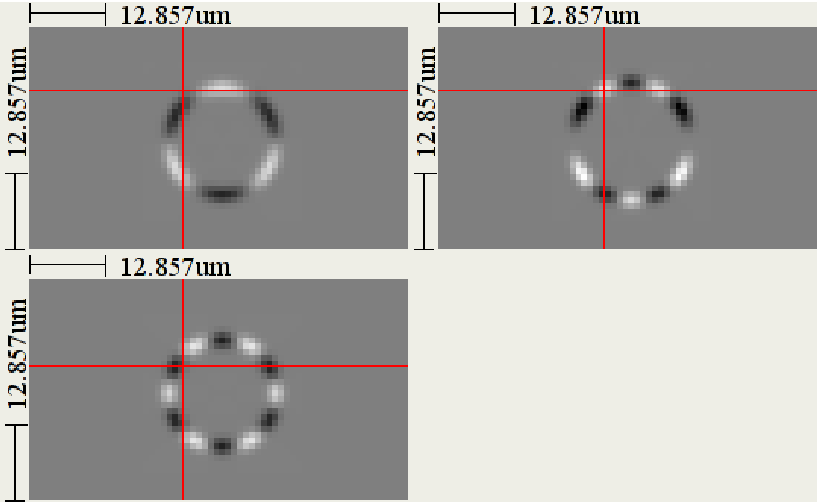}
\caption[Correct Sampling.]{\label{fig:feature:SHtemplatesampling} Example of sampling problems for small radii: {\bf Top:} $r=4$ and $b_{\max} = 2,7$. {\bf Bottom:} $r=10$ and $b_{\max} = 2,7$. The orthoview visualization clearly shows that the small radius does not provide
enough sampling points for the higher frequencies in the 7th band.

}
\end{figure}

The construction of correct discrete base function templates $Y^l_m[r]$ is more challenging because due to the dot-product nature of the 
discrete transformation (\ref{eq:feature:discreteSH} ) the sampling rate is fixed by the 
resolution of the input data and dominantly by the radius $r$, e.g. we cannot simply increase the sampling for higher frequency bands 
$l$ (see figure \ref{fig:feature:SHtemplatesampling})\footnotemark.\\ 
This results in an insurmountable limitation for our discrete harmonic transformations: the maximum expansion band $b_{max}$ is bound by the 
radius: given small radii, the regarding spherical neighborhood ${\cal S}[r]$ only provides a sufficient number of sampling points for low 
frequent base
functions.\\ 
Further more, the discretization of convex structures like spheres easily causes aliasing effects we have to avoid. We cope with this problem
by a Gaussian smoothing in radial direction. Figure \ref{fig:feature:SHtemplate} shows an example of a discrete base function template.
\footnotetext{Thanks to O. Ronneberger for the ``Volvim'' orthoviewer.} 

\subsection{\label{band_limit}Band Limitation $b_{\max}$ }
\begin{figure}[ht]
\begin{tabular}{lccr}
{\bf A}&\includegraphics[width=0.34\textwidth]{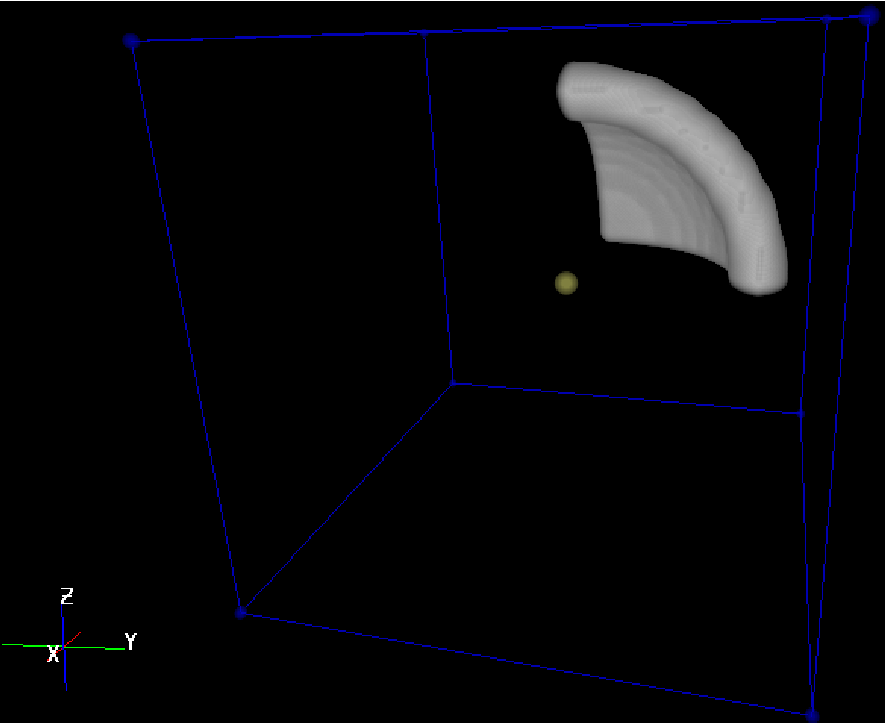}&
{\bf B}&\includegraphics[width=0.45\textwidth]{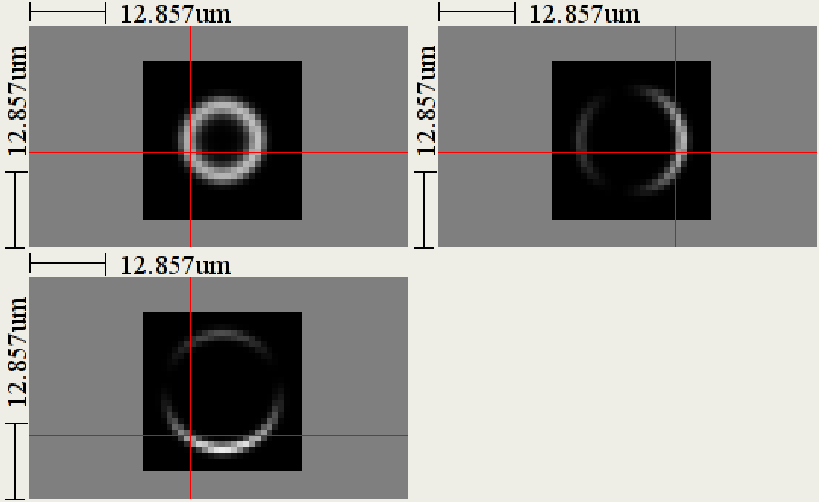}\\
{\bf C}&\includegraphics[width=0.45\textwidth]{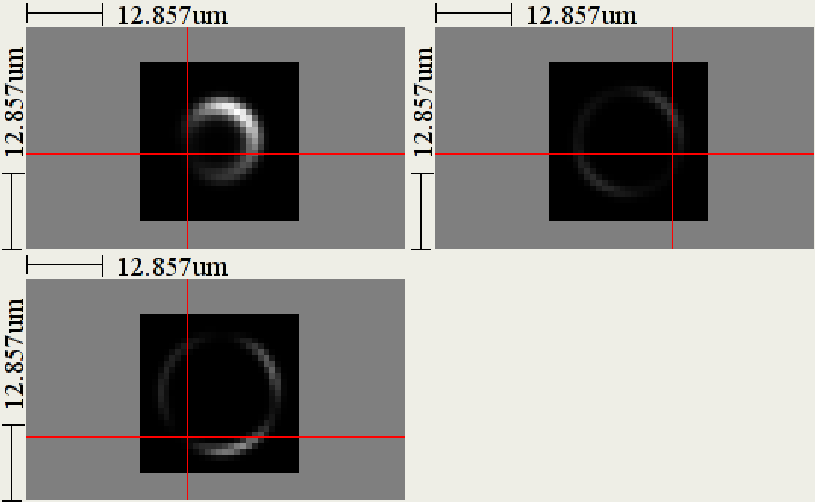}&
{\bf D}&\includegraphics[width=0.45\textwidth]{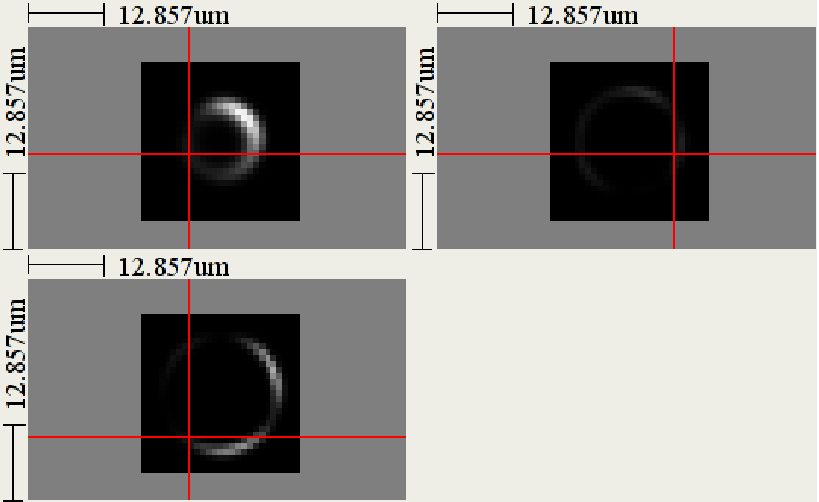}\\
{\bf E}&\includegraphics[width=0.45\textwidth]{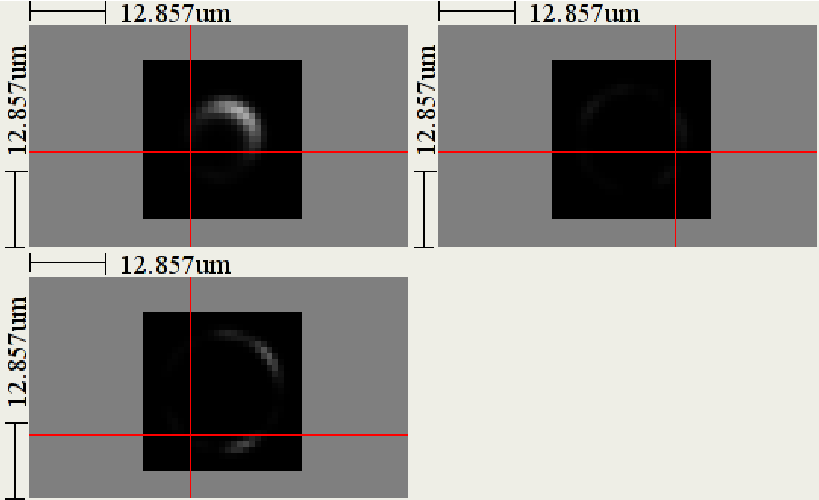}&
{\bf F}&\includegraphics[width=0.45\textwidth]{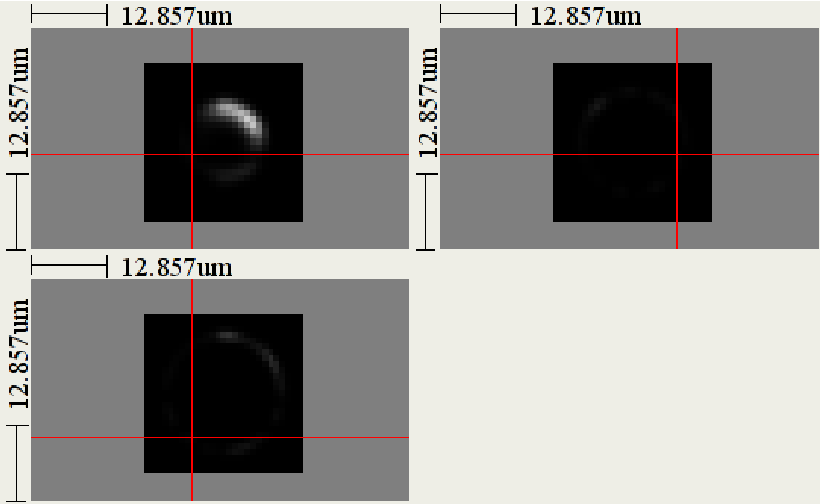}
\end{tabular}
\caption[Effects of the band-limitation on the reconstruction of a spherical signal.]{\label{fig:feature:SHreconst} Effects of the band-limitation on the reconstruction of a spherical signal: {\bf A}: volume rendering of the original binary signal on a sphere. {\bf B}-{\bf F}: 
orthoview of reconstructions with $b_{\max}=1,2,3,5,10$.}
\end{figure}

Assuming that we obey the sampling theorem during the construction of $Y^l_m[r]$ (see previous section), we still have to worry about 
the effect of the band limitation of the harmonic expansion and reasonable choice of $b_{\max}$ below the theoretic limit.\\ 
The good news is that reconstructions from the harmonic domain are strictly band-wise 
operations (e.g. see (\ref{eq:feature:SHbackward})). Hence, the actual band limitation has no effect on the correctness of the lower 
frequencies: the band limitation simply acts as low-pass filter on the spherical signal. Figure \ref{fig:feature:SHreconst}\footnotemark  
shows the effects of the band-limitation in a synthetic example.
\footnotetext{Thanks to H. Skibbe for his volume rendering tool.}
One should also keep in mind that a limitation of higher frequencies directly affects the angular resolution ${\cal SH}_{res}$
of the fast correlation and convolution in the harmonic domain (see section \ref{sec:feature:shcorr}).\\

In the end, the selection of $b_{\max}$ is always a tradeoff between computational speed and maximum resolution.

\subsection{Invariance}
Another practical aspect of the harmonic expansion is that we are able to obtain additional invariance or robustness properties directly
from the transformation implementation.
\subsubsection{Gray-Scale Robustness}
The most obvious example is the simple ``trick'' to become robust against gray-scale changes: 
As mentioned before in section \ref{sec:feature:sh}, one very convenient property of the spherical harmonic transformations is that analogous 
to the Fourier transform, the constant component of the
expanded signal is given by the 0th coefficient ${\cal SH}[r]\left(X\right)^0_0$. 
Hence, we can easily achieve invariance towards shift of the mean gray-value in scalar operations if
we simply normalize all coefficients by the 0th component.\\
Usually we denote this invariance only as ``gray-scale robustness'' since most practical applications include more complex 
gray-scale changes as this approach can handle. 

\subsubsection{Scale Normalization}
It is also very easy to normalize the ${\cal SH}$ coefficients to compensate known changes in the scale of the data. In case we need to compute 
comparable features for data of different scale, we can normalize the coefficients  
${\cal SH}[r]$ by the surface of the base functions, which is $4\pi r^2$ in a continuous setting. In the discrete case, we have to take the
Gaussian smoothing into account: we simply use the sum over $Y_0^0$ as normalization coefficient.

\subsubsection{\label{sec:learning:sampling_resinvar}Resolution Robustness}
\begin{figure}[ht]
\centering
\includegraphics[width=0.45\textwidth]{FIGURES/example_SHbase.eps}
\includegraphics[width=0.42\textwidth]{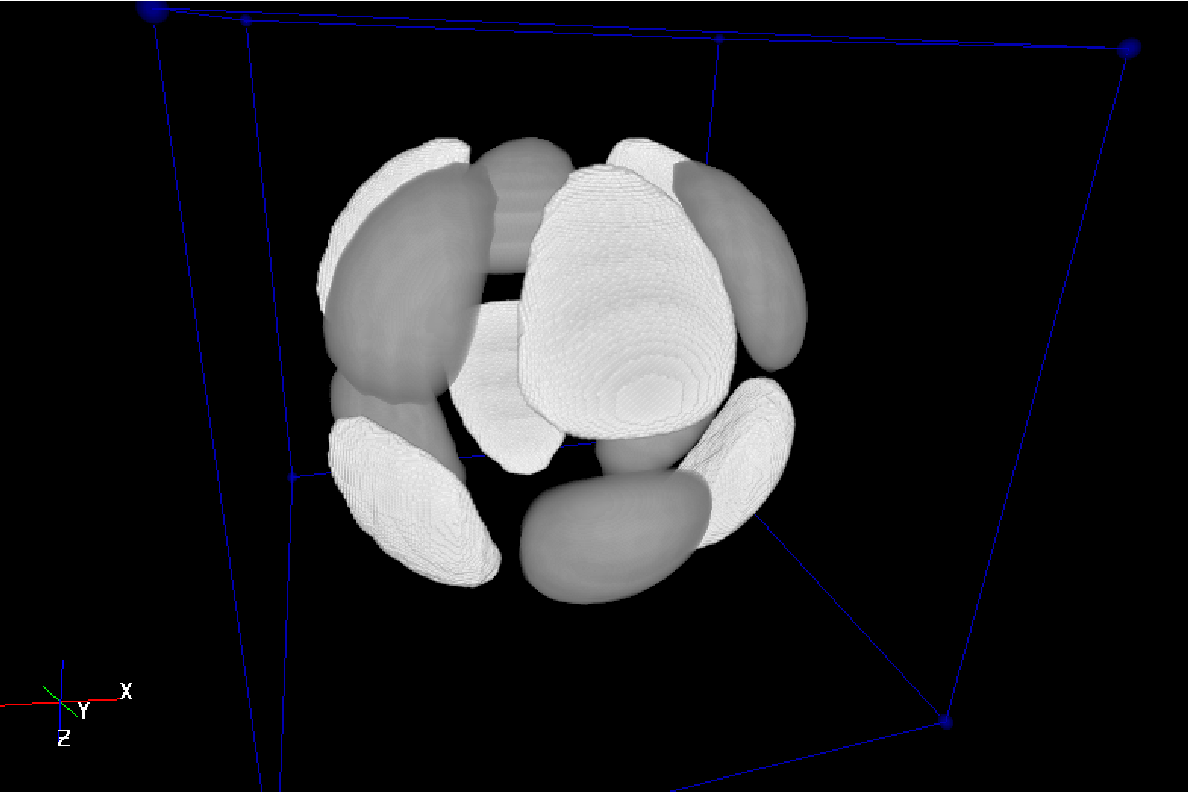}
\includegraphics[width=0.45\textwidth]{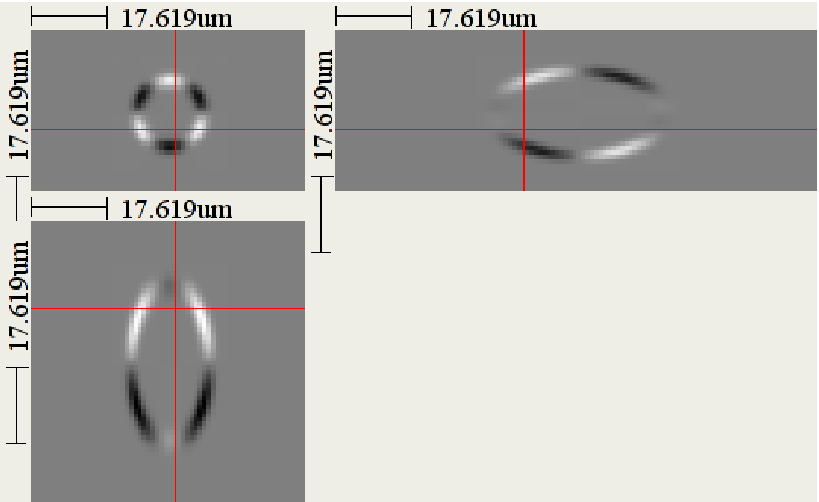}
\includegraphics[width=0.42\textwidth]{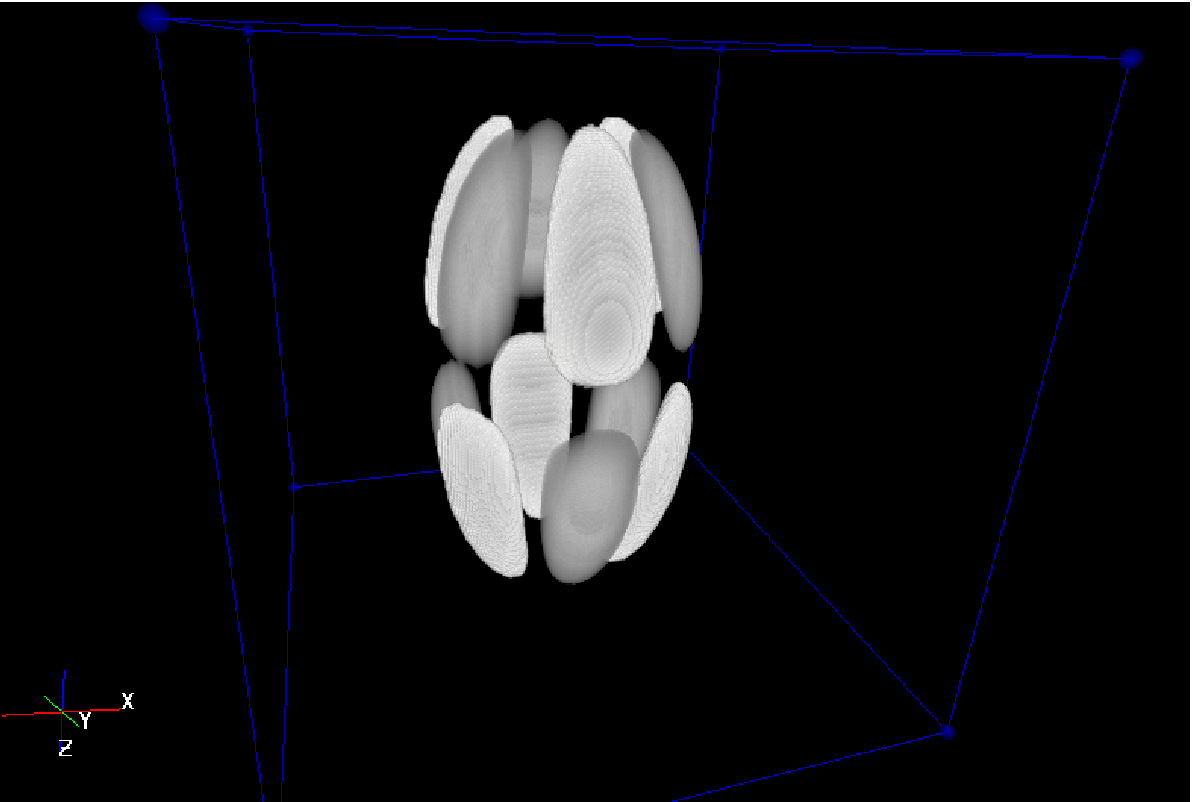}
\caption[Non-cubic voxels.]{\label{fig:feature:non-cube}Resolution Robustness: {\bf Top:} orthoview and volume rendering of the $Y^4_3$ base 
function ($r=10$) 
with the same resolution in all spatial directions. {\bf Bottom:} orthoview and volume rendering of a base function with higher $z$-resolution.
}
\end{figure}

A typical problem which arises in the context of ``real world'' volume data is that we sometimes have to deal with non-cubic voxels, i.e.
the input data is the result of a sampling of the real world which has not been equidistant in all spatial directions.\\
%This problem is usually caused by the imaging devices, e.g. confocal laser scanning microscopes (LSM) (see section \ref{sec:app:imageing_prob})
%typically have a lower resolution in
%$z$-direction than in $x$ and $y$.\\
Such non-cubic voxels cause huge problems when we try to obtain rotation invariant features. Fortunately, we can cope with this problem
during the construction of the base function templates $Y^l_m[r]$: as figure \ref{fig:feature:non-cube} shows, we simply adapt the voxel 
resolution
of the input data to the templates. Usually, we can obtain the necessary voxel resolution information directly from the imaging device. 

\subsection{Complexity}
Concerning the voxel-wise local transformation for a single radius  ${\cal SH}[r]\left(X\right)$ of a 3D volume $X$ with $m$ voxels,
we obtain the harmonic expansion to band $b_{\max}$ in $O(m(b_{\max})^2 + (m \log m))$ if we follow the fast convolution approach (
\ref{eq:feature:discreteSHfull}) and assume the base function templates are given.\\
Since we have to extract $n=b_{\max}(b_{\max}-1)$ coefficients, the memory consumption lies in $O(m(b_{\max})^2)$. 

\subsection{Parallelization}
Further speed-up can be achieved by parallelization (see section \ref{sec:feature:parallelization}): the data can be transformed into
the harmonic domain by parallel computation of the coefficients.
For ${\cal C}$ CPU cores with ${\cal C}\leq (b_{\max})^2$ and ${\cal C}\leq m$ we obtain:
$O(\frac{m(b_{\max})^2}{\cal C}) +O( \frac{(m \log m)}{\cal C})$.

\subsection{\label{sec:feature:sampling:fsh}Fast Spherical Harmonic Transform}
Recently, there has been an approach towards a fast Spherical Harmonic transform (fSHT) \cite{fastSH} for discrete signals.
The fSHT uses a similar approach as in the FFT speed-up of the DFT and performs the computation of the entire inverse transformation in 
$O(N \log^2 N)$, where $N$ is the number of sampling points.\\
Since we hardly need the inverse transformation and only a small set of different extraction radii throughout this work, we prefer
a simple caching of the pre-computed base functions to achieve faster transformations over of the quite complex fSHT method. Additionally,
for real valued input data, we can exploit the symmetry properties (\ref{eq:feature:SHpropSym}):   
\begin{equation}
\overline{Y^l_m} = (-1)^m Y^l_{-m},
\end{equation}
allowing us to actually compute only the positive half of the harmonic coefficients. 

\section{\label{sec:feature:VHimplement}Discrete Vectorial Harmonic Transform}
\index{Vectorial Harmonics}
\index{Implementation}
For the extraction of features on 3D vector fields,
we need a discrete version of the Vectorial Harmonic transform (see section \ref{sec:feature:VH}), i.e. we need to obtain the frequency 
decomposition of 3D vectorial signals at discrete positions on the discrete spherical neighborhoods
${{\cal S}[r]({\bf x})}$ (\ref{eq:feature:subpatterndef}) in ${\bf X}:\mathbb{Z}^3 \rightarrow
\mathbb{R}^3$.\\
As for the discrete Spherical Harmonic transform, we pre-compute
discrete approximations of the orthonormal harmonic base functions ${\bf Z}^l_{k,m}[r,{\bf x}]$ (\ref{eq:feature:VecHarmonicBase}) 
which are centered in $\bf x$.
In their discrete version, the ${\bf Z}^l_{k,m}$ are parameterized in Euclidean coordinates ${\bf x} \in \mathbb{Z}^3$ rather then Euler 
angles:
\begin{equation}
{\cal VH}\big({\bf X}|_{{\cal S}[r]({\bf x})}\big)^l_{k,m} := \sum\limits_{ {\bf x_i} \in {\cal S}[r]({\bf x})} X({\bf x_i})
{\bf Z}^l_{k,m}[r,{\bf x}]({\bf x_i}).
\label{discreteVH}
\end{equation}

\begin{figure}[ht]
\begin{tabular}{lccr}
{\bf A}&\includegraphics[width=0.45\textwidth]{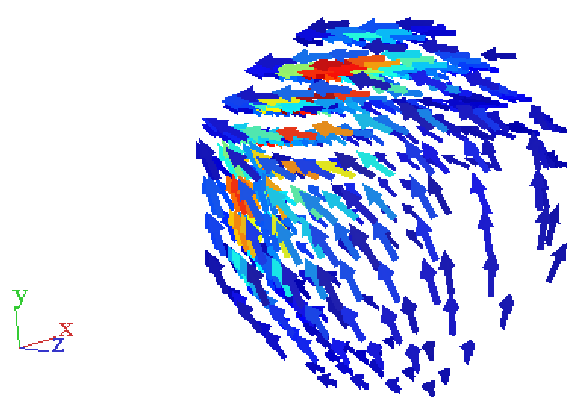}&
{\bf B}&\includegraphics[width=0.45\textwidth]{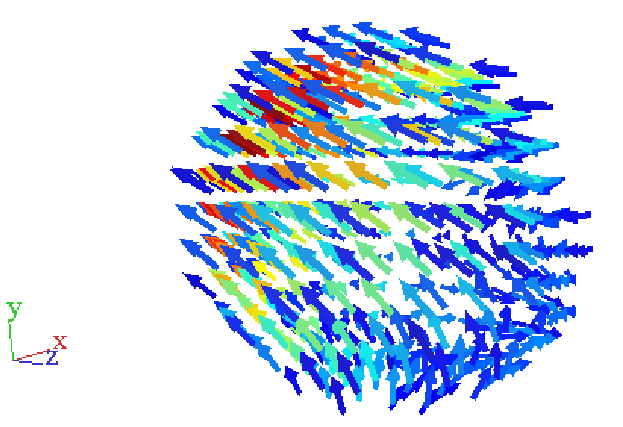}\\
{\bf C}&\includegraphics[width=0.45\textwidth]{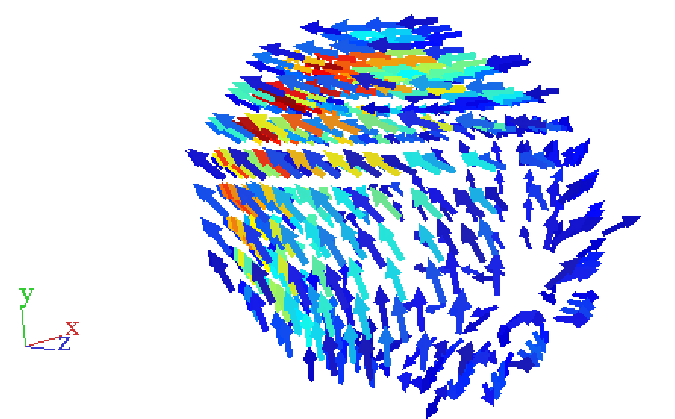}&
{\bf D}&\includegraphics[width=0.45\textwidth]{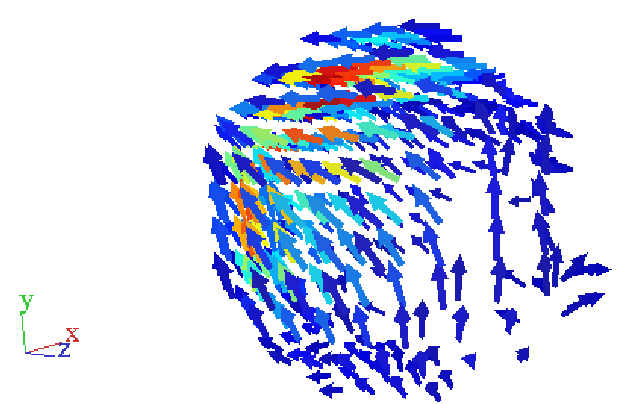}
\end{tabular}
\caption[Effects of the band-limitation on the reconstruction of a vectorial signal.]{\label{fig:feature:VHreconst} Effects of the 
band-limitation on the reconstruction of a vectorial signal: {\bf A}: original vectorial signal on a sphere. {\bf B}-{\bf D}:
reconstructions with $b_{\max}=1,2,5$.}
\end{figure}

For most practical applications we have to compute the harmonic transformation of the neighborhoods around each voxel ${\bf x}$,
which can be computed very efficiently: since (\ref{eq:feature:discreteVH}) is actually determined via convolution, we can apply the
standard convolution theorem ``trick'' and perform a fast convolution via FFT to obtain ${\cal VH}^l_{k,m}({\bf X}): \mathbb{R}^3 
\rightarrow \mathbb{C}^b$:
\begin{equation}
 {\cal VH}[r]\left({\bf X}\right)^l_{k,m} = {\bf X} * {\bf Z}^l_{k,m}[r].
 \label{eq:feature:discreteVH}
 \end{equation}
\begin{figure}[ht]
\centering
\includegraphics[width=0.45\textwidth]{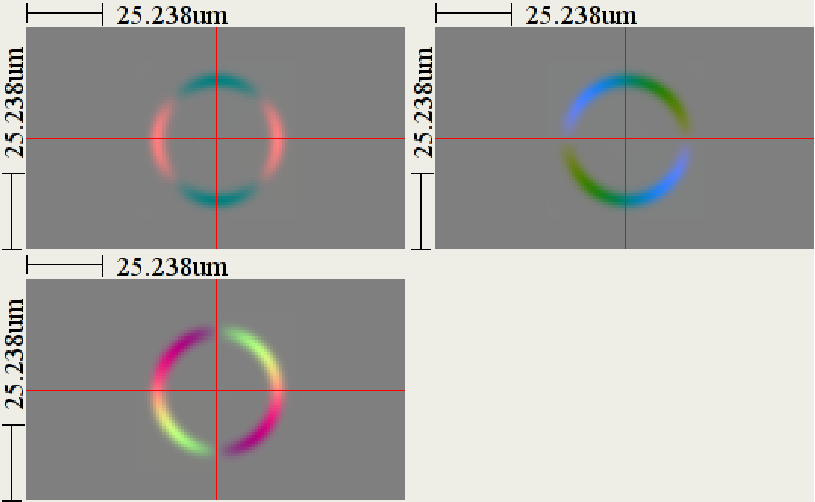}
\includegraphics[width=0.45\textwidth]{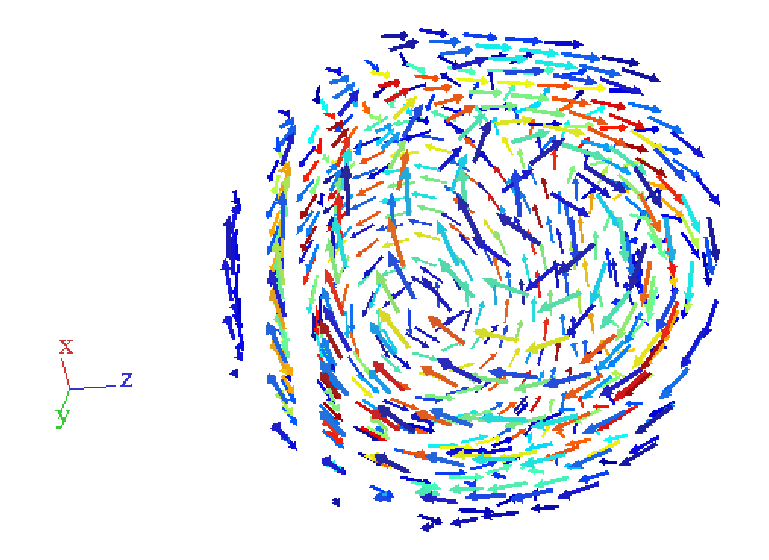}
\caption[Construction of discrete ${\cal VH}$ base functions.]{\label{fig:feature:VHbasediscrete}{\bf Left:} Color coded orthoview 
visualization of the ${\bf Z}^2_{1,4}$ base function. {\bf Right:} 3D vector visualization of the same base function. }
\end{figure}
The sampling and non-cubic voxel problems can be solved in the very same way as for the Spherical Harmonics. Figure 
\ref{fig:feature:VHreconst} shows an artificial reconstruction example.\\
The complexity of a vectorial transformation grows by factor three compared to the Spherical Harmonics, but we are able to apply the 
same parallelization techniques.

\subsection{\label{sec:feature:VHgrayscale}Gray-Scale Invariance}
The notion of gray-scale invariance might appear a bit odd, since vector fields are not directly associated with scalar gray values.
But it is common practice to obtain the 3D vector fields by the gradient evaluation of 3D scalar data (see part III). Hence, it is 
of major interest to know if and how a 3D gradient vector field changes under gray-scale changes of the underlying data.\\  
\cite{dissolaf} showed that the gradient direction is in fact invariant under additive and multiplicative gray-scale changes. Therefore, 
we consider features based on Vectorial Harmonics to be gray-scale invariant - which is an important property for many applications.

\section{\label{sec:feature:parallelization}Parallelization}
\index{Implementation}
\index{Parallelization}
Modern computing architectures come with an increasing number of general computing units: standard PCs have multi-core CPUs
and more specialized computing servers combine several of these multi-core CPUs in a single system. This endorses the use of 
parallel algorithms. \\
In this work, parallel computing is only a little side aspect - but one with great speed-up potential. We restrict ourself 
to very simple cases of parallelization algorithms: first, we only consider systems with shared memory where all computing units (we
refer to them as cores) share the same memory address space of a single system - hence, we explicitly disregard clusters.
Second, we only consider algorithmically very simple cases of parallelization where the individual threads run independently, i.e. we avoid
scenarios which would require a mutual exclusion handling, while still going beyond simplest cases data parallelization.\\
We give more details on the actual parallelization at the individual description of each feature implementation.

\cleardoublepage
\chapter{\label{sec:feature:SH}${\cal SH}$-Features}
\index{${\cal SH}$-Features}
In this chapter, we derive a set of local, rotation invariant features which are directly motivated by the sound
mathematical foundation for operations on the 2-sphere introduced in chapter \ref{sec:feature:mathbg}. We take advantage of the 
nice properties of Spherical Harmonics (\ref{eq:feature:SHbase}) which allow us to perform fast feature computations in the
frequency domain.\\

Given scalar 3D volume data $X$, the transformation ${\cal SH}\big(X|_{S[r]({\bf x})}\big)$ 
(\ref{eq:feature:discreteSH}) of local data on a sphere with radius $r$ around the 
center point ${\bf x}$ in Spherical Harmonics is nothing more than a change of the base-functions representing the initial data.
So the new base might provide us with a nice framework to operate on spheres, but we still have to perform the actual feature construction. 
Primarily, we want to obtain rotation and possibly gray-scale invariance.\\

First we introduce a simple method to obtain rotational invariance: In section \ref{sec:feature:SHabs} we review ${\cal SH}_{abs}$ features,
which use the fact that the band-wise energies of a ${\cal SH}$ representation does not change under rotation. This method is well
known from literature (i.e. \cite{shabs}), but has its limitations.\\ 

To cope with some of the problems with ${\cal SH}_{abs}$ features, we introduced a novel rotation and gray-scale 
invariant feature based on the ${\cal SH}$
phase information \cite{shphase}. We derive the ${\cal SH}_{phase}$ feature in section \ref{sec:feature:SHphase}.\\

The third member of the ${\cal SH}$-Feature class is a fast and also rotation invariant auto-correlation feature ${\cal SH}_{autocorr}$ 
(section \ref{sec:feature:SHautocorr}) which
is based on the fast correlation in Spherical Harmonics from section \ref{sec:feature:shcorr}.\\

Finally, in section \ref{sec:feature:SHbispectrum}, we derive a complete local rotation invariant 3D feature from a global 2D image 
feature introduced in \cite{shbispectrum}. The ${\cal SH}_{bispectrum}$ feature.

\newpage
%--------------------------------------------
% SHabs
%-------------------------------------------
\section{\label{sec:feature:SHabs}${\cal SH}_{abs}$}
\index{${\cal SH}$-Features}\index{${\cal SH}_{abs}$}
The feature we chose to call ${\cal SH}_{abs}$ throughout this work is also known as ``Spherical Harmonic Descriptor'' and has been 
used by several previous publications e.g. for 3D shape retrieval in \cite{shabs}.  
We use ${\cal SH}_{abs}$ as one of our reference features to evaluate the 
properties and performance of our methods (see chapter \ref{sec:feature:experiments}).

\subsection{Feature Design}
${\cal SH}_{abs}$ achieves rotation invariance by exploiting some basic principals of the Spherical Harmonic (\ref{eq:feature:SHbase}) 
formulation. Analogous to the Fourier transformation, where we can use the power spectrum as a feature, we use the absolute values of each 
harmonic expansion band $l$ as power of the $l$-th frequency in the Spherical Harmonic power spectrum: 
\begin{equation}
\label{eq:feature:SHabsvec}
\big({\cal SH}_{abs}[r]({\bf x})\big)^l :=  \sqrt{\sum\limits_{m=-l}^{l} \left(\left({\cal SH}\big(X|_{S[r]({\bf x})}\big)\right)^l_m\right)^2}.
\end{equation}

\paragraph{Rotation Invariance}
Rotations ${\cal R(\phi,\theta, \psi)} \in {\cal SO}(3)$ (see section \ref{sec:feature:shrot}) are represented in the harmonic 
domain in terms of band-wise multiplications of the expansions $\widehat{f^l}$ with the orthonormal Wigner D-Matrices 
$D^l$ (\ref{eq:feature:shrot}).\\
The power spectrum of a signal $f$ in Spherical Harmonics is given as (also see section \ref{sec:feature:SHbispectrum} for more details):
\begin{equation}
\label{eq:feature:SHpowerspec}
q(f,l) := \big(\overline{\widehat{f^l}}\big)^T  \widehat{f^l}.
\end{equation}
The $D^l$ are orthonormal (\ref{eq:feature:dwigOrthogonal}), hence it is easy to show the rotation invariance of the band-wise 
${\cal SH}_{abs}$ entries of the power spectrum:
\begin{eqnarray}
\begin{tabular}{r@{~ ~}lr }
${\cal SH}_{abs}\big(D^l({\cal R})\widehat{f^l}\big)$ &$=  \big(\overline{D^l({\cal R})\widehat{f^l}}\big)^T D^l({\cal R})\widehat{f^l}$ & \\[0.125cm]
&$= \big(\overline{\widehat{f^l}}\big)^T \big(\overline{D^l({\cal R})}\big)^T D^l({\cal R}) \widehat{f^l}$ &\\[0.125cm]
&$= \big(\overline{\widehat{f^l}}\big)^T  \widehat{f^l}$& 
\end{tabular}.\nonumber
\end{eqnarray}

So, we note that a rotation has only a band-wise effect on the expansion but does not change the respective absolute values. 
Hence, the approximation of the original data via harmonic expansion can be cut off at an
arbitrary band, encoding just the level of detail needed for the application.

\paragraph{Gray-Scale Robustness:}
We can obtain invariance towards additive gray-scale changes by normalization by the $0$th harmonic coefficient as described in section
\ref{sec:feature:implement}.

\subsection{Implementation}
The implementation of the ${\cal SH}_{abs}$ is straightforward. We follow the implementation of the Spherical Harmonic transformation as 
described in chapter \ref{sec:feature:implement}. 

\paragraph{Multi-Channel Data:} ${\cal SH}_{abs}$ cannot directly combine data from several channels into a single feature. In case of 
multi-channel data, we have to separately compute features for each channel.

\subsubsection{Complexity}
Following the implementation given in section \ref{sec:feature:implement}, we obtain the harmonic expansion to band $b_{\max}$ at each
point of a volume with $m$ voxels in $O(m(b_{\max})^2 + (m \log m))$. The computation of the absolute values takes another $O((b_{\max})^3)$.

\paragraph{Parallelization}
Further speed-up can be achieved by parallelization (see section \ref{sec:feature:implement}): the data can be transformed into
the harmonic domain by parallel computation of the coefficients and the computation of the absolute values can also be split into several 
threads.
For ${\cal C}$ CPU cores with ${\cal C}\leq (b_{\max})^2$ and ${\cal C}\leq m$ we obtain:
$$O(\frac{m(b_{\max})^3}{\cal C}) +O( \frac{m(b_{\max})^2 +(m \log m)}{\cal C})$$

\subsection{\label{sec:feature:ShabsDiss}Discussion }
The ${\cal SH}$-Features are a simple and straightforward approach towards local 3D rotation invariant features. They are computationally 
efficient and easy to implement, however, the discriminative properties are quite limited. The band-wise absolute values only capture 
the energy of the respective frequencies in the overall spectrum. Hence, we loose all the phase information which leads to strong ambiguities
within the feature mappings. In many applications it is possible to reduce these ambiguities by the combination of ${\cal SH}$-Features
which were extracted at different radii.

\begin{figure}[ht]
\centering
\includegraphics[width=0.45\textwidth]{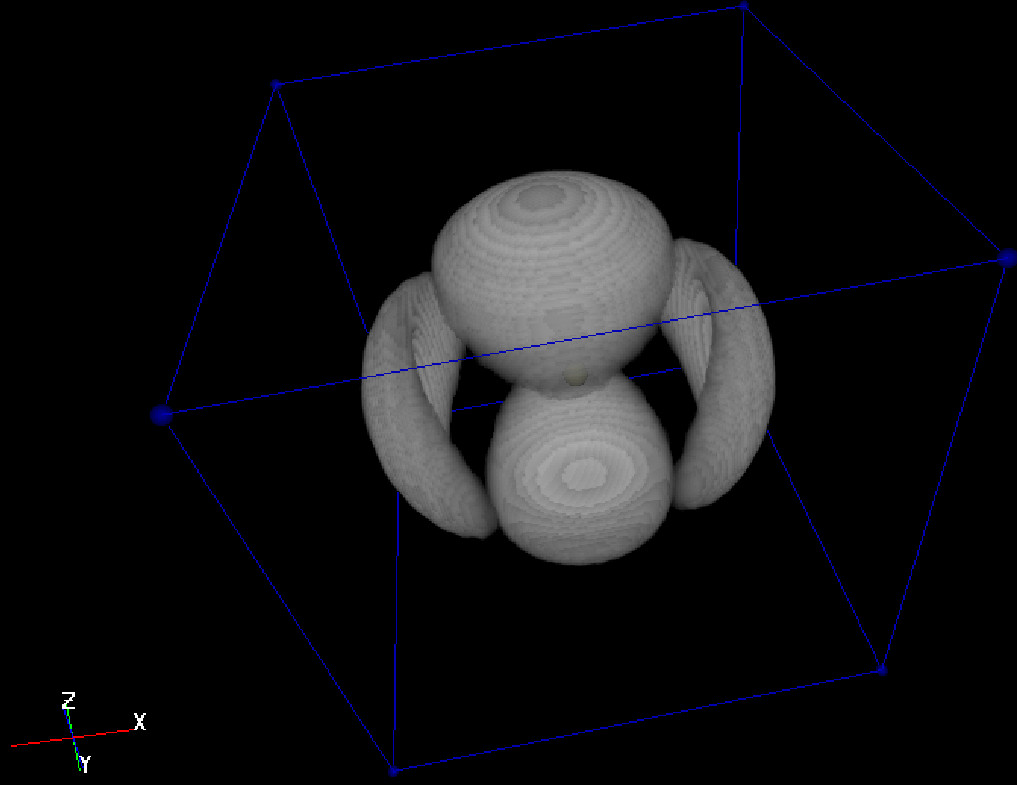}
\includegraphics[width=0.45\textwidth]{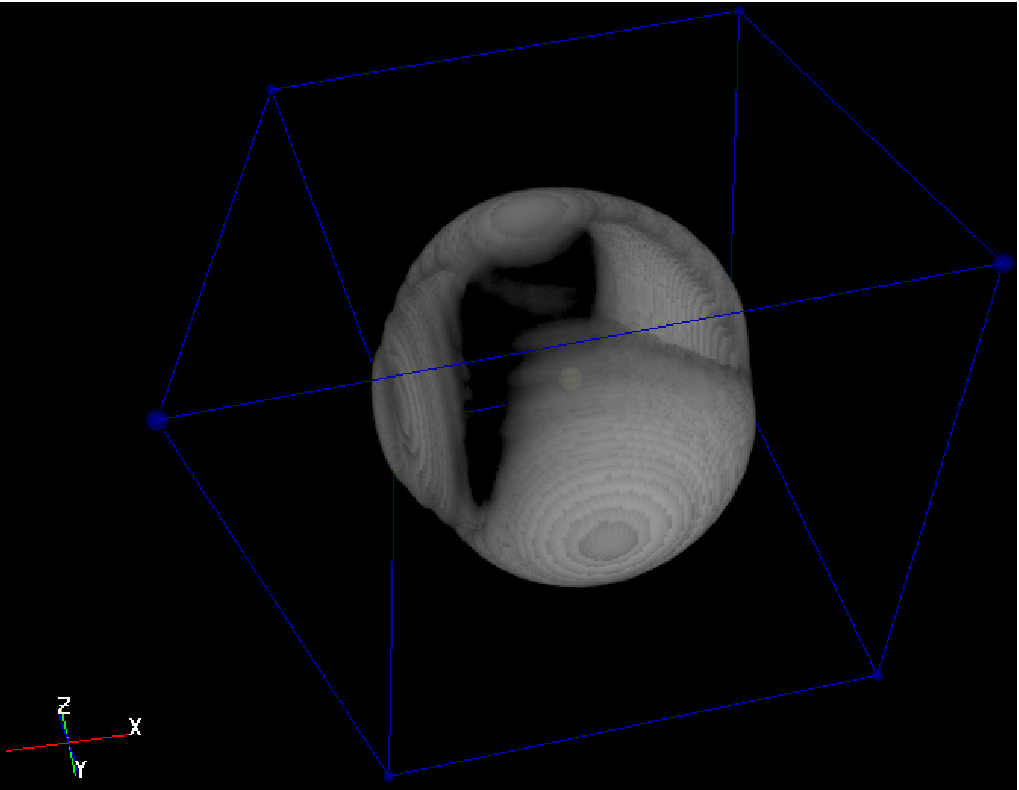}
\caption[${\cal SH}_{abs}$ ambiguities.]{\label{fig:feature:SHabs_abiguity} 3D volume rendering of two quite different signals on spheres which
have exactly the same ${\cal SH}_{abs}$ value.}
\end{figure}

\paragraph{${\cal SH}_{abs}$ Ambiguities:} in theory, there is an infinite number of input patterns which are mapped on the same 
${\cal SH}$-Feature just as there is an infinite number of possible phase shifts in harmonic expansions. However, one might argue that
this does not prevent a practically usage of the ${\cal SH}$-Feature since we generally do not need completeness (see section \ref{sec:feature:invariance}).\\
But we still need discriminative features, and there are practical relevant problems where ${\cal SH}_{abs}$ is not powerful enough, as
figure \ref{fig:feature:SHabs_abiguity} shows.

\newpage
%---------------------------------------------
% SHphase
%---------------------------------------------
\section{\label{sec:feature:SHphase}${\cal SH}_{phase}$}
\index{${\cal SH}$-Features}\index{${\cal SH}_{phase}$}
Motivated by the ambiguity problems caused by neglecting the phase information in the ${\cal SH}_{abs}$-Features 
(see discussion in section \ref{sec:feature:ShabsDiss}) we presented an
oppositional approach in \cite{shphase}. ${\cal SH}_{phase}$-Features preserve only the phase information of the Spherical Harmonic
representation and disregard the amplitudes.
This approach is
further motivated by results known from Fourier transform, which showed that the characteristic information is dominant in the phase of 
a signal's spectrum rather than in the pure magnitude of it's coefficients \cite{FTphase}. 
Following a phase-only strategy has the nice side-effect that since the overall gray-value
intensity is only encoded in the amplitude, the ${\cal SH}_{phase}$ method is gray-scale invariant.
Like the ${\cal SH}_{abs}$-Features (from section \ref{sec:feature:SHabs}) ${\cal SH}_{phase}$-Features are computed band-wise, but instead
of a single radius ${\cal SH}_{phase}$ combines expansions at different radii $r_1, r_2$ into a feature.
 
\subsection{Feature Design}
\begin{figure}[htbp]
\centering
\psfrag{Y10}{$Y_0^1[r_1]$}
\psfrag{Y11r}{$Y_1^{1}[r_1]$}
\psfrag{Y11c}{$Y_{-1}^{1}[r_1]$}
\psfrag{rY10}{$Y_0^1[r_2]$}
\psfrag{rY11r}{$Y_{1}^{1}[r_2]$}
\psfrag{rY11c}{$Y_{-1}^{1}[r_2]$}
\psfrag{data}{data}
\psfrag{alpha}{$\alpha$}
\includegraphics[width=0.3\textwidth]{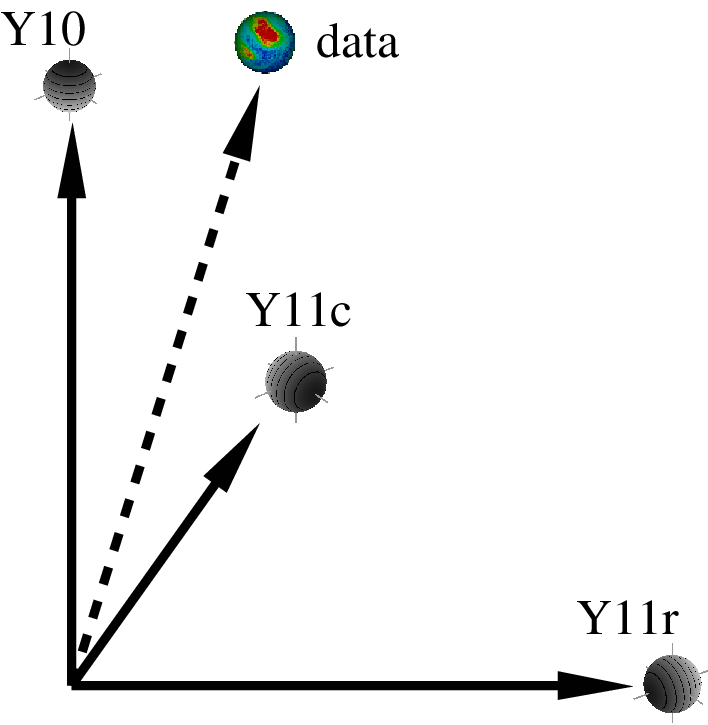}
\includegraphics[width=0.3\textwidth]{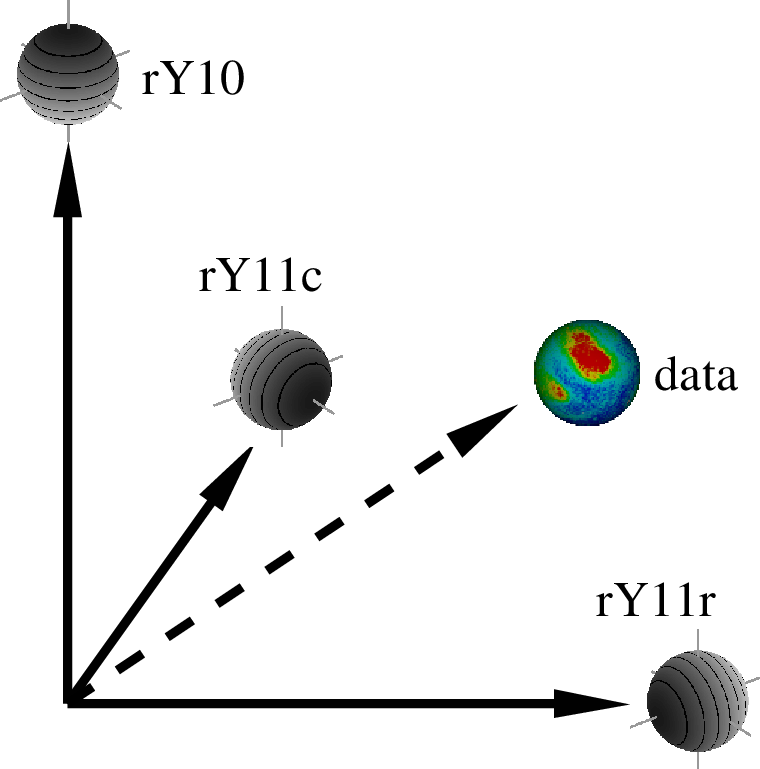}
\includegraphics[width=0.3\textwidth]{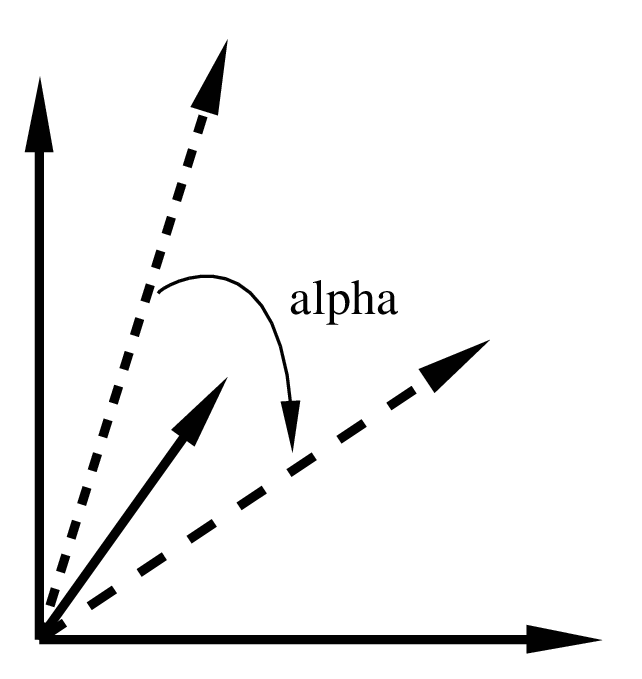}
\caption[phase base]{\label{phase_base} Schematic example of the phase based feature calculation. Left: representation of the original data 
as combination of the 3D base functions of an expansion in the 1st band at radius $r_1$. Center: representation at radius $r_2$. Right: 
the feature is encoding the 1st band phase angle $\alpha$ between the two concentric harmonic expansions.}
\end{figure}

The phase of a local harmonic expansion in band $l$ at radius $r$ is given by the orientation of the vector ${\bf p}^l[r]$, which contains 
the $2l+1$ harmonic coefficient
components of the band-wise local expansion (\ref{eq:feature:SHphase_vec}). Since the coefficients are changing when the underlying data is rotated,
the phase itself is not a rotational invariant feature.\\
\begin{equation}
{\bf p}^{l}_m[r]\left({\bf x}\right) := \frac{\left({\cal SH}\big(X|_{S[r]({\bf x})}\big)\right)^l_m }{\big({\cal SH}_{abs}[r]\left({\bf x}\right)\big)^l}
\label{eq:feature:SHphase_vec}
\end{equation}
Since we are often interested in encoding the neighborhood at several concentric radii, we can take advantage
of this additional information and construct a phase-only rotational invariant feature based on the band-wise relations of phases
between the different concentric harmonic series.\\
\noindent Fig. (\ref{phase_base}) illustrates the basic idea: for a fixed band $l$, the relation (angle) between phases of harmonic expansions 
at different radii are invariant towards rotation. Phases in the same harmonic band undergo the 
same changes under rotation of the underlying data (see section \ref{sec:feature:shrot} for details), keeping the angle between the phases 
of different radii constant. We encode this angle in terms of the dot-product of band-wise Spherical Harmonic expansions at radii 
$r_1, r_2$:
\begin{equation}
\label{eq:feature:SHphase}
\left({\cal SH}_{phase}[r_1,r_2]\left({\bf x}\right)\right)^l := \langle {\bf p}^l[r_1] ,  {\bf p}^l[r_2] \rangle. 
\end{equation}
\paragraph{Rotation Invariance:} the proof of the rotation invariance is rather straightforward basic linear algebra:
\begin{eqnarray}
\begin{tabular}{r@{~ ~}lr }
Rotations ${\cal R}X$ acting on \ref{eq:feature:SHphase}: &$ \langle D^l{\bf p}^l[r_1] , D^l{\bf p}^l[r_2] \rangle$ & \\[0.125cm]
&$= \big(\overline{D^l{\bf p}^l[r_1]}\big)^T(D^l{\bf p}^l[r_2]) $ &  rewrite as matrix multiplication\\[0.125cm]
&$= \big(\overline{{\bf p}^l[r_1]}\big)^T(\overline{D^l})^T (D^l{\bf p}^l[r_2]) $ &  resolve transposition\\[0.125cm]
&$= \big(\overline{{\bf p}^l[r_1]}\big)^T\left((\overline{D^l})^T D^l\right)({\bf p}^l[r_2]) $ &  commutativity\\[0.125cm]
&$= \big(\overline{{\bf p}^l[r_1]}\big)^T\underbrace{\left((\overline{D^l})^TD^l\right)}_{= I}({\bf p}^l[r_2]) $ & use orthogonality of $D^l$\\[0.125cm]
&$= \big((\overline{{\bf p}^l[r_1]})^T{\bf p}^l[r_2]\big) $ &\\[0.125cm]
&$= \langle{\bf p}^l[r_1],{\bf p}^l[r_2]\rangle $ &
\end{tabular}.\nonumber
\end{eqnarray}
The rotation ${\cal R}$ of the underlying data can now be expressed
in terms of matrix multiplications with the same Wigner-D matrix $D^l$ (\ref{eq:feature:shrot}).
Since the rotational invariance is achieved band-wise, the approximation of the original data via harmonic expansion can be cut off at an 
arbitrary band, encoding just the level of detail needed for the application.

\subsection{Implementation}
The implementation of the ${\cal SH}_{phase}$ is straightforward. We follow the implementation of the Spherical Harmonic transformation as
described in section \ref{sec:feature:implement} for the two radii $r_1$ and $r_2$. The band-wise computation of the phases and the evaluation 
of the dot-product is also very simple. 

\paragraph{Multi-Channel Data:} ${\cal SH}_{phase}$-Features can also directly combine data from several channels into a single feature: 
we simply extract the harmonic expansions for the different radii from different data channels.

\subsubsection{Complexity}
Following the implementation given in section \ref{sec:feature:implement}, we obtain the harmonic expansion to band $b_{\max}$ at each
point of a volume with $m$ voxels in $O(m(b_{\max})^2 + (m \log m))$. The computation of the dot-products and the phase vectors  takes 
another $O((b_{\max})^3)$.

\paragraph{Parallelization}
Further speed-up can be achieved by parallelization (see section \ref{sec:feature:parallelization}): the data can be transformed into
the harmonic domain by parallel computation of the coefficients and the computation of the absolute values can also be split into several
threads.
For ${\cal C}$ CPU cores with ${\cal C}\leq (b_{\max})^2$ and ${\cal C}\leq m$ we obtain:
$$O(\frac{m(b_{\max})^3}{\cal C}) +O( \frac{m(b_{\max})^2 +(m \log m)}{\cal C})$$

\subsection{Discussion}
Event though the ${\cal SH}_{phase}$-Features are not complete either, their discrimination abilities tend to be better than those
of the ${\cal SH}_{abs}$-Features (see section \ref{sec:feature:SHabs}). Also, the additional gray-scale invariance is very useful in many
applications.\\
Intuitively, ${\cal SH}_{phase}$ encodes local changes between the different radii. This property is especially applicable for texture
classification or to find 3D interest points (see part III).

\newpage
%------------------------------
% SHautocorr
%-------------------------------
\section{\label{sec:feature:SHautocorr}${\cal SH}_{autocorr}$}
\index{${\cal SH}$-Features}\index{${\cal SH}_{autocorr}$}
The next approach to compute invariant features directly from the harmonic representation is motivated by the introduction of the 
fast normalized cross-correlation in the harmonic domain (see introduction of chapter \ref{sec:feature:shcorr}). 
The cross-correlation ${\cal SH}_{corr}(f,g)$
on two signals $f,g \in S^2$ is a binary operation ${\cal SH}_{corr}: S^2\times S^2 \rightarrow \mathbb{R}$. Hence, it cannot be used 
directly as a feature, where we require a mapping of individual local signals $f\in S^2 \rightarrow {\cal H}$ into some feature space 
${\cal H} \subseteq \mathbb{R}^n$ (see section \ref{sec:feature:intro}).\\
A general and widely known method to obtain features from correlations is to compute the auto-correlation, e.g. \cite{autocorr}. In our case, 
we propose the local ${\cal SH}_{autocorr}$-Feature, which performs a fast auto-correlation of $f \in S^2$.\\ 

The auto-correlation under a given rotation $\cal R$ in Euler angles $\phi, \theta, \psi$ is defined as: 
\begin{equation}
(f \# f)({\cal R}) := \int\limits_{S^2}  f ({\cal R} f) \text{\quad} \sin{\Theta}d\Phi d\Theta.
\label{eq:feature:shautocorr0}
\end{equation}

\subsection{Feature Design}
As for most of our other features, we first expand the local neighborhood $f$ at radius $r$ around the point ${\bf x}$ in
Spherical Harmonics, $\widehat{f}:={\cal SH}\big(X|_{S[r]({\bf x})}\big)$.\\
Then we follow the fast correlation method which we introduced in section \ref{sec:feature:shcorr} to obtain the full correlation
$C^{\#}$ from equation (\ref{eq:feature:SHcorrFinal}). 

\paragraph{Invariance:}
In order to obtain rotation invariant features, we follow the Haar-Integration approach (see chapter 
\ref{sec:feature:invariance_via_groupintegration}) and integrate
over the auto-correlations at all possible rotations $\cal R$. $C^{\#}$ holds the necessary auto-correlation results in a 3D
$(\phi,\theta,\psi)$-space (\ref{eq:feature:SHcorrFT}), hence we simply integrate over $C^{\#}$,
\begin{equation}
{\cal SH}_{autocorr} := \int\limits_{\phi,\theta,\psi} \kappa \left(C^{\#}(\phi,\theta,\psi)\right) \sin{\theta}d\phi d\theta d\psi
\label{eq:feature:autocorrinvariant}
\end{equation}
and obtain a scalar feature. Additionally, we insert a non-linear kernel function $\kappa$ to increase the separability. Usually, very
simple non-linear functions, such as $\kappa(x):=x^2, \kappa(x):=x^3$ or $\kappa(x):= \sqrt{x}$, are sufficient.\\

Like in the case of the ${\cal SH}_{abs}$-Features, we can obtain invariance towards additive gray-scale changes by normalization by the
$0$th harmonic coefficient. If we additionally normalize $C^{\#}$ as in (\ref{eq:feature:SHcorrnormFinal}), ${\cal SH}_{autocorr}$
becomes completely gray-scale invariant.

\subsection{Implementation}
We follow the implementation of the Spherical Harmonic transformation as described in chapter \ref{sec:feature:implement} and the 
implementation of the fast correlation from (\ref{eq:feature:SHcorrPad}).\\
In practice, where the harmonic expansion is bound by a maximal expansion band $b_{\max}$, the integral (\ref{eq:feature:autocorrinvariant})
is reduce to the sum over the then discrete angular space $C^{\#}$: 
\begin{equation}
{\cal SH}_{autocorr} = \sum\limits_{\phi,\theta,\psi} \kappa \left(C^{\#}(\phi,\theta,\psi)\right). 
\label{eq:feature:autocorrinvariant_sum}
\end{equation}

\paragraph{Multi-Channel Data:} It is straightforward to combine the information from several data channels into a single 
${\cal SH}_{autocorr}$-Feature: We simply use the same approach as described in section \ref{sec:feature:shcorrradii}, where we
correlated the information of several different radii. 

\subsubsection{Complexity}
Following the implementation given in chapter \ref{sec:feature:implement}, we obtain the harmonic expansion to band $b_{\max}$ at each
point of a volume with $m$ voxels in $O(m(b_{\max})^2 + (m \log m))$. The complexity of the auto-correlation depends on $b_{\max}$ and
the padding parameter $p$ (\ref{eq:feature:SHcorrPad}) and can be computed in $O(m(b_{\max}+p)^3 \log (b_{\max}+p)^3))$. The sum over
$C^{\#}$ takes another $O((b_{\max}+p)^3)$ at each point.

\paragraph{Parallelization:}
Further speed-up can be achieved by parallelization (see section \ref{sec:feature:implement}): the data can be transformed into
the harmonic domain by parallel computation of the coefficients and the computation of the absolute values also be split into several
threads.
For ${\cal C}$ CPU cores with ${\cal C}\leq (b_{\max})^2$ and ${\cal C}\leq m$ we obtain:
$$O(\frac{m\left( (b_{\max}+p)^3 + (b_{\max}+p)^3 \log (b_{\max}+p)^3\right)}{\cal C}) +O( \frac{m(b_{\max})^2 +(m \log m)}{\cal C})$$

\subsection{Discussion}
Auto-correlation can be a very effective feature to encode texture properties. The discriminative power of ${\cal SH}_{autocorr}$ can
be further increased by we combining the correlation a several different radii to a single 
correlation result $C^{\#}$, as described in section \ref{sec:feature:shcorr}.

\newpage
%----------------------------------
% SHbispec
%---------------------------------
\section{\label{sec:feature:SHbispectrum}${\cal SH}_{bispectrum}$}
\index{${\cal SH}$-Features}\index{${\cal SH}_{bispectrum}$}
The final member of the class of features which are directly derived from the Spherical Harmonic representation is the 
so-called ${\cal SH}_{bispectrum}$-Feature. The approach to obtain invariant features via the computation of the bispectrum 
of the frequency representation is well known (e.g. see \cite{bispectrum}), hence, we review the basic concept in a simple 1D
setting before we move on to derive it in Spherical Harmonics.\\

Given a discrete complex 1D signal $f:\{0,1,\dots,n-1\}\rightarrow \mathbb{C}$ and its DFT $\widehat{f}$, the power 
spectrum $q(f,\omega)$ of $f$ at frequency $\omega$
is:
\begin{equation}
\label{eq:feature:powerspec}
q(f,\omega) := \overline{\widehat{f}(\omega)} \cdot \widehat{f}(\omega).
\end{equation}
The power spectrum is translation invariant since a translation $z$ of $f$ only affects the phases of the Fourier 
coefficients which are canceled 
out by $\overline{\widehat{f}(\omega)} \cdot \widehat{f}(\omega)$: 
\begin{eqnarray}
\overline{e^{-i2\pi z\omega/n}\widehat{f}(\omega)} \cdot e^{-i2\pi z\omega/n}\widehat{f}(\omega) &=&
e^{i2\pi z\omega/n}\overline{\widehat{f}(\omega)} \cdot e^{-i2\pi z\omega/n}\widehat{f}(\omega)\\
&=& \overline{\widehat{f}(\omega)} \cdot \widehat{f}(\omega).
\end{eqnarray}

We use the same principle to construct the ${\cal SH}_{abs}$-Features
(see section \ref{sec:feature:SHabs}). As mentioned in the context of ${\cal SH}_{abs}$, neglecting the valuable phase information makes
the power spectrum not a very discriminative feature.\\
The basic idea of the bispectrum is to couple two frequencies $\omega_1, \omega_2$ in order to implicitly preserve the phase information:
\begin{equation}
\label{eq:feature:bispec}
q(f,\omega_1,\omega_2) := \overline{\widehat{f}(\omega_1)} \cdot \overline{\widehat{f}(\omega_2)} \cdot \widehat{f}(\omega_1 + \omega_2).
\end{equation}
While the invariance property is the same as for the power spectrum:
\begin{eqnarray}
e^{i2\pi z\omega_1/n}\overline{\widehat{f}(\omega_1)}\cdot e^{i2\pi z\omega_2/n}\overline{\widehat{f}(\omega_2)}\cdot e^{-i2\pi z(\omega_1+\omega_2)/n} \widehat{f}(\omega_1 + \omega_2) =
\overline{\widehat{f}(\omega_1)} \cdot \overline{\widehat{f}(\omega_2)} \cdot \widehat{f}(\omega_1 + \omega_2),
\end{eqnarray}
it has been shown \cite{bispectrum} that the phases $\omega_i$ can be reconstructed from the bispectra. Hence, the bispectrum is a complete 
feature if $f$ is band limited and we extract the bispectrum at all frequencies.\\

Due to the analogy of the Spherical Harmonic and the Fourier domain, it is intuitive that the concept of the bispectrum is portable to signals
in $S^2$. This step was derived by \cite{shbispectrum} who constructed a global invariant feature for 2D images by projecting the images on 
the 2-sphere and then computing features in the harmonic domain. We adapt the methods from \cite{shbispectrum} to construct local rotation 
invariant features for 3D volume data.

\subsection{Feature Design}
In our case, we are interested in the extraction of invariant features of the local neighborhood $f$ at radius $r$ around the point 
${\bf x}$. Just as in the 1D example, we transform $f$ into the frequency space - i.e. in the Spherical Harmonic domain:
 $\widehat{f}:={\cal SH}\big(X|_{S[r]({\bf x})}\big)$.\\
Now, the individual frequencies $\omega$ correspond to the harmonic bands $\widehat{f}^l$, and \cite{shbispectrum} showed that the bispectrum
can be computed from the tensor product $(\widehat{f})^{l_1} \otimes (\widehat{f})^{l_2}$. \\
Further, we want to obtain invariance towards rotation instead of translation: given rotations ${\cal R}\in {\cal SO}(3)$, the tensor product
is affected by ${\cal R}$ in terms of:
\begin{equation}
\label{eq:feature:bispec0}
{\cal R}\left((\widehat{f})^{l_1} \otimes (\widehat{f})^{l_2}\right) = \left(D^{l_1}({\cal R}) \otimes D^{l_2}({\cal R})\right) 
\left((\widehat{f})^{l_1} \otimes (\widehat{f})^{l_2}\right), 
\end{equation}
where $D^l$ is the Wigner-D matrix for the $l$-th band (see section \ref{sec:feature:shrot}).\\
Just like in the 1D case, \cite{shbispectrum} proved that the bispectrum (\ref{eq:feature:bispec0}) will cancel out the impact of the rotation
$\cal R$. So, for the $l$-th band of expansion we can compute the bispectrum of the $l_1$-th and $l_2$-th band with $l_1,l_2\leq l$ by: 
\begin{equation}
\label{eq:feature:bispec1}
\left({\cal SH}_{bispectrum}\right)^{l_1,l_2,l} := \sum\limits_{m=-l}^l\sum\limits_{m_1=-l_1}^{l_1} \langle lm|l_1m_1,l_2m_2\rangle 
\overline{\widehat{f}^{l_1}_{m_1}} \cdot \overline{\widehat{f}^{l_2}_{(m-m_1)}}\cdot \widehat{f}^{l}_m,
\end{equation}
where the Clebsch-Gordan coefficients (see section \ref{sec:feature:clebschgordan}) $\langle lm|l_1m_1,l_2m_2 \rangle$ determine the impact of the 
frequency couplings in the tensor product computing the bispectrum. Refer to \cite{shbispectrum} for full proof.  

\subsection{Implementation}
As before, we follow the implementation of the Spherical Harmonic transformation as described in chapter \ref{sec:feature:implement} and
stop the expansion at an arbitrary band $b_{\max}$ (depending on the application) which has no effect on the rotation invariance.\\
The actual computation of bispectrum from (\ref{eq:feature:bispec1}) can be optimized by removing the $\widehat{f}^{l}_m $ term
to the outer iteration and limiting the inner iteration to values which form possible Clebsh-Gordan combinations:
\begin{equation}
\label{eq:feature:bispec2}
\left({\cal SH}_{bispectrum}\right)^{l_1,l_2,l}= \sum\limits_{m=-l}^l \widehat{f}^{l}_m \times 
\sum\limits_{m_1=\max{-l_1,(m-l_2)}}^{\min{l_1,(m+l_2)}} \langle lm|l_1m_1,l_2m_2 \rangle \overline{\widehat{f}^{l_1}_{m_1}} \cdot 
\overline{\widehat{f}^{l_2}_{(m-m_1)}}.
\end{equation}

\paragraph{Multi-Channel Data:} It is straightforward to combine the information from two different data channels into a single
${\cal SH}_{bispectrum}$-Feature: we can simply choose the coefficients $\widehat{f}^{l_1}$ and $\widehat{f}^{l_2}$ from two t
expansions of the data from two different channels.

\subsubsection{Complexity}
The computational complexity of a singe $\left({\cal SH}_{bispectrum}\right)^{l_1,l_2,l}({\bf x})$ feature lies in $O(l^3)$. To obtain 
completeness we need all $O(b_{\max}^2)$ features at all $m$ positions of $X$. The harmonic expansion to band $b_{\max}$ at each
point takes another $O(m(b_{\max})^2 + (m \log m))$.
\paragraph{Parallelization}
It is straightforward to get further speed-up by parallelization (see chapter \ref{sec:feature:implement}). Since the computation of each 
single feature 
 $\left({\cal SH}_{bispectrum}\right)^{l_1,l_2,l}({\bf x})$ is independent from all others, we can split the overall process in   
parallel computations.  

\subsection{Discussion}
The basic concept of the ${\cal SH}_{bispectrum}$-Features is quite similar to what we did for the ${\cal SH}_{phase}$-Features (see section
\ref{sec:feature:SHphase}): we try to obtain a better discrimination performance than ${\cal SH}_{abs}$-Features by implicit preservation 
of the phase information. In case of the ${\cal SH}_{phase}$-Features we do this by considering the relation of phases over different radii of the expansion,
here we relate different frequencies of the expansion. In theory, the completeness property makes the ${\cal SH}_{bispectrum}$ approach very 
competitive, but this comes at high computational costs.

\cleardoublepage
\chapter{\label{sec:feature:SHHaar}Scalar Haar-Features}
\index{Haar-Feature}\index{Scalar Haar-Feature}
In this chapter we derive several features operating on scalar data which obtain invariance via Haar-Integration. As discussed in 
section \ref{sec:feature:invariance}, one canonical approach to construct invariant features is to perform a Haar-Integration
over the transformation group.\\
Before we turn to the specific feature design, we first review the general framework of Haar-Integration in section 
\ref{sec:feature:invariance_via_groupintegration} and discuss some aspects of the construction of suitable feature kernels in
section \ref{sec:feature:haarkernels}. Then we introduce $2p$-Haar features \ref{sec:feature:2p} and $3p$-Haar features \ref{sec:feature:3p}
which are based on the class of separable kernel functions, before we derive the generalized $np$-Haar features.\\
 
It should be noted that we also use Haar-Integration methods for the computation of the auto-correlation features ${\cal SH}_{autocorr}$
(see section \ref{sec:feature:SHautocorr}) and ${\cal VH}_{autocorr}$ (see section \ref{sec:feature:VH_autocorr}).

\paragraph{Related Work:} 
Based on the general group-integration framework (\ref{eq:feature:general_haar_integral}) which was introduced by \cite{DissMirbach}, \cite{Schulz:Mirbach95} and
\cite{Burkhardt2001}, several invariant features were introduced for scalar data in 2D \cite{Schulz:Mirbach95} and in 3D volumetric data
\cite{schael3D} \cite{Ronneberger2002} \cite{dagm1} \cite{iasted}.
We will discuss these methods in the next section when we take a closer look at the class of sparse and separable
kernels \cite{Ronneberger2002} \cite{dagm1} \cite{iasted} which form the basis of our features.

\subsection{\label{sec:feature:invariance_via_groupintegration} Invariance via Group-Integration}
\index{Haar-Feature}\index{Scalar Haar-Feature}\index{General Haar-Framework}
Following the general objectives of feature extraction (see \ref{sec:feature:extraction}) we apply the Haar-Intergration approach to obtain
invariant features. This method is generally bound to the sub-class of compact group transformations (see \ref{sec:feature:grouptrans}),
where for a given transformation group ${\cal G}$, the individual transformations $g \in {\cal G}$ differ only by their associated set of 
parameters $\vec{\lambda}$, which cover the degrees of freedom under ${\cal G}$.\\
 
In this chapter we derive features from the canonical group integration approach (see section \ref{sec:feature:invariance}) which generates invariant features 
via Haar-Integration over all degrees of freedom of the transformation group $G$:
\begin{equation}
\label{eq:feature:haar_integral}
T(X) = \int\limits_{\cal G} (g_{\vec{\lambda}}{X})dg_{\vec{\lambda}},
\end{equation}
eliminating the influence of $\vec{\lambda}$. Equation (\ref{eq:feature:haar_integral}) is also referred to as the ``group-average''.
For the cause of simplicity, we denote individual transformation $g_{\vec{\lambda}}$ just by $g$.\\

It has to be noted that even though the Haar-Integration approach  (\ref{eq:feature:haar_integral}) meets the necessary
condition of invariance (\ref{eq:feature:Necessary}), the sufficient condition (\ref{eq:feature:sufficient}) is anything but guaranteed. 
In fact, a simple group-averaging itself produces incomplete features which often tend to have a weak separability performance. 
This can be overcome by embedding non-linear kernel functions $\kappa$
into the integral \cite{DissMirbach}: it cannot be stressed enough that the use of such non-linear mappings is essential for any feature design
\cite{habilBu} \cite{DissMirbach} \cite{Schulz:Mirbach95} \cite{Ronneberger2002} \cite{iasted}, 
and is the key element of the group-integration framework. The resulting general framework for invariant
feature generation via group-integration (\ref{eq:feature:general_haar_integral}) then embeds an arbitrary 
non-linear kernel function $\kappa$.

\begin{eqnarray}
  T(X) := \int\limits_{\cal G}
  \kappa(g{X})dg
  \label{eq:feature:general_haar_integral}
\end{eqnarray}
  \begin{tabular}{r@{~:~}l}
    ${\cal G}$          & transformation group\\
    $g$          & one element of the transformation group \\
    $dg$         & Haar measure\\
    $\kappa$          & nonlinear kernel function\\
    ${X}$ & $n$-dim, multi-channel data set \\
    $g{X}$  & the transformed $n$-dim data set
  \end{tabular}
\\[0.25cm]
Within this framework, features can be generated for data of arbitrary dimensionality and from multiple input channels.
This reduces the key design issue to the selection of appropriate kernel functions. 

\subsection{\label{sec:feature:haarkernels} Local, Sparse and Separable Kernel Functions}
Since we are interested in the construction of local features, we restrict the kernel functions $\kappa$ in the general group-integration
approach (\ref{eq:feature:general_haar_integral}) to functions of local support.
\begin{figure}[htbp]
\centering
\psfrag{k(x1)}{\tiny$\kappa(X(s_g(\phi)({\bf x}_1)))$} 
\psfrag{k(x2)}{\tiny$\kappa(X(s_g(\phi)({\bf x}_2)))$}
\psfrag{k(x3)}{\tiny$\kappa(X(s_g(\phi)({\bf x}_3)))$}
\psfrag{x1}{${\bf x}_1$}
\psfrag{x2}{${\bf x}_2$}
\psfrag{x3}{${\bf x}_3$}
\psfrag{2pi}{$2\pi$}
\psfrag{0}{$0$}
\includegraphics[width=0.75\textwidth]{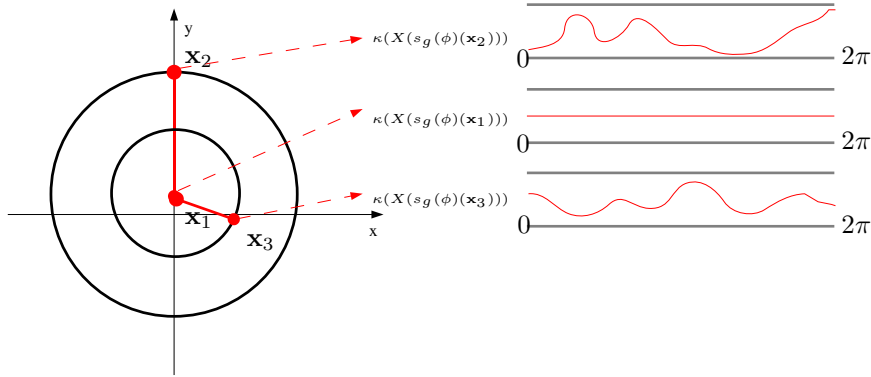}
  \caption[Sparse kernels for Haar-Integration]{ Using a local sparse kernel with three points on a continuous 2D image.
    Regarding the rotation group, this kernel returns the gray values
    sensed by the kernel points ${\bf x}_1,{\bf x}_2$ and
    ${\bf x}_3$ as one-dimensional functions $X(s_g(\phi)({\bf x}_1))$,
    $X(s_g(\phi)({\bf x}_2))$ and $X(s_g(\phi)({\bf x}_1))$. Invariant features are 
    computed by combining the kernel points
    with a nonlinear kernel function $\kappa$ which is then integrated over all
    possible rotations (parameterized in $\phi$).}
  \label{fig:feature:sparse_kernel_example}
\end{figure}
\noindent Further, following the approach in \cite{Ronneberger2002} \cite{iasted}, we restrict these local kernels to sparse functions which
only depend on a few discrete points of the local continuous data. Hence,  $\kappa({X})$ can be rewritten  as
  $\kappa\left(
  {X}({\bf x}_1), {X}({\bf x}_2), {X}({\bf x}_3),
  \dots\right)$ \cite{iasted}.\\
This way, we can reformulate (\ref{eq:feature:general_haar_integral}) and perform
the group transformation only on the local kernel support,
instead of the whole data set ${X}$ (see Fig. \ref{fig:feature:sparse_kernel_example}). This local transformation
is denoted as $s_g({\bf x}_i)$ such that
\begin{equation}
  (g{X})({\bf x}_i) =: {X}(s_g({\bf x}_i))   \quad \forall g,
  {\bf x}_i.
\end{equation}

For these local kernels, (\ref{eq:feature:general_haar_integral}) can be rewritten as
\begin{eqnarray}
  T(X) := \int\limits_G
  \kappa\left(
{X}(s_g({\bf x}_1)),\; {X}(s_g({\bf x}_2)),
 \dots \right)dg.
\label{eq:feature:haar_gray_scale_invariant}
\end{eqnarray}
Fig. (\ref{fig:feature:sparse_kernel_example}) shows how a {\bf sparse kernel} with a local support of three discrete points can be applied 
to ``sense'' the local continuous data.\\

For kernels with a larger support it does not make much sense to combine single data points over a certain distance. Instead we are interested
in combining larger structures, i.e. in having a kernel over regions rather than over single points. One very simple way to achieve this was
suggested in \cite{Ronneberger2002}: by applying a Gaussian smoothing of the input data which directly depends on the selected size of the 
local support, we can define a ``{\bf multi-scale}'' kernel which has different sizes of local support in each point.\\ 

This class of {\bf local sparse kernel} functions provides a more structured but still very flexible framework for the design of local 
invariant features.
However, even with a support reduced to $n$ discrete points, naive kernel computation is still very expensive since the local support has to be
integrated over the entire transformation group.
\cite{Schulz:Mirbach95}\cite{Schael2000}\cite{Burkhardt2001} suggested to overcome this problem by the use of
Monte Carlo methods, but this approach is only effective
when features are computed via integration over the entire dataset (i.e. integration over the translation group). For the
computation of local features, i.e. a Monte Carlo integration over the rotation group is not suitable.\\

To make group-integral features applicable to large data sets, \cite{Ronneberger2002} introduced
a sub-class of sparse kernel-functions. For these so called {\bf separable kernels}, the kernel can be split into a
linear combination of non-linear sub-kernels such that:
\begin{eqnarray}
    \kappa\left({X}(s_g({\bf x}_1)),\; {X}(s_g({\bf x}_2)), \dots \right) = \kappa_1\left({X}(s_g({\bf x}_1))\right)\cdot \kappa_2\left({X}(s_g({\bf x}_2))\right) \cdot \dots .
     \label{eq:feature:haar_separable_kernel}
\end{eqnarray}
This separability constraint is not a strong limitation of the original class of sparse kernel-functions since in many cases it is possible to find
approximative decompositions of non-separable kernels via Taylor series expansion.\\
Besides the non-linearity, the choice of the sub-kernels $\kappa_i$ is theoretically not further constrained, but in most cases very simple
non-linear mappings such as $\kappa(x) = x^2, \kappa(x) = x^3,\dots$ or $\kappa(x) = \sqrt{x}$ are powerful enough (see experiments in part 
III).\\

Based on these separable kernels, \cite{Ronneberger2002} derived a fast convolution method for the evaluation of kernels with a support of 
only two sparse points on continuous 2D images - so called ``2-point'' kernels (see section \ref{sec:feature:2p}).

\newpage
%------------------------
% 2p
%------------------------
\section{\label{sec:feature:2p}2-Point Haar-Features ($2p$)}
\index{Haar-Feature}\index{Scalar Haar-Feature}\index{2p-Feature}
Our first feature which makes use of the general group integration framework (\ref{eq:feature:general_haar_integral}) is the so-called
2-Point or $2p$-Haar feature. It was first introduced as a global feature for 2D images in \cite{Ronneberger2002}. We later extended this
approach 
to local features on scalar 3D volume data in \cite{dagm1} and \cite{ro:fehr:bu:report05} with an application to biomedical 3D image
analysis in \cite{dagm2} (see part III).\\

$2p$-Features use a sub-class of the previously introduced separable kernel functions (\ref{eq:feature:haar_separable_kernel}). The name 2-Point
derives from the constraint that we restrict kernels to have just two separable kernel points ${\bf x}_1, {\bf x}_2$.
This restriction allows a reformulation of the initial parameterization $\bf \lambda$ of the rotation group ${\cal SO}(3)$, 
which is drastically 
reducing the computational complexity necessary to obtain rotation invariance. However, this comes at the price of reduced 
discrimination power as we discuss at the end of this section. 
 
\subsection{\label{sec:feature:2p_design}Feature Design}
\begin{figure}[ht]
\centering
\psfrag{r}{$r$}
\psfrag{x1}{${\bf x}_1$}
\psfrag{x2}{${\bf x}_2$}
\includegraphics[width=0.3\textwidth]{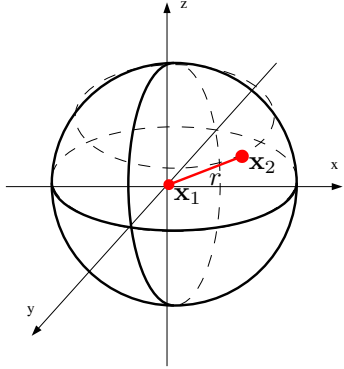}
\caption[$2p$-Haar Features.]{\label{fig:feature:2phaarfeature} Schematic example of a $2p$-Haar Feature: the first kernel point is set to the
center of the local features extraction ${\bf x}_1 := X({\bf x})$, while the second kernel point lies on the spherical neighborhood with 
radius $r$: ${\bf x}_2 \in S[r]\left({\bf x}\right)$}
\end{figure}

The selection of the kernel points ${\bf x}_1$ and ${\bf x}_2$ is bound by the following design principle for the 2-Point features: 
For the extraction of a local $2p$-Feature at a given point ${\bf x}$ in $X$ of the scalar (or possibly multi-channel) 3D input volume $X$, 
${\bf x}_1$ is fixed at the center of the neighborhood, i.e. ${\bf x}_1 := X({\bf x})$. The second kernel point is chosen from the local
neighborhood: ${\bf x}_2 \in S[r]\left({\bf x}\right)$ (see \ref{eq:feature:subpatterndef} for the neighborhood definition).\\
Since ${\bf x}_1$ is fixed, we only have to choose the parameters for ${\bf x}_2$: the local neighborhood $r$ and the spherical coordinates
$\Phi, \Theta$ which can be neglected later on.\\
We are using the scheme for separable kernels (\ref{eq:feature:haar_separable_kernel}), we can write the $2p$-Kernels as:     
\begin{equation}
  \kappa\big(X({\bf x}_1), X({\bf x}_2)\big) =
\kappa_1\big(X({\bf x}_1)\big)\cdot
\kappa_2\big(X({\bf x}_2)\big).
\label{eq:feature:2p_formular}
\end{equation}
Figure \ref{fig:feature:2phaarfeature} shows a schematic example of a local 3D $2p$ kernel on volume data. 

\subsubsection{Rotation Invariance}
As for all other local features, we want to obtain rotation invariance. If we plug the $2p$ kernel (\ref{eq:feature:2p_formular}) into
the general Haar framework (\ref{eq:feature:general_haar_integral}), we can achieve invariance regarding rotations 
${\cal R(\phi,\theta, \psi)} \in {\cal SO}(3)$ parameterized in Euler angles (see section \ref{sec:feature:shrot}) with local transformations
(\ref{eq:feature:haar_separable_kernel}) $s_{\cal R}(\phi,\theta, \psi) \in {\cal SO}(3)$. Since ${\bf x}_1$ is by definition always in
the rotation center, it is not affected by any rotation. Hence we can simplify the Haar-Integration for the multiplicative and separable
$2p$-Kernel functions:
\begin{equation} 
T[r,{\bf x}_2]({\bf x}) := \kappa_1\big(X({\bf x})\big)\cdot \int\limits_{{\cal SO}(3)} 
\kappa_2\big(X( s_{\cal R(\phi,\theta,\psi)}({\bf x}_2))\big) \sin{\theta}d\phi d\theta d\psi.
\label{eq:feature:2prot}
\end{equation}

\subsubsection{Fast Computation}
In order to compute (\ref{eq:feature:2prot}) we have to evaluate the integral over all possible rotations at each point $X({\bf x})$, which 
turns out to be quite expensive in terms of computational complexity. At this point, the restriction of 
(\ref{eq:feature:haar_separable_kernel}) to two points provides us with a fast solution: due to the fact that we have to integrate only over
the position of a single point ${\bf x}_2 \in S[r]\left(X({\bf x})\right)$, the integral over $\psi$ becomes a constant factor and we can 
rewrite (\ref{eq:feature:2prot}) as:
\begin{equation}
T[r,{\bf x}_2]({\bf x}) = \kappa_1\big(X({\bf x})\big)\cdot \int\limits_{\phi,\theta} 
\kappa_2\big(X(s_{\cal R(\phi,\theta,\psi)}({\bf x}_2))\big) \sin{\theta}d\phi d\theta.
\label{eq:feature:2protfast}
\end{equation}
Since ${\bf x}_2 \in S[r]\left({\bf x}\right)$ is also parameterized in $\phi,\theta$, we can further reformulate the integral
and simply solve:
\begin{equation}
T[r,{\bf x}_2]({\bf x}) = \kappa_1\big(X({\bf x})\big)\cdot \int\limits_{{\bf x_i} \in {\cal S}[r]({\bf x})} 
\kappa_2\big(X\big|_{{\cal S}[r]({\bf x})} ({\bf x}_i)\big). 
\label{eq:feature:2protfast2}
\end{equation}
Finally, the integration over a spherical neighborhood $S[r]\left({\bf x}\right)$ can easily be formulated as convolution of 
$X\big|_{{\cal S}[r]({\bf x})}$ with a 
spherical template $S_t[r]$ with $S_t[r](\Phi,\Theta) = 1,\text{\quad} \forall \Phi \in [0,\dots, 2\pi], \Theta \in [0,\dots, \pi]$: 
\begin{equation}
T[r]({\bf x}) = \kappa_1\big(X({\bf x})\big)\cdot \big(\kappa_2(X\big|_{{\cal S}[r]({\bf x})}) * S_t[r]({\bf x})\big).
\label{eq:feature:2protfast3}
\end{equation}

In the same way, we
can evaluate the $2p$-Feature at all positions in $X$ at once, using fast convolution in the Fourier domain:
\begin{equation}
T[r](X) = \kappa_1(X)\cdot \big(\kappa_2(X) * S_t[r]\big).
\label{eq:feature:2protconvolve}
\end{equation}
\begin{figure}[ht]
\centering
\psfrag{k(x1)}{\tiny$\kappa(X(s_g(\phi)({\bf x}_1)))$}
\psfrag{k(x2)}{\tiny$\kappa(X(s_g(\phi)({\bf x}_2)))$}
\psfrag{k(x3)}{\tiny$\kappa(X(s_g(\phi)({\bf x}_3)))$}
\psfrag{x1}{${\bf x}_1$}
\psfrag{x2}{${\bf x}_2$}
\psfrag{x3}{${\bf x}_3$}
\psfrag{2pi}{$2\pi$}
\psfrag{0}{$0$}
\includegraphics[width=0.75\textwidth]{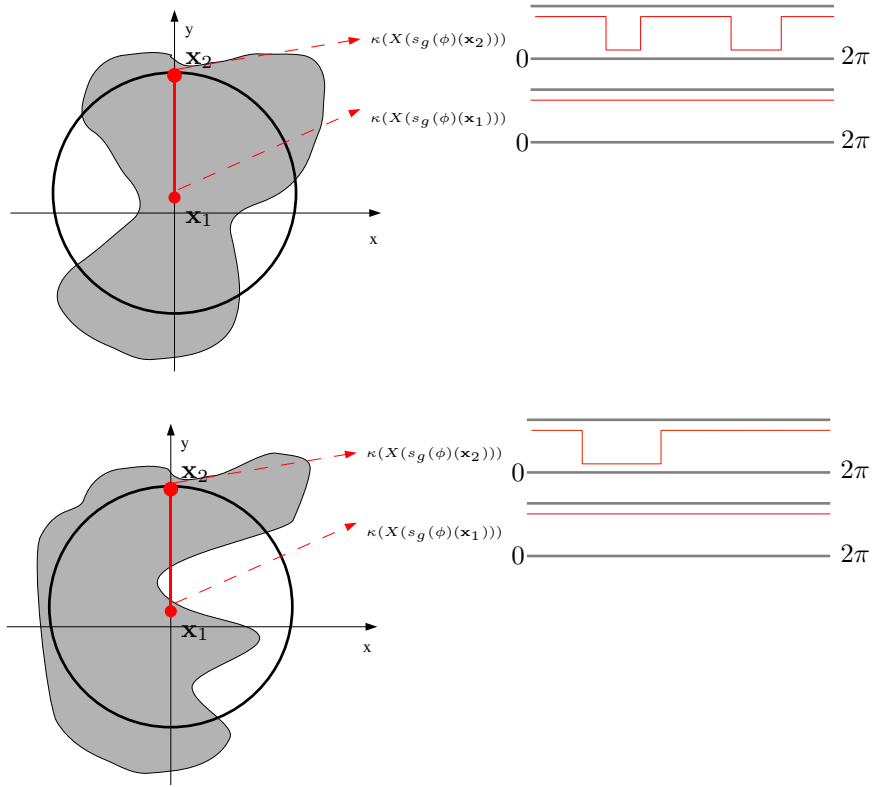}
\caption[Ambiguities of $2p$-Features.]{\label{fig:feature:2pambiguiety} Ambiguities of $2p$-Features: This binary toy example 
illustrates the rather weak separability performance of $2p$-Features. The integral $\int\limits_{\phi=0}^{2\pi}\kappa(X(s_g(\phi)({\bf x}_2)))
d\phi$ returns equal values for different local neighborhoods.}
\end{figure} 

\subsection{Implementation}
The implementation is straightforward: given discrete input data, we apply the convolution theorem to compute the convolution via
FFT:
\begin{equation}
T[r](X) = \kappa_1(X)\cdot { FFT^{-1}}\big({ FFT}(\kappa_2(X)) \cdot { FFT}(S_t[r])\big).
\label{eq:feature:2pimplement}
\end{equation}
The only thing we have to handle with some care is the implementation of the spherical template $S_t[r]$. To avoid sampling issues, we apply
the same implementation strategies as in the case of the Spherical Harmonic base functions (see section \ref{sec:feature:SHimplement}
for details).

\begin{figure}[ht]
\centering
\psfrag{C1}{\tiny$X[c_1]]$}
\psfrag{C2}{\tiny$X[c_2]$}
\psfrag{f1}{$\kappa_1$}
\psfrag{f2}{$\kappa_2$}
\psfrag{FT}{\tiny$ FFT$}
\psfrag{FT-1}{\tiny${FFT}^{-1}$}
\includegraphics[width=0.8\textwidth]{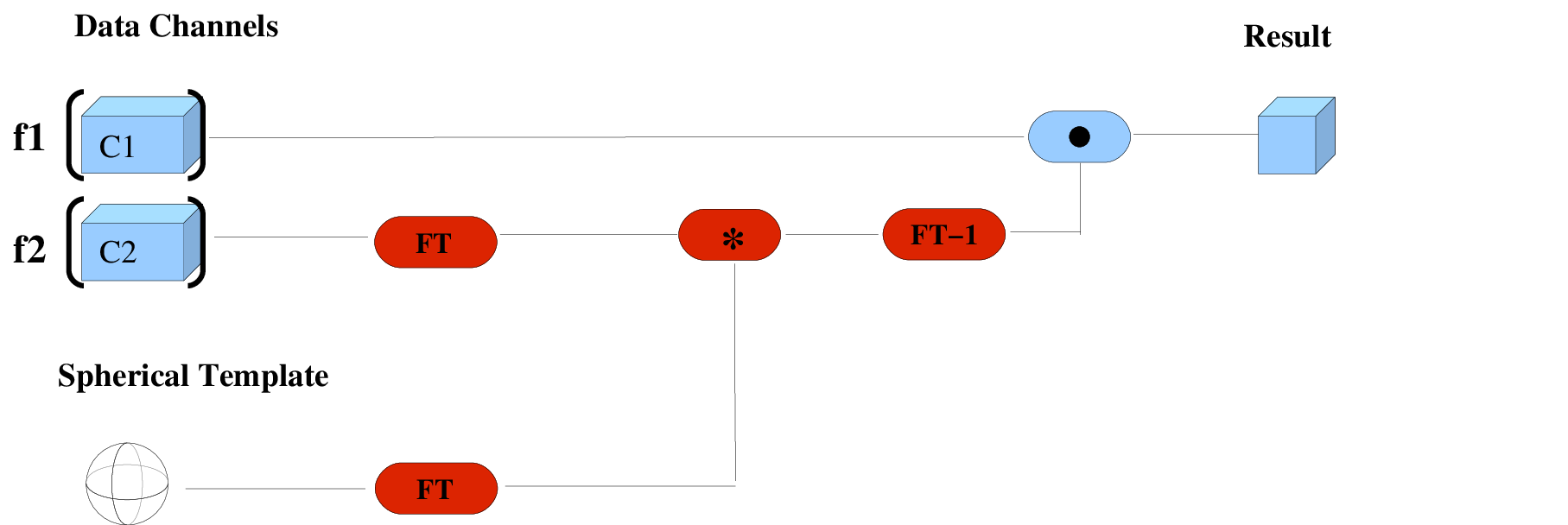}
\caption[$2p$-Feature: implementation.]{\label{fig:feature:2pimplement} Schematic overview of the implementation of $2p$-Features.  }
\end{figure}

\paragraph{Multi-Channel Data:} Naturally, the application of $2p$-Features to multi-channel data is limited to two channels per feature,
but this is straightforward: we can simply set the kernel points to be on different data channels $c_i$: 
\begin{equation}
T[r](X) = \kappa_1\big(X[c_1]\big)\cdot \big(\kappa_2(X[c_2]\big|_{{\cal S}[r]({\bf x})}) * S_t[r]\big).
\label{eq:feature:2pmultichannel}
\end{equation}
\paragraph{Complexity:} By reducing the feature computation to a fast convolution, we end up with a complexity of $O( m \log m)$ for
an input volume with $m$ voxels.

\paragraph{Parallelization:} Since there is no easy way to parallelize the Fourier Transformation, we do not further parallelize the computation
of $2p$-Features. However, because $2p$-Features can be computed so fast anyway, this is not a real drawback. 

\subsection{\label{sec:feature:2p_discussion}Discussion}
The best property of $2p$-Features is their computational speed: no other spherical local 3D feature, neither in the context of this work 
nor in
the literature can be computed this fast. However, the speed comes at the price of a rather low discrimination power and the lack of
gray-scale robustness. While one might try to compensate the missing gray-scale robustness by pre-normalization of the input data, the 
discrimination power hardly can be improved.\\  
The problem is caused by the fact that $2p$-Features are not only invariant under rotations, but also under arbitrary permutations of
signals on the sphere.
This causes problematic ambiguities, as illustrated in figure \ref{fig:feature:2pambiguiety}.

\newpage
%---------------------------------------------------------
% 3p
%---------------------------------------------------------
\section{\label{sec:feature:3p}3-Point Haar-Features ($3p$)}
\begin{figure}[ht]
\centering
\psfrag{k(x1)}{\tiny$\kappa(X(s_g(\phi)({\bf x}_1)))$}
\psfrag{k(x2)}{\tiny$\kappa(X(s_g(\phi)({\bf x}_2)))$}
\psfrag{k(x3)}{\tiny$\kappa(X(s_g(\phi)({\bf x}_3)))$}
\psfrag{x1}{${\bf x}_1$}
\psfrag{x2}{${\bf x}_2$}
\psfrag{x3}{${\bf x}_3$}
\psfrag{2pi}{$2\pi$}
\psfrag{0}{$0$}
\includegraphics[width=0.75\textwidth]{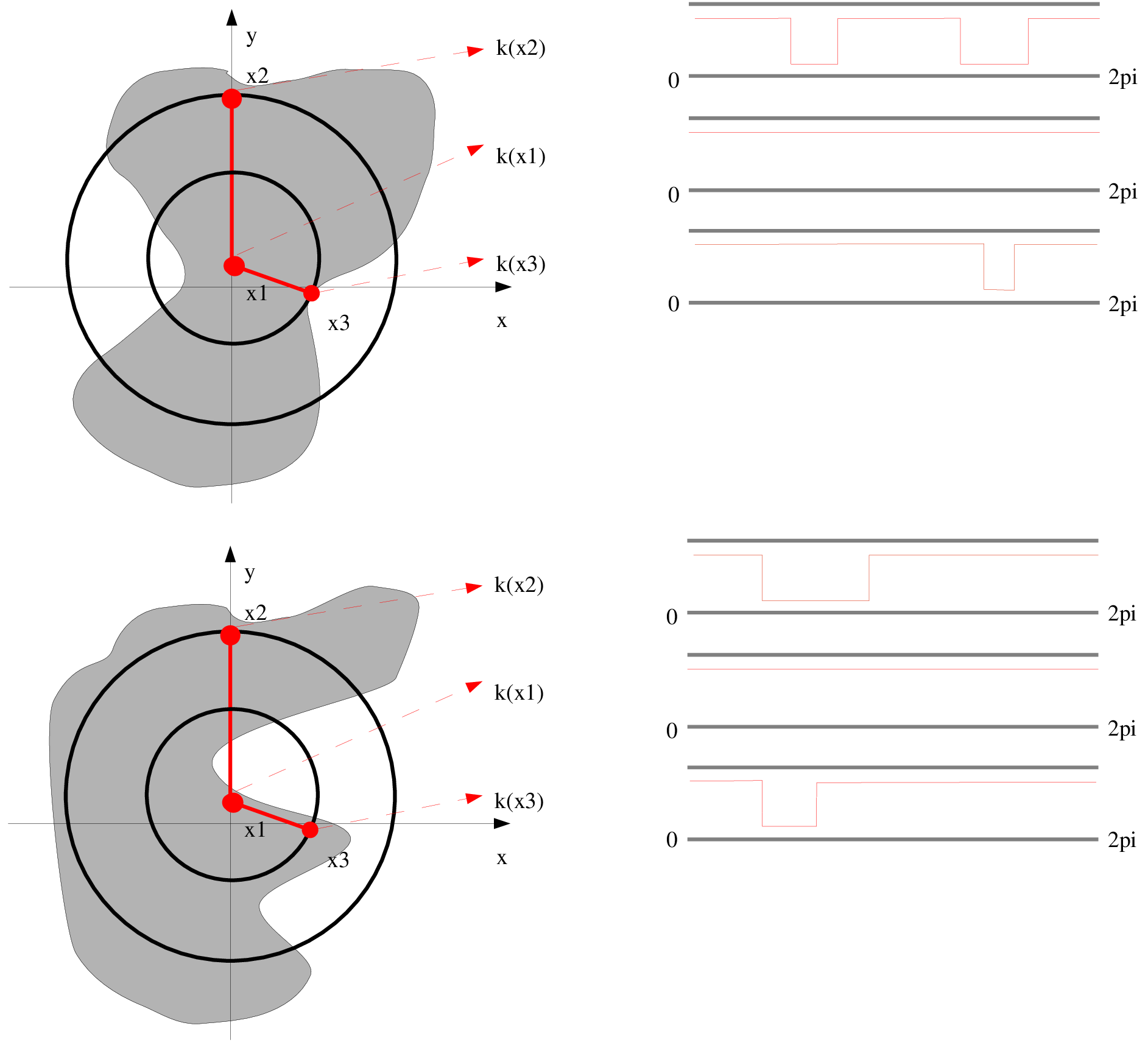}
\caption[Ambiguities of $2p$-Features.]{\label{fig:feature:2pambiguiety3p} Reducing ambiguities of $2p$-Haar features: for the same binary toy
example as in figure \ref{fig:feature:2pambiguiety}, the addition of a third kernel point leads to different results of the integral
$\int\limits_{\phi=0}^{2\pi}\kappa(X(s_g(\phi)({\bf x}_2)))\kappa(X(s_g(\phi)({\bf x}_3))) d\phi$ for different local neighborhoods.}
\end{figure}

\index{Haar-Feature}\index{Scalar Haar-Feature}\index{3p-Feature}
3-Point Haar-Features (or $3p$-Features) are a direct extension of separable kernels (\ref{eq:feature:haar_separable_kernel}) 
from two (\ref{eq:feature:2p_formular}) to three kernel points. The main motivation for this extension derives from 
the discussion of the $2p$-Features (see section \ref{sec:feature:2p_discussion}), where we pointed out that even though 2-point kernels 
(\ref{eq:feature:2p_formular})
provide computationally very efficient features, the resulting discrimination power is flawed by the fact that these kernels are also
invariant to arbitrary permutations.\\
To overcome this major drawback, we introduced the $3p$-Features in \cite{dagm1} and \cite{iasted}. The basic idea is to add a third kernel
point ${\bf x}_3$ to the separable kernel function $\kappa$ (\ref{eq:feature:haar_separable_kernel}) (see figure 
\ref{fig:feature:2pambiguiety3p}), which cancels out the permutation ambiguities:  

\begin{equation}
  \kappa\big(X({\bf x}_1), X({\bf x}_2), X({\bf x}_3)\big) =
\kappa_1\big(X({\bf x}_1)\big)\cdot
\kappa_2\big(X({\bf x}_2)\big)\cdot
\kappa_3\big(X({\bf x}_3)\big).
\label{eq:feature:3p_formular}
\end{equation}

\subsection{\label{sec:feature:3p_design}Feature Design}

\begin{figure}[ht]
\centering
\psfrag{r2}{$r_2$}
\psfrag{r3}{$r_3$}
\psfrag{x1}{${\bf x}_1$}
\psfrag{x2}{${\bf x}_2$}
\psfrag{x3}{${\bf x}_3$}
\includegraphics[width=0.3\textwidth]{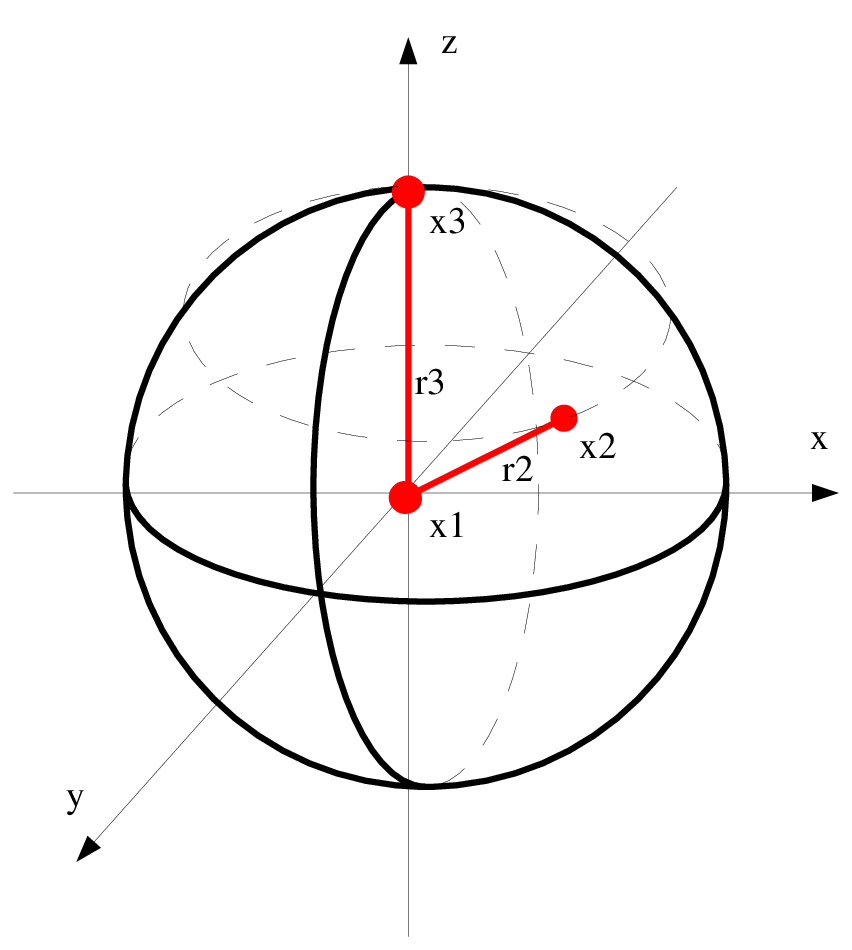}
\includegraphics[width=0.3\textwidth]{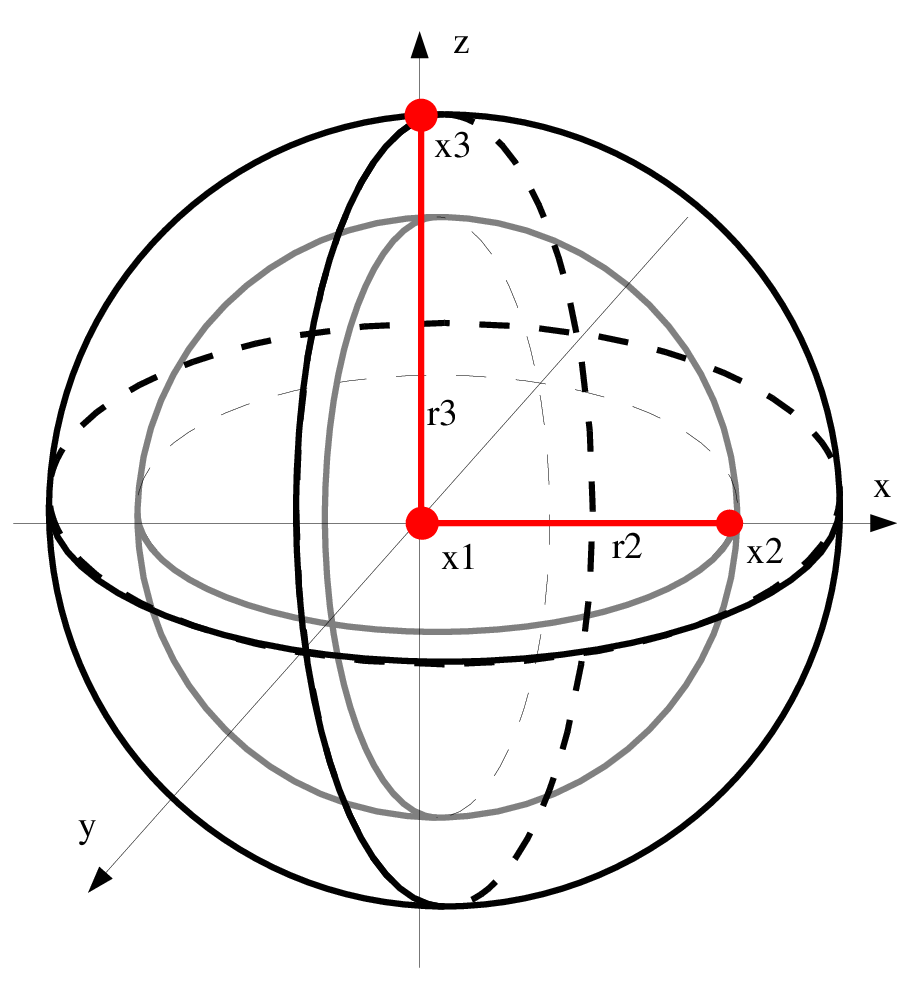}
\caption[$3p$ kernels: a schematic overview.]{\label{fig:feature:3p:scheme}Schematic examples of  $3p$-Kernels. {\bf Left:} both kernel 
points ${\bf x}_2,{\bf x}_3$ are located on the same concentric sphere ($r_2=r_3$). {\bf Right:} ($r_2\neq r_3$). The first kernel point is set to 
the center of the local features extraction ${\bf x}_1 := X({\bf x})$, while the other kernel points lie on the spherical neighborhoods with 
radii $r_2,r_3$: ${\bf x}_i \in S[r_i]\left({\bf x}\right)$ }
\end{figure}

As in the $2p$ case, we fix the first kernel point ${\bf x}_1 := X({\bf x})$ at the point of the local feature extraction,
while the other two points are placed at the concentric spherical neighborhoods surrounding the first point:
${\bf x}_2  \in S[r_2]\left({\bf x}\right),{\bf x}_3 \in S[r_3]\left({\bf x}\right)$.\\
Of course, both kernel points ${\bf x}_2, {\bf x}_3$ can be on the same sphere, resulting in $r_2 = r_3$, and are parameterized in spherical 
coordinates $\Phi_2,\Phi_3$ and $\Theta_2, \Theta_3$. Figure \ref{fig:feature:3p:scheme} shows examples of such $3p$-Kernels. 

\subsubsection{Rotation Invariance}
\begin{figure}[ht]
\centering
\psfrag{r2}{$r_2$}
\psfrag{r3}{$r_s$}
\psfrag{x1}{${\bf x}_1$}
\psfrag{x2}{${\bf x}_2$}
\psfrag{x3}{${\bf x}_3$}
\psfrag{phi}{$\phi$}
\psfrag{theta}{$\theta$}
\psfrag{psi}{$\psi$}
\includegraphics[width=0.3\textwidth]{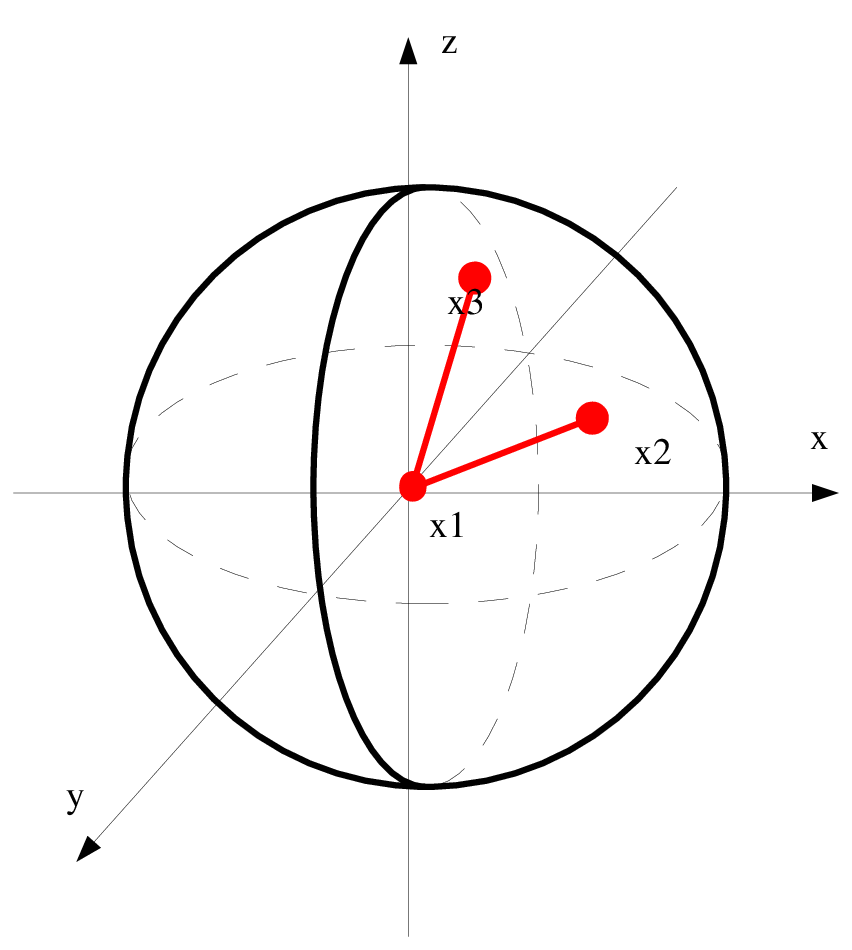}
\includegraphics[width=0.3\textwidth]{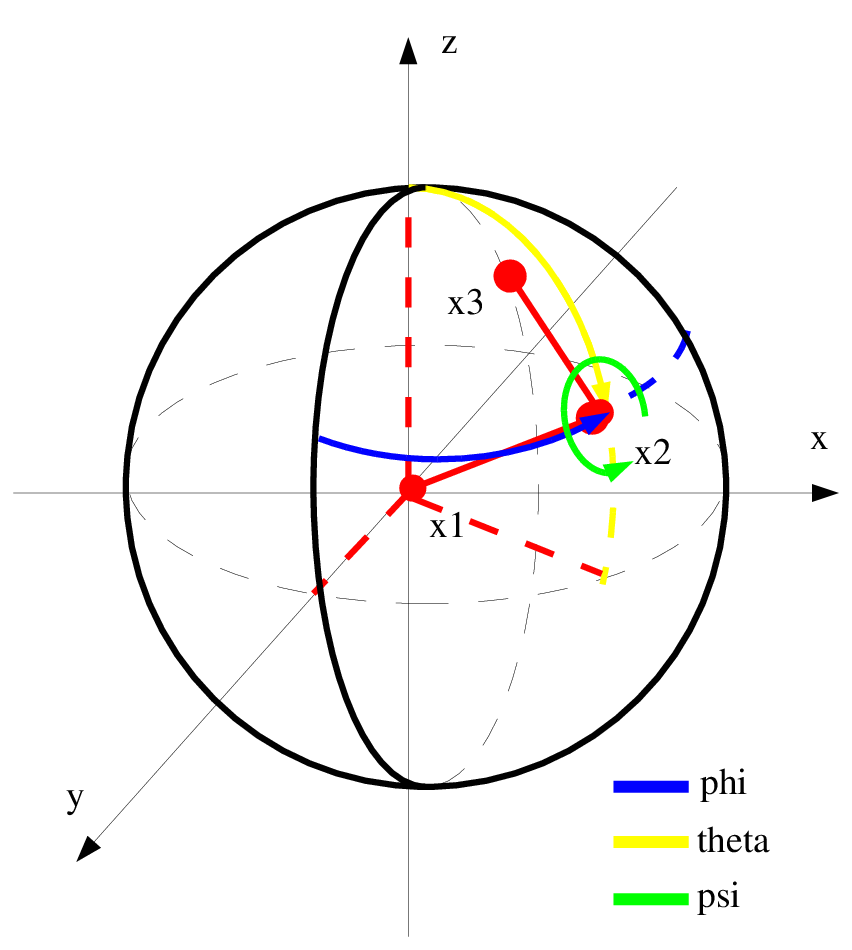}
\caption[$3p$-Feature: Dependency of the kernel points.]{\label{fig:feature:3pkernelpoints} {\bf Left:} Arbitrary $3p$ kernel with an 
independent formulation of the points ${\bf x}_2,{\bf x}_3$. {\bf Right:} formulation ${\bf x}_3$ in dependency of ${\bf x}_2$.}
\end{figure}

If we plug the $3p$ kernel (\ref{eq:feature:3p_formular}) into
the general Haar framework (\ref{eq:feature:general_haar_integral}), we can achieve invariance regarding rotations
${\cal R(\phi,\theta, \psi)} \in {\cal SO}(3)$ parameterized in Euler angles (see section \ref{sec:feature:shrot}) with local transformations
(\ref{eq:feature:haar_separable_kernel}) $s_{\cal R}(\phi,\theta, \psi) \in {\cal SO}(3)$. As in the $2p$ case, ${\bf x}_1$ is by definition 
always in the rotation center, hence it is not affected by any rotation. This way, we end up with the Haar-Integration approach for the 
separable $3p$-kernel functions:
\begin{eqnarray}
\label{eq:feature:3prot}
T[r_1,r_2,{\bf x}_2,{\bf x}_3]({\bf x}) &:=& \kappa_1\big(X({\bf x})\big)\cdot\\
&&\int\limits_{{\cal SO}(3)} 
 \kappa_2\big(X(s_{\cal R(\phi,\theta,\psi)}({\bf x}_2))\big) \cdot
 \kappa_2\big(X(s_{\cal R(\phi,\theta,\psi)}({\bf x}_3))\big).
\sin{\theta}d\phi d\theta d\psi\nonumber
\end{eqnarray}
We can further simplify this integral by the same considerations we made in (\ref{eq:feature:2protfast}): since the kernel points 
${\bf x}_2,{\bf x}_3$ are not rotated independently, we express (without loss of generality) ${\bf x}_3$ in dependency of ${\bf x}_2$
(see Figure \ref{fig:feature:3pkernelpoints}).
The integral over $\psi$ is a constant factor in ${\bf x}_2$ (as shown in (\ref{eq:feature:2protfast})), but for each position of ${\bf x}_2$
the dependency of ${\bf x}_3$ is expressed in terms of the angle $\psi$. Hence we have to integrate over all $\psi$ in ${\bf x}_3$:
\begin{eqnarray}
\label{eq:feature:3protfast}
T[r_1,r_2,{\bf x}_2, {\bf x}_3]({\bf x}) &:=& \kappa_1\big(X({\bf x})\big)\cdot\\
&&\int\limits_{\phi,\theta}
\kappa_2\big(X(s_{\cal R(\phi,\theta)}({\bf x}_2))\big) \int\limits_{\psi}
 \kappa_2\big(X(s_{\cal R(\phi,\theta)}({\bf x}_3))\big)
\sin{\theta}d\phi d\theta d\psi.\nonumber
\end{eqnarray}
\subsubsection{Fast Computation}
\begin{figure}[ht]
\centering
\psfrag{r3}{$r_c$}
\psfrag{x1}{${\bf x}_1$}
\psfrag{x2}{${\bf x}_2$}
\psfrag{x3}{${\bf x}_3$}
\includegraphics[width=0.3\textwidth]{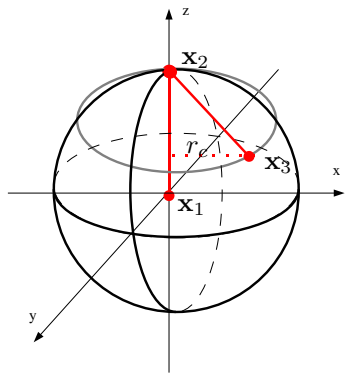}
\caption[$3p$-Feature: re-parameterization of the kernel points.]{\label{fig:feature:3pparam} Re-parameterization of the kernel points. Given
rotations in ${\cal SO}(3)$, ${\bf x}_3$ is parameterized by the radius $r_c$ of a circle centered in ${\bf x}_2$}
\end{figure}

It is obvious that the introduction of the 3rd kernel point makes it impossible to solve (\ref{eq:feature:3protfast}) by the same convolution
approach as in (\ref{eq:feature:2protconvolve}). But the formulation of (\ref{eq:feature:3protfast}) leads us to an intuitive 
re-parameterization of the original problem. Without loss of generality, we consider the case where both kernel points ${\bf x}_2, {\bf x}_3$
are located on the same sphere, i.e. $r_2 = r_3$. Further we can fix ${\bf x}_2$ at the ``north pole'' ${\bf x}_N$ and reduce its parameterization 
to the radius $r_2$.\\
Since ${\bf x}_3$ is bound to ${\bf x}_2$ by the angle $\psi$, we can express the possible positions of ${\bf x}_3$ in terms of the points
on the circle which lies on the same sphere as ${\bf x}_2$ and is centered in ${\bf x}_2$. As figure \ref{fig:feature:3pparam} shows, this way we can
reduce the parameterization of ${\bf x}_3$ to the radius $r_c$ of this circle (Note: if we assume $r_2 \neq r_3$, the circle simply lies on a 
sphere with radius $r_3$). 

Given this re-parameterization, we can give a fast algorithm for the evaluation of (\ref{eq:feature:3protfast}): the integral over $\psi$
can be expressed as a convolution of a circular template on a sphere (analogous to (\ref{eq:feature:2protconvolve})) in spherical coordinates 
(we denote this operation by $*$):
\begin{eqnarray}
\label{eq:feature:3protconvolve}
T[r,r_c]({\bf x}) &=& \kappa_1\left(X({\bf x})\right)\cdot \int\limits_{S^2} \left(\kappa_2\left(X(s_{\cal R(\phi,\theta)}({\bf x}_2 ))\right)
\right)
\cdot\\
&&\left(\kappa_3\left(X\big|_{S[r]({\bf x})}\right)* C_t[r_c] \right)\sin{\theta} d\phi d\theta.\nonumber
\end{eqnarray}
The key step towards a fast algorithm is to transfer the evaluation of (\ref{eq:feature:3protconvolve}) to the Spherical Harmonic domain: 
we expand the kernelized spherical neighborhoods 
$$\widehat{\bf x}_2 := {\cal SH}[r]\left(\kappa_2\left(X\big|_{S[r]({\bf x})}\right)\right), \text{\quad}
\widehat{\bf x}_3:={\cal SH}[r]\left(\kappa_3\left(X\big|_{S[r]({\bf x})}\right)\right)$$ 
and the circle template 
$\widehat{C_t}:={\cal SH}[r]\big(C_t[r_c]\big)$ into the harmonic domain. Hence, we can apply the methods for fast convolution 
(see section \ref{sec:feature:shconvolve}), or ``left-convolution'' (see section \ref{sec:feature:shleftconvolve})
in case of the convolution with the circle template, in order to evaluate (\ref{eq:feature:3protconvolve}).\\
Using these the techniques and exploiting the orthonormal dependencies of the harmonic base functions, we can directly derive a fast algorithm 
for the computation of the $3p$ integral \cite{dagm1}: 
\begin{eqnarray}
\label{eq:feature:3p_final}
T[r,r_c]({\bf x}) &=& \kappa_1\big(X({\bf x})\big)\cdot \sum\limits_{l=0}^{\infty}\sum\limits_{m=-l}^l
\big(\widehat{\bf x_2}\big)^l_m \cdot \big(\widehat{\bf x_3} * \widehat{C_t}\big)^l_m.
\end{eqnarray}

\subsection{Implementation}
The transformation into the harmonic domain is implemented as described in section \ref{sec:feature:SHimplement}. Hence, 
we can also obtain the expansions at all points in $X$ at once using the convolution approach (\ref{eq:feature:discreteSHfull}).\\
The implementation of the circular template $C_t[r_c]$ has to be handled with some care: to avoid sampling issues, we apply
the same implementation strategies as in the case of the Spherical Harmonic base functions (see section \ref{sec:feature:SHimplement}
for details).\\
Finally, we can even further simplify the computation of the ``left convolution'' (\ref{eq:feature:shleftconvolve}), 
\begin{equation}
\big(\widehat{{\bf x}_3} * \widehat{C_t}\big)^l_m = 2\pi\sqrt{\frac{4\pi}{2l+1}}(\widehat{{\bf x}_3})^l_m \widehat{C_t}^l_0.
\label{eq:feature:lconv3p}
\end{equation}
Since the $0$th order of the harmonic base functions $Y^l_0$ always has constant values for a fixed
latitude $\Theta$ (\ref{eq:feature:SHbase}), given by the Legendre Polynomials $P^l_0(\sin{\Theta})$ (\ref{eq:feature:legendere}),  
and $C_t$ only holds ones on a fixed latitude, we can compute (\ref{eq:feature:lconv3p}) by a simple multiplication with a scalar value.\\

Figure \ref{fig:feature:3pimplement} gives a schematic overview of the implementation of $3p$-Features:

\begin{figure}[ht]
\centering
\psfrag{C1}{\tiny$X[c_1]]$}
\psfrag{C2}{\tiny$X[c_2]$}
\psfrag{C3}{\tiny$X[c_3]$}
\psfrag{c1}{\tiny$\widehat{X[c_1]}$}
\psfrag{c2}{\tiny$\widehat{X[c_2]}$}
\psfrag{c3}{\tiny$\widehat{X[c_3]}$}
\psfrag{f1}{$\kappa_1$}
\psfrag{f2}{$\kappa_2$}
\psfrag{f3}{$\kappa_3$}
\psfrag{SH(r3)}{\tiny${\cal SH}[{r_3}]$}
\psfrag{SH(r2)}{\tiny${\cal SH}[{r_2}]$}
\psfrag{SH}{\tiny${\cal SH}$}
\psfrag{rtd}{\tiny$\mathbb{R}^n$}
\psfrag{t}{\tiny$\widehat{C_t}$}
\includegraphics[width=0.8\textwidth]{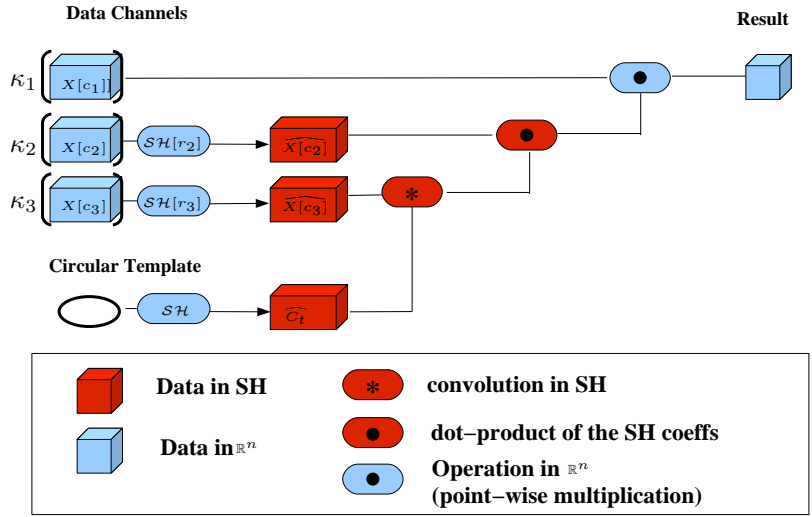}
\caption[$3p$-Feature: implementation.]{\label{fig:feature:3pimplement} Schematic overview of the implementation of $3p$-Features.  }
\end{figure}

\paragraph{Multi-Channel Data:}
Naturally, the application of $3p$-Features to multi-channel data is limited to three channels per feature
but straightforward: we can simply set the kernel points to be on different data channels as shown in the $2p$ case.
\paragraph{Complexity:}
Given input data $X$ with $m$ voxels, we need to compute the Spherical Harmonic transformation three times, to obtain $\widehat{{\bf x}_2}$
,$\widehat{{\bf x}_3}$ and $\widehat{C_t}$. Depending on the maximum expansion band $b_{\max}$, this lies in $O(b_{\max} \cdot m \log m)$ 
(see section \ref{sec:feature:SHimplement}). The convolution with the circular template and the dot-product take another $O(m\cdot b_{\max}^2)$,
followed by the voxel-wise multiplication with $\kappa_1(X)$ in $O(m)$.

\paragraph{Parallelization:}
As stated in section \ref{sec:feature:SHimplement}, we can gain linear speed-up in the number of cores for the parallelization of the 
harmonic transformation. Further, we could also split the computation of the convolution and the dot-product into several threads, but 
in practice this speed-up hardly falls into account.

\subsection{Discussion}
The $3$-Point Haar-Features solve the permutation invariance problem of the $2$-Point Features. However, this comes at the price of increased 
computational complexity, where the transformation to the harmonic domain makes up most of the additional cost.\\
Another issue is the growing parameter set: for $3p$ kernels we have to set $\kappa_1, \kappa_2, \kappa_3, r$ and $r_c$. which rises the 
question of an appropriate feature selection. 
%We tackle this problem in depth in section \ref{sec:feature:featureselect}, but it is evident that 
%such a selection will become more difficult with a growing number of parameters.

\newpage
%-------------------------------------
% np
%-------------------------------------
\section{\label{sec:feature:np}$n$-Point Haar-Features ($np$)}
\index{Haar-Feature}\index{Scalar Haar-Feature}\index{$np$-Feature}
In this section, we introduce a generic algorithm for the implementation of the general scheme for separable kernels 
(\ref{eq:feature:haar_separable_kernel}) which can handle an arbitrary number of kernel points ${\bf x}_1,\dots, x_n$. 
Just as we obtain an increase in discrimination power by going from two to three kernel points (see section \ref{sec:feature:3p}),
we motivate the strategy to add further points to the kernel by the goal of deriving even more selective features.\\
The actual number of needed kernel points depends on the application: i.e. for a single feature, the use of four points might deliver more 
discriminative texture
features than $3p$ kernels, while one might use kernels with eight or more points to locate very specific structures in an object detection task
(see part III).\\   

As in (\ref{eq:feature:2p_formular}) and (\ref{eq:feature:3p_formular}), we formalize the $n$-Point kernels as given by (\ref{eq:feature:haar_separable_kernel}): 
\begin{eqnarray}
\kappa := 
    \kappa_1\left({X}(s_g({\bf x}_1))\right) \cdot
    \kappa_2\left({X}(s_g({\bf x}_2))\right) \cdot
    \dots \cdot
    \kappa_n\left({X}(s_g({\bf x}_n))\right).
\label{eq:feature:haarnp_formular}
\end{eqnarray}
\subsection{\label{sec:feature:np_design}Feature Design}
As in the case of local 2- and 3-Point features, the primary goal is to achieve rotation invariance. 
Hence, the transformation group ${\cal G}$ is
given by the group of 3D rotations ${\cal SO}(3)$. If we parameterize these global rotations ${\cal R}\in {\cal SO}(3)$ as local rotations 
of the kernel points in Euler angles $s_g(\phi,\theta,\psi)$ (see Fig. \ref{fig:feature:eulerschema}), we can rewrite 
(\ref{eq:feature:haarnp_formular}) as:
\begin{eqnarray}
T[\Lambda](X) := \int\limits_{{\cal SO}(3)}
    \kappa_1\left({ s_g}_{(\phi,\theta,\psi)}X({\bf x}_1)\right) \cdot
    \kappa_2\left({ s_g}_{(\phi,\theta,\psi)}X({\bf x}_2)\right) \cdot
    \dots \nonumber\\
    \cdot
    \kappa_n\left({ s_g}_{(\phi,\theta,\psi)}X({\bf x}_n)\right)
    \sin{\theta}d\phi d\theta d\psi.
\label{eq:np_paramterized}
\end{eqnarray}
where $\Lambda$ is the set of parameters, i.e. including $\kappa_1,\dots,\kappa_n$ - we define $\Lambda$ in detail when we present
the parameterization of the kernel in the next section (\ref{sec:feature:np_parameterization}).\\

It is obvious that a direct and naive computation of these $n$-Point features is hardly tractable in terms of computational costs. For the
computation of every single (voxel-wise) feature, we would have to evaluate the kernel at all possible combinations of $\phi,\theta,\psi$
while transforming the $n$ kernel points respectively.\\

To cope with this massive computational complexity, we generalize the methods for the fast computation of 3D 2- and 3-Point features
\cite{dagm1} via fast convolution in the harmonic domain. The main challenge for this generalization is that we need to be able to couple
the $n$ sparse kernel points during the rotation in order to meet the separability criteria (\ref{eq:feature:haar_separable_kernel}) 
in (\ref{eq:np_paramterized}).\\
Previously, we were able to avoid the coupling problem:  
in the case of ``2-point'' kernels no coupling is needed, and the 3-Point kernels take advantage of 
the exception that the third point always lies on a circle centered in the second point (see section \ref{sec:feature:3p_design}).\\
For the general $np$ case, we need to derive a new approach which actually solves the coupling problem.\\

As in the previous sections,
we will first derive the theory in a continuous setting before we deal with the implementation issues for actual applications in a discrete
world (see section \ref{sec:feature:np_implement}).

\subsubsection{\label{sec:feature:np_parameterization}Parameterization}
\begin{figure}[ht]
\centering
\psfrag{x0}{\footnotesize${\bf x}_0$}
\psfrag{x1}{\footnotesize${\bf x}_1$}
\psfrag{x2}{\footnotesize${\bf x}_2$}
\psfrag{x3}{\footnotesize${\bf x}_3$}
\psfrag{x4}{\footnotesize${\bf x}_4$}
\includegraphics[width=0.2\textwidth]{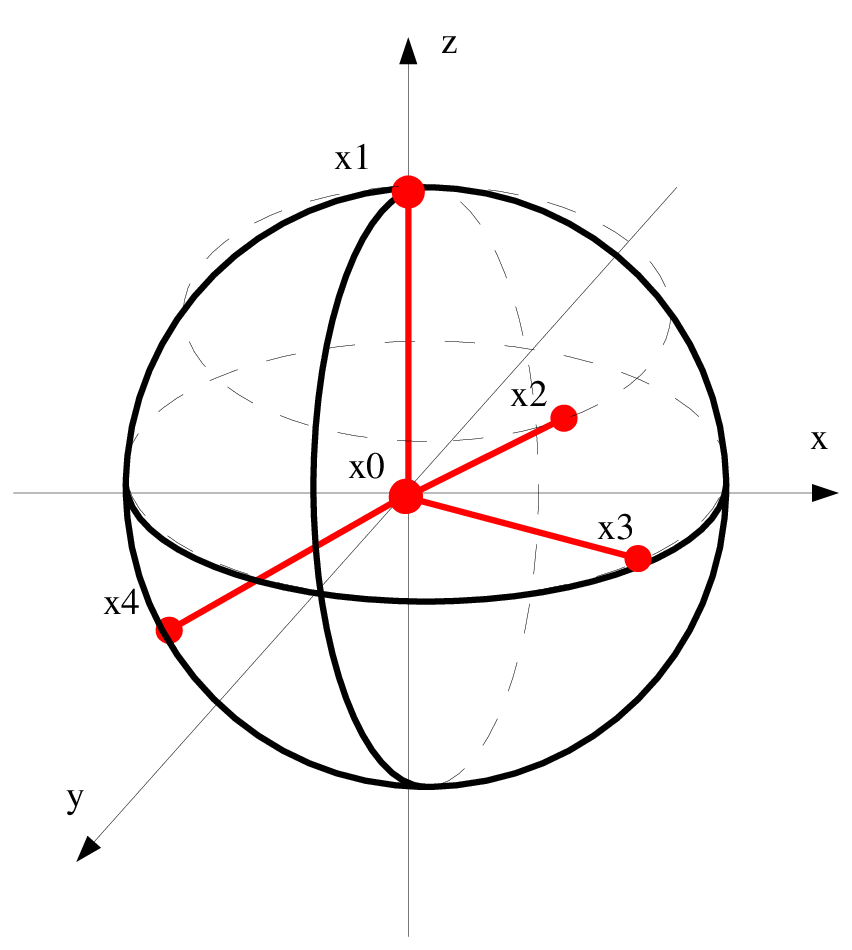}
\psfrag{phi}{\footnotesize$\Phi$}
\psfrag{theta}{\footnotesize$\Theta$}
\psfrag{radius}{\footnotesize$r$}
\psfrag{xi}{\footnotesize${\bf x}_i$}
\includegraphics[width=0.2\textwidth]{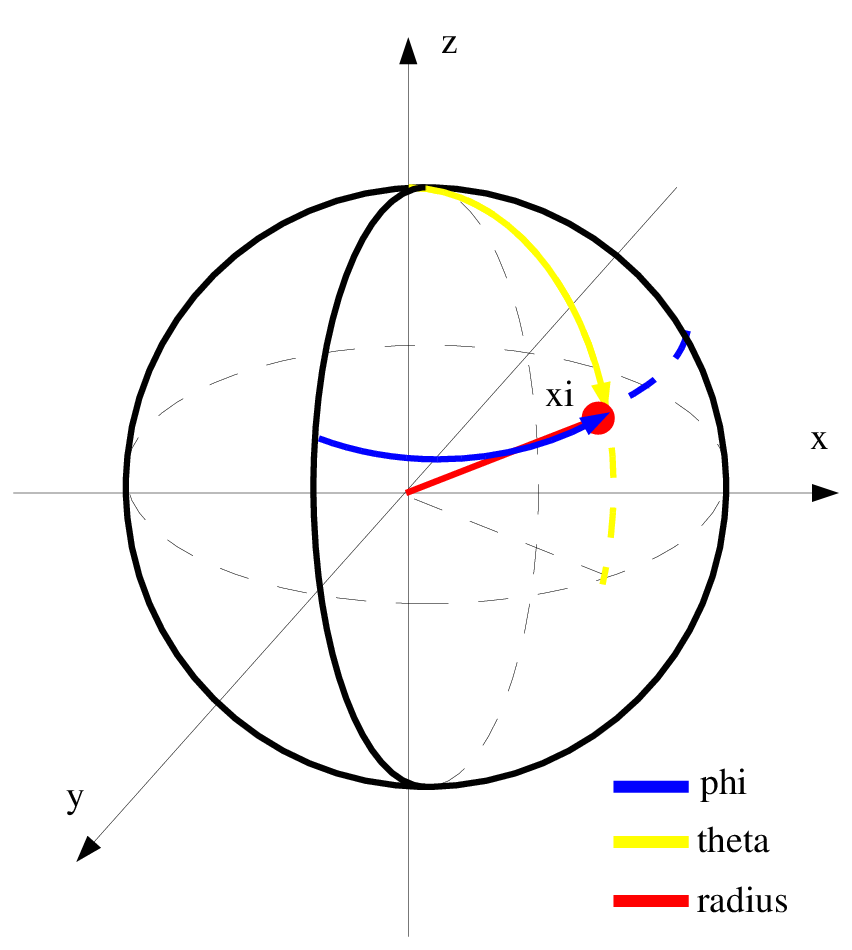}
\caption[Parameterization of $n$-Point kernels.]{\label{fig:feature:npparameter} Parameterization of $n$-Point kernels. {\bf Left}: The $n$ kernel points are parameterized 
as points on a sphere with exception of  ${\bf x}_1 = {\bf x}$ which is by definition placed in the origin of the local kernel support
${\bf x}$.
Note: the points must not necessarily lie on the same sphere as indicated in this scheme.
{\bf Right}: Each kernel point ${\bf x}_i\in S[r_i]\left({\bf x}\right)$ is parameterized by it's spherical coordinates in $
\Phi_i,\Theta_i$ and distance to the origin (= radius of the sphere) $r_i$.}
\end{figure}

As in the $2p$ and $3p$ case, we fix the first kernel point ${\bf x}_1 := X({\bf x})$ at the point of the local feature extraction,
while the other points ${\bf x_i}, i\in\{2,\dots,n\}$ are placed at concentric spherical neighborhoods with radii $r_i$:  
${\bf x}_i  \in S[r_i]\left({\bf x}\right)$. Hence, each ${\bf x}_i $ is parameterized by the spherical angles 
$\Phi_i \in [0,\dots, 2\pi], \Theta_i \in [0,\dots, \pi]$, the input data channel $c_i$ and the radius $r_i \in \mathbb{R}$ 
(also see figure \ref{fig:feature:npparameter}).

Overall, we end up with set of kernel parameters:
\begin{equation}
\Lambda := \big\{\kappa_1, \{\kappa_2,r_2,c_2,\Phi_2,\Theta_2\},\dots,
\{\kappa_n,r_n,c_n,\Phi_n,\Theta_n\}\big\}.
\label{eq:feature:np_lambda}
\end{equation}

Given this spherical parameterization, we first treat each ${\bf x}_i$ independently and perform the angular coupling of all points later on. 
We represent the ${\bf x}_i$ by a spherical delta-function ${\cal T}_i[r_i] \in S^2$  with radius $r_i$:
\begin{equation}
{\cal T}_i[r_i](\Phi,\Theta) := \delta(\Phi-\Phi_i)\delta(\Theta-\Theta_i).
\label{point_template}
\end{equation}
In its harmonic representation, ${\cal T}_i[r_i]$ is given by the according Spherical Harmonic base functions:
\begin{equation}
{\cal T}_i[r_i](\Phi,\Theta) = \sum\limits_{l=0}^\infty\sum\limits_{m=-l}^l \overline{Y_m^l}(\Phi,\Theta)Y_m^l(\Phi_i,\Theta_i).
\end{equation}
Hence, we can obtain the Spherical Harmonic transformation of  ${\cal T}_i[r_i]$ directly from the harmonic base functions:
\begin{equation}
\big(\widehat{\cal T}_i[r_i, \Phi_i,\Theta_i]\big)_m^l = Y_m^l(\Phi_i,\Theta_i).
\end{equation}

In the next step, we evaluate the contribution of the kernel points at the constellation of the ${\cal T}_i[r_i]$ 
given the local support of each feature extraction point ${\bf x}$.\\
\begin{figure}[th]
\centering
\psfrag{f1}{$\kappa_2$}
\psfrag{f2}{$\kappa_3$}
\psfrag{f3}{$\kappa_4$}
\psfrag{fn}{$\kappa_n$}
\psfrag{f0}{$\kappa_1$}
\psfrag{f4}{$\kappa()$}
\psfrag{SH(r1)}{\tiny${\cal SH}_{r_1}$}
\psfrag{SH(r2)}{\tiny${\cal SH}_{r_2}$}
\psfrag{SH(r3)}{\tiny${\cal SH}_{r_3}$}
\psfrag{SH(rn)}{\tiny${\cal SH}_{r_n}$}
\psfrag{SH}{\tiny${\cal SH}$}
\psfrag{C1}{\tiny$X_2$}
\psfrag{C2}{\tiny$X_3$}
\psfrag{C3}{\tiny$X_4$}
\psfrag{Cn}{\tiny$X_n$}
\psfrag{C0}{\tiny$X_1$}
\psfrag{Ct}{\tiny$X$}
\psfrag{c1}{\tiny$\widehat{X}_2$}
\psfrag{c2}{\tiny$\widehat{X}_3$}
\psfrag{c3}{\tiny$\widehat{X}_4$}
\psfrag{cn}{\tiny$\widehat{X}_n$}
\psfrag{t1}{\tiny$\widehat{\cal T}_2$}
\psfrag{t2}{\tiny$\widehat{\cal T}_3$}
\psfrag{t3}{\tiny$\widehat{\cal T}_4$}
\psfrag{tn}{\tiny$\widehat{\cal T}_n$}
\psfrag{FFT}{\tiny$\ \ {\cal F}^{-1}$}
\psfrag{rtd}{\tiny$\mathbb{R}^3$}
\psfrag{mi}{$\cdot$ \ $ \int$}
\includegraphics[width=0.75\textwidth]{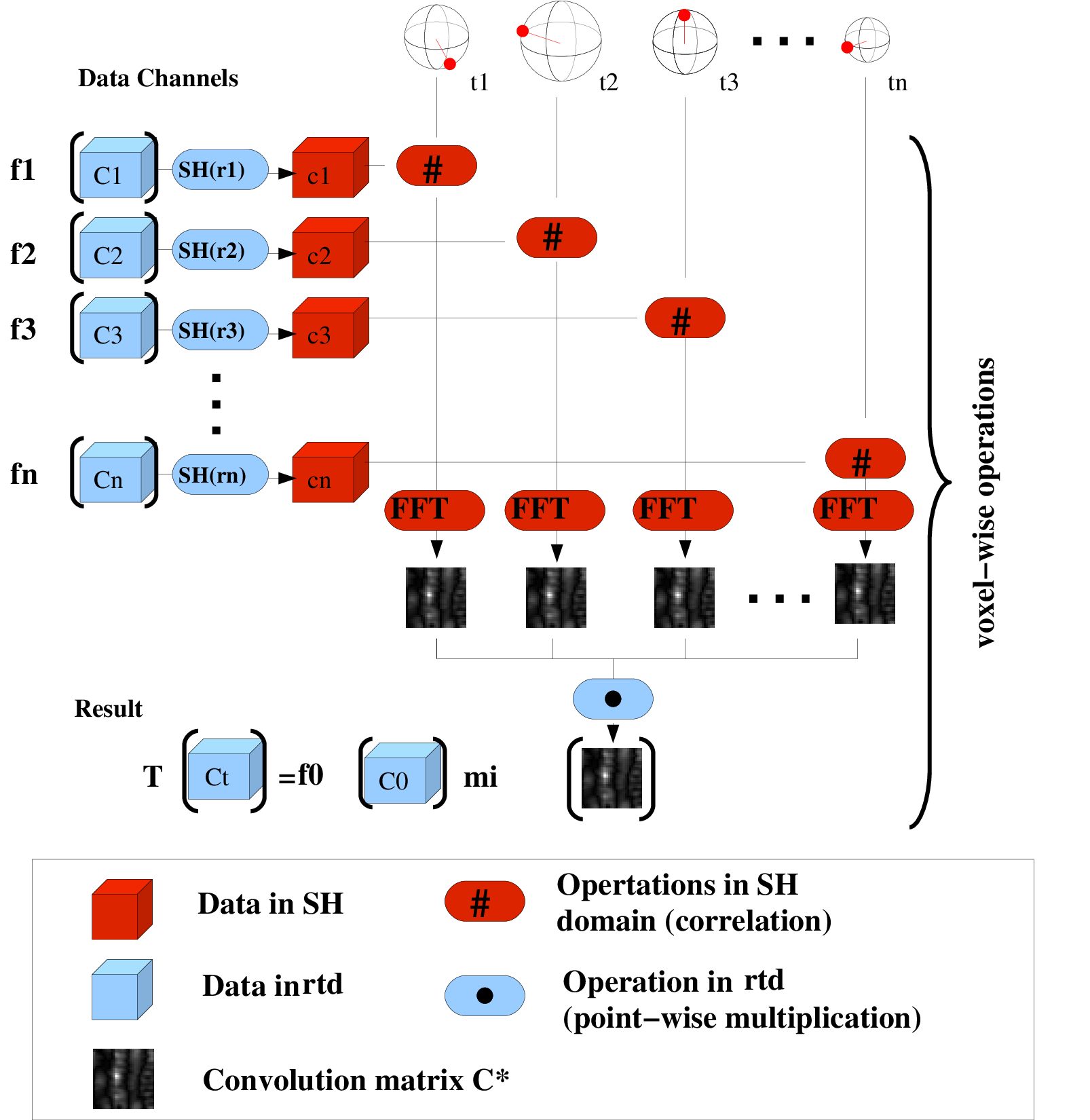}
\caption{\label{fig:feature:npscheme} Schematic overview of the fast computation of $n$p-Features. }
\end{figure}

Due to the separability of our kernels (\ref{eq:feature:haar_separable_kernel}), each kernel point is associated with a potentially different
 non-linear sub-kernel $\kappa_i$ and might operate on a different data channel $c_i$. For each feature evaluation,
we perform Spherical Harmonic expansions around the center voxel at the radii $r_i$ (associated with the respective kernel points)
of the non-linearly transformed input data $\kappa_i(X[c_i])$:
\begin{equation}
\widehat{ X[r_i,c_i]}({\bf x}) = {\cal SH}[r_i]\big(\kappa_i(X[c_i]\big|_{{\cal S}[r_i]({\bf x})})\big).
\end{equation}
With the data and the kernel points represented in the harmonic domain, we can now apply a fast correlation to evaluate the contribution
of each kernel point on the local data and perform this evaluation over all rotations.
Given a point at position ${\bf x}$, we compute the result $C^{\#}_i$ of this fast correlation over all spherical angles for the i-th kernel
point as shown in (\ref{eq:feature:SHcorrFinal}):
\begin{equation}
C^{\#}_i = \widehat{X[r_i,c_i]}({\bf x}) \# \widehat{\cal T}_i. 
\label{eq:feature:np_kernel_shtrans}
\end{equation}

\subsubsection{\label{sec:feature:np_rotinvariance}Rotation Invariance}
The key issue regarding the construction of $n$-Point" kernels is that we need to couple the contributions of the individual kernel 
points in such 
a way that only the chosen kernel constellation (given by the $\Phi_i,\Theta_i, r_i$) has a contribution when we rotate over all possible 
angles, i.e. the kernel points must not rotate independently.\\
Since the correlation matrices $C^{\#}_i$ hold the contribution at each possible angle in a 3D Euclidean space with a ($\phi,\theta,\psi$) 
coordinate-system
(see section \ref{sec:feature:shcorr}), we can perform the multiplicative
coupling of the separate sub-kernels (\ref{eq:feature:haar_separable_kernel}) by an angle-wise multiplication of the point-wise
correlation results: $\prod_{i=2}^n C^{\#}_i$.\\

Finally, by integrating over the resulting Euclidean space of this coupling, we easily obtain rotation invariance as in (\ref{eq:np_paramterized}):
\begin{eqnarray}
\label{eq:feature:np_almost_final}
\int\limits_{{\cal SO}(3)} \left( \prod\limits_{i=2}^n  {\cal C}^{\#}_i \right) \sin{\theta}d\phi d\theta d\psi.
\end{eqnarray}
With the additional coupling of $x_1$, we are now able to compute the $n$-Point Haar-Feature as shown in figure (\ref{fig:feature:npscheme}):
\begin{eqnarray}
\label{eq:feature:np_final}
T[\Lambda]({\bf x}) &:=& \kappa_1\big(X({\bf x})\big) \cdot \int\limits_{{\cal SO}(3)} \left( \prod\limits_{i=2}^n  {\cal C}^{\#}_i \right) \sin{\theta} d\phi d\theta d\psi.
\end{eqnarray}

\subsubsection{\label{sec:feature:np_grayinvariance}Gray-Scale Invariance}
A nice side effect of the kernel point coupling via fast correlation (\ref{eq:feature:np_final}) is the fact that we can obtain
real invariance towards additive and multiplicative gray-value changes: we simply use the normalized cross-correlation (\ref{eq:feature:SHcorrnormFinal}) to compute the 
$$C^{\#}_i = \widehat{X[r_i,c_i]}({\bf x}) \# \widehat{\cal T}_i$$
where the individually normalized correlations are independent of gray-scale changes.  

\subsection{\label{sec:feature:np_implement}Implementation}
The transformation into the harmonic domain is implemented as described in section \ref{sec:feature:SHimplement}. Hence,
we can also obtain the expansions at all points in $X$ at once using the convolution approach (\ref{eq:feature:discreteSHfull}).\\
The implementation of the template ${\cal T}_t$ has to be handled with some care: to avoid sampling issues, we apply
the same implementation strategies as in the case of the Spherical Harmonic base functions (see section \ref{sec:feature:SHimplement}
for details).\\
The computation of the correlation matrices ${\cal C}^{\#}$ follows the algorithm given in section \ref{sec:feature:ncrosscorr}. The 
size of the padding $p$ we need to apply strongly depends on the angular resolution necessary to resolve the given configuration of the kernel 
points.\\
Finally, the evaluation of the Haar-Integration over all possible rotations is approximated by the sum over the combined 
$(\phi,\theta,\psi)$-space: 
\begin{eqnarray}
\label{np_final_discrete}
T[\Lambda]({\bf x}) &\approx& \kappa_1\big(X({\bf x})\big) \cdot \sum \left( \prod\limits_{i=2}^n  {\cal C}^{\#}_i \right).
\end{eqnarray}

\paragraph{Multi-Channel Data:}
As in the other cases of scalar Haar-Features, the application of $np$-Features to multi-channel data is limited to $n$ channels per feature,
but straightforward: we can simply set the kernel points to be on different data channels as shown in the $2p$ case.

\paragraph{\label{sec:feature:np_complexity}Complexity}
The computational complexity of the $np$-Feature is dominated by the $n$ Spherical Harmonic expansions needed to transform the kernelized
input data into the harmonic domain which takes $O(n \cdot b_{\max} \cdot m \log m)$ for input data of size $m$. The costs for the
correlation and multiplication of the correlation matrices are negligible.

\paragraph{\label{sec:feature:np_parallelize}Parallelization}
As stated in section \ref{sec:feature:SHimplement}, we can gain linear speed-up in the number of cores for the parallelization of the
harmonic transformation. Further, we could also split the computation of the correlation matrices into several threads, but
as mentioned before, this speed-up hardly falls into account.

\subsection{\label{sec:feature:np_speedup}Further Speed-up}
Concerning computational complexity, the main bottleneck of the $np$-Feature is actually the transformation to the Spherical Harmonic domain.
Due to the non-linear mappings $\kappa_i$ of the separable kernel, we have to compute the harmonic expansion at all points ${\bf x}$ in $X$ for
each kernel point independently (\ref{eq:feature:np_kernel_shtrans}). Without the $\kappa_i$, we would only need a single transformation
for all kernel points which lie on the same radius and the same data channel (a setting which is very common in practice). However,  
we cannot simply neglect the non-linear kernel mappings.\\
\begin{figure}[ht]
\centering
\psfrag{f1}{$\kappa_2$}
\psfrag{f2}{$\kappa_3$}
\psfrag{f3}{$\kappa_4$}
\psfrag{fn}{$\kappa_n$}
\psfrag{f0}{$\kappa_1$}
\psfrag{f4}{$\kappa()$}
\psfrag{SH(r1)}{\tiny${\cal SH}_{r_2}$}
\psfrag{SH(r2)}{\tiny${\cal SH}_{r_3}$}
\psfrag{SH(r3)}{\tiny${\cal SH}_{r_4}$}
\psfrag{SH(rn)}{\tiny${\cal SH}_{r_n}$}
\psfrag{SH}{\tiny${\cal SH}$}
\psfrag{C1}{\tiny$X_2$}
\psfrag{C2}{\tiny$X_3$}
\psfrag{C3}{\tiny$X_4$}
\psfrag{Cn}{\tiny$X_n$}
\psfrag{C0}{\tiny$X_1$}
\psfrag{Ct}{\tiny$X$}
\psfrag{c1}{\tiny$\widehat{X}_1$}
\psfrag{c2}{\tiny$\widehat{X}_2$}
\psfrag{c3}{\tiny$\widehat{X}_3$}
\psfrag{cn}{\tiny$\widehat{X}_n$}
\psfrag{t1}{\tiny$\widehat{\cal T}_2$}
\psfrag{t2}{\tiny$\widehat{\cal T}_3$}
\psfrag{t3}{\tiny$\widehat{\cal T}_4$}
\psfrag{tn}{\tiny$\widehat{\cal T}_n$}
\psfrag{FFT}{\tiny$\ \ {\cal F}^{-1}$}
\psfrag{rtd}{\tiny$\mathbb{R}^3$}
\psfrag{mi}{$\cdot$ \ $ \int$}
\includegraphics[width=0.75\textwidth]{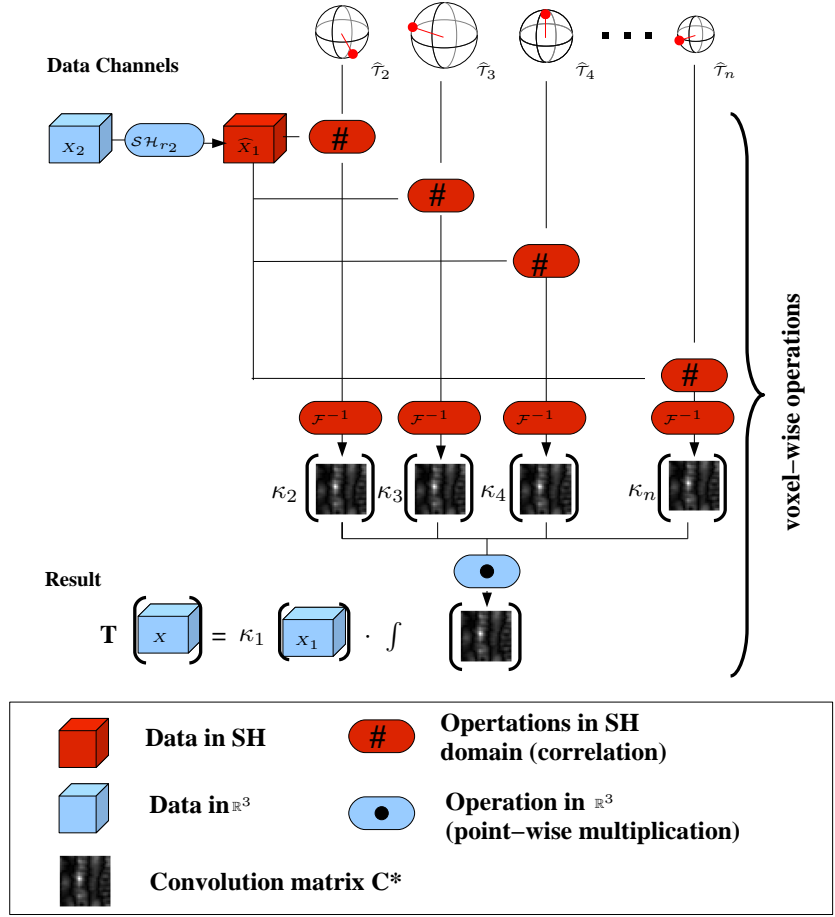}
\caption[Schematic overview of the optimized computation of $n$p-Features.]{\label{fig:feature:npscheme_fast} Schematic overview of the optimized computation of ``np''-features. }
\end{figure}
On the other hand, we are not bound to the class of separable kernels, which were only introduced to support the development of fast 
algorithms. Hence, we construct a new kernel, which is separating the kernel point ${\bf x}_1 = X({\bf x})$ in the center from the points 
${\bf x}_i \in S[r_i]({\bf x})$ in the local spherical neighborhood of ${\bf x}$: 

\begin{eqnarray}
\kappa :=
    \kappa_1\left({X}(s_g({\bf x}_1))\right) \cdot
    \kappa_s\left({X}(s_g({\bf x}_2)),
    \dots, 
    {X}(s_g({\bf x}_n))\right),
    \label{eq:feature:np_formularfast}
\end{eqnarray}
where $\kappa_s$ is some non-linear mapping of $(n-1)$ arguments (just like in (\ref{fig:feature:sparse_kernel_example})).\\

Instead of a non-linear weighting of the underlying data sensed by the kernel points (as before), we choose $\kappa_s$ to provide a 
non-linear weighting of the combination of the kernel points. Technically this is only a small change, but it enables us to
move the $\kappa_i$ into the harmonic domain, weighting the contribution of the kernel points to the Integral:  

\begin{eqnarray}
\label{eq:feature:np_fast}
T[\Lambda]({\bf x}) &:=& \kappa_1\big(X({\bf x})\big) \cdot \int\limits_{{\cal SO}(3)} \left( \prod\limits_{i=2}^n  \kappa_i\big(
{\cal C}^{\#}_i\big) \right) \sin{\theta} d\phi d\theta d\psi.
\end{eqnarray}

Figure \ref{fig:feature:npscheme_fast} shows the changes in the overall computation scheme. It should be noted that this optimized 
approach is similar but not equivalent to the original $np$ formulation.

\subsection{\label{sec:feature:np_discussion}Discussion}
The $np$-Features provide a powerful framework for the implementation of local features which are able obtain invariance towards  
rotations and multiplicative gray-scale changes via Haar-Integration.\\
In practice, $np$-Features are especially suitable for the design of highly specific features with a strong discriminative power used in 
challenging image analysis tasks justifying the higher computational costs. For less complex problems, we are better off using some
of the less complex feature methods.\\

A major problem concerning the application of $np$-Features is the huge set of kernel parameters $\Lambda$ (\ref{eq:feature:np_lambda}) 
we have to choose. In practice, it is infeasible to try all possible parameter combinations in a feature selection process, like we suggest 
for other features. Neither is it practically possible to select the best parameter settings by hand. 
%To overcome this problem, we introduce
%an automatic, data driven algorithm which learns the kernel parameters on given training samples (see section \ref{sec:feature:select_data}). 

%\newpage
%\section{\label{sec:feature:rkf}RKF Features }
%\index{Haar-Feature}\index{Scalar Haar-Feature}\index{RKF-Feature}

\cleardoublepage
\chapter{\label{sec:feature:VHfeature}${\cal VH}$-Features}
\index{${\cal VH}$-Features}
In this chapter, we derive a set of local, rotation invariant features which are directly motivated by the 
mathematical formulation of the Vectorial Harmonics (see section \ref{sec:feature:VH}). Analogous to the ${\cal SH}$-Features, we take 
advantage of the nice properties of the harmonic representation which allow us to perform fast feature computations in the
frequency domain.\\

Given 3D vector fields $\bf X$,  the transformation ${\cal VH}[r]({\bf x})$
(\ref{eq:feature:VHforward}) of local vectors on a sphere with radius $r$ around the
center ${\bf x}$ in  ${\bf X}$ in Vectorial Harmonics is nothing more than a change of the base-functions representing the initial data.
So the new base might provide us with a nice framework to operate on spheres, but we still have to perform the actual feature construction.
Primarily, we want to obtain rotation invariance.\\

First we introduce a method to obtain rotational invariance which is the simple extension of ${\cal SH}_{abs}$-Features 
(see section \ref{sec:feature:SHabs}) to vector fields: In section \ref{sec:feature:VHabs} we introduce ${\cal VH}_{abs}$-Features,
which use the fact that the band-wise energies of a ${\cal VH}$ representation do not change under rotation.\\

The second member of the ${\cal VH}$-Feature class is also derived from its ${\cal SH}$ counter part: the fast and also rotation invariant 
auto-correlation feature ${\cal VH}_{autocorr}$ (section \ref{sec:feature:VH_autocorr}) 
is based on the fast correlation in Vectorial Harmonics introduced in section \ref{sec:feature:vhcorr}.\\

Finally, since we transfer all $\cal VH$-Features directly from the class of $\cal SH$-Features, one might ask if the other two 
$\cal SH$-Features, ${\cal SH}_{phase}$ and ${\cal SH}_{bispectrum}$ could also be extended to the $\cal VH$ domain. And in fact, 
theoretically both extension could be done without much effort, but practically, none of them make much sense: the bispectrum features
(see section \ref{sec:feature:SHbispectrum}) simply become exceptionally expensive when we have to add additional couplings over the 
sub-bands $k$. For the vectorial phase, we could simply somehow define a phase in $\cal VH$, however, it is actually not evident how such a 
phase should be chosen and what it actually represents with respect to the mapping of the input data.

\newpage
%-----------------------------------------------------
% VHabs
%-----------------------------------------------------
\section{\label{sec:feature:VHabs}${\cal VH}_{abs}$}
\index{${\cal VH}$-Features}\index{${\cal VH}_{abs}$}
${\cal VH}_{abs}$-Features are the direct extension of ${\cal SH}_{abs}$-Features (see section \ref{sec:feature:SHabs}) to vector fields.
Again, we use the fact that the band-wise energies of a ${\cal VH}$ representation does not change under rotation.

\subsection{Feature Design}
Rotations ${\cal R(\phi,\theta, \psi)} \in i{\cal SO}(3)$ on 3D vector fields $\mathbb{R}^3 \times \mathbb{R}^3$ (see section \ref{sec:feature:vhrot})
are represented in the Vectorial Harmonic
domain in terms of band-wise multiplications of the expansions $\widehat{f^l}$ with Wigner D-Matrices $D^l$ (\ref{eq:feature:VHrot}).
Hence, we can directly follow the very same power spectrum approach as for the ${\cal SH}_{abs}$-Features.
This way we easily obtain a rotation invariant scalar entry for the $l$-th frequency in the power spectrum:

\begin{equation}
\label{eq:feature:VHabsvec}
\left({\cal VH}_{abs}[r]({\bf x})\right)^l :=  \sqrt{\sum\limits_{k=-1}^1\sum\limits_{m=-(l+k)}^{(l+k)} 
\left(\left({\cal VH}[r]({\bf x})\right)^l_{k,m}\right)^2}.
\end{equation}

Since the rotation invariance is achieved band wise, the approximation of the original data via harmonic expansion can be cut off at an
arbitrary band, encoding just the level of detail needed for the application.
\subsection{Implementation}
The implementation of the ${\cal VH}_{abs}$ is straightforward. We follow the implementation of the Vectorial Harmonic transformation as
described in section \ref{sec:feature:VHimplement}.

\paragraph{Multi-Channel Data:} ${\cal VH}_{abs}$-Feature cannot directly combine data from several channels into a single feature. In case of
multi-channel data, we would have to compute features for each channel separately.

\subsubsection{Complexity}
Following the implementation given in section \ref{sec:feature:VHimplement}, we obtain the harmonic expansion to band $b_{\max}$ at each
point of a volume with $m$ voxels in $O(m(b_{\max})^2 + (m \log m))$. The computation of the absolute values takes another $O((b_{\max})^3)$.\\
The additional loop over $k$ does not effect the $O$-Complexity, but in practice, ${\cal VH}_{abs}$ takes about factor three longer
to compute than ${\cal SH}_{abs}$.

\paragraph{Parallelization}
Further speed-up can be achieved by parallelization (see section \ref{sec:feature:implement}): the data can be transformed into
the harmonic domain by parallel computation of the coefficients and the computation of the absolute values can also be split into several
threads.
For ${\cal C}$ CPU cores with ${\cal C}\leq (b_{\max})^2$ and ${\cal C}\leq m$ we obtain:
$$O(\frac{m(b_{\max})^3}{\cal C}) +O( \frac{m(b_{\max})^2 +(m \log m)}{\cal C}).$$

\subsection{\label{sec:feature:vhabsDiss}Discussion }
The ${\cal VH}$-Features are a simple and straightforward extension of ${\cal SH}_{abs}$ to 3D vector fields. They are computationally
efficient and easy to implement. However, the discriminative properties are even more limited than the ${\cal SH}_{abs}$-Features. 
The band-wise absolute values capture only
the energy of the respective frequencies in the overall spectrum. Hence, we loose all the phase information which leads to strong ambiguities
within the feature mappings. The additional sub-bands $k$ further increase this problem compared to ${\cal SH}_{abs}$.
In many applications it is possible to reduce these ambiguities by combining ${\cal VH}$-Features
extracted at different radii.

\newpage
%-----------------------------------------------------
% VHautocorr
%-----------------------------------------------------
\section{\label{sec:feature:VH_autocorr}${\cal VH}_{autocorr}$}
\index{${\cal VH}$-Features}\index{${\cal VH}_{autocorr}$}
The second member of the ${\cal VH}$-Feature class is also derived from its ${\cal SH}$ counter part: based
on the auto-correlation feature ${\cal SH}_{autocorr}$ (section \ref{sec:feature:SHautocorr})
we compute invariant features directly from the Vectorial Harmonic representation. Again, this is motivated by the introduction of the
fast normalized cross-correlation in the Vectorial Harmonic domain (see introduction of chapter \ref{sec:feature:vhcorr}).
The cross-correlation ${\cal VH}_{corr}(f,g)$
of two vectorial signals ${\bf f,g} \in S^2 $ is a binary operation ${\cal VH}_{corr}: S^2\times S^2  
\rightarrow \mathbb{R}$. Hence, it cannot be used
directly as a feature, where we require a mapping of individual local signals ${\bf f}\in S^2 \rightarrow {\cal H}$ into some feature space
${\cal H} \subseteq \mathbb{R}^n$.\\
A general and widely known method for obtaining features from correlations is to compute the auto-correlation, e.g. \cite{autocorr}. In our case, we
propose the local ${\cal VH}_{autocorr}$-Feature, which performs a fast auto-correlation of ${\bf f} \in (S^2\times \mathbb{R}^3)$ with itself.\\

We use local dot-products of vectors to define the auto-correlation under a given rotation $\cal R$ in Euler angles $\phi, \theta, \psi$ as:
\begin{equation}
({\bf f} \# {\bf f})({\cal R}) := \int\limits_{\Phi,\Theta}  \langle{\bf f}(\Phi,\Theta),{\cal R} {\bf f}(\Phi,\Theta)\rangle \text{\quad} \sin{\Theta}d\Phi d\Theta.
\label{eq:feature:vhautocorr0}
\end{equation}

\subsection{Feature Design}
We first expand the local neighborhood $f$ at radius $r$ around the point ${\bf x} \in X$ in
Vectorial Harmonics, $\widehat{\bf f}:={\cal VH}[r]({\bf X}({\bf x}))$.\\
Then we follow the fast correlation method which we introduced in section \ref{sec:feature:vhcorr} to obtain the full correlation
$C^{\#}$ from equation (\ref{eq:feature:VHcorrFinal}).

\paragraph{Invariance:}
In order to obtain rotation invariant features, we follow the Haar-Integration approach (see section 
\ref{sec:feature:invariance_via_groupintegration}) and integrate
over the auto-correlations at all possible rotations $\cal R$. $C^{\#}$ holds the necessary auto-correlation results in a 3D
$(\phi,\theta,\psi)$-space (\ref{eq:feature:SHcorrFT}), hence we simply integrate over $C^{\#}$,
\begin{equation}
{\cal VH}_{autocorr} := \int\limits_{\phi,\theta\psi} \kappa \left(C^{\#}(\phi,\theta,\psi)\right) \sin{\theta}d\phi d\theta d\psi
\label{eq:feature:autocorrinvariantVH}
\end{equation}
and obtain a scalar feature. Additionally, we insert a non-linear kernel function $\kappa$ to increase the separability. Usually, very
simple non-linear functions, such as $\kappa(x):=x^2, \kappa(x):=x^3$ or $\kappa(x):= \sqrt{x}$, are sufficient.\\

\subsection{Implementation}
We follow the implementation of the Vectorial Harmonic transformation as described in section \ref{sec:feature:VHimplement} and the
implementation of the fast correlation from (\ref{eq:feature:SHcorrPad}).\\
In practice, where the harmonic expansion is bound by a maximal expansion band $b_{\max}$, the integral (\ref{eq:feature:autocorrinvariantVH})
is reduce to the sum over the then discrete angular space $C^{\#}$:
\begin{equation}
{\cal VH}_{autocorr} = \sum\limits_{\phi,\theta\psi} \kappa \left(C^{\#}(\phi,\theta,\psi)\right).
\label{eq:feature:autocorrinvariant_sumVH}
\end{equation}

\paragraph{Multi-Channel Data:} ${\cal VH}_{autocorr}$ cannot directly combine data from several channels into a single feature. In case of
multi-channel data, we would have to compute features for each channel separately.

\subsubsection{Complexity}
Following the implementation given in section \ref{sec:feature:VHimplement}, we obtain the harmonic expansion to band $b_{\max}$ at each
point of a volume with $m$ voxels in $O(m(b_{\max})^2 + (m \log m))$. The complexity of the auto-correlation depends on $b_{\max}$ and
the padding parameter $p$ (\ref{eq:feature:SHcorrPad}) and can be computed in $O(m(b_{\max}+p)^3 \log (b_{\max}+p)^3))$. The summ over
$C^{\#}$ takes another $O((b_{\max}+p)^3)$ at each point.

\paragraph{Parallelization:}
Further speed-up can be achieved by parallelization (see section \ref{sec:feature:implement}): the data can be transformed into
the harmonic domain by parallel computation of the coefficients and the computation of the absolute values can also be split into several
threads.
For ${\cal C}$ CPU cores with ${\cal C}\leq (b_{\max})^2$ and ${\cal C}\leq m$ we obtain:
$$O(\frac{m\left( (b_{\max}+p)^3 + (b_{\max}+p)^3 \log (b_{\max}+p)^3\right)}{\cal C}) +O( \frac{m(b_{\max})^2 +(m \log m)}{\cal C})$$

\subsection{Discussion}
Auto-correlation can be a very effective feature to encode texture properties. The discriminative power of ${\cal VH}_{autocorr}$ can
be further increased by combining the correlation at several different radii into a correlation result $C^{\#}$ as described in section \ref{sec:feature:shcorrradii}.

\cleardoublepage
\chapter{\label{sec:feature:VHHaar}Vectorial Haar-Features}
\index{Haar-Feature}\index{Vectorial Haar-Feature}
In this chapter we derive several features operating on vectorial data which obtain invariance via Haar-Integration. All of the methods
are strongly related to the features presented in the chapter \ref{sec:feature:SHHaar} and are based on the Haar-Integration framework 
\ref{eq:feature:general_haar_integral}.
In the case of vectorial data, we take advantage of the Vectorial Harmonic (see section \ref{sec:feature:VH}) representation of local 
spherical neighborhoods ${\cal S}[r]({\bf x})$ of radii $r$ at position ${\bf x} \in \mathbb{R}^3$ of the 3D vector fields ${\bf X}:  
\mathbb{R}^3\rightarrow \mathbb{R}^3$ with vectorial elements ${\bf X}({\bf x}) \in \mathbb{R}^3$.\\

Please refer to the sections \ref{sec:feature:invariance_via_groupintegration} and \ref{sec:feature:haarkernels} for an in-depth introduction
of the Haar approach. It also might be useful to take a look at the scalar kernels in \ref{sec:feature:2p}, \ref{sec:feature:3p} and 
\ref{sec:feature:np} first.\\

Analogical to the $2p,3p$ and $np$ kernels, where the name indicated the number of scalar kernel points in a local, sparse and separable 
kernel (\ref{eq:feature:haar_separable_kernel}), we also denote the local, sparse and separable vectorial kernels by $1v, 2v$ and $nv$:\\

The $1v$-Feature (section \ref{sec:feature:1v}) uses a kernel with a single vector component and acts as vectorial extension of 
the $2p$-Feature. Basically, it integrates
the local similarities of the data vectors with the normal vectors of a spherical neighborhood template. The $1v$ kernel is especially
suitable for the detection of sphere like convex structures and is primarily a shape feature, not a texture feature.\\

The $2v$-Feature (section \ref{sec:feature:2v}) applies a variation of the $1v$ kernel: instead of using the normal vectors of a spherical 
neighborhood template, 
the $2v$ kernel integrates over the similarities of the data vectors with the centering vector ${\bf X}({\bf x})$. $2v$ kernels return 
more texture based and less shape based features.\\

Finally, we introduce the $nv$-Feature (section \ref{sec:feature:nv}) where we apply the direct extension of the $np$ kernel 
(section \ref{sec:feature:np}) to 3D vector fields in order to derive highly specific local features. 

\paragraph{Related Work:}
In general, there have not been many publications on local invariant features for 3D vector fields. One exception is the work of 
\cite{schulz}, which uses a voting scheme in a 3D gradient vector field to detect spherical structures. The results of this
feature are practically identical to those of our $1v$-Features - both just follow different approaches to implement a detector which
could be considered as Hough-Transform for spheres.

\newpage
%---------------------------------------
% 1v
%---------------------------------------
\section{\label{sec:feature:1v}1-Vector Features ($1v$)}
\index{Haar-Feature}\index{Vectorial Haar-Feature}\index{$1v$-Feature}
The $1v$-Feature uses a kernel with a single vector component and acts as vectorial extension of the $2p$-Feature. Basically, it integrates
the local similarities of the data vectors with the normal vectors of a spherical neighborhood template.
\subsection{Feature Design}
\begin{figure}[ht]
\centering
\psfrag{v1}{${\bf x}_i^\bot$}
\psfrag{x0}{$\bf x$}
\psfrag{x1}{${\bf X}({\bf x}_i)$}
\psfrag{x2}{${\bf x}_i$}
\psfrag{r}{$r$}
\includegraphics[width=0.4\textwidth]{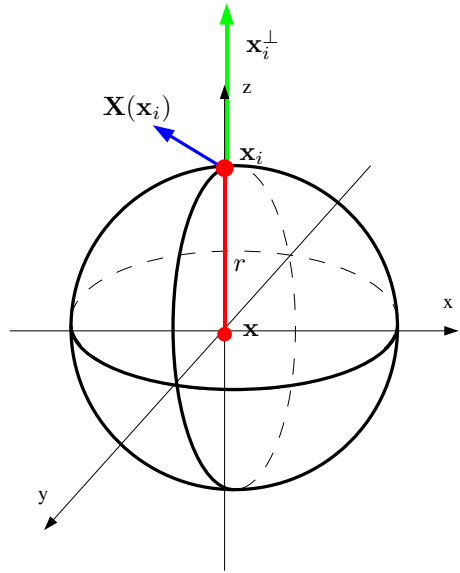}
\caption[$1v$-Feature scheme.]{\label{fig:feature:1vscheme} Example for the basic design of the $1v$ kernel with $\alpha = 1$.}
\end{figure}

Given a 3D vector field ${\bf X}:\mathbb{R}^3\rightarrow \mathbb{R}^3$, we extract local features from the spherical neighborhoods 
${\cal S}[r]({\bf x})$ at positions ${\bf x}$. We integrate over the dot-products between the vectorial data ${\bf X}({\bf x}_i)$ and
the normal vectors ${\bf x}_i^\bot$ at all positions $ {\bf x}_i \in {\cal S}[r]({\bf x})$ on the sphere around ${\bf x}$.
The normal vectors are defined as:
\begin{equation}
{\bf x}_i^\bot := \alpha\left({\bf x} -  {\bf x}_i \right),
\label{eq:feature:1vdirect}
\end{equation}
where the $\alpha \in \{-1,1\}$ factor determines whether the normal vector points towards or away from the feature extraction point.
Figure \ref{fig:feature:1vscheme} illustrates the basic kernel design

\subsubsection{Rotation Invariance}
If we plug the dot-product into the general Haar framework (\ref{eq:feature:general_haar_integral}), we can achieve invariance regarding 
rotations ${\cal R(\phi,\theta, \psi)} \in {\cal SO}(3)$ parameterized in Euler angles (see section \ref{sec:feature:shrot}).\\
It is obvious that all possible positions  ${\bf x}_i$ lie on the spherical neighborhood 
${\cal S}[r]({\bf x})$ with the radius 
\begin{equation}
r=\big|{\bf x} -  {\bf x}_i\big|, 
\end{equation}
whereas the normal vector ${\bf x}_i^\bot$ changes with the position according to (\ref{eq:feature:1vdirect}). Because we are considering
a singe kernel vector, we can reduce the integral over all rotations to an integral over all points of the spherical neighborhood
parameterized by the angles $\phi$ and $\theta$ (see \ref{eq:feature:2protfast2} for a detailed justification).
The final formulation of the $1v$-Feature is then:
\begin{equation}
T[r, \alpha]({\bf x}) := \int\limits_{ x_i \in {\cal S}[r]({\bf x})} \big\langle {\bf x}_i^\bot , {\bf X}({\bf x}_i) 
\big\rangle \sin\theta d\phi d\theta.
\label{eq:feature:1vformular}
\end{equation}
\subsection{\label{sec:feature:1vimplement}Implementation}
\begin{figure}[ht]
\centering
\includegraphics[width=0.4\textwidth]{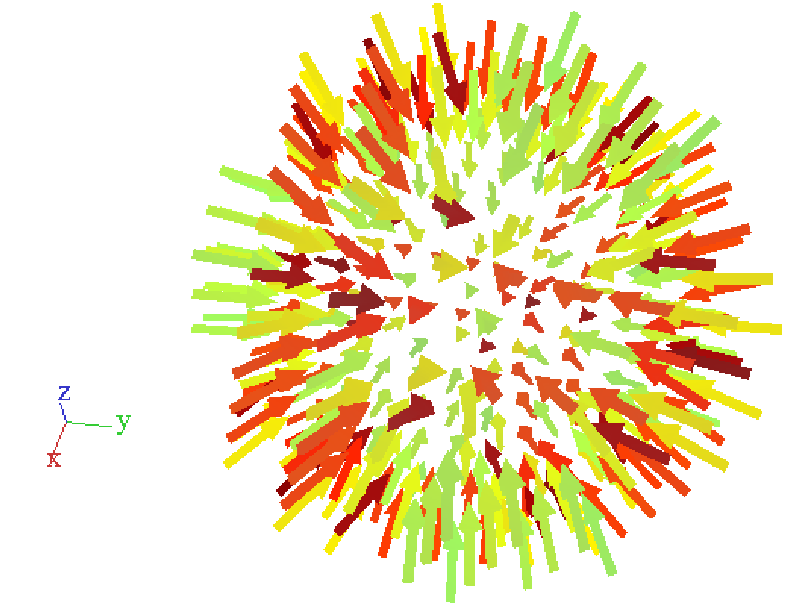}
\caption[$1v$ template.]{\label{fig:feature:1vtemplate} Sample ${\bf \cal T}[r,\alpha]$, with $\alpha = -1$.}
\end{figure}

The evaluation of (\ref{eq:feature:1vformular}) could be implemented straightforward. However, usually we want to compute features at
all voxels ${\bf x}$ simultaneously. Therefore, we propose an optimized algorithm: we pre-compute a vectorial template 
${\bf \cal T}[r,\alpha]$, which simply holds the normal vectors of the spherical neighborhood ${\cal S}[r]({\bf x})$ weighted by $\alpha$. 
Figure \ref{fig:feature:1vtemplate} shows such a template.
We then reformulate the dot-product in (\ref{eq:feature:1vformular}) as component-wise convolution of  
${\bf \cal T}[r,\alpha]$ with ${\cal S}[r]({\bf x})$:
\begin{equation}
T[r, \alpha]({\bf x}) := \sum\limits_{c=0}^2  {\bf X}[c]\big|_{{\cal S}[r]({\bf x})} * {\bf \cal T}[r,\alpha][c],
\label{eq:feature:1vfastformular}
\end{equation}
where ${\bf X}({\bf x})[c]$ returns the $c$th directional component of ${\bf X}({\bf x})$. Hence, we can apply a fast convolution to 
simultaneously evaluate (\ref{eq:feature:1vfastformular}) at all voxels ${\bf x}$:
\begin{equation}
T[r, \alpha]({\bf X}) := \sum\limits_{c=0}^2 FFT^{-1}\big( FFT({\bf X}[c]) \cdot FFT({\bf \cal T}[r,\alpha][c])\big).
\label{eq:feature:1vfastconvolve}
\end{equation}

For discrete input data we have to handle the implementation of the spherical template $T[r,\alpha]$ with some care. To avoid sampling issues,
we apply the same implementation strategies as in the case of the Spherical Harmonic base functions (see section \ref{sec:feature:SHimplement}
for details).

\paragraph{Multi-Channel Data:}
$1v$-Features cannot directly combine data from several channels into a single feature. In case of
multi-channel data, we would have to compute features for each channel separately.

\subsubsection{Complexity}
Using the convolution approach, we  end up with a complexity of $O(m \cdot m \log m)$ for an input volume of size $m$.

\paragraph{Parallelization:}
Since there is no easy way to parallelize the Fourier Transform, we do not further parallelize the computation
of $1v$-Features. But since $1v$-Features can be computed so fast anyway, this is not a real drawback.

\subsection{\label{sec:feature:1vDiss}Discussion }
The $1v$-Feature provides a very fast and rotation invariant method for the extraction of local features from 3D vector fields. 
The nature of the kernel vectors given as normals of the spherical neighborhood makes the $1v$ kernel an optimal detector for spherical
structures which relies mostly on shape and hardly on texture properties of the underlying data. The $\alpha$ factor then indicates if
we detect the inner or the outer surface of a spherical shape.\\
In an alternative interpretation, the $1v$ approach could be seen as Hough-Transform \cite{hough} for spheres. This could be reinforced by
an additional integration over several radii.

\newpage
%-----------------------------------------------
% 2v
%-----------------------------------------------
\section{\label{sec:feature:2v}2-Vector Features ($2v$)}
\index{Haar-Feature}\index{Vectorial Haar-Feature}\index{$2v$-Feature}
The $2v$-Feature uses a variation of the $1v$ kernel: instead of using the normal vectors of a spherical
neighborhood template, the $2v$ kernel integrates over the similarities of the data vectors of the centering vector ${\bf X}({\bf x})$.
\subsection{Feature Design}
Given a 3D vector field ${\bf X}:  \mathbb{R}^3\rightarrow \mathbb{R}^3$, we extract local features from the spherical neighborhoods
${\cal S}[r]({\bf x})$ at positions ${\bf x}$.\\ 
The basic idea of the $2v$ kernel is to compute the similarity (in terms of the dot-product) of the direction of the data vectors 
${\bf X}({\bf x}_i), \forall {\bf x}_i \in {\cal S}[r]({\bf x})$ with the direction of the center vector ${\bf X}({\bf x})$. 

\begin{figure}[ht]
\centering
\psfrag{v1}{${\bf X}({\bf x}_i)$}
\psfrag{x0}{$\bf X(x)$}
\psfrag{x2}{}
\psfrag{r}{$r$}
\includegraphics[width=0.4\textwidth]{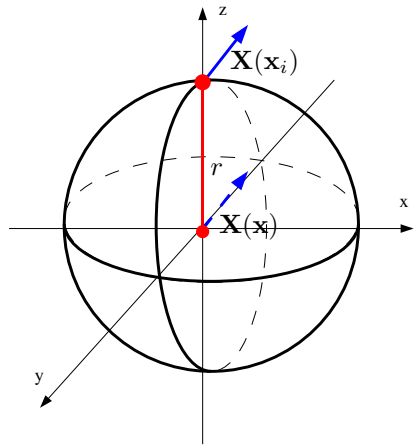}
\caption[$2v$-Feature scheme.]{\label{fig:feature:2vscheme} Basic design of the $2v$ kernel.}
\end{figure}
\paragraph{Rotation Invariance:}
If we plug the $2v$ kernel into the general Haar framework (\ref{eq:feature:general_haar_integral}), we can achieve invariance regarding
rotations ${\cal R(\phi,\theta, \psi)} \in {\cal SO}(3)$ which are parameterized in Euler angles (see section \ref{sec:feature:shrot}).\\
Just as in the $1v$ case, we use the fact that all possible positions of the ${\bf x}_i$ lie on the spherical neighborhood
${\cal S}[r]({\bf x})$ with the radius:
\begin{equation}
r=\big|{\bf x} -  {\bf x}_i\big|,
\end{equation}
And again, since we are considering
only a singe kernel vector, we can reduce the integral over all rotations to an integral over all points of the spherical neighborhood
parameterized by the angles $\phi$ and $\theta$ (see \ref{eq:feature:2protfast2} for a detailed justification).
The final formulation of the $2v$-Feature is then:
\begin{equation}
T[r]({\bf x}) := \int\limits_{ x_i \in {\cal S}[r]({\bf x})} \big\langle {\bf X}({\bf x}) , {\bf X}({\bf x}_i) \big\rangle \sin\theta d\phi d\theta.
\label{eq:feature:2vformular}
\end{equation}

\subsection{Implementation}
The implementation strictly follows the convolution based algorithm introduced for the $1v$ case 
(see section \ref{sec:feature:1vimplement}). The only difference
is that the vectors in the template $\cal T$ are oriented in the same direction as $\bf X(x)$.

\paragraph{Multi-Channel Data:}
$v2$-Features can combine data from two channels into a single feature: we can simply extract the kernel direction ${\bf X}[c_1]({\bf x})$
and the neighborhood data ${\bf X}[c_2]({\bf x}_i)$ from different channels.

\subsubsection{Complexity}
Using the convolution approach, we  end up with a complexity of $O(m \cdot m \log m)$ for an input volume of size $m$.

\paragraph{Parallelization:}
Since there is no easy way to parallelize the Fourier Transformation, we do not further parallelize the computation
of $2v$-Features. But as $2v$-Features can be computed so fast anyway, this is not a real drawback.

\subsection{\label{sec:feature:2vDiss}Discussion }
The fast $2v$ kernels return more texture based and less shape based features. Intuitively, $2v$-Features are an indicator for the 
local homogeneity of the vector field. The name $2v$-Feature might miss leading to some degree, since we only consider a single kernel vector.
But in contrast to the $1v$-Feature approach, we actually combine two vectors from the input data ${\bf X}({\bf x})$ and ${\bf X}({\bf x}_i)$.

\newpage
%-----------------------------------------------
% nv
%-----------------------------------------------
\section{\label{sec:feature:nv}n-Vector Features ($nv$)}
\index{Haar-Feature}\index{Vectorial Haar-Feature}\index{$nv$-Feature}
The $nv$-Features are the direct extension of the $np$-Features (see section \ref{sec:feature:np}) to 3D vector fields. Analogous to the 
properties of $np$ kernels on scalar (multi-channel) data, the goal is to derive highly specific local features for the detection of local
structures (objects) in 3D vector fields.\\
To obtain a strong discrimination power, we introduce a vectorial kernel which is able to handle an arbitrary number of kernel vectors 
${\bf v}_1,\dots, {\bf v}_n$ instead of only one or two (as for $1v,2v$-Features).\\
Since the entire derivation of the $nv$ kernel strongly relies on the very same methods and algorithms that were introduced for the derivation
 of the $np$ kernel, the reader may to refer to section \ref{sec:feature:np} for some technical details.\\

Given a 3D vector field ${\bf X}: \{\mathbb{R}^3\rightarrow \mathbb{R}^3\}$, we extract local features from the spherical neighborhoods
${\cal S}[r]({\bf x})$ at positions ${\bf x}$. For the kernel vectors ${\bf v}_i \in \{\mathbb{R}^3\times \mathbb{R}^3\}$,
we write $\dot{\bf v}_i \in \mathbb{R}^3$ and $\overrightarrow{{\bf v}_i} \in \mathbb{R}^3$ to address their position and direction.\\
 
\begin{figure}[ht]
\centering
\psfrag{v1}{$\overrightarrow{{\bf v}_1}$}
\psfrag{v2}{$\overrightarrow{{\bf v}_2}$}
\psfrag{v3}{$\overrightarrow{{\bf v}_3}$}
\psfrag{v4}{$\overrightarrow{{\bf v}_4}$}
\psfrag{vv1}{$\dot{\bf v}_1$}
\psfrag{vv2}{$\dot{\bf v}_2$}
\psfrag{vv3}{$\dot{\bf v}_3$}
\psfrag{vv4}{$\dot{\bf v}_4$}
\psfrag{x0}{${\bf X}(x)$}
\psfrag{x2}{}
\psfrag{r}{$r$}
\includegraphics[width=0.4\textwidth]{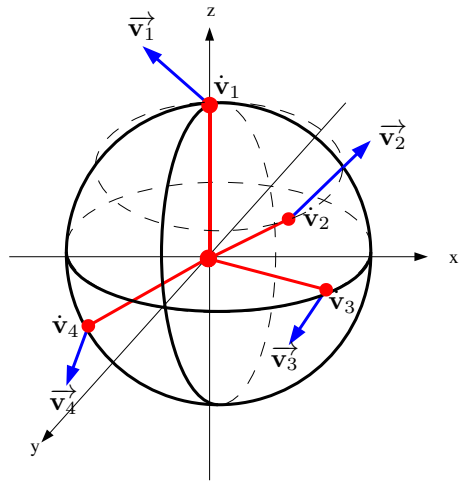}
\caption[Example ``n-vector'' kernel.]{\label{fig:feature:nvec_example} Example for the basic design of the $nv$ kernel with four kernel 
vectors. Note that the kernel vectors are not necessarily bound to the same radius as illustrated in this example.}
\end{figure}

There are two major differences in the basic formulation between general sparse and local scalar ($np$) and vectorial $(nv)$ kernels:
first, we do not explicitly consider a center vector for $nv$ kernels (even though the framework would
allow such a constellation). The main reason to do so is given by the second difference: since non-linear mappings (like the $\kappa_i$ 
in the $np$ kernel) are not well defined on vectorial data, we do not use a separable kernel approach (\ref{eq:feature:haar_separable_kernel}) 
for the construction of the $nv$ kernel.\\
Instead, we are following the alternative (fast) approach (\ref{eq:feature:np_formularfast}), which allows us to formalize the $n$-Vector 
kernels in a more abstract way:
(\ref{eq:feature:haar_separable_kernel}):
\begin{eqnarray}
    \kappa\left(s_g({\bf v}_1),
    \dots ,
    s_g({\bf v}_n)\right).
\label{eq:feature:haarvp_formular}
\end{eqnarray}
Figure \ref{fig:feature:nvec_example} shows an example $nv$ kernel. Later on, we give the actual kernel mapping $\kappa$, which is still 
non-linear, but does not operate directly on vectorial data.
\subsection{\label{sec:feature:vp_design} Feature Design}
The primary goal is to achieve rotation invariance.
Hence, the transformation group ${\cal G}$ is
given by the group of 3D rotations ${\cal SO}(3)$. If we parameterize these global rotations ${\cal R}\in {\cal SO}(3)$ as local rotations
of the kernel vectors in Euler angles $s_g(\phi,\theta,\psi)$ (see Fig. \ref{fig:feature:eulerschema}), we can rewrite
(\ref{eq:feature:haarvp_formular}) as:
\begin{eqnarray}
T[\Lambda]({\bf x}) &:=& \int\limits_{{\cal SO}(3)}
    \kappa\left({ s_g}_{(\phi,\theta,\psi)}({\bf v}_1),\dots, 
    { s_g}_{(\phi,\theta,\psi)}({\bf v}_n)\right)
    \sin{\theta}d\phi d\theta d\psi,
\label{eq:vp_paramterized}
\end{eqnarray}
where $\Lambda$ is the set of parameters, i.e. including $\kappa$ - we define $\Lambda$ in detail when we present
the parameterization of the kernel in the next sub-section(\ref{sec:feature:nv_parameterization}).\\

\subsubsection{\label{sec:feature:nv_parameterization}Parameterization}
The key for a fast computational evaluation of (\ref{eq:vp_paramterized}) is the smart parameterization of the kernel. Following the 
approach for the $np$ kernels, we parameterize the position of the kernel vectors as points $\dot{\bf v}_i$ with $i\in\{1,\dots,n\}$ located 
at concentric spherical neighborhoods ${\cal S}[r_i]\left({\bf x}\right)$ with radii $r_i$ surrounding the point of the feature extraction ${\bf x}$. 
Hence, each $\dot{\bf v}_i $ is parameterized by the spherical angles
$\Phi_i \in [0,\dots, 2\pi], \Theta_i \in [0,\dots, \pi]$ and $r_i \in \mathbb{R}$ (also see figure \ref{fig:feature:npparameter}).\\
The overall parameter set $\Lambda$ thus includes the parameterized position $\dot{\bf v}_i $, the direction $\overrightarrow{{\bf v}_i} $ 
(which is normalized to $|\overrightarrow{{\bf v}_i} |=1$) and the non-linear mapping $\kappa$ which will be split into $\kappa_1,\dots,\kappa_n$
later on:
\begin{equation}
\Lambda := \big\{\{\kappa_1,r_1,\Phi_1,\Theta_1,\overrightarrow{{\bf v}_1} \},\dots,
\{\kappa_n,r_n,\Phi_n,\Theta_n,\overrightarrow{{\bf v}_n} \}\big\}.
\label{eq:feature:vp_lambda}
\end{equation}

Given this parameterization, we further follow the approach from the $np$ derivation and introduce  ``delta like'' vectorial template 
functions ${\cal T}_i$ which represent the kernel vectors ${\bf \cal T}_i[r_i]$ in the harmonic domain:
\begin{equation}
\big(\widehat{\bf \cal T}_i[r_i, \Phi_i,\Theta_i, \overrightarrow{{\bf v}_i}]\big)_{km}^l = \overrightarrow{{\bf v}_i}^T {\bf Z}^l_{km}(\Phi,\Theta).
\end{equation}

Now we have a frequency representation of the individual kernel vectors. In the next step, we evaluate the contribution of the input data at 
these kernels.
For each feature evaluation,
we perform Vectorial Harmonic expansions around ${\bf x}$ at the radii $r_i$ (associated with the position of the respective kernel vectors)
 of the input vector field $X$:
 \begin{equation}
 \widehat{ {\cal S}[r_i]}({\bf x}) = {\cal VH}[r_i]({\bf x}).
 \end{equation}
 With the data and the kernel vectors represented in the harmonic domain, we can apply a fast correlation to evaluate the contribution
 of each kernel point on the local data and perform this evaluation over all rotations.
 Given a vector at position ${\bf x}$, we compute the result $C^{\#}_i$ of this fast correlation over all spherical angles for 
 the i-th kernel vector as shown in (\ref{eq:feature:VHcorrFinal}):
 \begin{equation}
 C^{\#}_i = \widehat{ {\cal S}[r_i]}({\bf x}) \# \widehat{\cal T}_i.
 \label{eq:feature:vp_kernel_vhtrans}
 \end{equation}

\subsubsection{\label{sec:feature:vp_rotinvariance}Rotation Invariance}
As in the case of $n$-Point kernels,  we need to couple the contributions of the individual kernel vectors in such
a way that only the chosen kernel constellation (given by the $\Phi_i,\Theta_i, r_i$) has a contribution to the feature while rotating over 
all possible angles, i.e. the positions of the kernel vectors must not rotate independently. 
Note that the correct orientation of the kernel vectors
under the rotation is guaranteed by the Vectorial Harmonic formulation.\\
Since the $C^{\#}_i$ hold the contribution at each possible angle in a 3D Euclidean space with a ($\phi,\theta,\psi$) coordinate-system
(see section \ref{sec:feature:vhcorr}), we can perform the multiplicative
coupling of the separate sub-kernels (\ref{eq:feature:haar_separable_kernel}) by a angle-wise multiplication of the point-wise
correlation results: $\prod_{i=2}^n C^{\#}_i$.\\

By integrating over the resulting Euclidean space of this coupling, we easily obtain rotation invariance as in 
(\ref{eq:vp_paramterized}):
\begin{eqnarray}
\label{eq:feature:vp_almost_final}
\int\limits_{{\cal SO}(3)} \left( \prod\limits_{i=2}^n  {\cal C}^{\#}_i \right) \sin{\theta}d\phi d\theta d\psi.
\end{eqnarray}

\begin{figure}[th]
\centering
\psfrag{f1}{$\kappa_1$}
\psfrag{f2}{$\kappa_2$}
\psfrag{f3}{$\kappa_3$}
\psfrag{fn}{$\kappa_n$}
\psfrag{f0}{$\kappa_0$}
\psfrag{f4}{$\kappa()$}
\psfrag{VH}{${\cal VH}$}
\psfrag{SH(r1)}{\tiny${\cal VH}_{r_1}$}
\psfrag{SH(r2)}{\tiny${\cal VH}_{r_2}$}
\psfrag{SH(r3)}{\tiny${\cal VH}_{r_3}$}
\psfrag{SH(rn)}{\tiny${\cal VH}_{r_n}$}
\psfrag{SH}{\tiny${\cal VH}$}
\psfrag{C1}{\tiny$C_1$}
\psfrag{C2}{\tiny$C_2$}
\psfrag{C3}{\tiny$C_3$}
\psfrag{Cn}{\tiny$C_n$}
\psfrag{C0}{\tiny$\hat{\bf C}_0$}
\psfrag{c1}{\tiny$\hat{\bf C}_1$}
\psfrag{c2}{\tiny$\hat{\bf C}_2$}
\psfrag{c3}{\tiny$\hat{\bf C}_3$}
\psfrag{cn}{\tiny$\hat{\bf C}_n$}
\psfrag{t1}{\tiny$\hat{\cal T}_1$}
\psfrag{t2}{\tiny$\hat{\cal T}_2$}
\psfrag{t3}{\tiny$\hat{\cal T}_3$}
\psfrag{tn}{\tiny$\hat{\cal T}_n$}
\psfrag{Ct}{\tiny$C$}
\psfrag{FFT}{\tiny$\ \ {\cal F}^{-1}$}
\psfrag{rtd}{\tiny$\mathbb{R}^3$}
\psfrag{mi}{\quad$ \int$}
\includegraphics[width=0.8\textwidth]{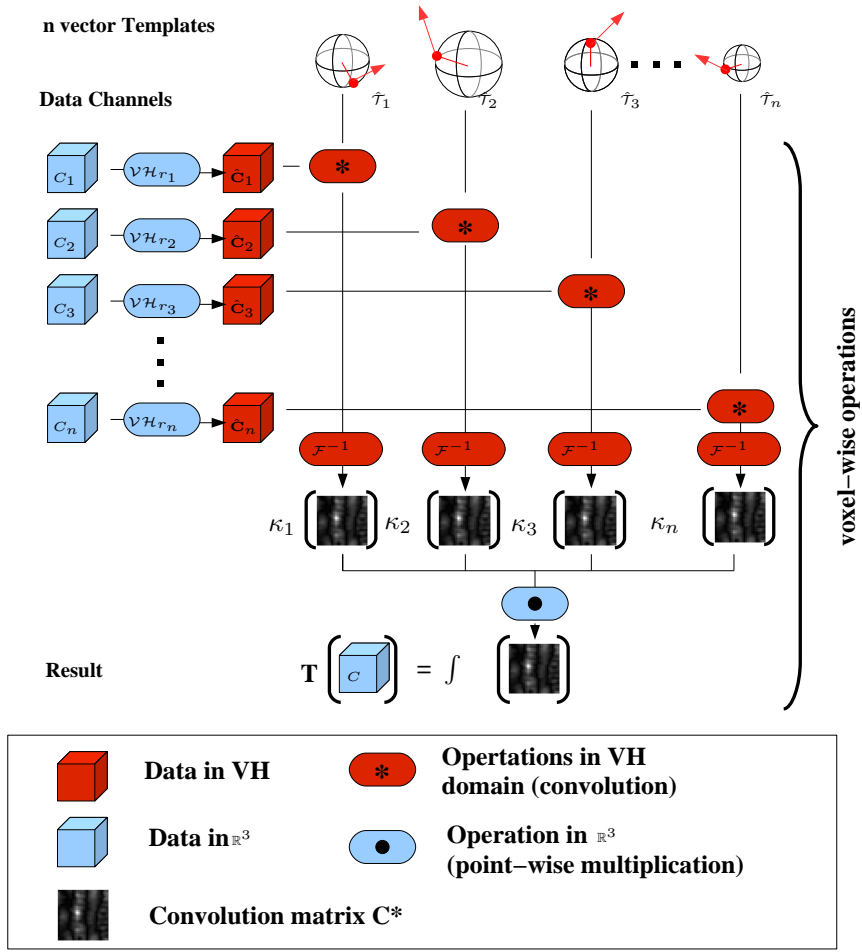}
\caption[Schematic overview of the computation of ``nv''-features.]{\label{fig:feature:nv_scheme} Schematic overview of the computation of $nv$-Features.}
\end{figure}

Finally, we still have to introduce the non-linear mapping into (\ref{eq:feature:vp_almost_final}) to satisfy (\ref{eq:feature:haarvp_formular})
. We follow the fast approach from (\ref{eq:feature:np_formularfast}), where we split $\kappa$ into $n$ non-linear mappings $\kappa_1,\dots,
\kappa_2$ which act directly on the correlation matrices. This leads to the final formulation of the $nv$-Feature:
\begin{eqnarray}
\label{eq:feature:vp_final}
T[\Lambda]({\bf x}) &:=&  \int\limits_{{\cal SO}(3)} \left( \prod\limits_{i=1}^n  \kappa_i\big(
{\cal C}^{\#}_i\big) \right) \sin{\theta} d\phi d\theta d\psi.
\end{eqnarray}
Figure \ref{fig:feature:nv_scheme} shows a schematic overview of the computation of $nv$-Features.

\subsection{Implementation}
The transformation into the harmonic domain is implemented as described in section \ref{sec:feature:VHimplement}. Hence, we
can also obtain the expansions at all points in $X$ at once using the convolution approach analogous to (\ref{eq:feature:discreteSHfull}).\\
The implementation of the template ${\cal T}_t$ has to be handled with some care: to avoid sampling issues, we apply
the same implementation strategies as in the case of the Spherical Harmonic base functions (see section \ref{sec:feature:VHimplement}
for details).\\
The computation of the correlation matrices ${\cal C}^{\#}$ follows the algorithm given in section \ref{sec:feature:vhcorr}. The
size of the padding $p$ we need to apply strongly depends on the angular resolution necessary to resolve the given configuration of the kernel
points.\\
Finally, the evaluation of the Haar-Integration over all possible rotations is approximated by the sum over the combined
$(\phi,\theta,\psi)$-space:
\begin{eqnarray}
\label{vp_final_discrete}
T[\Lambda]({\bf x}) &\approx&  \sum \left( \prod\limits_{i=2}^n  {\cal C}^{\#}_i \right).
\end{eqnarray}
\paragraph{Multi-Channel Data:}
$nv$-Features cannot directly combine data from several channels into a single feature. In case of
multi-channel data, we have to compute features for each channel separately.

\subsubsection{Complexity}
The computational complexity of the $nv$-Feature is dominated by the Vectorial Harmonic expansions needed to transform the 
input data and the kernel vector templates into the harmonic domain. This takes $O( b_{\max} \cdot m \log m)$ for input data of size 
$m$ and $O(n \cdot b_{\max} \cdot m' \log m')$ for a template size of $m'$. The costs for the correlation and multiplication of the 
correlation matrices are negligible.
\paragraph{Parallelization:}
As stated in section \ref{sec:feature:SHimplement}, we can gain linear speed-up in the number of cores for the parallelization of the
harmonic transformation. Further, we could also split the computation of the correlation matrices into several threads, but
as mentioned before, this speed-up hardly falls into account.

\subsection{\label{sec:feature:nvDiss}Discussion }
The $nv$-Features provide a powerful framework for the implementation of local features which are able obtain invariance towards 
rotation via Haar-Integration.\\
In practice, $nv$-Features are especially suitable for the design of highly specific features with a strong discriminative power used in
challenging image analysis tasks justifying the higher computational costs. For less complex problems, we are better off using some
of the less complex feature methods.\\

A major problem concerning the application of $nv$-Features is the huge set of kernel parameters $\Lambda$ (\ref{eq:feature:vp_lambda})
we have to choose. In practice, it is infeasible to try all possible parameter combinations in a feature selection process, as we suggest
for other features. Neither is it practically possible to select the best parameter settings by hand.
%To overcome this problem, we use
%the same automatic, data driven algorithm which learns the kernel parameters on given training samples as for the selection of the 
%$np$-Features (see section \ref{sec:feature:select_data}).

%\input{feature_selection}
\cleardoublepage
\chapter{\label{sec:feature:experiments} Experiments}
In the final chapter of the first part, we evaluate the feature methods which were introduced in the previous chapters. We start with the 
evaluation of the speed and accuracy of our fast correlation in Spherical Harmonics in section \ref{sec:feature:Shcorrexperiment} and
 the correlation in Vectorial Harmonics in section\ref{sec:feature:VHcorrexperiment}.\\
Section \ref{sec:feature:eval_complex} evaluates the computational complexity of our features on real world data. Then we use a database of 
semi-artificial 3D textures (see Appendix \ref{sec:app:texture_db}) to perform a series of 3D texture classification (see section 
\ref{sec:feature:eval_texture}).

\section{\label{sec:feature:Shcorrexperiment} Evaluating ${\cal SH}$-Correlation}
\index{Rotation}\index{Experiment}\index{Correlation}
Unlike previous publications \cite{Rot1}\cite{Rot2}\cite{Rot3}, which only performed a small set of experiments with a fixed
number of predefined example rotations, we evaluate our methods with a series of large scale experiments on real word data.\\
%We use the ``Princeton Shape Benchmark'' (PSB) (see Appendix \ref{sec:app:psb}) \cite{psb} dataset (which contains about 1800 3D objects)
%for our experiments.
If not mentioned otherwise, all experiments have the same basic setup: for each parameter set, we evaluate the error statistics of 100
random rotations of random objects. We generate the rotations over all possible angles $\phi,\psi \in [0, 2\pi[$ and $\theta \in
[0, \pi[$ with a resolution of $0.001 \approx 0.1^{\circ}$. Note that an error of $1^{\circ} \approx 0.017$. All given error rates are
the sums over the errors of all three angles.

\subsubsection{\label{sec:feature:Shcorrexperiment1}Rotating Objects in the Harmonic Domain}
In this first series of experiments, we extract a harmonic expansion with a fixed radius around the object center and then
rotate this expansion using (\ref{eq:feature:shrot}).\\

%\begin{floatingfigure}[r]{55mm}
\begin{figure}
\centering
\includegraphics[width=0.49\textwidth]{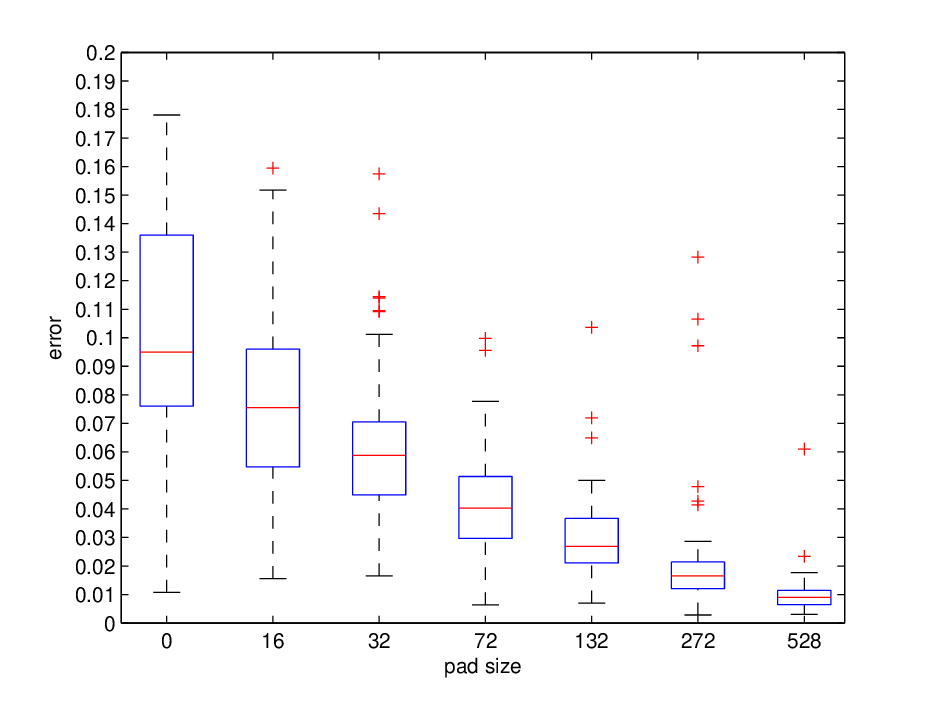}
\caption[Rotation estimation in ${\cal SH}$, error by pad size]{\label{fig:feature:SHcortest1} Estimation errors with $b=24$ and increasing pad
size $p$.}
\end{figure}
%\end{floatingfigure}
\paragraph{Pad Size:} In a first experiment, we are able to show the effect of our padding method on the estimation accuracy.
Figure (\ref{fig:feature:SHcortest1}) clearly shows the correlation of the pad size and the expected error.
It is also evident that we are able to achieve a high precision with errors below 1 degree. Hence, the experimental errors are found to be well within
the
theoretical bounds given in (\ref{eq:feature:Corr_err_pad}).

\paragraph{Maximum Band:} The next two experiments investigate the practical influence of the maximum expansion band on the
estimation errors.
\begin{figure}
\includegraphics[width=0.49\textwidth]{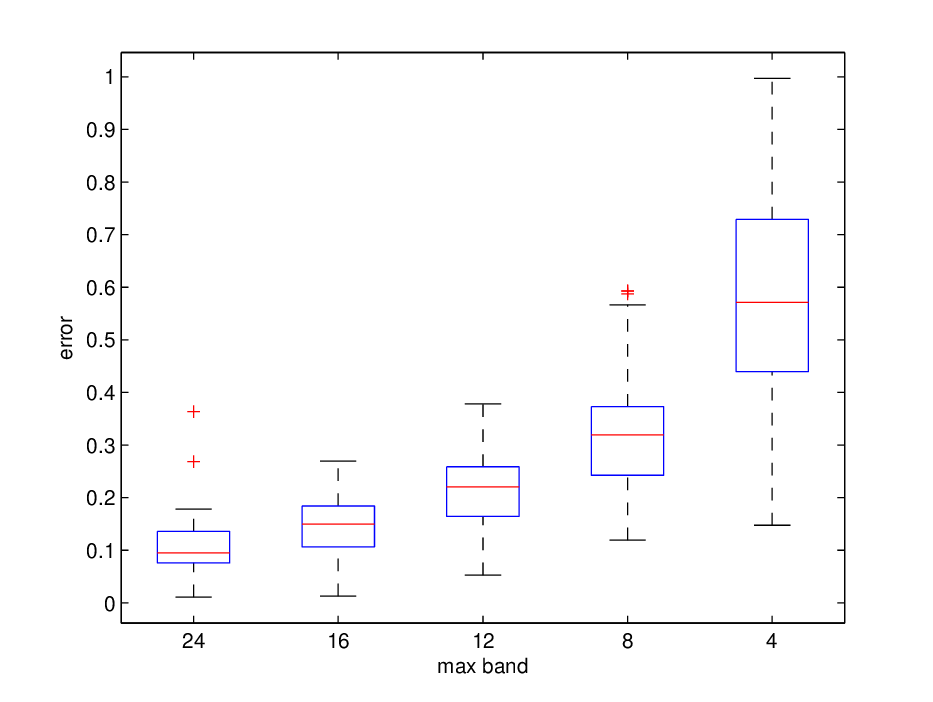}
\includegraphics[width=0.49\textwidth]{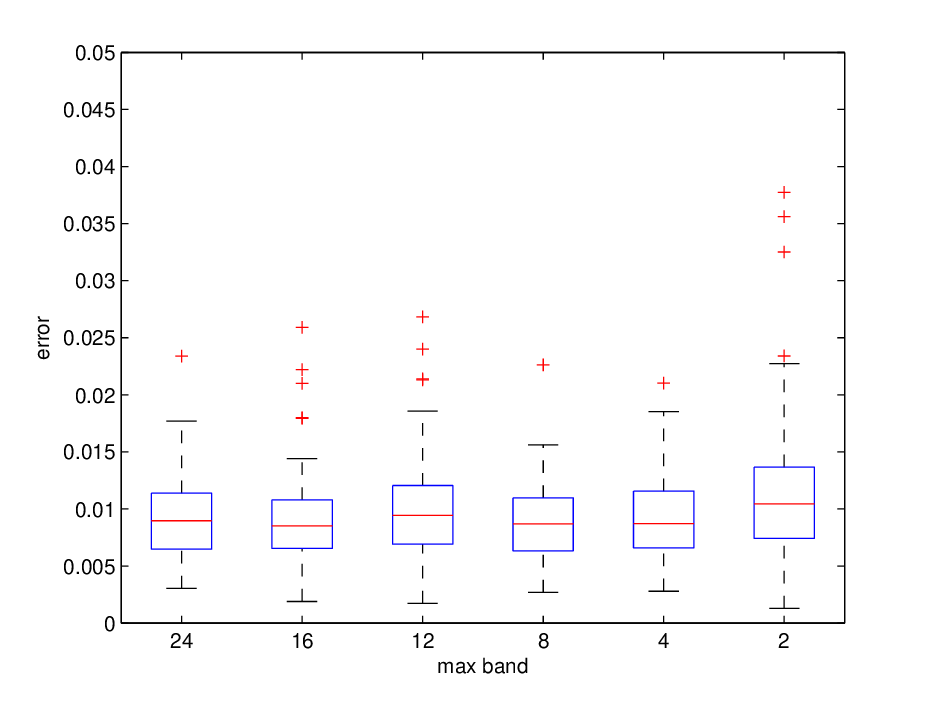}
\caption[Rotation estimation in ${\cal SH}$, error by max expansion band]{\label{fig:feature:SHcortest2} Estimation errors with
increasing maximum expansions. Left: $p=0$. Right: $p>512$ ($p$ is not fix due to the padding
to optimal FFT sizes). Note that the experiment with $p=0, b=2$ is left out because the result was so poor that it did not fit into the
the chosen error scale.}
\end{figure}
Figure (\ref{fig:feature:SHcortest2}) strongly supports our initial assumption that the original formulation is not able to achieve
accurate estimates for
low expansions. Our method on the other hand achieves very low error rates even for extremely low expansions with $b=2$.

\begin{figure}[ht]
\includegraphics[width=0.49\textwidth]{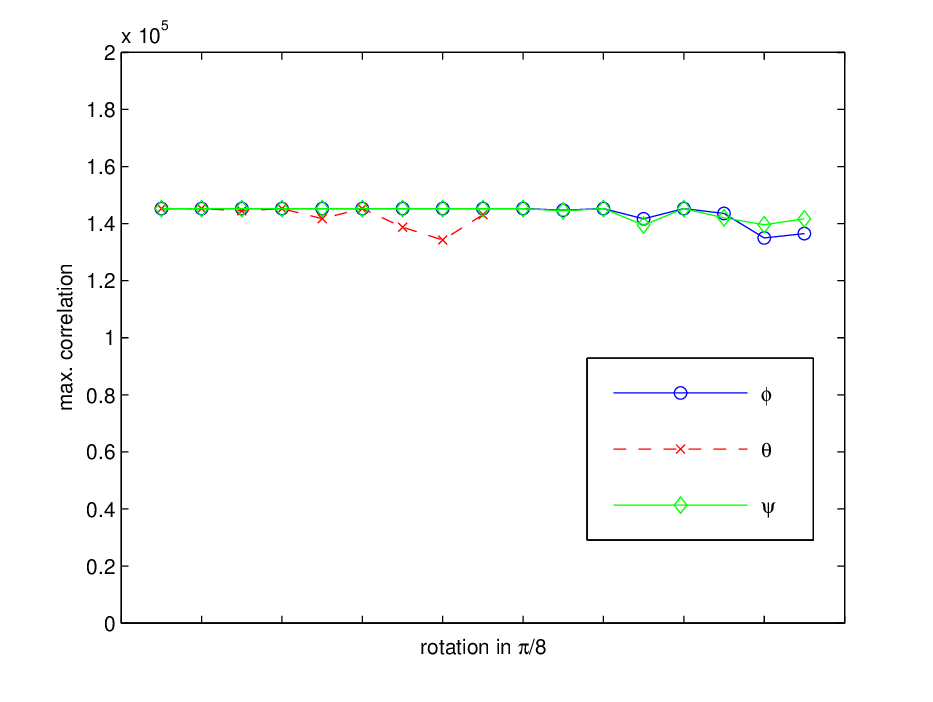}
\includegraphics[width=0.49\textwidth]{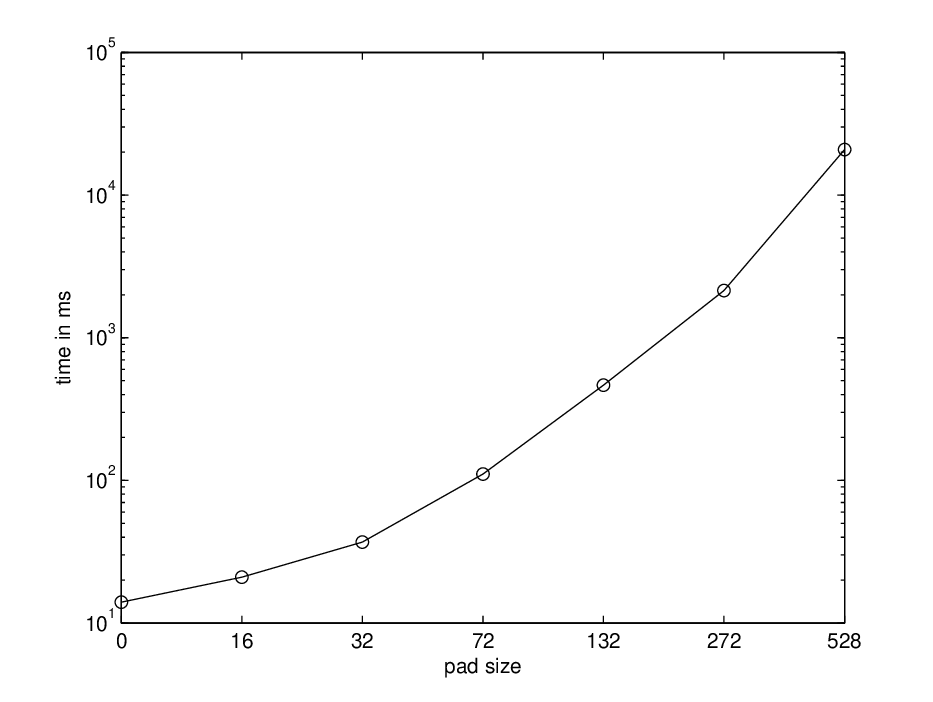}
\caption[Rotation estimation in ${\cal SH}$, costs]{\label{fig:feature:SHcortest3} Left: Maximum correlation for separate rotations in each angle.
Right: Computational costs in ms on a standard 2GHz PC.}
\end{figure}

\paragraph{Rotational Invariance and Computational Costs:}
Rotational Invariance and Computational Costs are investigated in the last two experiments
(figure (\ref{fig:feature:SHcortest3})) of
the first series. We rotate the object in $\pi/8$ steps in every angle to show that the correlation maximum is stable and
indeed independent of the rotation.
The computational complexity is largely dominated by the costs for the inverse FFT, hence growing with the pad size. So accuracy comes at
some cost but reasonable accuracy can still be achieved well within 1 second.
\subsubsection{\label{sec:feature:Shcorrexperiment2}Rotating Objects in $\mathbb{R}^3$}
The results of figure (\ref{fig:feature:SHcortest2}) suggest that the maximum expansion band has no influence on the quality
of the rotation
estimation - of course, this is only true if we are considering input signals that are limited to the very same maximum band. This is very
unlikely for very low bands in the case of real data.\\
In order to evaluate the actual influence of the maximum expansion band, we need to rotate the objects in $\mathbb{R}^3$ and extract
a second harmonic expansion after the rotation.
As mentioned before, the usability of our sinc interpolation approach is limited to correctly sampled (concerning the Sampling Theorem)
input signals (also see section \ref{sec:feature:implement} for more details on sampling issues). Hence, one must not expect to obtain precise rotation
estimates for low band expansions, which act as a low pass filter, of high frequent input signals.
Luckily, for most input data, we are not depending on the high frequent components in order to find the maximum correlation. Hence, we
can apply a low pass filter (Gaussian) on the input data prior to the harmonic expansion.
\begin{figure}[ht]
\includegraphics[width=0.49\textwidth]{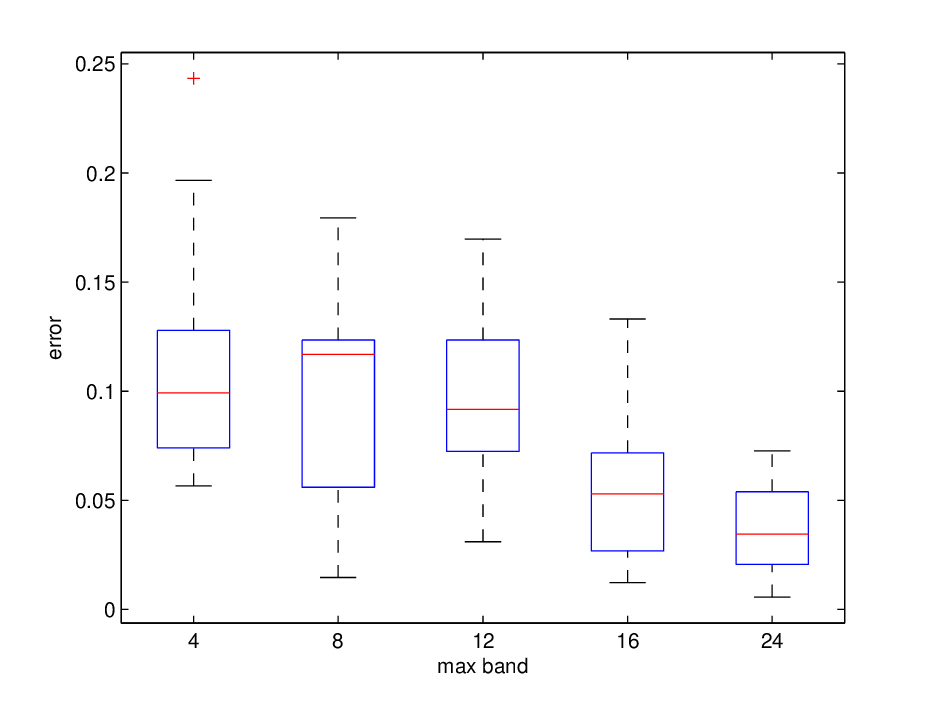}
\includegraphics[width=0.49\textwidth]{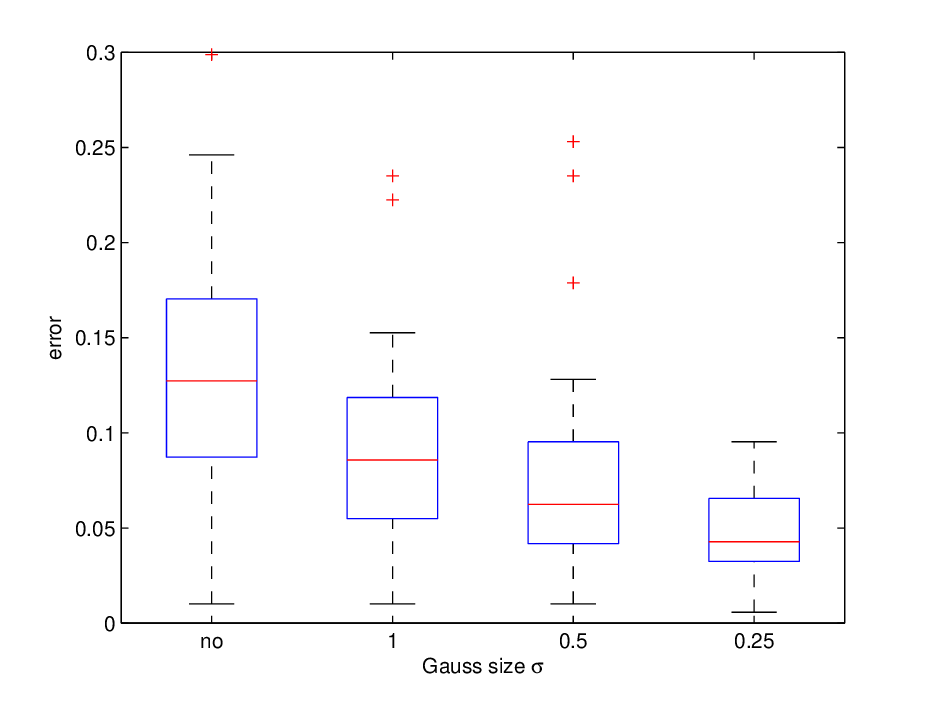}
\caption[Rotation estimation in ${\cal SH}$, real world error by max expansion band ]{\label{fig:feature:SHcortest4} Accuracy for rotations
in $\mathbb{R}^3$. Left: Influence of the maximum band $b$, $p>512, \sigma=0.25$. Right: Correct
sampling does matter! Without Gaussian smoothing and with different values for $\sigma$, $p>512, b=24$}
\end{figure}
Figure (\ref{fig:feature:SHcortest4}) shows the impact of the maximum band and smoothing for rotations in $\mathbb{R}^3$. Overall,
the estimation results
are slightly worse than before, but are still quite reasonable.

\section{\label{sec:feature:eval_complex}Evaluating the Feature Complexity}
\index{Experiments}\index{Complexity}
We evaluated the computational complexity on dummy volume data. All experiments were conducted on a 3GHz machine with 16 CPU cores and 128GB Ram. 
However, only a single CPU was used if not noted otherwise.
\subsection{\label{sec:feature:eval_complex_sh}Complexity of the Spherical Harmonic Transformation}
\begin{figure}[ht]
\centering
\psfrag{bmax}{$b_{max}$}
\psfrag{SH}{\tiny ${\cal SH}$}
\psfrag{SH with caching}{\tiny ${\cal SH}$ caching}
\psfrag{SH parallel}{\tiny ${\cal SH}$ parallel}
\psfrag{volume size}{volume size in $x^3$}
\epsfig{file=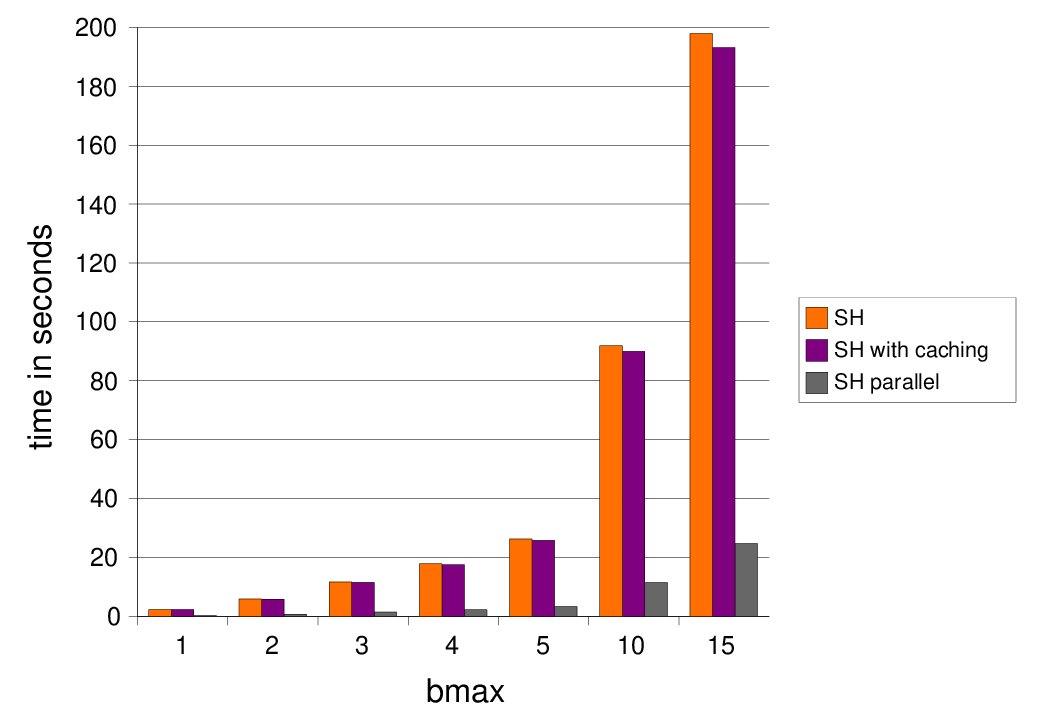, width=8cm}
\epsfig{file=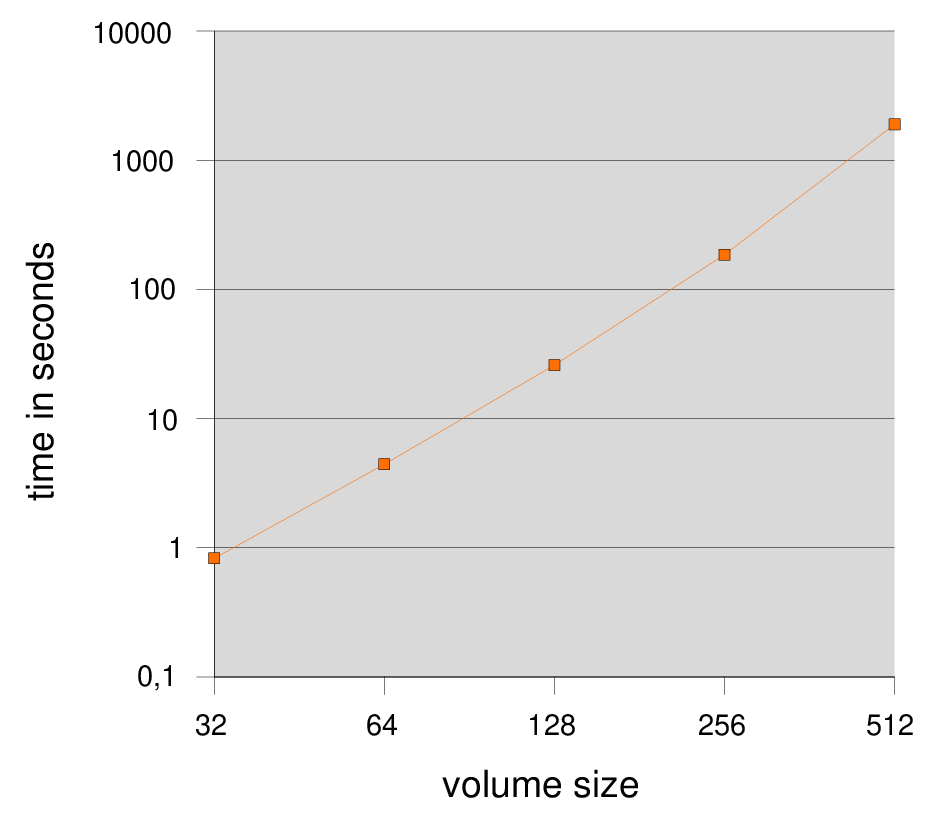, width=6.4cm}
\caption[Computational Complexity of the ${\cal SH}$ Transform]{\label{fig:feature:exp_shspeed} {\bf Left:} Computational complexity of the ${\cal SH}$ transformation:
voxel-wise computation at radius $r=10$ on a $(128\times 128 \times 128)$ test volume. The parallel computation was performed with 8 cores and 
shows an almost linear speed-up, while the caching of the ${\cal SH}$ coefficients has only a small effect.\\
{\bf Right:} Computational complexity in dependency of the volume size (single core results with $b_{max}=5$). The logarithmic increase in
the complexity clearly indicates that the underlying Fourier Transform, which is used for the convolution of the base functions with the 
data, is dominating the overall complexity of the harmonic transformation.}
\end{figure}

\index{Experiments}\index{Complexity}\index{Spherical Harmonics}
We conducted two experiments to show the complexity of the voxel-wise Spherical Harmonic transformation of 3D volume data. The complexity
is independent of the actual data and is only influenced by the maximum expansion band ($b_{max}$) and the data size as shown in 
figure \ref{fig:feature:exp_shspeed}.

\section{\label{sec:feature:VHcorrexperiment} Evaluating ${\cal VH}$-Correlation}
\begin{figure}[ht]
\centering
\includegraphics[width=0.4\textwidth]{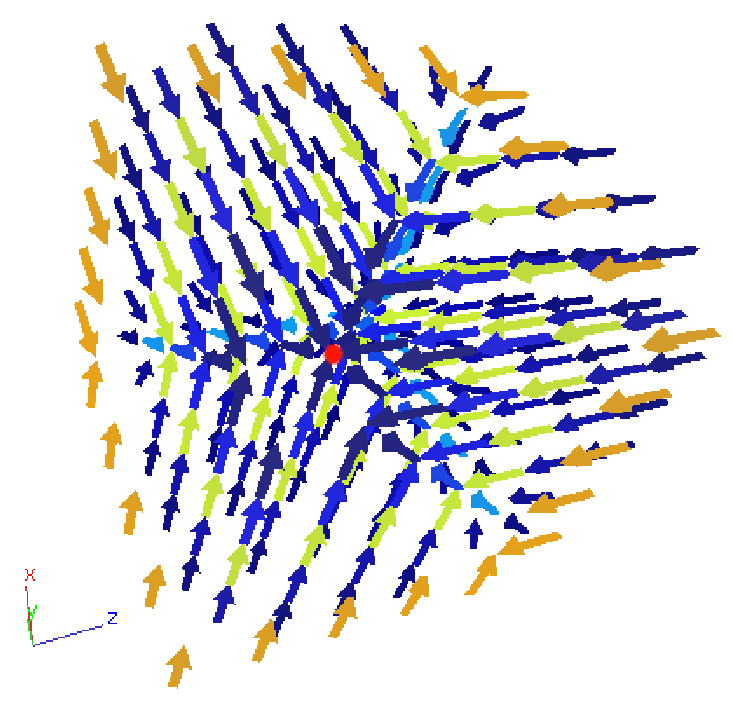}
\caption[Test sample for the accuracy of the ${\cal VH}$ correlation.]{\label{fig:feature:exp:vec_corner} Artificial 3D vector field used for the rotation estimation experiments. 
The red dot indicates the position ${\bf x}$ of the rotation center, at which the spherical test
patches have been extracted.}
\end{figure}

\index{Rotation}\index{Experiment}\index{Correlation}
We use a sample 3D
vector field (see figure \ref{fig:feature:exp:vec_corner}) which is rotated around the center of one
spherical patch parameterized by a single radius of $r=10$.

For each experiment, we evaluate the error statistics of 100
random rotations of this vector field. We generate the rotations over all possible angles
$\varphi,\psi \in [0, 2\pi[$ and $\theta \in [0, \pi[$ with a resolution of $0.001
\approx 0.1^{\circ}$. Note that an error of $1^{\circ} \approx 0.017$. All given error rates
are the accumulated errors of all three angles.
\begin{figure}[ht]
\centering
\psfrag{bmax}{$b_{max}$}
\psfrag{Error}{error}
\includegraphics[width=0.31\textwidth]{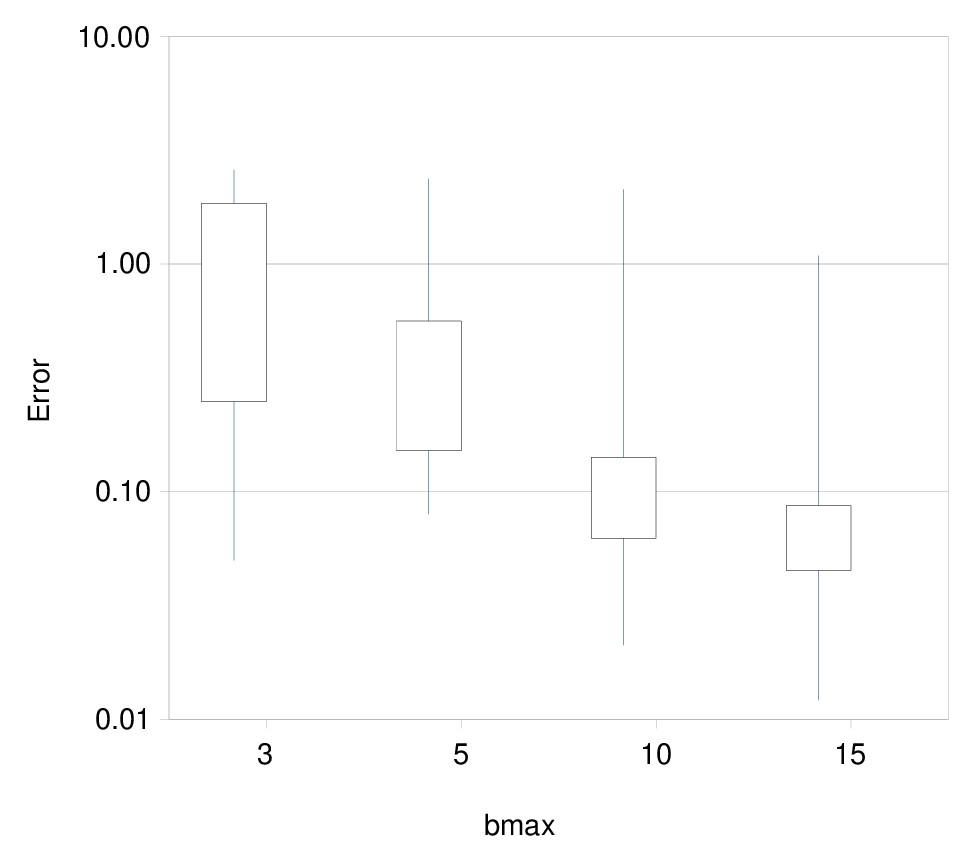}
\includegraphics[width=0.31\textwidth]{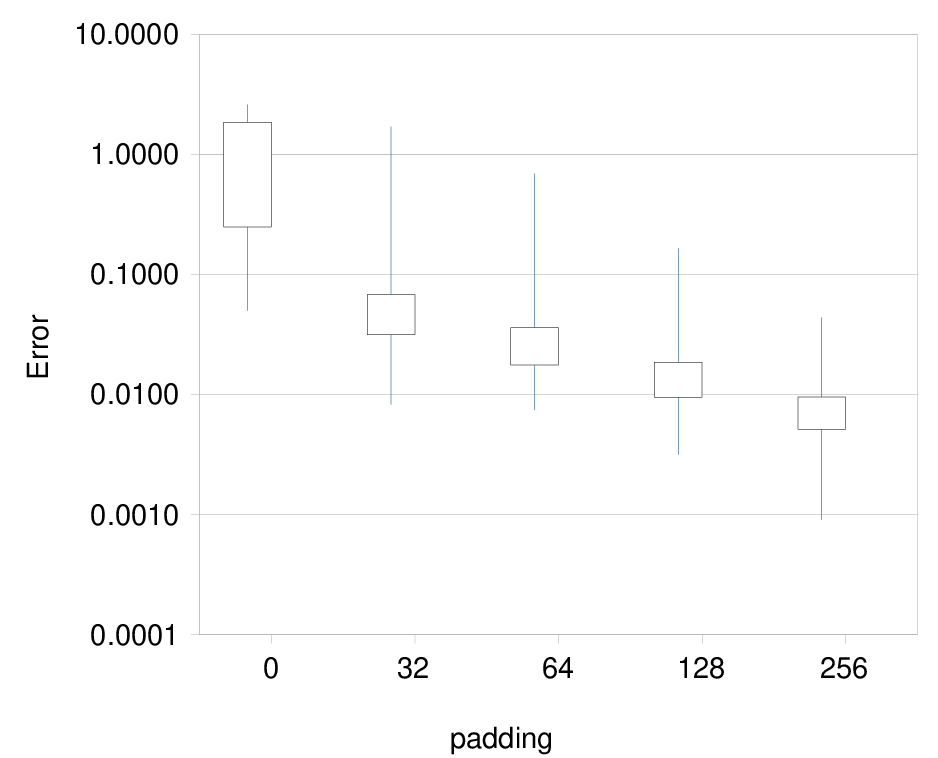}
\includegraphics[width=0.31\textwidth]{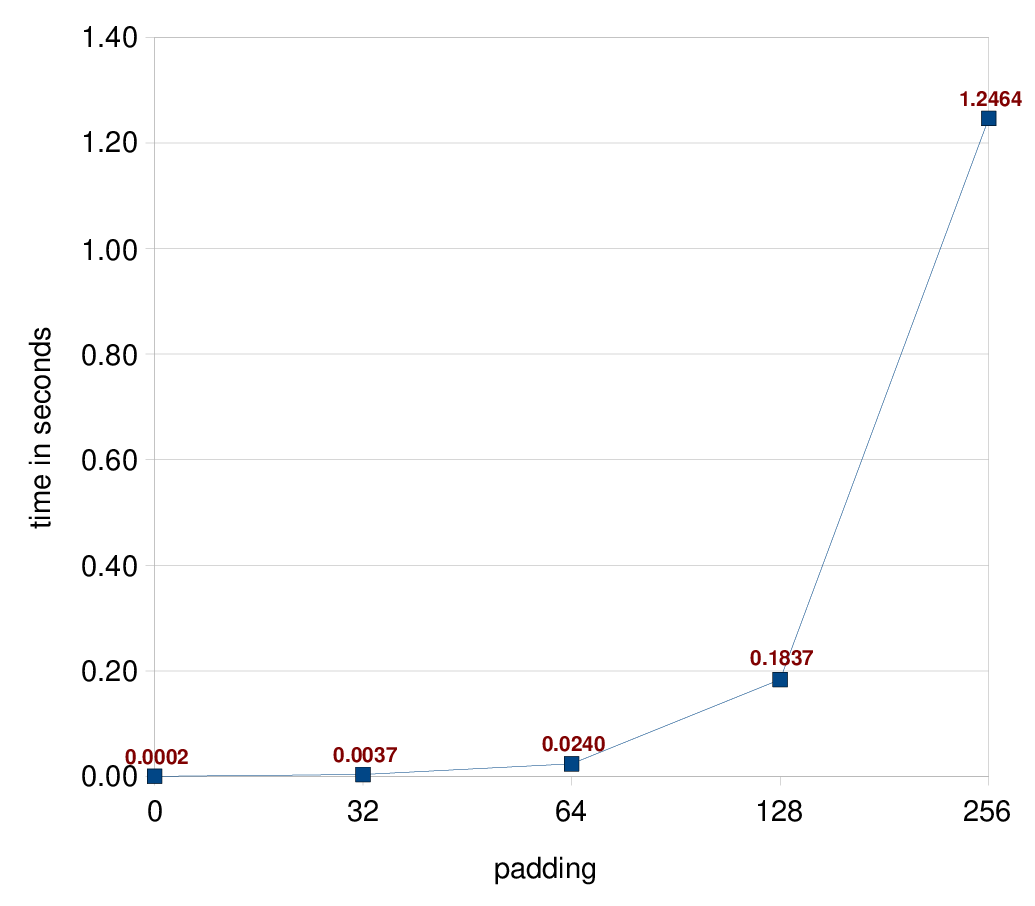}
\caption[Accuracy of the  ${\cal VH}$ correlation.]{\label{fig:feature:exp:VHcorr_band} {\bf Left:} Accumulated rotation estimation error for increasing $b_{max}$ and
without using the Sinc interpolation method ($p=0$). {\bf Center:} Accumulated rotation estimation error for increasing pad size $p$
of the Sinc interpolation with $b_{max}=5$. {\bf Right:} Computational complexity for increasing pad size $p$
of the Sinc interpolation with $b_{max}=5$. The experiments were performed on a standard 2GHz PC,
using the FFTW \cite{FFTW} implementation of the inverse FFT.}
\end{figure}
Figure \ref{fig:feature:exp:VHcorr_band} shows the direct effect of the maximum expansion band $b_{max}$ on the
rotation estimate. But even for expensive ``higher band'' expansions, we encounter strong outliers
and a rather poor average accuracy.\\
This can be compensated by our Sinc interpolation approach (\ref{eq:feature:sinc}): Figure
\ref{fig:feature:exp:VHcorr_band} shows how we can reduce the rotation estimation error well below $1^{\circ}$,
just by increasing the pad size $p$. The additional computational costs caused by the padding
are also given in figure \ref{fig:feature:exp:VHcorr_band}.

Summarizing these first experiments, we are able to show that our proposed method is able
to provide a fast and accurate rotation estimation even for rather low band expansions, e.g.
if we choose $p=64$ and $b_{max}=5$, we can expect an average estimation error below $1^{\circ}$
at a computation time of less than $25$ms.

\begin{figure}[ht]
\centering
\includegraphics[width=0.4\textwidth]{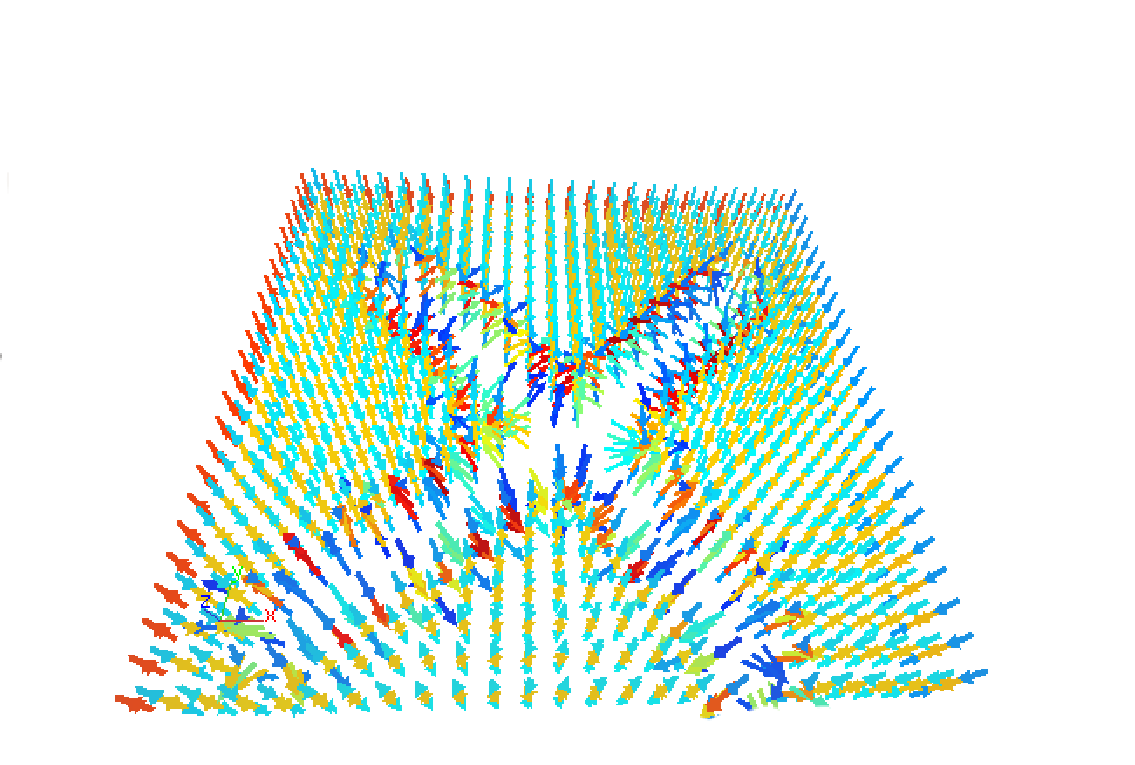}
\includegraphics[width=0.4\textwidth]{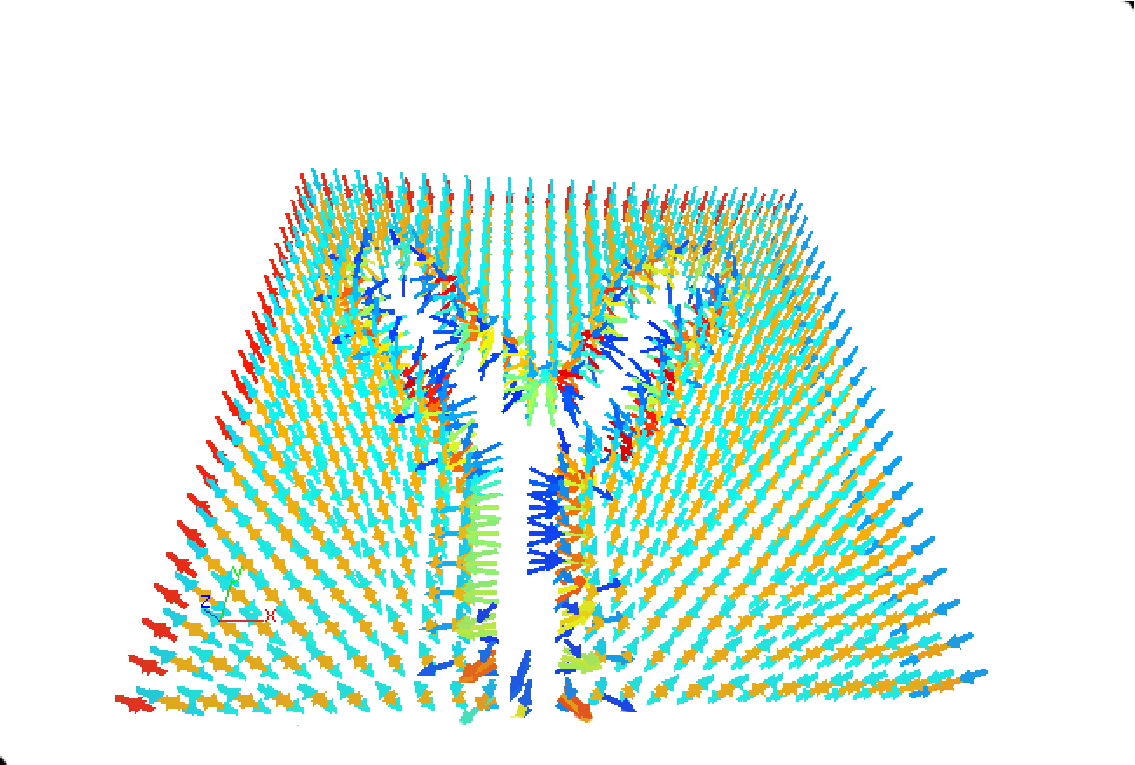}
\caption[Sample target structures for the ${\cal VH}$ correlation.]{\label{fig:feature:exp:Exp_1} Sample target structures for the detection problem: 3D vector fields of
$X$- and $Y$-like shaped bifurcations.}
\end{figure}

\begin{figure}[ht]
\centering
\includegraphics[width=0.4\textwidth]{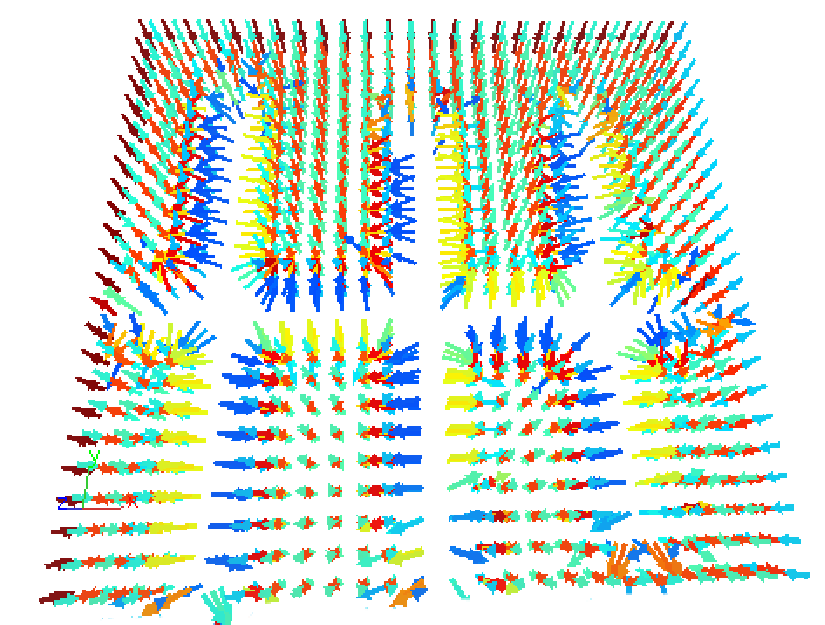}
\includegraphics[width=0.3\textwidth]{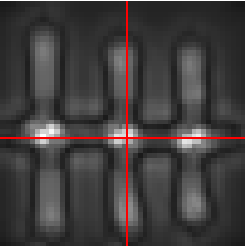}
\caption[Sample test data for the ${\cal VH}$ correlation.]{\label{fig:feature:exp:Exp_2} {\bf Left:} Sample test data. {\bf Right:} xy-slice of a sample correlation
result for the $X$-bifurcation target. The red cross indicates the position of the maximum
correlation value.}
\end{figure}

\begin{figure}[ht]
\centering
\includegraphics[width=0.4\textwidth]{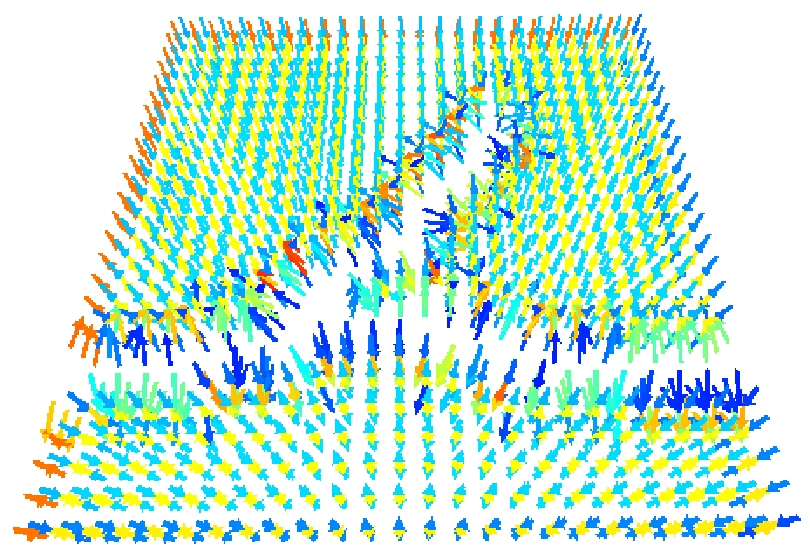}
\includegraphics[width=0.3\textwidth]{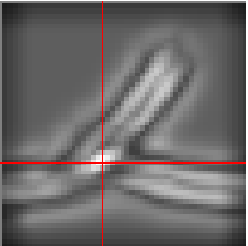}
\caption[Sample test data for the ${\cal VH}$ correlation.]{\label{fig:feature:exp:Exp_3} {\bf Left:}  Sample test data. {\bf Right:} xy-slice of the correlation
result for the $Y$-bifurcation target. The red cross indicates the position of the maximum
correlation value.}
\end{figure}

\paragraph{Key Point Detection.}
In a second series of experiments, we evaluate the performance of
our methods in a key point (or object) detection problem on artificial data. Figure \ref{fig:feature:exp:Exp_1}
shows the 3D vector fields of two of our target structures. Our goal is to detect the center
of such $X$- and $Y$-like shaped bifurcations under arbitrary rotations in larger vector fields.
For each target structure, we extract a single patch, parameterized in four different radii with
$b_{max}=3$, at the center of the bifurcations.\\
Using (\ref{eq:feature:discreteVH}), we extract patches with the same parameterization at each point of the
test samples and apply our fast, combined (see section \ref{sec:feature:shcorrradii}) and normalized
(\ref{eq:feature:SHcorrnormFinal} )
cross-correlation to detect the target structures in the test vector fields.
Figures \ref{fig:feature:exp:Exp_2} and \ref{fig:feature:exp:Exp_3} show some example test data together with the correlation results.

It should be noted that the test bifurcations are only similar in terms of a $X$ or $Y$ shape,
but not identical to the given target structures. We also rotate the test data in a randomized
procedure over all angles.


Applying a threshold of $0.9$ to the correlation results, we were able to detect the correct
target structures in all of our test samples without false positives.

\subsection{\label{sec:feature:eval_complex_vh}Complexity of the Vectorial Harmonic Transformation}
\index{Experiments}\index{Complexity}\index{Vectorial Harmonics}

\begin{figure}[ht]
\centering
\psfrag{bmax}{$b_{max}$}
\psfrag{VH}{\tiny ${\cal VH}$}
\psfrag{VH paralell}{\tiny ${\cal VH}$ parallel}
\epsfig{file=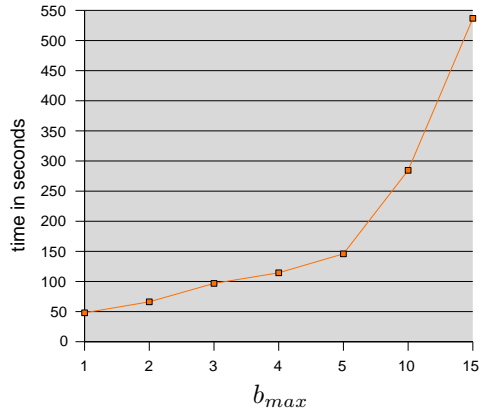, width=6.4cm}
\caption[Computational Complexity of the ${\cal VH}$ Transform]{\label{fig:feature:exp_vhspeed} Computational complexity of the ${\cal VH}$ transformation:
voxel-wise computation at radius $r=10$ on a $(128\times 128 \times 128)$ test vector field.
The computation was performed using 8 cores in parallel.}
\end{figure}
We also performed the experiment measuring the complexity in dependency of the maximum expansion band ($b_{max}$) for the voxel-wise
Vectorial Harmonic transformation of 3D volume data. Figure \ref{fig:feature:exp_vhspeed} clearly shows that the complexity of the
transformation in the vectorial case is much higher than in the scalar case. This can only be compensated by the parallelization of
the transformation.

\subsection{\label{sec:feature:eval_complex_feature}Complexity of a voxel-wise Feature Extraction }
\index{Experiments}\index{Complexity}\index{${\cal SH}_{abs}$}\index{${\cal SH}_{autocorr}$}\index{${\cal SH}_{bispectrum}$}
\index{${\cal SH}_{phase}$}\index{$2p$-Feature}\index{$3p$-Feature}\index{$np$-Feature}\index{$1v$-Feature}\index{$2v$-Feature}
\index{$nv$-Feature}\index{3D LBP}\index{${\cal VH}_{abs}$}\index{${\cal VH}_{autocorr}$}
In the final experiment regarding the computational complexity, we evaluated all features on a $(250\times 250 \times 250)$ volume texture
sample. We extracted voxel-wise features simultaneously at all voxels. We used a fixed radius of $r=10$ and evaluated the computation time
on a single core with $b_{max}=\{3,5,8\}$.\\

\begin{figure}[ht]
\centering
\psfrag{Shabs  }{\tiny${\cal SH}_{abs}$}
\psfrag{Shphase  }{\tiny${\cal SH}_{phase}$}
\psfrag{Shautocorr  }{\tiny${\cal SH}_{autocorr}$}
\psfrag{Shbispectrum  }{\tiny${\cal SH}_{bispectrum}$}
\psfrag{SH  }{\tiny${\cal SH}$}
\psfrag{LBP  }{\tiny LBP}
\psfrag{2p  }{\tiny$2p$}
\psfrag{3p  }{\tiny$3p$}
\psfrag{4p4  }{\tiny$4p$ (4)}
\psfrag{4p1  }{\tiny$4p$ (1)}
\psfrag{5p5  }{\tiny$5p$ (5)}
\psfrag{5p1  }{\tiny$5p$ (1)}
\psfrag{Vhabs  }{\tiny${\cal VH}_{abs}$}
\psfrag{Vhautocorr  }{\tiny${\cal VH}_{autocorr}$}
\psfrag{1v  }{\tiny$1v$}
\psfrag{2v  }{\tiny$2v$}
\psfrag{3v  }{\tiny$3v$}
\psfrag{4v  }{\tiny$4v$}
\psfrag{VH  }{\tiny${\cal VH}$}
\psfrag{bmax = 3}{$b_{max}=3$}
\psfrag{bmax = 5}{$b_{max}=5$}
\psfrag{bmax = 8}{$b_{max}=8$}
\includegraphics[width=0.9\textwidth]{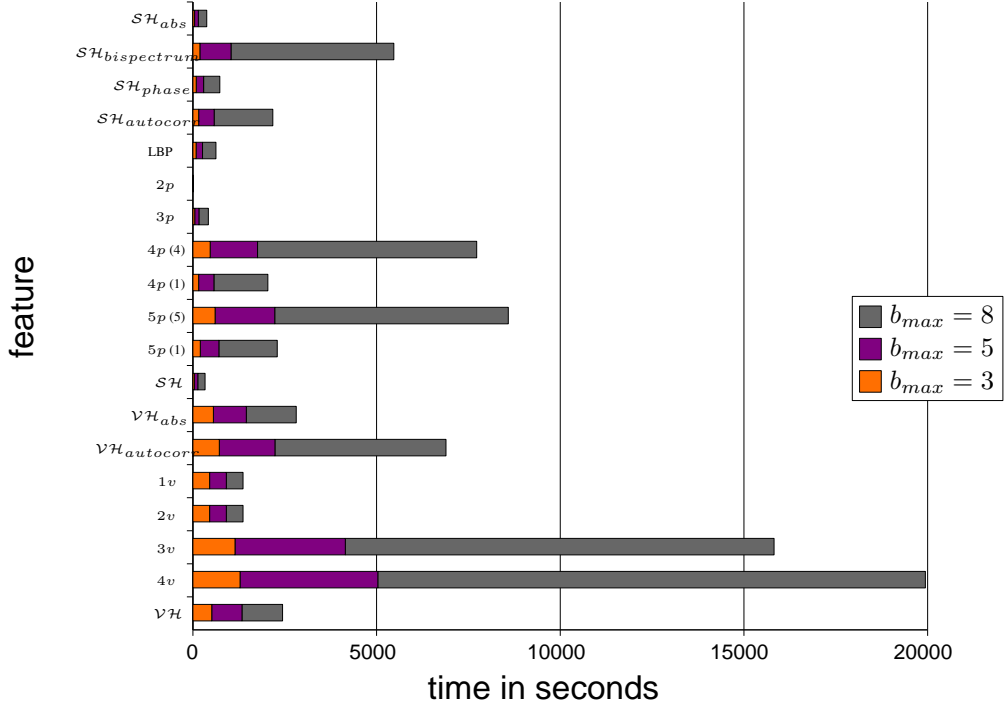}
\caption[Comparing the Computational Complexity of a voxel-wise Feature Extraction for all Features]{\label{fig:feature:exp_featurespeed} 
Illustrates the computational complexity of the individual features as given in table \ref{tab:feature:exp_featurespeed}. The complexity
was measured on a  $(250\times 250 \times 250)$ volume texture sample with $r=10$ and $b_{max}=\{3,5,8\}$ using only a single CPU core. 
We also give the complexity for the ${\cal SH}$ and ${\cal VH}$ transformations as reference values. $4p$ (4) indicates that the $4p$-Feature 
was computed with kernel points in 4 different channels
(\ref{eq:feature:np_final} ), whereas $4p$ (1) indicates the fast $np$-Feature version
(\ref{eq:feature:np_formularfast}), where all kernel points 
are located in the same channel.}
\end{figure}

Figure \ref{fig:feature:exp_featurespeed} illustrates the computation time for all features which is also given in table
\ref{tab:feature:exp_featurespeed}. The complexity of the individual features has a wide range: from about 3 seconds for the
computation of the simple $2p$-Feature (which is not based on a Spherical Harmonic transformation), to almost 4 hours needed to compute
a $4v$-Feature with $b_{max}=8$ at every voxel of the $(250\times 250 \times 250)$ volume.

\begin{figure}[ht]
\psfrag{corr    }{\tiny corr}
\psfrag{mult    }{\tiny mult}
\psfrag{SH    }{\tiny $\cal SH$}
\centering
\subfigure[1 core vs 8]{
\includegraphics[width=0.29\textwidth]{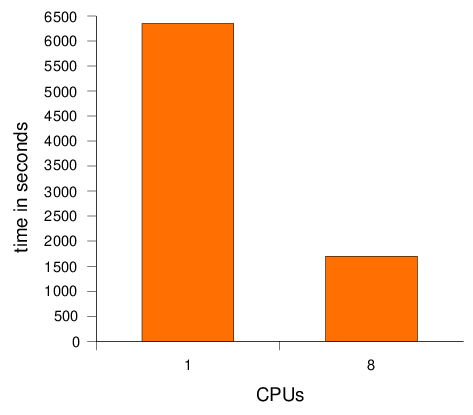}
}
\subfigure[1 core]{
\includegraphics[width=0.29\textwidth]{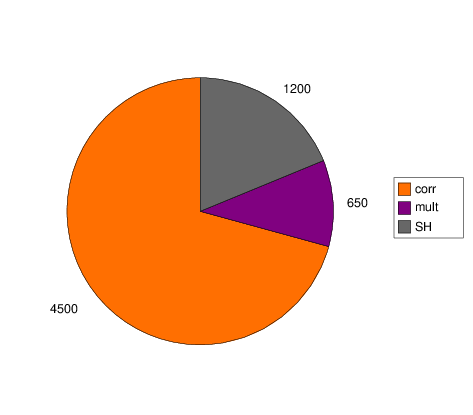}
}
\subfigure[8 cores]{
\includegraphics[width=0.29\textwidth]{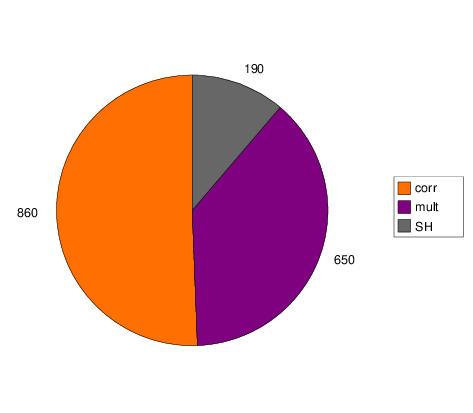}
}
\caption[Computational Complexity of a $5p$-Feature on a single and a 8 core system.]{\label{fig:feature:exp_5pspeed} Comparing the 
computational complexity of a $5p$-Feature on a single and a 8 core system. (a) Parallelization speed-up. (b) Distribution of the computation 
time with a single core. (c) Distribution of the computation time at the parallelization to 8 cores.   }
\end{figure}

It is obvious that some of the features are too complex to be of practical use in such a setting as presented here. Especially, a computation 
of the highly specialized $np$ and
$vp$-Features at all voxels and at a high expansion band $b_{max}$ appears to be practically intractable.\\ However, it turns out that this is 
not a major drawback in practice: First of all, as figure \ref{fig:feature:exp_5pspeed} shows, the features are well suited for 
parallelization, and second, it is usually not necessary to compute such specific features at all $256^3$ voxels. Typically, it is very easy
to reduce the number of candidate voxels drastically, if one uses the response of ``cheap'' features to perform a rough 
pre-segmentation.

\begin{table}[ht]
\centering
\begin{tabular}{||l|c|c|c||}
\hline
\hline
feature & $b_{max}=3$ & $b_{max}=5$ & $b_{max}=8$ \\
\hline
${\cal SH}_{abs}$	&46  &102	&229\\
${\cal SH}_{phase}$	&190 &848	&4431\\
${\cal SH}_{autocorr}$	    &90	&201 &441\\
${\cal SH}_{bispectrum}$	&157 &422	&1596\\
LBP	&87  &175	&365\\
$2p$	&3   &3	&3\\
$3p$	&50  &114	&257\\
$4p$ (4)	&470 &1290    &5965\\
$4p$ (1)	&156 &416	&1468\\
$5p$ (5)	&604 &1624    &6358\\
$5p$ (1)	&201 &506	&1589\\
${\cal SH}$	&40  &93	&198\\
${\cal VH}_{abs}$	&558 &896	&1359\\
${\cal VH}_{autocorr}$	&719 &1515    &4655\\
$1v$	&455 &455	&455\\
$2v$	&455 &455	&455\\
$3v$	&1146	&3003	&11671\\
$4v$	&1283	&3751	&14905\\
${\cal VH}$ &513   &823	&1105\\
\hline
\hline
\end{tabular}
\caption[Comparing the Computational Complexity of a voxel-wise Feature Extraction for all Features.]{\label{tab:feature:exp_featurespeed} 
Computational complexity of the individual features as illustrated in figure \ref{fig:feature:exp_featurespeed}.
The complexity
was measured on a  $(250\times 250 \times 250)$ volume texture sample with $r=10$ and $b_{max}=\{3,5,8\}$ using only a single CPU core.
}
\end{table}
\paragraph{Multicore Speed-up:} We examined the potential speed-up of a parallelization of the feature computation at the example of the
$5p$ (5) feature (see table \ref{tab:feature:exp_featurespeed}).
Using 8 instead of a single CPU core, the complexity drops from 6350s (almost
2 hours) to 1700s ($\approx$ 30min). Figure \ref{fig:feature:exp_5pspeed} shows how the parallelization affects the different computation steps
like the ${\cal SH}$ transformation, the correlation step or the non-linear transformations and multiplications.
\clearpage
\section{\label{sec:feature:eval_texture}Evaluating 3D Texture Discrimination}
\index{Experiments}\index{Texture Discrimination}
In a final experiment, we evaluated the texture discrimination performance of our proposed features. The experiments were conducted
on our artificial 3D volume texture database (see appendix \ref{sec:app:texture_db} for details on this database).\\
\begin{figure}[ht]
\centering
\includegraphics[width=0.3\textwidth]{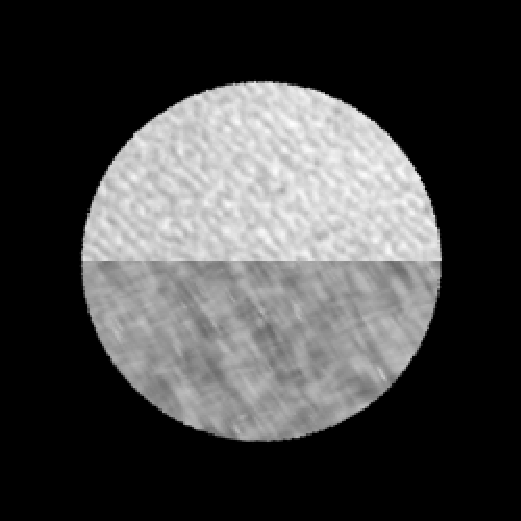}
\includegraphics[width=0.3\textwidth]{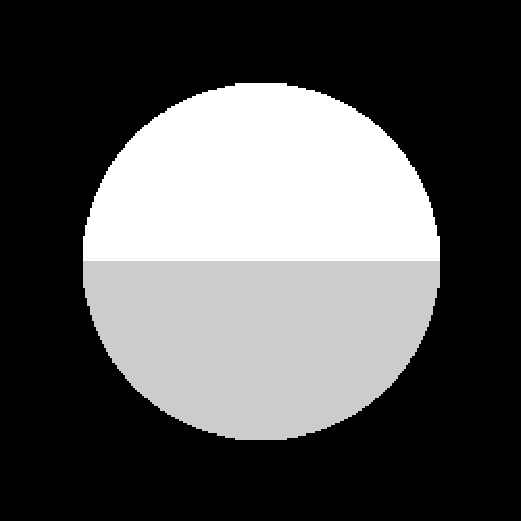}
\caption[Volume Texture Segmentation Test Samples]{\label{fig:app:db_test} Example 3D texture database entry. {\bf Left:} the xy-slice shows 
how two randomly selected textures are combined in one test sample. {\bf Right:} ground-truth labeling.}
\end{figure}
Using the SIMBA feature selection algorithm, we extracted the top 10 parameter combinations 
for each of our features. Scalar features were expanded to the 5th band, vectorial features were computed on the gradient field of the scalar
input data and expanded to the 3rd band.\\
Given these feature vectors, we used a voxel-wise SVM classification to evaluate the 3D texture segmentation 
performance of the individual features. 

\begin{table}[ht]
\centering
\begin{tabular}{||l|c|c||}
\hline
\hline
feature & rotations & rotations and gray-scale changes \\
\hline
${\cal SH}_{abs}$	    & 91\% & 82\% \\
${\cal SH}_{phase}$	    & 90\% & 90\% \\
${\cal SH}_{autocorr}$      & 93\% & 94\% \\
${\cal SH}_{bispectrum}$    & 94\% & 89\% \\
LBP			    & 91\% & 91\% \\
$2p$			    & 85\% & 78\% \\
$3p$			    & 86\% & 79\% \\
$4p$ (4)		    & 93\% & 93\% \\
$4p$ (1)		    & 92\% & 92\% \\
$5p$ (5)		    & 91\% & 91\% \\
$5p$ (1)		    & 91\% & 91\% \\
${\cal VH}_{abs}$	    & 94\% & 94\% \\
${\cal VH}_{autocorr}$	    & \bf{95}\% & \bf{95}\% \\
$3v$			    & 91\% & 91\% \\
$4v$			    & 93\% & 93\% \\
\hline
\hline
\end{tabular}
\caption[Comparing the Computational Complexity of a voxel-wise Feature Extraction for all Features.]{\label{tab:feature:exp_texture}
3D texture segmentation benchmark. Voxel-vise error rate in percent. $4p$ (4) indicates that the $4p$-Feature was computed with kernel points in
4 different channels(\ref{eq:feature:np_final} ), whereas $4p$ (1) indicates the fast $np$-Feature version
(\ref{eq:feature:np_formularfast}), where all kernel points are located in the same channel.}
\end{table}

Our evaluation clearly shows that those features that are not invariant towards gray-scale changes strongly suffer in the case of such changes.
The vectorial features appear to be very stable, however this comes at the cost of higher computational complexity (see table 
\ref{tab:feature:exp_featurespeed}).\\
The highly specific $np$ and $vp$-Features are not able to outperform the other approaches. These features are probably too selective to be able to
describe the large variations in the textures by just 10 parameter settings. However, these features anyway have been designed for key point and object 
detections (see section \ref{sec:feature:VHcorrexperiment}) rather than texture description.

\begin{appendix}
%\chapter*{Appendix}
\cleardoublepage
\chapter{\label{sec:app:texture_db}Artificial 3D Volume Texture Database}
\index{Texture DB}
The following tables show a few sample images of xy-slices taken from the training samples of our artificial 3D volume texture database.

\subsection{Texture Generation}
\begin{figure}[ht]
\centering
\includegraphics[width=0.6\textwidth]{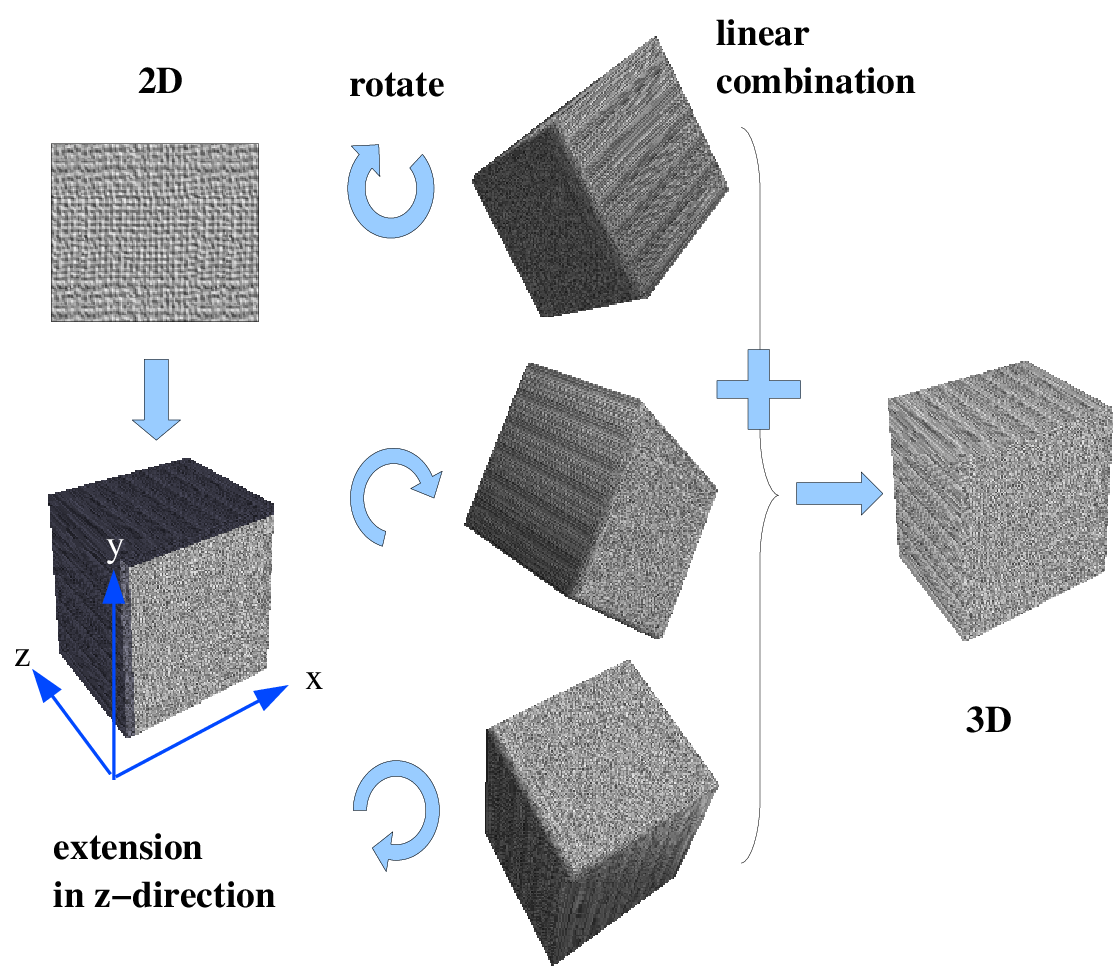}
\caption[Generation of the volume texture database]{\label{fig:app:db-train}Generation of the volume texture database: in a first step, we extend 2D texture images $X$ from the 
{\bf BFT texture data base}  into volumes $V$, such that $\forall z: V(x,y,z) = X(x,y)$. We then generate volume textures $VT$ as linear combinations of arbitrary rotations
of these volumes: $VT := \alpha_1{\cal R}_1 V_1 + \dots +  \alpha_n{\cal R}_n V_n$.
}
\end{figure}

The volume textures were generated from 2D texture samples which were taken from the {\bf BFT texture data base} provided by the University
of Bonn ({\it http://btf.cs.uni-bonn.de/download.html}).  Figure \ref{fig:app:db-train} gives an overview of our very simple volume texture generation process.

The number of linear combinations $n$, as well as the rotations ${\cal R}_i$ and factors $\alpha_i \in [0,1]$ are chosen randomly.

\begin{table}[ht] 
\centering 
\begin{tabular}{||ccccc||} 
\hline
\multicolumn{5}{||c||}{texture 1} \\ 
\hline
\includegraphics[width=0.15\textwidth]{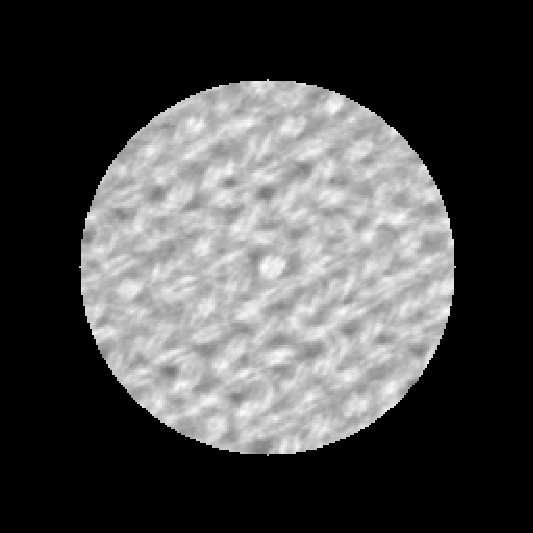}&\includegraphics[width=0.15\textwidth]{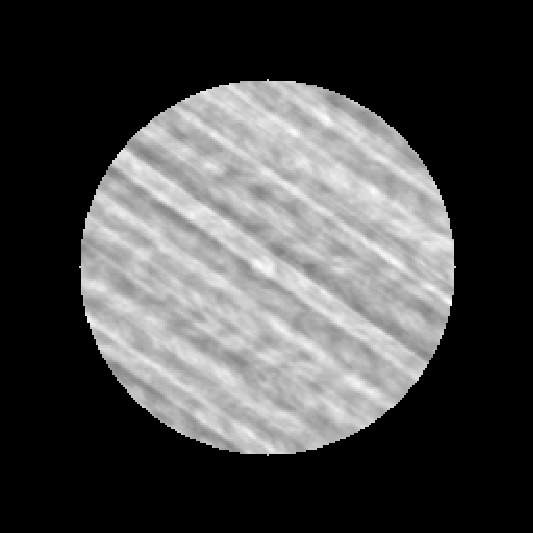}&\includegraphics[width=0.15\textwidth]{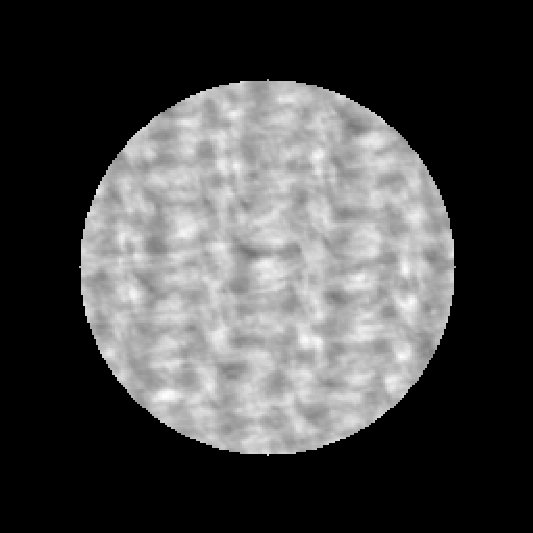}&\includegraphics[width=0.15\textwidth]{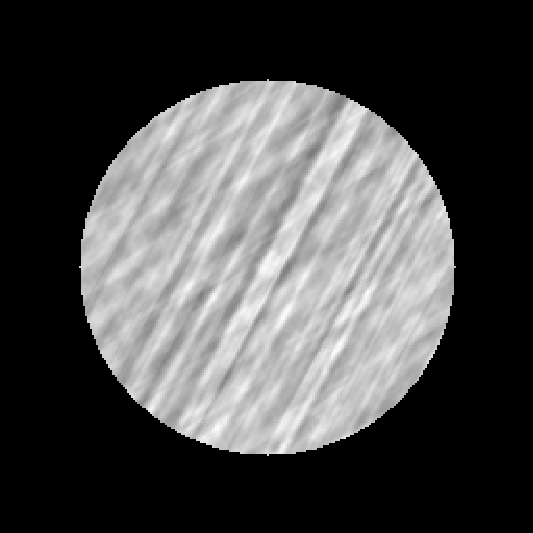}&\includegraphics[width=0.15\textwidth]{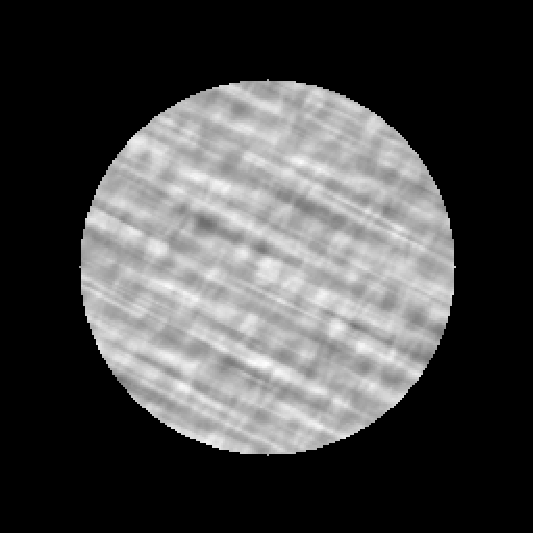}\\ 
\hline
\includegraphics[width=0.15\textwidth]{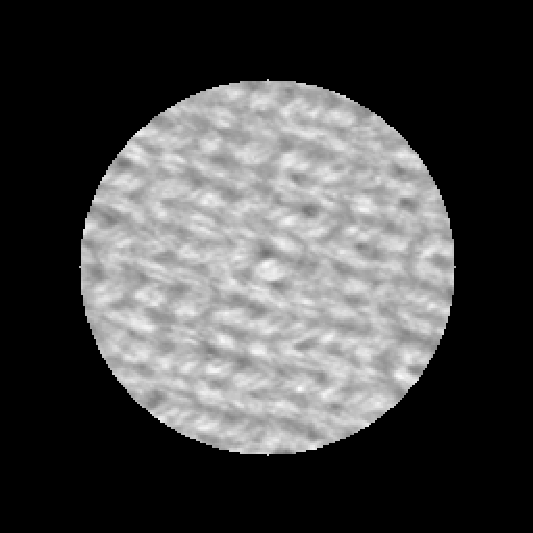}&\includegraphics[width=0.15\textwidth]{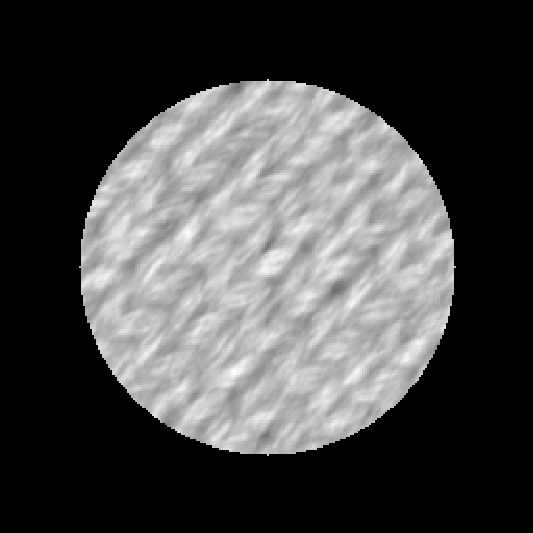}&\includegraphics[width=0.15\textwidth]{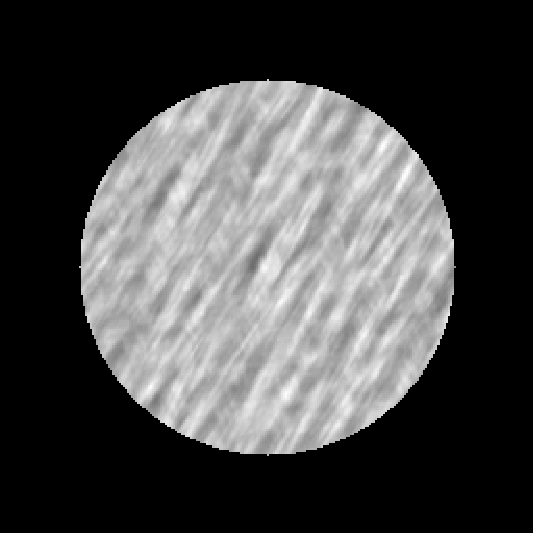}&\includegraphics[width=0.15\textwidth]{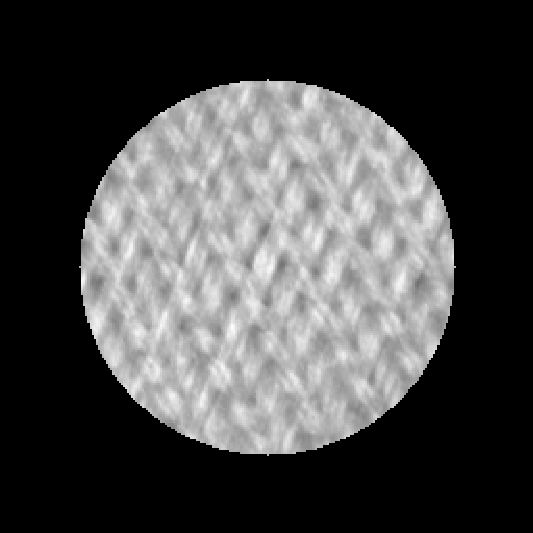}&\includegraphics[width=0.15\textwidth]{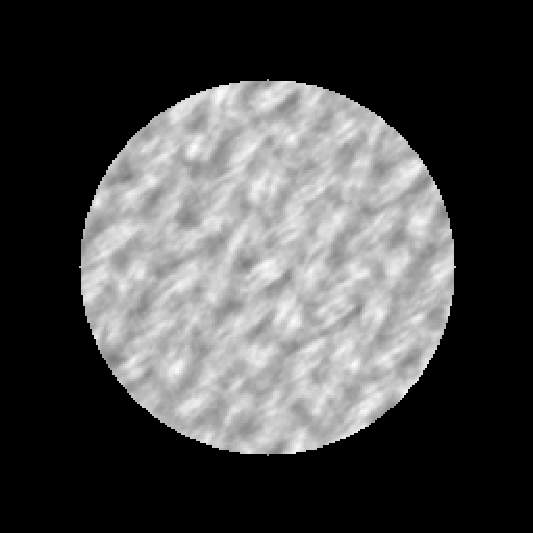}\\ 
\hline
\end{tabular} 
\end{table} 
\begin{table}[ht] 
\centering 
\begin{tabular}{||ccccc||} 
\hline
\multicolumn{5}{||c||}{texture 2} \\ 
\hline
\includegraphics[width=0.15\textwidth]{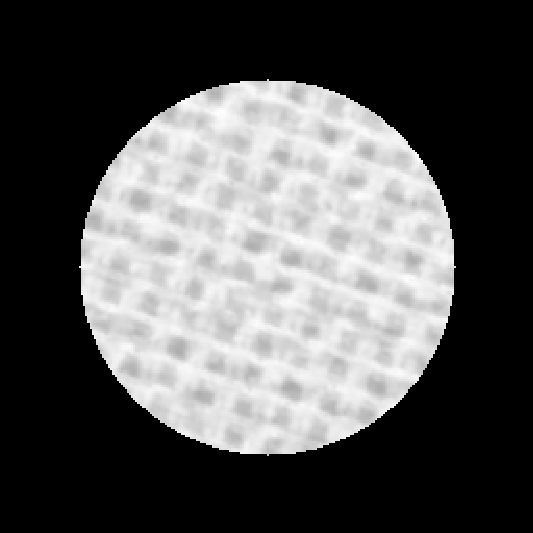}&\includegraphics[width=0.15\textwidth]{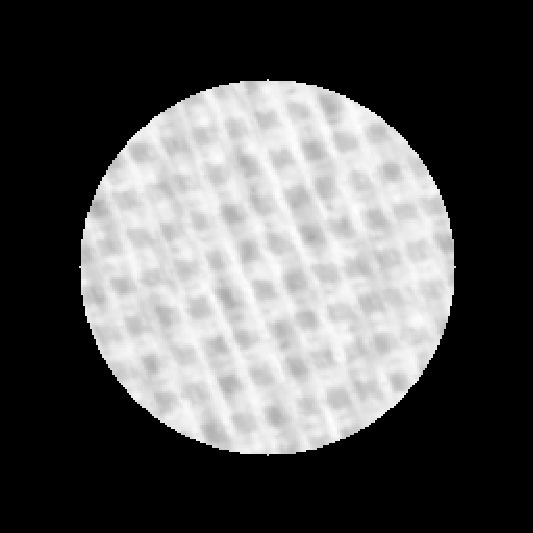}&\includegraphics[width=0.15\textwidth]{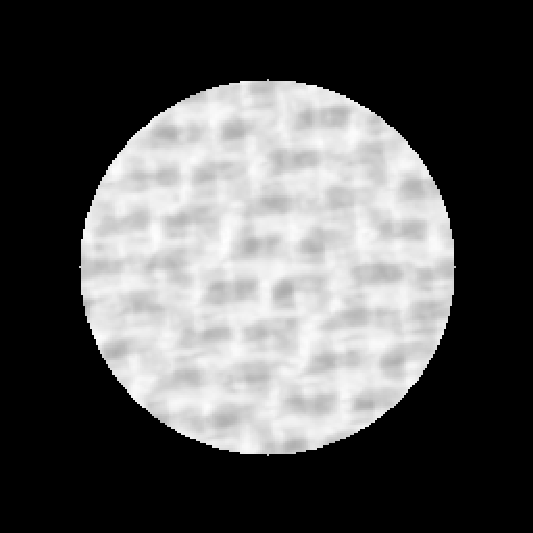}&\includegraphics[width=0.15\textwidth]{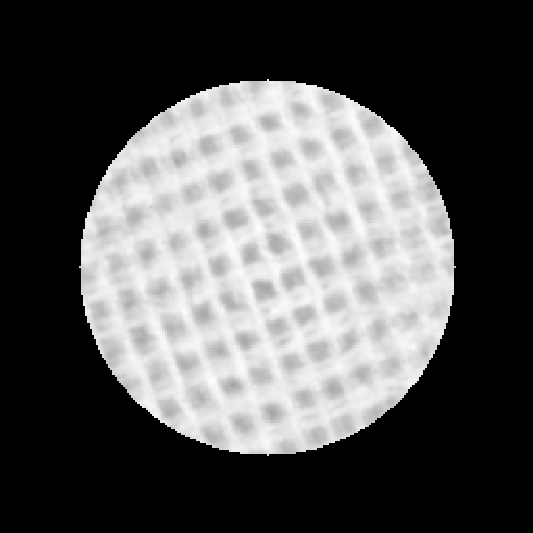}&\includegraphics[width=0.15\textwidth]{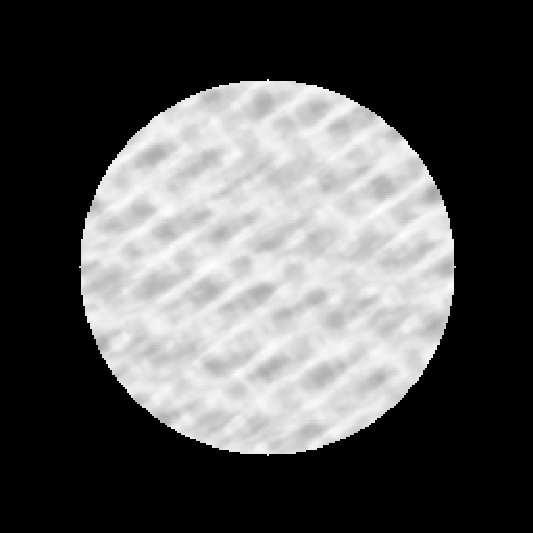}\\ 
\hline
\includegraphics[width=0.15\textwidth]{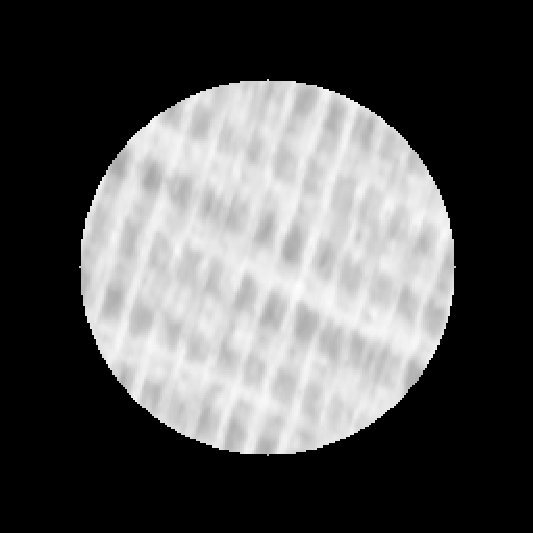}&\includegraphics[width=0.15\textwidth]{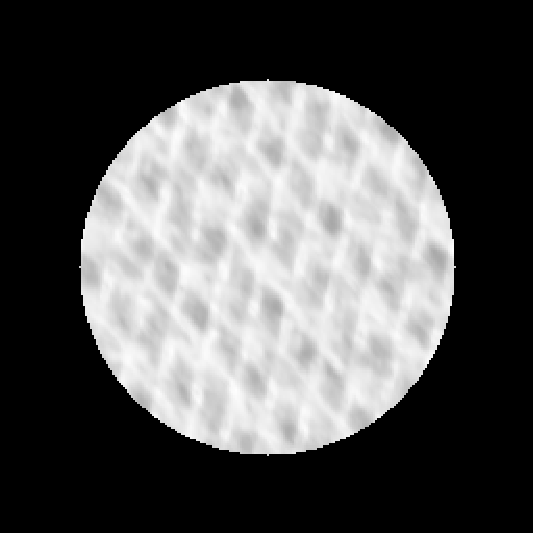}&\includegraphics[width=0.15\textwidth]{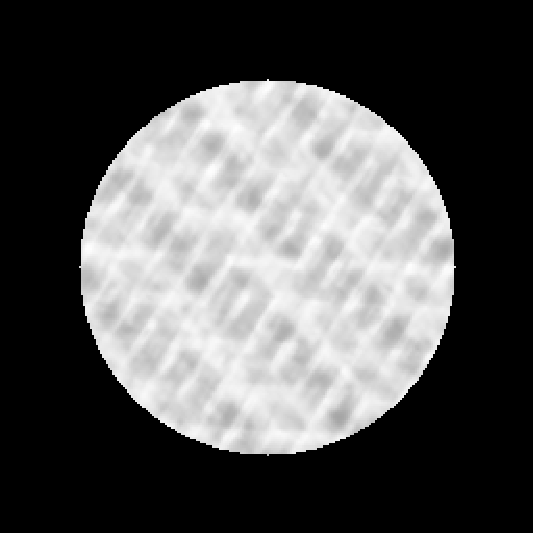}&\includegraphics[width=0.15\textwidth]{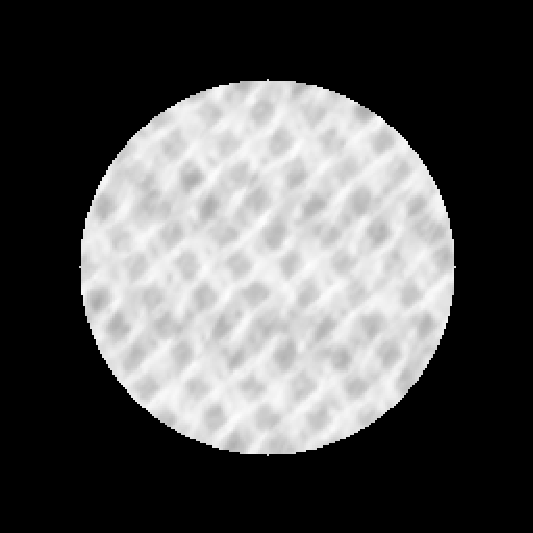}&\includegraphics[width=0.15\textwidth]{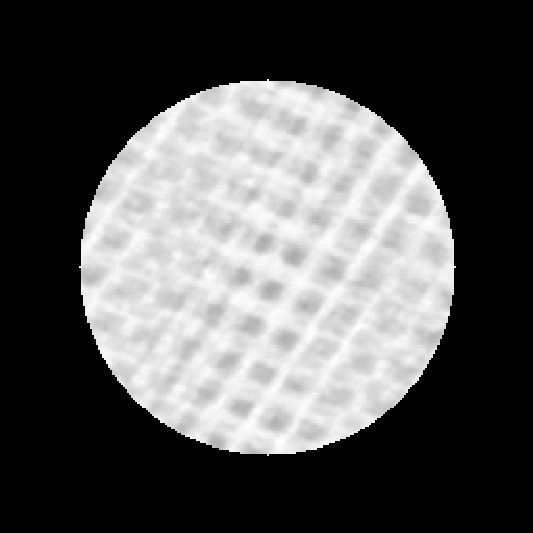}\\ 
\hline
\end{tabular} 
\end{table} 
\begin{table}[ht] 
\centering 
\begin{tabular}{||ccccc||} 
\hline
\multicolumn{5}{||c||}{texture 3} \\ 
\hline
\includegraphics[width=0.15\textwidth]{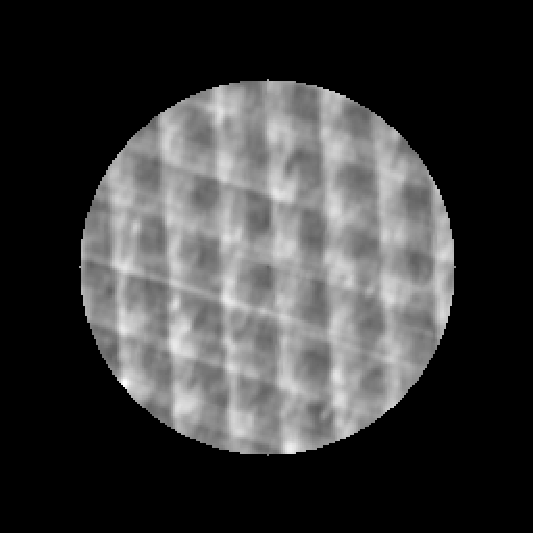}&\includegraphics[width=0.15\textwidth]{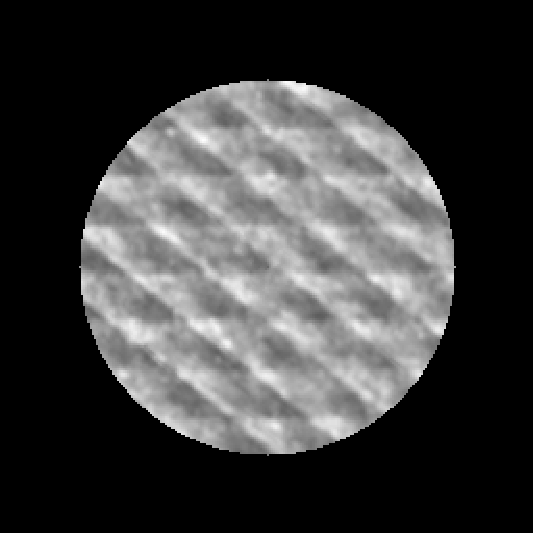}&\includegraphics[width=0.15\textwidth]{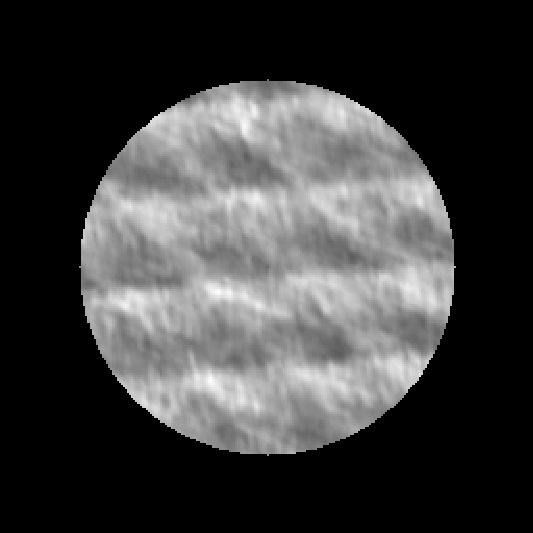}&\includegraphics[width=0.15\textwidth]{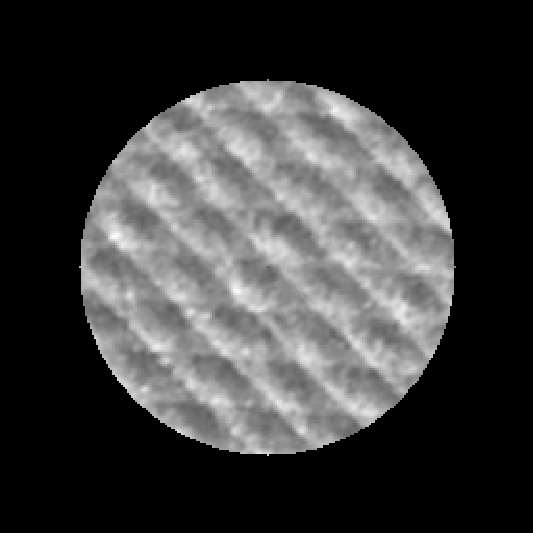}&\includegraphics[width=0.15\textwidth]{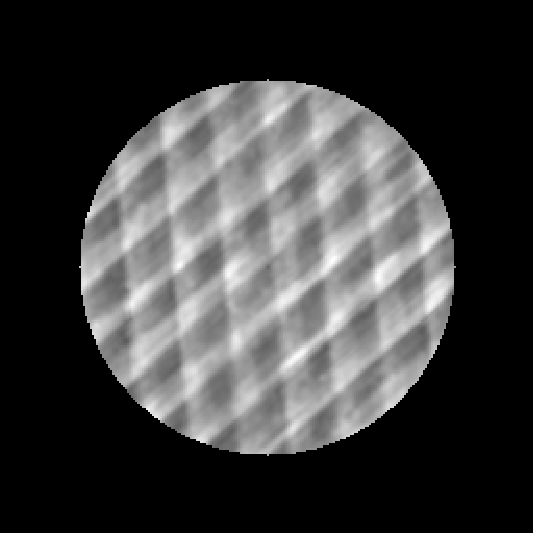}\\ 
\hline
\includegraphics[width=0.15\textwidth]{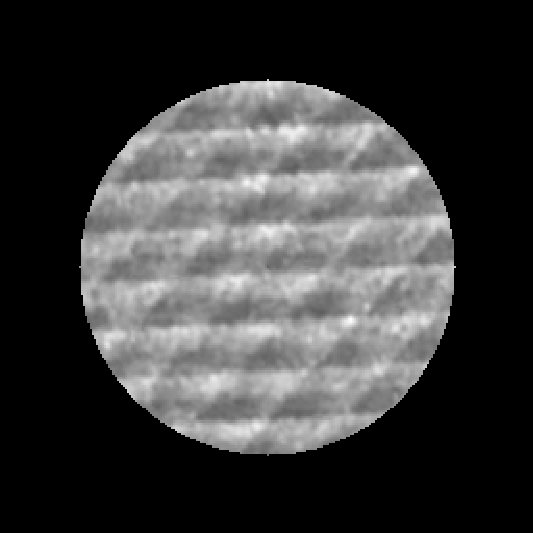}&\includegraphics[width=0.15\textwidth]{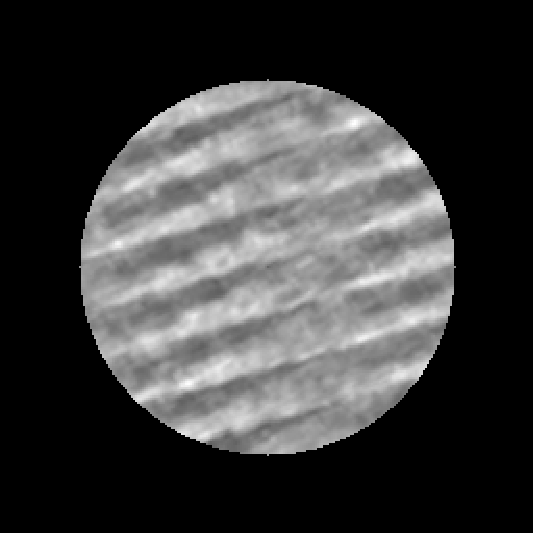}&\includegraphics[width=0.15\textwidth]{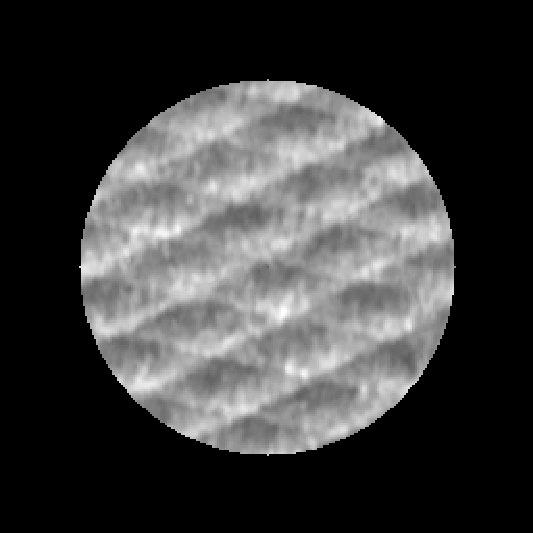}&\includegraphics[width=0.15\textwidth]{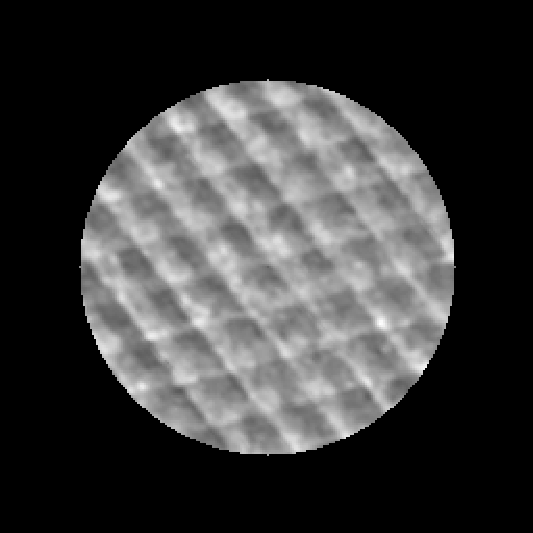}&\includegraphics[width=0.15\textwidth]{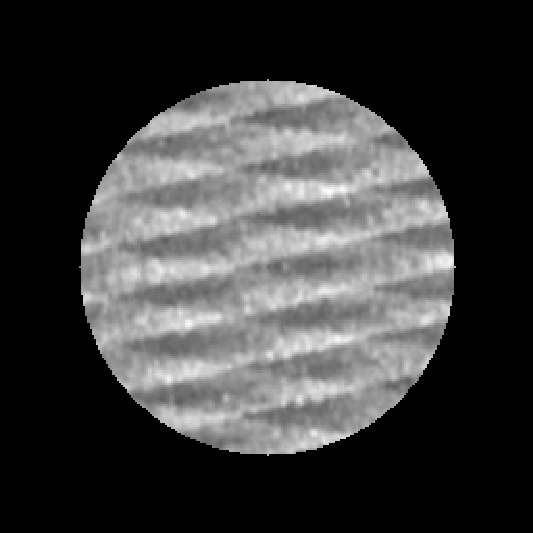}\\ 
\hline
\end{tabular} 
\end{table} 
\begin{table}[ht] 
\centering 
\begin{tabular}{||ccccc||} 
\hline
\multicolumn{5}{||c||}{texture 4} \\ 
\hline
\includegraphics[width=0.15\textwidth]{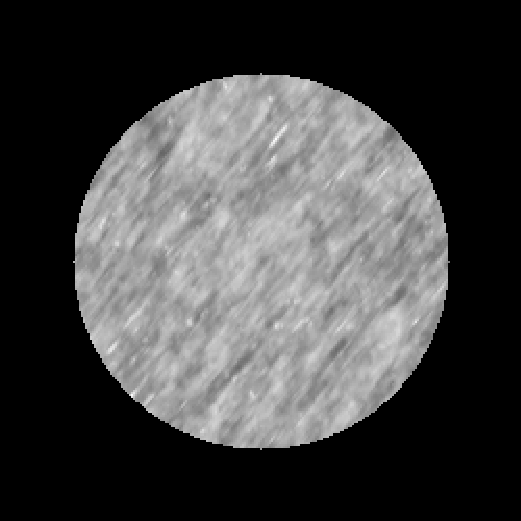}&\includegraphics[width=0.15\textwidth]{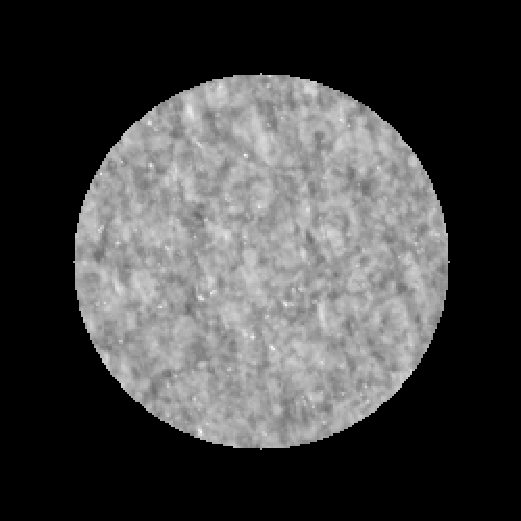}&\includegraphics[width=0.15\textwidth]{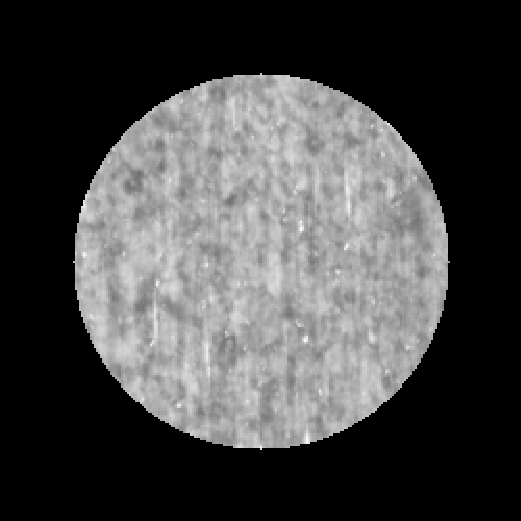}&\includegraphics[width=0.15\textwidth]{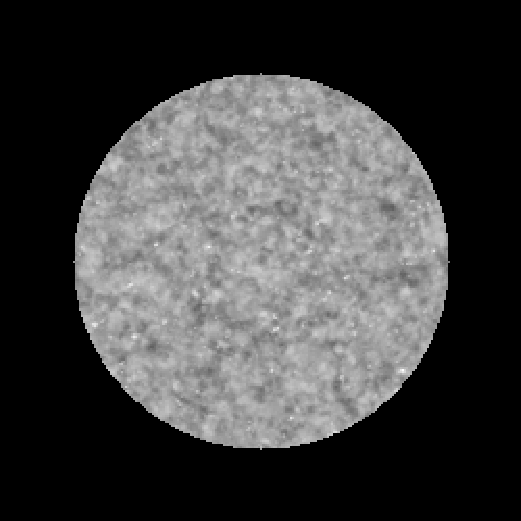}&\includegraphics[width=0.15\textwidth]{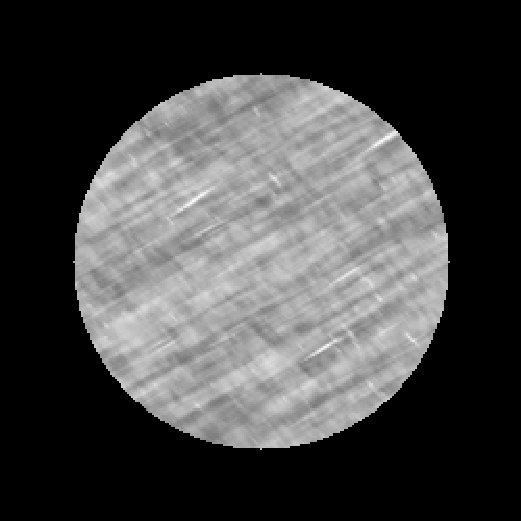}\\ 
\hline
\includegraphics[width=0.15\textwidth]{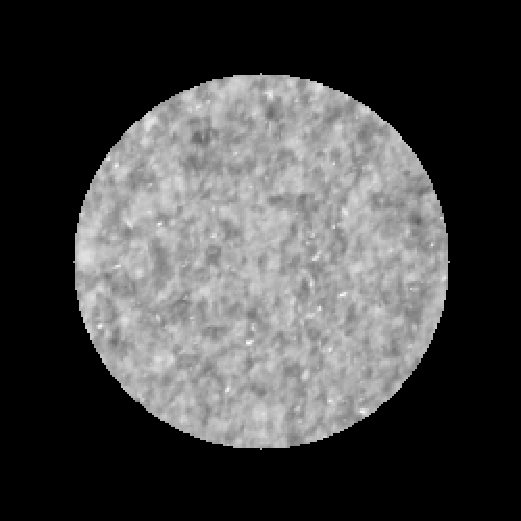}&\includegraphics[width=0.15\textwidth]{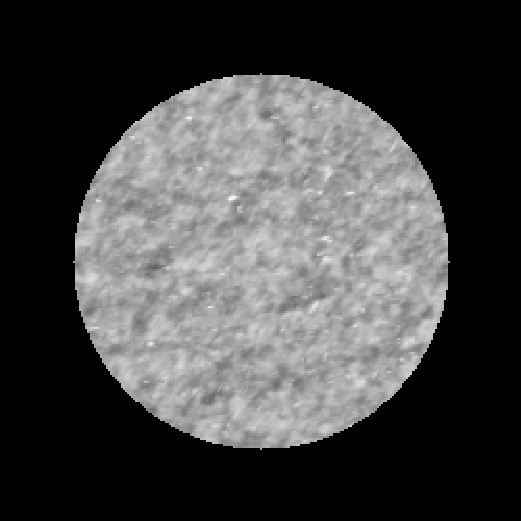}&\includegraphics[width=0.15\textwidth]{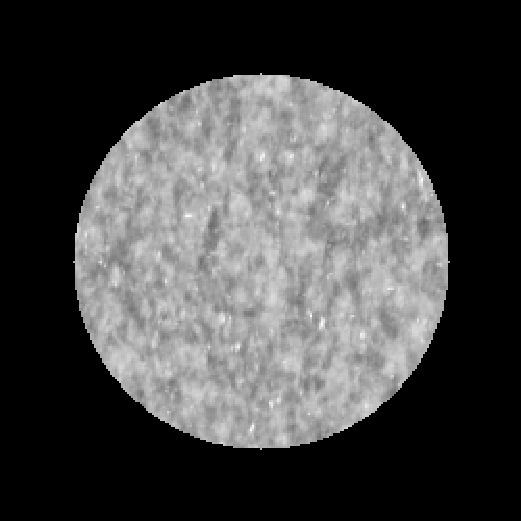}&\includegraphics[width=0.15\textwidth]{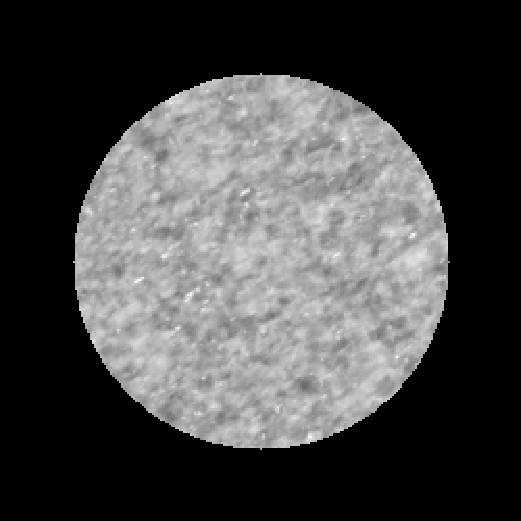}&\includegraphics[width=0.15\textwidth]{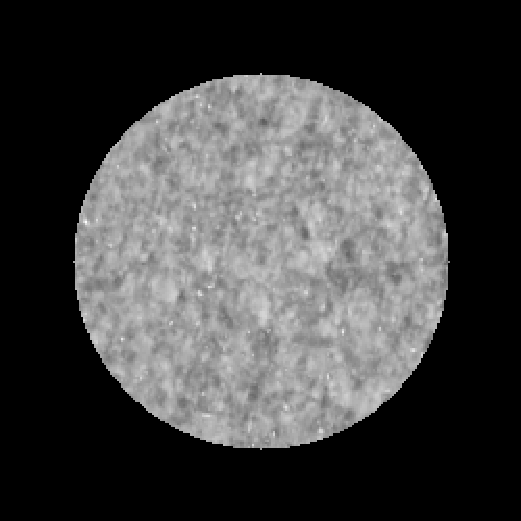}\\ 
\hline
\end{tabular} 
\end{table} 

\begin{table}[ht] 
\centering 
\begin{tabular}{||ccccc||} 
\hline
\multicolumn{5}{||c||}{texture 5} \\ 
\hline
\includegraphics[width=0.15\textwidth]{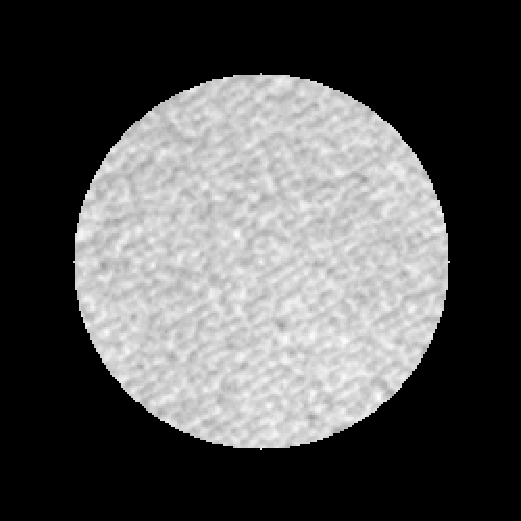}&\includegraphics[width=0.15\textwidth]{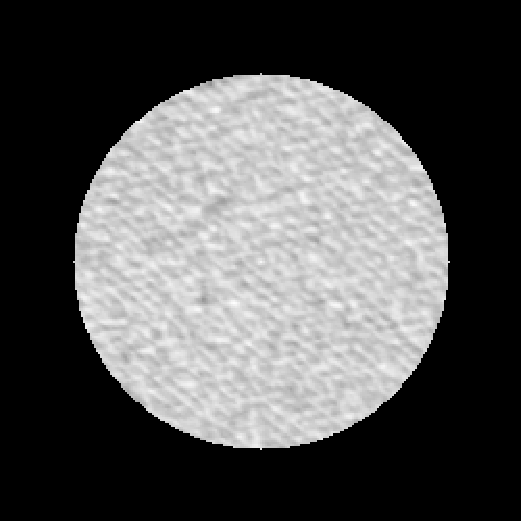}&\includegraphics[width=0.15\textwidth]{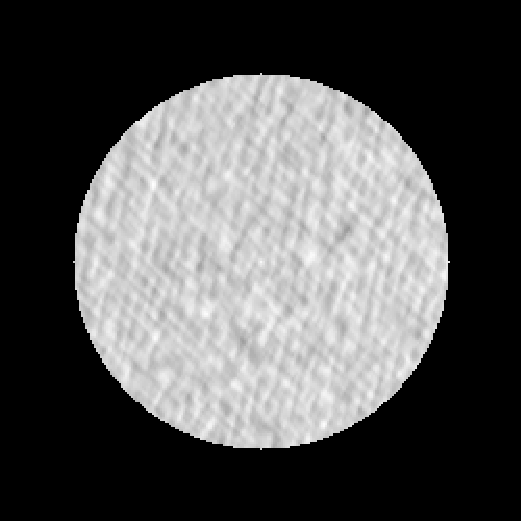}&\includegraphics[width=0.15\textwidth]{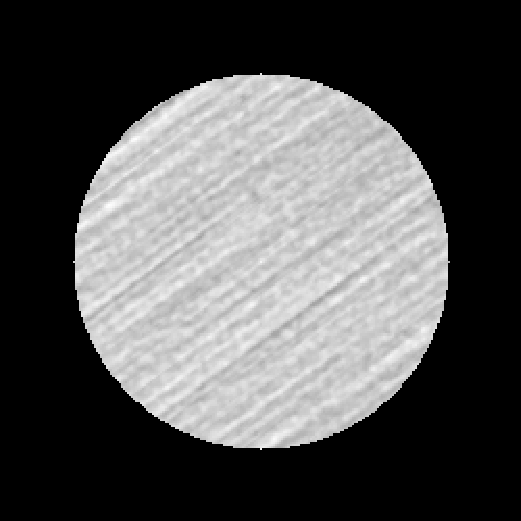}&\includegraphics[width=0.15\textwidth]{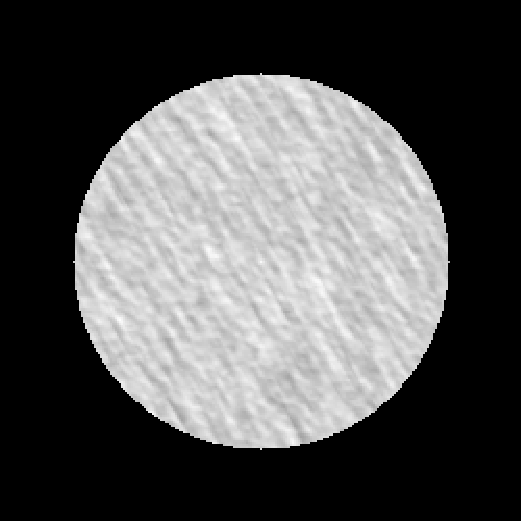}\\ 
\hline
\includegraphics[width=0.15\textwidth]{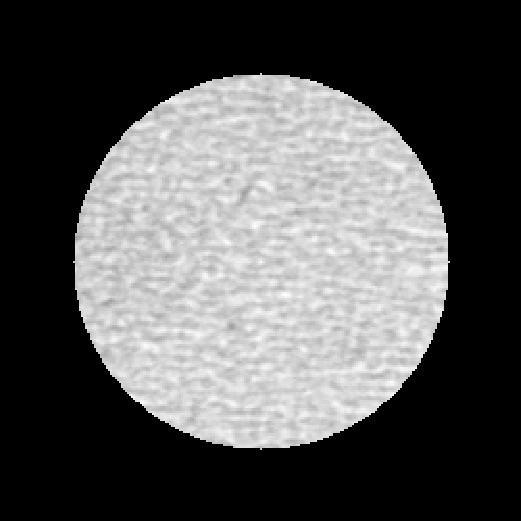}&\includegraphics[width=0.15\textwidth]{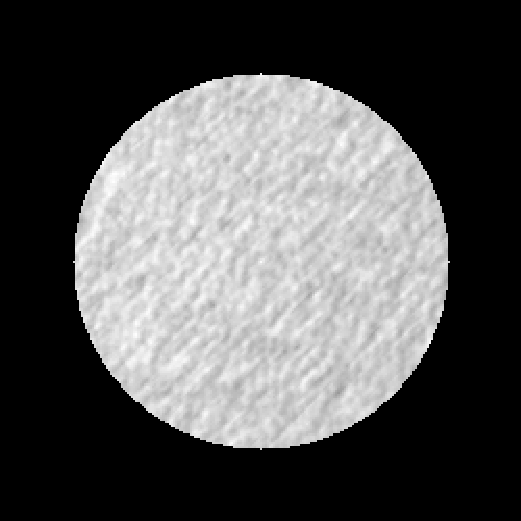}&\includegraphics[width=0.15\textwidth]{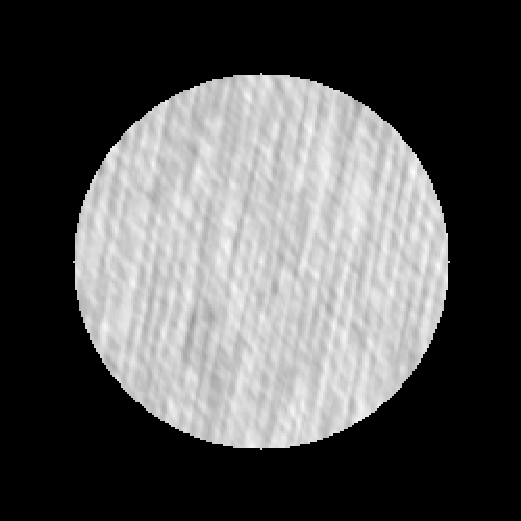}&\includegraphics[width=0.15\textwidth]{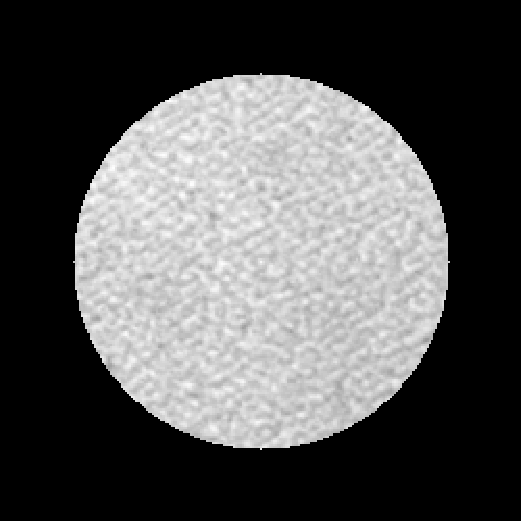}&\includegraphics[width=0.15\textwidth]{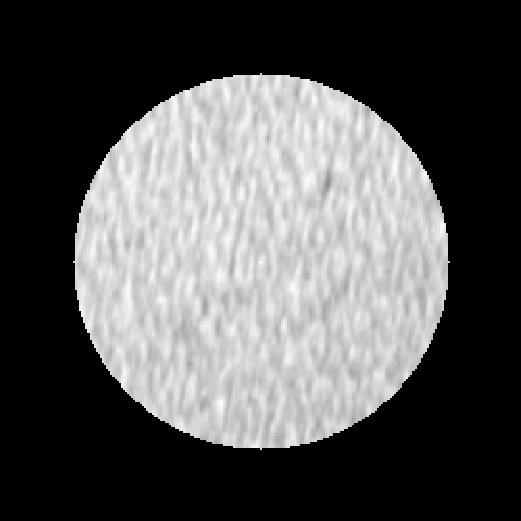}\\ 
\hline
\end{tabular} 
\end{table} 
\begin{table}[ht] 
\centering 
\begin{tabular}{||ccccc||} 
\hline
\multicolumn{5}{||c||}{texture 6} \\ 
\hline
\includegraphics[width=0.15\textwidth]{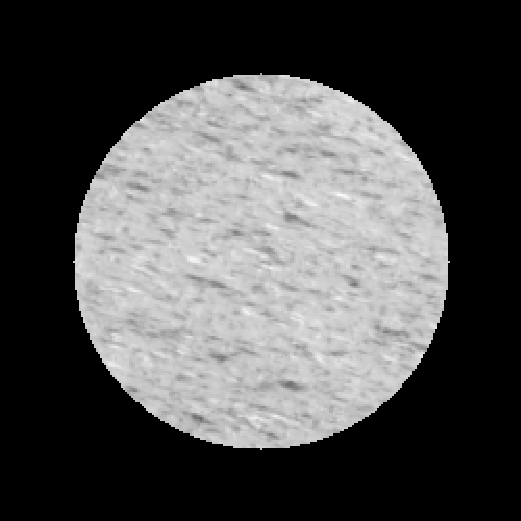}&\includegraphics[width=0.15\textwidth]{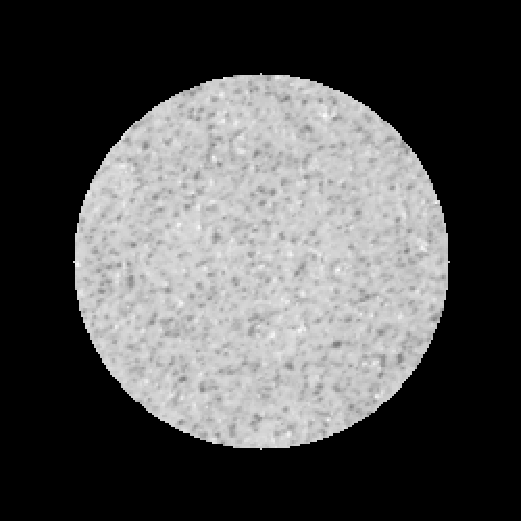}&\includegraphics[width=0.15\textwidth]{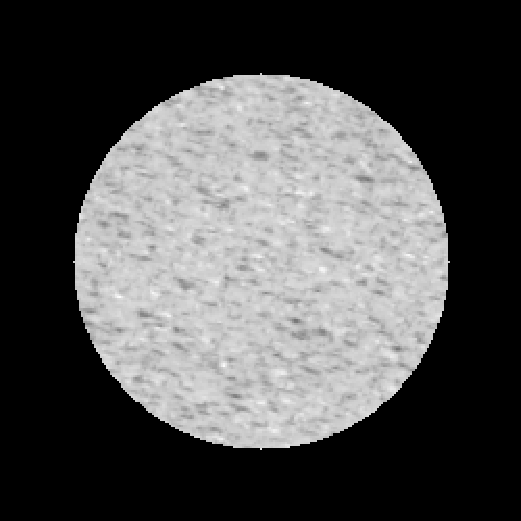}&\includegraphics[width=0.15\textwidth]{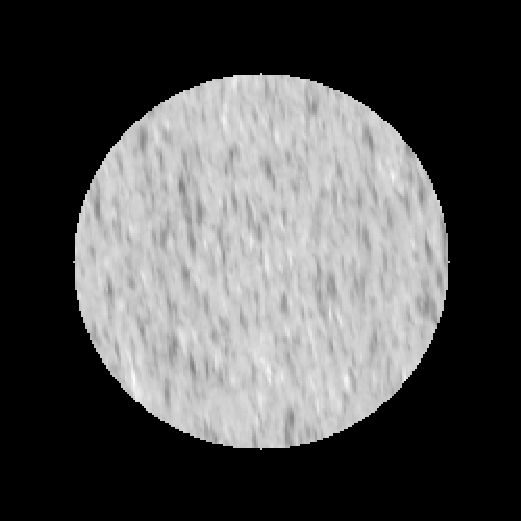}&\includegraphics[width=0.15\textwidth]{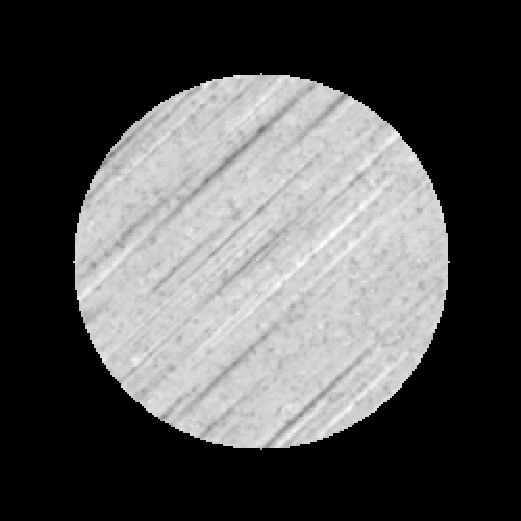}\\ 
\hline
\includegraphics[width=0.15\textwidth]{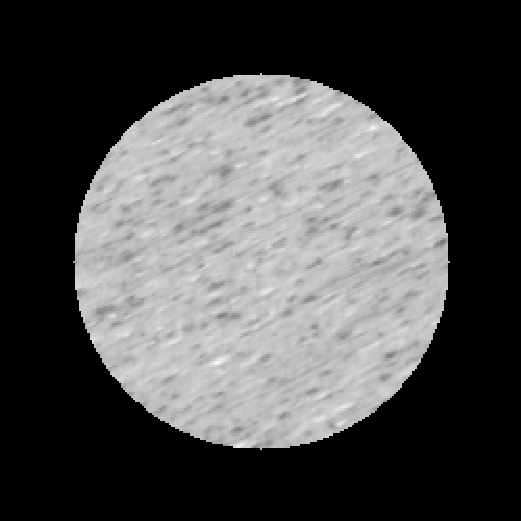}&\includegraphics[width=0.15\textwidth]{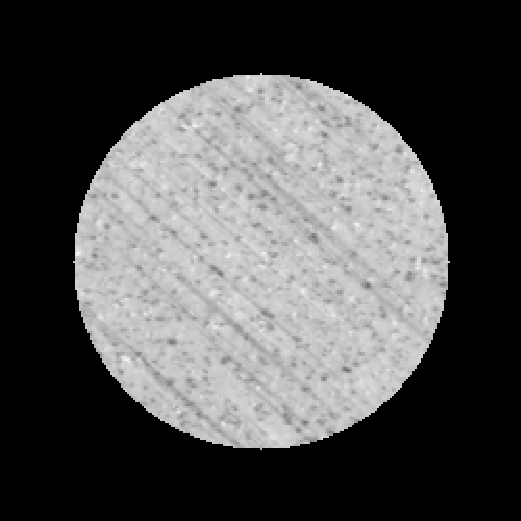}&\includegraphics[width=0.15\textwidth]{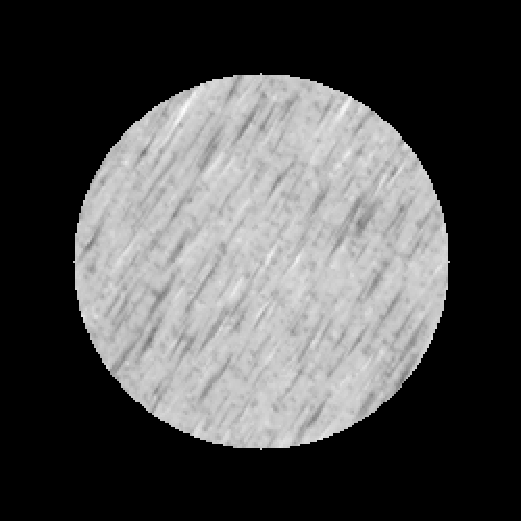}&\includegraphics[width=0.15\textwidth]{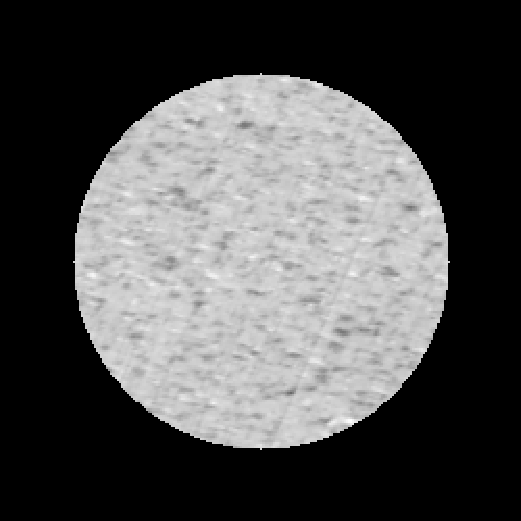}&\includegraphics[width=0.15\textwidth]{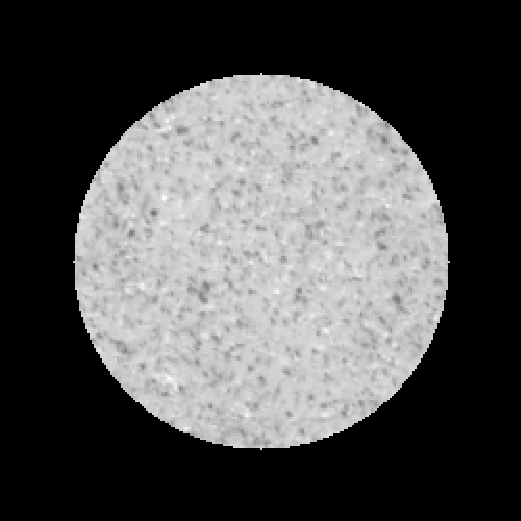}\\ 
\hline
\end{tabular} 
\end{table} 

\subsection{Base Textures}
The database contains 10 ``base samples'' for each of the six different textures ({\it texture 1-6}), all of which have a normalized average
gray-value. These ``base samples'' are used to generate separate training and test sets using arbitrary rotations and additive 
gray-value changes.
\clearpage

\subsection{\label{sec:app:texture_db_seg}Texture Segmentation}
\begin{figure}[ht]
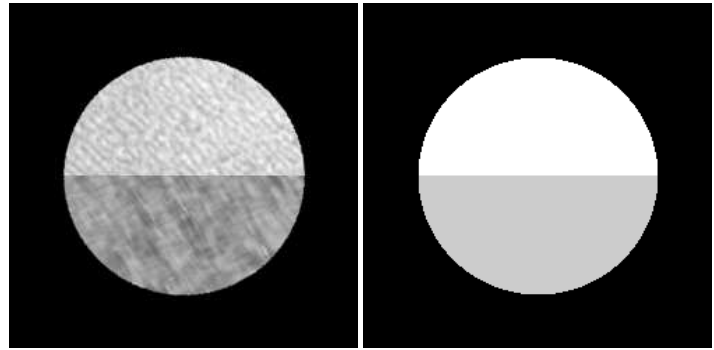

\centering
\includegraphics[width=0.3\textwidth]{FIGURES/example_test_texture.eps}
\includegraphics[width=0.3\textwidth]{FIGURES/example_test_label.eps}
\caption[Volume Texture Segmentation Test Samples]{\label{fig:app:db_test_texture} Test sample. {\bf Left:} the xy-slice shows how two randomly 
selected textures are combined in one test sample. {\bf Right:} ground-truth labeling.}
\end{figure}

\begin{figure}[ht]
\centering
\includegraphics[width=0.3\textwidth]{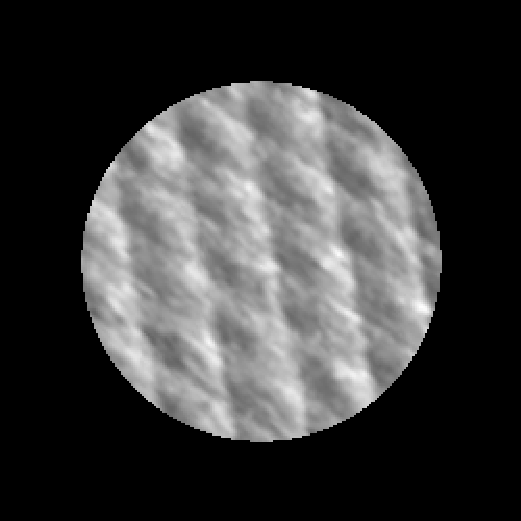}
\includegraphics[width=0.3\textwidth]{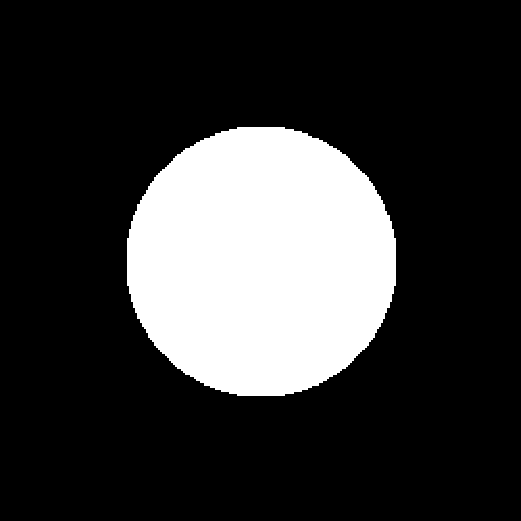}
\caption[Volume Texture Segmentation Training Samples]{\label{fig:app:db_train} Training sample. {\bf Left:} the xy-slice shows the volume
texture of a training sample. {\bf Right:} ground-truth labeling.}
\end{figure}

Given the ``base samples'' of the 3D volume textures, we generated a simple texture segmentation benchmark. One half of the ``base samples''
was used to build $60$ labeled training samples (see figure \ref{fig:app:db_train}) and the other half was used for the test samples.\\
The $200$ test samples consist of random combinations of two textures with a ground-truth labeling, where each the textures was 
rotated randomly and subject to an additive gray-value change (see figure \ref{fig:app:db_test_texture}).

\end{appendix}

\newpage
\addcontentsline{toc}{chapter}{Bibliography}
\bibliographystyle{abbrv}
\bibliography{main}

\end{document}